\documentclass[letterpaper]{article}
\usepackage[margin=1in,dvips]{geometry}
\usepackage{graphicx,psfrag,amsmath,amsthm,amssymb}
\usepackage{natbib}
\usepackage{url,color}
\usepackage{algorithm,algorithmic}

\newcommand{\A}{\ensuremath{\mathbf{A}}}
\newcommand{\B}{\ensuremath{\mathbf{B}}}

\newcommand{\I}{\ensuremath{\mathbf{I}}}

\newcommand{\X}{\ensuremath{\mathbf{X}}}
\newcommand{\Y}{\ensuremath{\mathbf{Y}}}
\newcommand{\Z}{\ensuremath{\mathbf{Z}}}

\newcommand{\f}{\ensuremath{\mathbf{f}}}

\newcommand{\h}{\ensuremath{\mathbf{h}}}

\newcommand{\x}{\ensuremath{\mathbf{x}}}

\newcommand{\z}{\ensuremath{\mathbf{z}}}

\newcommand{\balpha}{\ensuremath{\boldsymbol{\alpha}}}

\newcommand{\bbN}{\ensuremath{\mathbb{N}}}
\newcommand{\bbR}{\ensuremath{\mathbb{R}}}

\newcommand{\calB}{\ensuremath{\mathcal{B}}}

\newcommand{\calL}{\ensuremath{\mathcal{L}}}

\newcommand{\calO}{\ensuremath{\mathcal{O}}}

\newcommand{\norm}[1]{\left\lVert#1\right\rVert}

\newcommand{\caja}[4][1]{{%
    \renewcommand{\arraystretch}{#1}%
    \begin{tabular}[#2]{@{}#3@{}}%
      #4%
    \end{tabular}%
    }}

{%
\begin{list}{#1}{
\vspace{-\topsep}
\vspace{-\partopsep}
\setlength{\itemindent}{0cm}
\setlength{\rightmargin}{0cm}
\setlength{\listparindent}{0cm}
\settowidth{\labelwidth}{#1}
\setlength{\leftmargin}{\labelwidth}
\addtolength{\leftmargin}{\labelsep}
\setlength{\itemsep}{0cm}
}%
}%
{%
\end{list}
\vspace{-\topsep}
\vspace{-\partopsep}
}

{\begin{enumerate}%
}%
{\end{enumerate}}

\theoremstyle{plain}

\newtheorem*{lemma*}{Lemma}

\newtheorem*{prop*}{Proposition}

\theoremstyle{definition}

\newtheorem*{defn*}{Definition}

\newtheorem*{exmp*}{Example}

\newtheorem*{conj*}{Conjecture}

\theoremstyle{remark}

\newtheorem*{rmk*}{Remark}

\graphicspath{{grf/},{grf-miguel/}}

\bibpunct[, ]{(}{)}{;}{a}{,}{,} 

\title{Optimizing affinity-based binary hashing using auxiliary coordinates}
\author{
  Ramin Raziperchikolaei\hspace{5ex} Miguel \'A. Carreira-Perpi\~n\'an \\
  Electrical Engineering and Computer Science, University of California, Merced \\
  {\url{http://eecs.ucmerced.edu}}
}
\date{February 4, 2016}

\begin{document}

\maketitle

\begin{abstract}
  
  In supervised binary hashing, one wants to learn a function that maps a high-dimensional feature vector to a vector of binary codes, for application to fast image retrieval. This typically results in a difficult optimization problem, nonconvex and nonsmooth, because of the discrete variables involved. Much work has simply relaxed the problem during training, solving a continuous optimization, and truncating the codes a posteriori. This gives reasonable results but is quite suboptimal. Recent work has tried to optimize the objective directly over the binary codes and achieved better results, but the hash function was still learned a posteriori, which remains suboptimal. We propose a general framework for learning hash functions using affinity-based loss functions that uses auxiliary coordinates. This closes the loop and optimizes jointly over the hash functions and the binary codes so that they gradually match each other. The resulting algorithm can be seen as a corrected, iterated version of the procedure of optimizing first over the codes and then learning the hash function. Compared to this, our optimization is guaranteed to obtain better hash functions while being not much slower, as demonstrated experimentally in various supervised datasets. In addition, our framework facilitates the design of optimization algorithms for arbitrary types of loss and hash functions.

\end{abstract}

\section{Introduction}
\label{s:intro}

Information retrieval arises in several applications, most obviously web search. For example, in image retrieval, a user is interested in finding similar images to a query image. Computationally, this essentially involves defining a high-dimensional feature space where each relevant image is represented by a vector, and then finding the closest points (nearest neighbors) to the vector for the query image, according to a suitable distance \citep{Shakhn_06a}. For example, one can use features such as SIFT \citep{Lowe04a} or GIST \citep{OlivaTorral01a} and the Euclidean distance for this purpose. Finding nearest neighbors in a dataset of $N$ images (where $N$ can be millions), each a vector of dimension $D$ (typically in the hundreds) is slow, since exact algorithms run essentially in time $\calO(ND)$ and space $\calO(ND)$ (to store the image dataset). In practice, this is approximated, and a successful way to do this is \emph{binary hashing} \citep{GraumanFergus13a}. Here, given a high-dimensional vector $\x \in \bbR^D$, the hash function \h\ maps it to a $b$-bit vector $\z = \h(\x) \in \{-1,+1\}^b$, and the nearest neighbor search is then done in the binary space. This now costs $\calO(Nb)$ time and space, which is orders of magnitude faster because typically $b < D$ and, crucially, (1) operations with binary vectors (such as computing Hamming distances) are very fast because of hardware support, and (2) the entire dataset can fit in (fast) memory rather than slow memory or disk.

The disadvantage is that the results are inexact, since the neighbors in the binary space will not be identical to the neighbors in the original space. However, the approximation error can be controlled by using sufficiently many bits and by \emph{learning a good hash function}. This has been the topic of much work in recent years. The general approach consists of defining a supervised objective that has a small value for good hash functions and minimizing it. Ideally, such an objective function should be minimal when the neighbors of any given image are the same in both original and binary spaces. Practically in information retrieval, this is often evaluated using precision and recall. However, this ideal objective cannot be easily optimized over hash functions, and one uses approximate objectives instead. Many such objectives have been proposed in the literature. We focus here on \emph{affinity-based loss functions}, which directly try to preserve the original similarities in the binary space. Specifically, we consider objective functions of the form
\begin{equation}
  \label{e:objfcn}
  \min{ \calL(\h) = \sum^N_{n,m=1}{ L(\h(\x_n),\h(\x_m)\mathpunct{;}\ y_{nm}) } }
\end{equation}
where $\X = (\x_1,\dots,\x_N)$ is the high-dimensional dataset of feature vectors, $\min_{\h}$ means minimizing over the parameters of the hash function \h\ (e.g.\ over the weights of a linear SVM), and $L(\cdot)$ is a loss function that compares the codes for two images (often through their Hamming distance $\norm{\h(\x_n)-\h(\x_m)}$) with the ground-truth value $y_{nm}$ that measures the affinity in the original space between the two images $\x_n$ and $\x_m$ (distance, similarity or other measure of neighborhood; \citealp{GraumanFergus13a}). The sum is often restricted to a subset of image pairs $(n,m)$ (for example, within the $k$ nearest neighbors of each other in the original space), to keep the runtime low. Examples of these objective functions (described below) include models developed for dimension reduction, be they spectral such as Laplacian Eigenmaps \citep{BelkinNiyogi03b} and Locally Linear Embedding \citep{RoweisSaul00a}, or nonlinear such as the Elastic Embedding \citep{Carreir10a} or $t$-SNE \citep{MaatenHinton08a}; as well as objective functions designed specifically for binary hashing, such as Supervised Hashing with Kernels (KSH) \citep{Liu_12c}, Binary Reconstructive Embeddings (BRE) \citep{KulisDarrel09a} or Semi-supervised sequential Projection Learning Hashing (SPLH) \citep{Wang_12a}.

If the hash function \h\ was a continuous function of its input \x\ and its parameters, one could simply apply the chain rule to compute derivatives over the parameters of \h\ of the objective function~\eqref{e:objfcn} and then apply a nonlinear optimization method such as gradient descent. This would be guaranteed to converge to an optimum under mild conditions (for example, Wolfe conditions on the line search), which would be global if the objective is convex and local otherwise \citep{NocedalWright06a}. Hence, optimally learning the function \h\ would be in principle doable (up to local optima), although it would still be slow because the objective can be quite nonlinear and involve many terms.

In binary hashing, the optimization is much more difficult, because in addition to the previous issues, the hash function must output binary values, hence the problem is not just generally nonconvex, but also nonsmooth. In view of this, much work has sidestepped the issue and settled on a simple but suboptimal solution. First, one defines the objective function~\eqref{e:objfcn} directly on the $b$-dimensional codes of each image (rather than on the hash function parameters) and optimizes it assuming continuous codes (in $\bbR^b$). Then, one binarizes the codes for each image. Finally, one learns a hash function given the codes. Optimizing the affinity-based loss function~\eqref{e:objfcn} can be done using spectral methods or nonlinear optimization as described above. Binarizing the codes has been done in different ways, from simply rounding them to $\{-1,+1\}$ using zero as threshold \citep{Weiss_09a,Zhang_10e,Liu_11a,Liu_12c}, to optimally finding a threshold \citep{Liu_11a,Strech_12a}, to rotating the continuous codes so that thresholding introduces less error \citep{YuShi03a,Gong_13a}. Finally, learning the hash function for each of the $b$ output bits can be considered as a binary classification problem, where the resulting classifiers collectively give the desired hash function, and can be solved using various machine learning techniques. Several works (e.g.\ \citealp{Zhang_10e,Lin_13a,Lin_14b}) have used this approach, which does produce reasonable hash functions (in terms of retrieval measures such as precision and recall).

In order to do better, one needs to take into account during the optimization (rather than after the optimization) the fact that the codes are constrained to be binary. This implies attempting directly the discrete optimization of the affinity-based loss function over binary codes. This is a daunting task, since this is usually an NP-complete problem with $Nb$ binary variables altogether, and practical applications could make this number as large as millions or beyond. Recent works have applied alternating optimization (with various refinements) to this, where one optimizes over a usually small subset of binary variables given fixed values for the remaining ones \citep{Lin_13a,Lin_14b}, and this did result in very competitive precision/recall compared with the state-of-the-art. This is still slow and future work will likely improve it, but as of now it provides an option to learn better binary codes.

Of the three-step suboptimal approach mentioned (learn continuous codes, binarize them, learn hash function), these works manage to join the first two steps and hence learn binary codes. Then, one learns the hash function given these binary codes. Can we do better? Indeed, in this paper \emph{we show that all elements of the problem (binary codes and hash function) can be incorporated in a single algorithm that optimizes jointly over them}. Hence, by initializing it from binary codes from the previous approach, this algorithm is guaranteed to achieve a lower error and learn better hash functions. In fact, our framework can be seen as an iterated, corrected version of the two-step approach: learn binary codes \emph{given the current hash function}, learn hash functions given codes, iterate (note the emphasis). The key to achieve this in a principled way is to use a recently proposed \emph{method of auxiliary coordinates (MAC)} for optimizing ``nested'' systems, i.e., consisting of the composition of two or more functions or processing stages. MAC introduces new variables and constraints that cause decoupling between the stages, resulting in the mentioned alternation between learning the hash function and learning the binary codes. Section~\ref{s:objfcn} reviews affinity-based loss functions, section~\ref{s:MAC} describes our MAC-based proposed framework, section~\ref{s:expts} evaluates it in several supervised datasets, using linear and nonlinear hash functions, and section~\ref{s:discussion} discusses implications of this work.

\paragraph{Related work}
\label{s:related}

Although one can construct hash functions without training data \citep{AndoniIndyk08a,KulisGrauman12a}, we focus on methods that learn the hash function given a training set, since they perform better, and our emphasis is in optimization. The learning can be unsupervised, which attempts to preserve distances in the original space, or supervised, which in addition attempts to preserve label similarity. Many objective functions have been proposed to achieve this and we focus on affinity-based ones. These create an affinity matrix for a subset of training points based on their distances (unsupervised) or labels (supervised) and combine it with a loss function \citep{Liu_12c,KulisDarrel09a,NorouzFleet11a,Lin_13a,Lin_14b}. Some methods optimize this directly over the hash function. For example, Binary Reconstructive Embeddings \citep{KulisDarrel09a} use alternating optimization over the weights of the hash functions. Supervised Hashing with Kernels \citep{Liu_12c} learns hash functions sequentially by considering the difference between the inner product of the codes and the corresponding element of the affinity matrix. Although many approaches exist, a common theme is to apply a greedy approach where one first finds codes using an affinity-based loss function, and then fits the hash functions to them (usually by training a classifier). The codes can be found by relaxing the problem and binarizing its solution \citep{Weiss_09a,Zhang_10e,Liu_11a}, or by approximately solving for the binary codes using some form of alternating optimization (possibly combined with GraphCut), as in two-step hashing \citep{Lin_13a,Lin_14b,Ge_14a}, or by using relaxation in other ways \citep{Liu_12c,NorouzFleet11a}.

\section{Nonlinear embedding and affinity-based loss functions for binary hashing}
\label{s:objfcn}

The dimensionality reduction literature has developed a number of objective functions of the form~\eqref{e:objfcn} (often called ``embeddings'') where the low-dimensional projection $\z_n\in\bbR^b$ of each high-dimensional data point $\x_n\in\bbR^D$ is a free, real-valued parameter. The neighborhood information is encoded in the $y_{nm}$ values (using labels in supervised problems, or distance-based affinities in unsupervised problems). A representative example is the elastic embedding \citep{Carreir10a}, where $L(\z_n,\z_m\mathpunct{;}\ y_{nm})$ has the form:
\begin{equation}
  \label{e:EE}
  y^+_{nm} \norm{\z_n-\z_m}^2 + \lambda y^-_{nm} \exp{(-\norm{\z_n-\z_m}^2)}, \ \lambda > 0
\end{equation}
where the first term tries to project true neighbors (having $y^+_{nm} > 0$) close together, while the second repels all non-neighbors' projections (having $y^-_{nm} > 0$) from each other. Laplacian Eigenmaps \citep{BelkinNiyogi03b} and Locally Linear Embedding \citep{RoweisSaul00a} result from replacing the second term above with a constraint that fixes the scale of \Z, which results in an eigenproblem rather than a nonlinear optimization, but also produces more distorted embeddings. Other objectives exist, such as $t$-SNE \citep{MaatenHinton08a}, that do not separate into functions of pairs of points. Optimizing nonlinear embeddings is quite challenging, but much progress has been done recently \citep{Carreir10a,VladymCarreir12a,Maaten13a,Yang_13a,VladymCarreir14a}. Although these models were developed to produce continuous projections, they have been successfully used for binary hashing too by truncating their codes \citep{Weiss_09a,Zhang_10e} or using the two-step approach of \citep{Lin_13a,Lin_14b}.

Other loss functions have been developed specifically for hashing, where now $\z_n$ is a $b$-bit vector (where binary values are in $\{-1,+1\}$). For example (see a longer list in \citealp{Lin_13a}), for Supervised Hashing with Kernels (KSH) $L(\z_n,\z_m\mathpunct{;}\ y_{nm})$ has the form
\begin{equation}
  \label{e:KSH}
  (\z_n^T \z_m - b y_{nm})^2
\end{equation}
where $y_{nm}$ is $1$ if $\x_n$, $\x_m$ are similar and $-1$ if they are dissimilar. Binary Reconstructive Embeddings \citep{KulisDarrel09a} uses $\smash{(\frac{1}{b} \norm{\z_n - \z_m}^2 -y_{nm})^2}$ where $y_{nm} = \smash{\frac{1}{2} \norm{\x_n-\x_m}^2}$. The exponential variant of SPLH \citep{Wang_12a} proposed by \citet{Lin_13a} (which we call eSPLH) uses $\exp(-\frac{1}{b} y_{nm}\z_n^T\z_n)$.
Our approach can be applied to any of these loss functions, though we will mostly focus on the KSH loss for simplicity. When the variables \Z\ are binary, we will call these optimization problems \emph{binary embeddings}, in analogy to the more traditional continuous embeddings for dimension reduction.

\section{Learning codes and hash functions using auxiliary coordinates}
\label{s:MAC}

The optimization of the loss $\calL(\h)$ in eq.~\eqref{e:objfcn} is difficult because of the thresholded hash function, \emph{which appears as the argument of the loss function $L$}. We use the recently proposed \emph{method of auxiliary coordinates (MAC)} \citep{CarreirWang12a,CarreirWang14a}, which is a meta-algorithm to construct optimization algorithms for nested functions. This proceeds in 3 stages. First, we introduce new variables (the ``auxiliary coordinates'') as equality constraints into the problem, with the goal of unnesting the function. We can achieve this by introducing one binary vector $\z_n \in \{-1,+1\}$ for each point. This transforms the original, unconstrained problem into the following, constrained problem:
\begin{equation}
  \label{e:MAC-constrained}
  \min_{\h,\Z}{ \sum^N_{n=1}{ L(\z_n,\z_m\mathpunct{;}\ y_{nm}) } } \ \text{ s.t.\ } \
  \left\{
  \begin{array}{@{}c@{}}
    \z_1 = \h(\x_1) \\[-0.7ex]
    \cdots \\[-0.7ex]
    \z_N = \h(\x_N)
  \end{array}
  \right.
\end{equation}
which is seen to be equivalent to~\eqref{e:objfcn} by eliminating \Z. We recognize as the objective function the ``embedding'' form of the loss function, except that the ``free'' parameters $\z_n$ are in fact constrained to be the deterministic outputs of the hash function \h.

Second, we solve the constrained problem using a penalty method, such as the quadratic-penalty or augmented Lagrangian \citep{NocedalWright06a}. We discuss here the former for simplicity. We solve the following minimization problem (unconstrained again, but dependent on $\mu$) while progressively increasing $\mu$, so the constraints are eventually satisfied:
\begin{equation}
  \label{e:MAC-QP}
  \min \calL_P(\h,\Z;\mu) = \sum^N_{n,m=1}{ L(\z_n,\z_m\mathpunct{;}\ y_{nm}) }\ + \mu \sum^N_{n=1}{ \norm{\z_n - \h(\x_n)}^2 } \qquad \text{s.t.} \qquad \z_1,\dots,\z_N \in \{-1,1\}^b.
\end{equation}
The quadratic penalty $\norm{\z_n - \h(\x_n)}^2$ is proportional to the Hamming distance between the binary vectors $\z_n$ and $\h(\x_n)$.

Third, we apply alternating optimization over the binary codes \Z\ and the hash function parameters \h. This results in iterating the following two steps (described in detail later):
\begin{itemize}
\item Optimize the binary codes $\z_1 ,\dots,\z_N$ given \h\ (hence, given the output binary codes $\h(\x_1),\dots,\h(\x_N)$ for each of the $N$ images). This can be seen as a \emph{regularized binary embedding}, because the projections \Z\ are encouraged to be close to the hash function outputs $\h(\X)$. Here, we try two different approaches \citep{Lin_13a,Lin_14b} with some modifications.
\item Optimize the hash function \h\ given binary codes \Z. This reduces to training $b$ binary classifiers using \X\ as inputs and \Z\ as targets.
\end{itemize}
This is very similar to the two-step (TSH) approach of \citet{Lin_13a}, except that the latter learns the codes \Z\ in isolation, rather than given the current hash function, so iterating the two-step approach would change nothing, and it does not optimize the loss \calL. More precisely, TSH corresponds to optimizing $\calL_P$ for $\mu\rightarrow 0^+$. In practice, we start from a very small value of $\mu$ (hence, initialize MAC from the result of TSH), and increase $\mu$ slowly while optimizing $\calL_P$, until the equality constraints are satisfied, i.e., $\z_n = \h(\x_n)$ for $n=1,\dots,N$.

Fig.~\ref{f:MAC-alg} gives the overall MAC algorithm to learn a hash function by optimizing an affinity-based loss function. We now describe the steps over \h\ and \Z, and the path followed by the iterates as a function of $\mu$.

\begin{figure}[t]
  \begin{center}
    \setlength{\fboxsep}{1ex}
    \framebox{%
      \begin{minipage}[c]{0.70\columnwidth}
        \begin{tabbing}
          n \= n \= n \= n \= n \= \kill
          \underline{\textbf{input}} $\X_{D \times N} = (\x_1,\dots,\x_N)$, $\Y_{N \times N} = (y_{nm})$, $b \in \bbN$ \\
          Initialize $\Z_{b \times N} = (\z_1,\dots,\z_N) \in \{0,1\}^{bN}$ \\
          \underline{\textbf{for}} $\mu = 0 < \mu_1 < \dots < \mu_{\infty}$ \+ \\
          \underline{\textbf{for}} $i = 1,\dots,b$ \` {\small\textsf{\h\ step}} \+ \\
          $h_i \leftarrow$ fit hash function to $(\X,\Z_{\cdot i})$ \- \\
          \underline{\textbf{repeat}} \` {\small\textsf{\Z\ step}} \+ \\
          \underline{\textbf{for}} $i = 1,\dots,b$ \+ \\
          $\Z_{\cdot i} \leftarrow$ approximate minimizer of $\calL_P(\h,\Z;\mu)$ over $\Z_{\cdot i}$ \- \- \\
          \underline{\textbf{until}} no change in \Z\ or \texttt{maxit} cycles ran \\
          \underline{\textbf{if}} $\Z = \h(\X)$ \underline{\textbf{then}} stop \- \\
          \underline{\textbf{return}} \h, $\Z = \h(\X)$
        \end{tabbing}
      \end{minipage}
    }
    \caption{MAC algorithm to optimize an affinity-based loss function for binary hashing.}
    \label{f:MAC-alg}
  \end{center}
\end{figure}

\subsection{Stopping criterion, schedule over $\mu$ and path of optimal values}

It is possible to prove that once $\Z = \h(\X)$ after a \Z\ step (regardless of the value of $\mu$), the MAC algorithm will make no further changes to \Z\ or \h, since then the constraints are satisfied. This gives us a reliable stopping criterion that is easy to check, and the MAC algorithm will stop after a finite number of iterations (see below).

It is also possible to prove that the path of minimizers of $\calL_P$ over the continuous penalty parameter $\mu \in [0,\infty)$ is in fact discrete, with changes to $(\Z,\h)$ happening only at a finite number of values $0 < \mu_1 < \dots < \mu_{\infty} < \infty$. Based on this and on our practical experience, we have found that the following approach leads to good schedules for $\mu$ with little effort. We use exponential schedules, of the form $\mu_i = \mu_1 \alpha^{i-1}$ for $i = 1,2,\dots$, so the user has to set only two parameters: the initial $\mu_1$ and the multiplier $\alpha > 1$. We choose exponential schedules because typically the algorithm makes most progress at the beginning, and it is important to track a good minimum there. The upper value $\mu_{\infty}$ past which no changes occur will be reached by our exponential schedule in a finite number of iterations, and our stopping criterion will detect that. We set the multiplier to a value $1< \alpha < 2$ that is as small as computationally convenient. If $\alpha$ is too small, the algorithm will take many iterations, some of which may not even change \Z\ or \h\ (because the path of minima is discrete). If $\alpha$ is too big, the algorithm will reach too quickly a stopping point, without having had time to find a better minimum. As for the initial $\mu_1$, we estimate it by trying values (exponentially spaced) until we find a $\mu$ for which changes to \Z\ from its initial value (for $\mu = 0$) start to occur. (It is also possible to find lower and upper bounds for $\mu_1$ and $\mu_{\infty}$, respectively, for a particular loss function, such as KSH, eSPH or EE.) Overall, the computational time required to estimate $\mu_1$ and $\alpha$ is comparable to running a few extra iterations of the MAC algorithm.

Finally, in practice we use a form of early stopping in order to improve generalization. We use a small validation set to evaluate the precision achieved by the hash function \h\ along the MAC optimization. If the precision decreases over that of the previous step, we ignore the step and skip to the next value of $\mu$. Besides helping to avoid overfitting, this saves computation, by avoid such extra optimization steps. Since the validation set is small, it provides a noisy estimate of the generalization ability at the current iterate, and this occasionally leads to skipping a valid $\mu$ value. This is not a problem because the next $\mu$ value, which is close to the one we skipped, will likely work. At some point during the MAC optimization, we do reach an overfitting region and the precision stops increasing, so the algorithm will skip all remaining $\mu$ values until it stops. In summary, using this validation procedure guarantees that the precision (in the validation set) is greater or equal than that of the initial \Z, thus resulting in a better hash function.

\subsection{$\h$ step}

Given the binary codes $\z_1,\dots,\z_N$, since \h\ does not appear in the first term of $\calL_P$, this simply involves finding a hash function \h\ that minimizes
\begin{equation*}
  \min_{\h}{ \sum^N_{n=1}{ \norm{\z_n - \h(\x_n)}^2 } } = \sum^b_{i=1}{ \min_{h_i}{ \sum^N_{n=1}{ (z_{ni} - h_{i}(\x_n))^2 } } }
\end{equation*}
where $z_{ni} \in \{-1,+1\}$ is the $i$th bit of the binary vector $\z_n$. Hence, we can find $b$ one-bit hash functions in parallel and concatenate them into the $b$-bit hash function. Each of these is a binary classification problem using the number of misclassified patterns as loss. This allows us to use a regular classifier for \h, and even to use a simpler surrogate loss (such as the hinge loss), since this will also enforce the constraints eventually (as $\mu$ increases). For example, we can fit an SVM by optimizing the margin plus the slack and using a high penalty for misclassified patterns. We discuss other classifiers in the experiments. 

\subsection{$\Z$ step}

Although the MAC technique has significantly simplified the original problem, the step over \Z\ is still complex. This involves finding the binary codes given the hash function \h, and it is an NP-complete problem in $Nb$ binary variables. Fortunately, some recent works have proposed practical approaches for this problem based on alternating optimization: a quadratic surrogate method \citep{Lin_13a}, and a GraphCut method \citep{Lin_14b}. In both cases, this would correspond to the first step in the two-step hashing of \citet{Lin_13a}. 

In both the quadratic surrogate and the GraphCut method, the starting point is to apply alternating optimization over the $i$th bit of all points given the remaining bits are fixed for all points (for $i = 1,\dots,b$), and to solve the optimization over the $i$th bit approximately. We describe this next for each method. We start by describing each method in their original form (which applies to the loss function over binary codes, i.e., the first term in $\calL_P$), and then we give our modification to make it work with our \Z\ step objective (the regularized loss function over binary codes, i.e., the complete $\calL_P$).

\paragraph{Solution using a quadratic surrogate method \citep{Lin_13a}}

This is based on the fact that any loss function that depends on the Hamming distance of two binary variables can be equivalently written as a quadratic function of those two binary variables \citep{Lin_13a}. Since this is the case for every term $L(\z_n,\z_m\mathpunct{;}\ y_{nm})$ (because only the $i$th bit in each of $\z_n$ and $\z_m$ is free), we can write the first term in $\calL_P$ as a binary quadratic problem. We now consider the second term (on $\mu$) as well. (We use a similar notation as that of \citealp{Lin_13a}.) The optimization for the $i$th bit can be written as:
\begin{equation}
  \label{e:MAC-alt}
  \min_{\z_{(i)}}{ \sum^N_{n,m=1}{ l_i(z_{ni},z_{mi}) } + \mu \sum_{n=1}^N{(z_{ni} - h_i(\x_n))^2} }
\end{equation}
where $l_i = L(z_{ni},z_{mi},\bar{\z}_n,\bar{\z}_m;y_{nm})$ is the loss function defined on the $i$th bit, $z_{ni}$ is the $i$th bit of the $n$th point, $\bar{\z}_n$ is a vector containing the binary codes of the $n$th point except the $i$th bit, and $h_i(\x_n)$ is the $i$th bit of the binary code of the $n$th point generated by the hash function $\h$. \citet{Lin_13a} show that $l(z_1,z_2)$ can be replaced by a binary quadratic function
\begin{equation}
  \label{e:TSH}
  l(z_1,z_2) = \textstyle\frac{1}{2} z_1 z_2 \big(l^{(11)}-l^{(-11)}\big) + \text{constant}
\end{equation}
as long as $l(1,1) = l(-1,-1) = l^{(11)}$ and $l(1,-1) = l(-1,1) = l^{(-11)}$, where $z_1, z_2 \in \{-1,1\}$. Equation ~\eqref{e:TSH} helps us to rewrite the optimization ~\eqref{e:MAC-alt} as the following:
\begin{equation}
  \label{e:MAC-alt2}
  \min_{\z_{(i)}} \sum^N_{n,m=1}{ \frac{1}{2}z_{ni} z_{mi} \big(l^{(11)}-l^{(-11)}\big) } + \mu \sum^N_{n=1}{(z_{ni} - h_i(\x_n))^2}.
\end{equation}
By defining $a_{nm} = \smash{\big(l_{(inm)}^{(11)}-l_{(inm)}^{(-11)}\big)}$ as the $(n,m)$ element of a matrix $\A \in \bbR^{N \times N}$ and ignoring the coefficients, we have the following optimization problem:
\begin{equation*}
  \min_{\z_{(i)}}{ \z_{(i)}^T \A \z_{(i)} + \mu \norm{\z_{(i)} - \h_i(\X)}^2 } \qquad \text{s.t.} \qquad \z_{(i)} \in \{-1,+1\}^N
\end{equation*}
where $\h_i(\X) = (h_i(\x_1),\dots,h_i(\x_N))^T$ is a vector of length $N$ (one bit per data point). Both terms in the above minimization are quadratic on binary variables. This is still an NP-complete problem (except in special cases), and we approximate it by relaxing it to a continuous quadratic program (QP) over $\z_{(i)} \in [-1,1]^N$ and binarizing its solution. In general, the matrix \A\ is not positive definite and the relaxed QP is not convex, so we need an initialization. (However, the term on $\mu$ adds $\mu \I$ to \A, so even if \A\ is not positive definite, $\A + \mu\I$ will be positive definite for large enough $\mu$, and the QP will be convex.) We construct an initialization by converting the binary QP into a binary eigenproblem:
\begin{equation}
  \min_{\balpha}{ \balpha^T \B \balpha } \qquad \text{s.t.} \qquad \alpha_0=1, \quad \z_{(i)} \in \{-1,1\}^N, \quad \balpha = \bigl( \begin{smallmatrix} \z_{(i)} \\ \alpha_0 \end{smallmatrix}\bigr), \quad \B = \left( \begin{smallmatrix} \A & -\frac{\mu}{2} \h_i(\X) \\ -\frac{\mu}{2} \h_i(\X)^T &  0  \end{smallmatrix} \right).
\end{equation}
To solve this problem we use spectral relaxation, where the constraints $\z_{(i)} \in \{-1,+1\}^N$ and $z_{i+1}=1$ are relaxed to $\norm{\balpha} = N + 1$. The solution to this problem is the eigenvector corresponding to the smallest eigenvalue of $\B$. We use the truncated eigenvector as the initialization for minimizing the relaxed, bound-constrained QP:
\begin{equation*}
  \min_{\z_{(i)}}{ \z_{(i)}^T \A \z_{(i)} + \mu \norm{\z_{(i)} - \h_i(\X)}^2 } \qquad \text{s.t.} \qquad \z_{(i)} \in [-1,1]^N.
\end{equation*}
which we solve using L-BFGS-B \citep{Zhu_97b}.

As noted above, the \Z\ step is an NP-complete problem in general, so we cannot expect to find the global optimum. It is even possible that the approximate solution could increase the objective over the previous iteration's \Z\ (this is likely to happen as the overall MAC algorithm converges). If that occurs, we simply skip the update, in order to guarantee that we decrease monotonically on $\calL_P$, and avoid oscillating around a minimum.

\paragraph{Solution using a GraphCut algorithm \citep{Lin_14b}}

To optimize over the $i$th bit (given all the other bits are fixed), we have to minimize eq.~\eqref{e:MAC-alt2}. In general, this is an NP-complete problem over $N$ bits (the $i$th bit for each image), with the form of a quadratic function on binary variables. We can apply the GraphCut algorithm \citep{BoykovKolmog03a,BoykovKolmog04a,KolmogZabih04a}, as proposed by the FastHash algorithm of \citet{Lin_14b}. This proceeds as follows. First, we assign all the data points to different, possibly overlapping groups (blocks). Then, we minimize the objective function over the binary codes of the same block, while all the other binary codes are fixed, then proceed with the next block, etc.\ (that is, we do alternating optimization of the bits over the blocks). Specifically, to optimize over the bits in block $\calB$, we define $a_{nm} = \smash{\big(l_{(inm)}^{(11)}-l_{(inm)}^{(-11)}\big)} $ and, ignoring the constants, we can rewrite equation~\eqref{e:MAC-alt2} as:
\begin{equation*}
  \min_{\z_{(i,\calB)}} \sum_{n \in \calB} { \sum_{m \in \calB}{ a_{nm}z_{ni} z_{mi} } } + 2\sum_{n \in \calB} { \sum_{m \not\in \calB}{ a_{nm}z_{ni} z_{mi} } } - \mu \sum_{n\in \calB}{z_{ni}h_i(\x_n)}.
\end{equation*}
We then rewrite this equation in the standard form for the GraphCut algorithm:
\begin{equation*}
  \min_{\z_{(i,\calB)}} \sum_{n \in \calB} { \sum_{m \in \calB}{ v_{nm}z_{ni} z_{mi} } } + \sum_{n \in \calB} { u_{nm}z_{ni}}
\end{equation*}
where $v_{nm}=a_{nm}$, $u_{nm}=2 \sum_{m \not\in \calB} { a_{nm} z_{mi} } - \mu h_i(\x_n)$. To minimize the objective function using the GraphCut algorithm, the blocks have to define a submodular function. For the objective functions that we explained in the paper, this can be easily achieved by putting points with the same label in one block (\citealp{Lin_14b} give a simple proof of this).

Unlike in the quadratic surrogate method, using the GraphCut algorithm with alternating optimization on blocks defining submodular functions is guaranteed to find a \Z\ that has a lower or equal objective value that the initial one, and therefore to decrease monotonically $\calL_P$.

\section{Experiments}
\label{s:expts}

We have tested our framework with several combinations of loss function, hash function, number of bits, datasets, and comparing with several state-of-the-art hashing methods (appendix~\ref{s:expts-additional} contains additional experiments). We report a representative subset to show the flexibility of the approach. We use the KSH~\eqref{e:KSH} \citep{Liu_12c} and eSPLH \citep{Wang_12a} loss functions. We test quadratic surrogate and GraphCut methods for the \Z\ step in MAC. As hash functions (for each bit), we use linear SVMs (trained with LIBLINEAR; \citealp{Fan_08a}) and kernel SVMs%
\footnote{To train a kernel SVM, we use $500$ radial basis functions with centers given by a random subset of the training points, and apply a linear SVM to their output. Computationally, this is fast because we can use a constant Gram matrix. Using as hash function a kernel SVM trained with LIBSVM gave similar results, but is much slower because the support vectors change when the labels change. We set the RBF bandwidth to the average Euclidean distance of the first $300$ points.}

We use the following labeled datasets (all using the Euclidean distance in feature space): (1) CIFAR \citep{Krizhev09a} contains $60\,000$ images in $10$ classes. We use $D=320$ GIST features \citep{OlivaTorral01a} from each image. We use $58\,000$ images for training and $2\,000$ for test. (2) Infinite MNIST \citep{Loosli_07a}. We generated, using elastic deformations of the original MNIST handwritten digit dataset, $1\,000\,000$ images for training and $2\,000$ for test, in 10 classes. We represent each image by a $D=784$ vector of raw pixels. Because of the computational cost of affinity-based methods, previous work has used training sets limited to a few thousand points \citep{KulisDarrel09a,NorouzFleet11a,Liu_12c,Lin_13a}. We train the hash functions in a subset of $10\,000$ points of the training set, and report precision and recall by searching for a test query on the entire dataset (the base set).

We report precision and precision/recall for the test set queries using as ground truth (set of true neighbors in original space) all the training points with the same label. In precision curves, the retrieved set contains the $k$ nearest neighbors of the query point in the Hamming space. We report precision for different values of $k$ to test the robustness of different algorithms. In precision/recall curves, the retrieved set contains the points inside Hamming distance $r$ of the query point. These curves show the precision and recall at different Hamming distances $r=0$ to $r=L$. We report zero precision when there is no neighbor inside Hamming distance $r$ of a query. This happens most of the time when $L$ is large and $r$ is small. In most of our precision/recall curves, the precision drops significantly for very small and very large values of $r$. For small values of $r$, this happens because most of the query points do not retrieve any neighbor. For large values of $r$, this happens because the number of retrieved points becomes very large.

The main comparison point are the quadratic surrogate and GraphCut methods of \citet{Lin_13a,Lin_14b}, which we denote in this section as \emph{quad} and \emph{cut}, respectively, regardless of the hash function that fits the resulting codes. Correspondingly, we denote the MAC version of these as \emph{MACquad} and \emph{MACcut}, respectively. We use the following schedule for the penalty parameter $\mu$ in the MAC algorithm (regardless of the hash function type or dataset). We initialize \Z\ with $\mu=0$, i.e., the result of \emph{quad} or \emph{cut}. Starting from $\mu_1 = 0.3$ (\emph{MACcut}) or $0.01$ (\emph{MACquad}), we multiply $\mu$ by $1.4$ after each iteration (\Z\ and \h\ step).

Our experiments show that the MAC algorithm indeed finds hash functions with a significantly and consistently lower objective function value than rounding or two-step approaches (in particular, \emph{cut} and \emph{quad}); and that it outperforms other state-of-the-art algorithms on different datasets, with \emph{MACcut} beating \emph{MACquad} most of the time. The improvement in precision makes using MAC well worth the relatively small extra runtime and minimal additional implementation effort it requires. In all our plots, the vertical arrows indicate the improvement of \emph{MACcut} over \emph{cut} and of \emph{MACquad} over \emph{quad}.

\subsection{The MAC algorithm finds better optima}
\label{s:optima}

\begin{figure}[p]
  \centering
  \psfrag{rerror}[][t]{loss function \calL}
  \psfrag{iteration}[t][]{iterations}
  \psfrag{K}[t][]{$k$}
  \psfrag{precision}[][t]{precision}
  \psfrag{recall}[][b]{recall}
  \begin{tabular}{@{}l@{}c@{}c@{}c@{}}
    & $b=16$ & $b=32$ & $b=48$\\
    \hspace{2ex}\rotatebox{90}{\hspace{7ex}loss function \calL} &
    \includegraphics[width=0.24\linewidth]{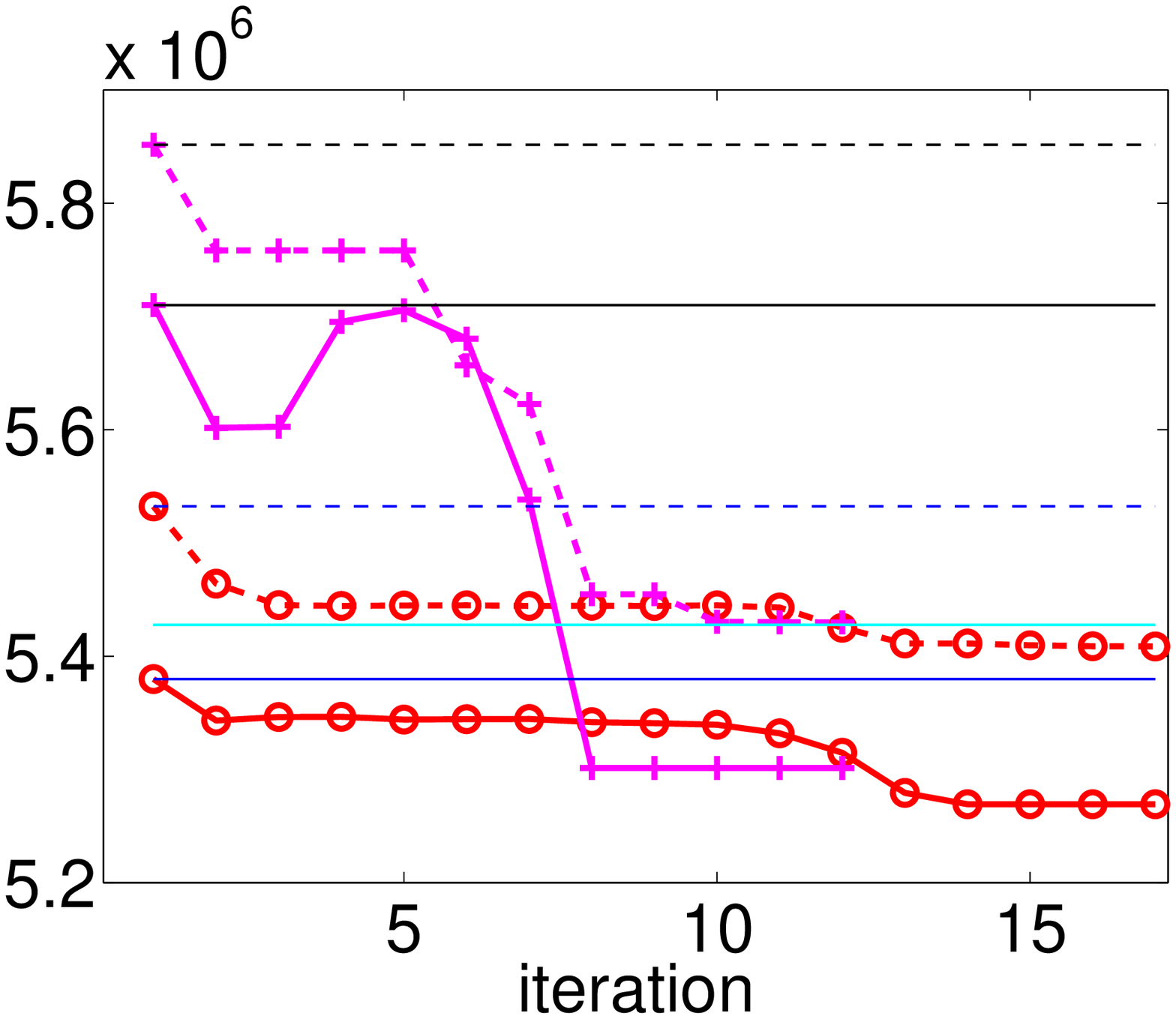} &
    \includegraphics[width=0.24\linewidth]{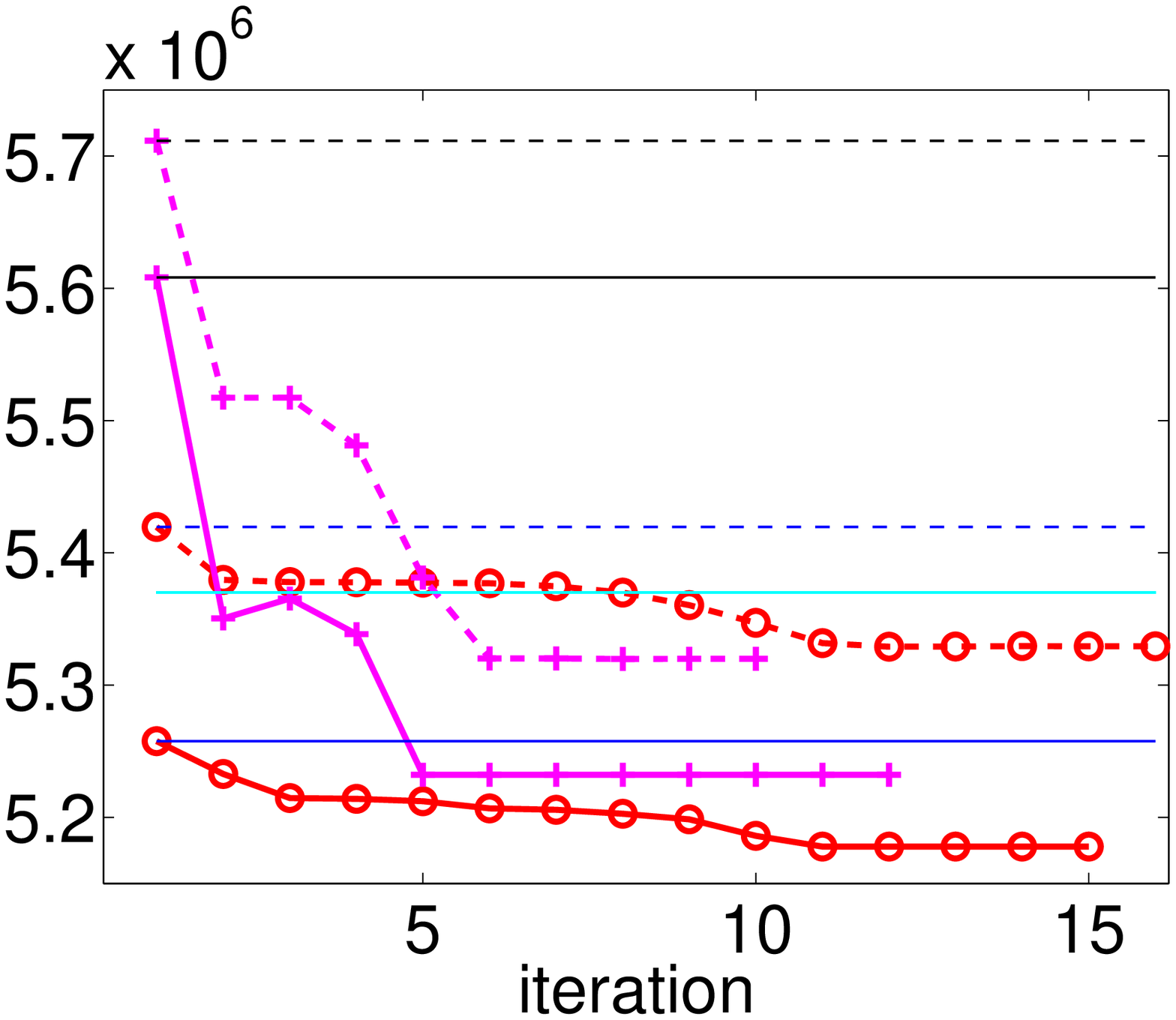} &
    \includegraphics[width=0.24\linewidth]{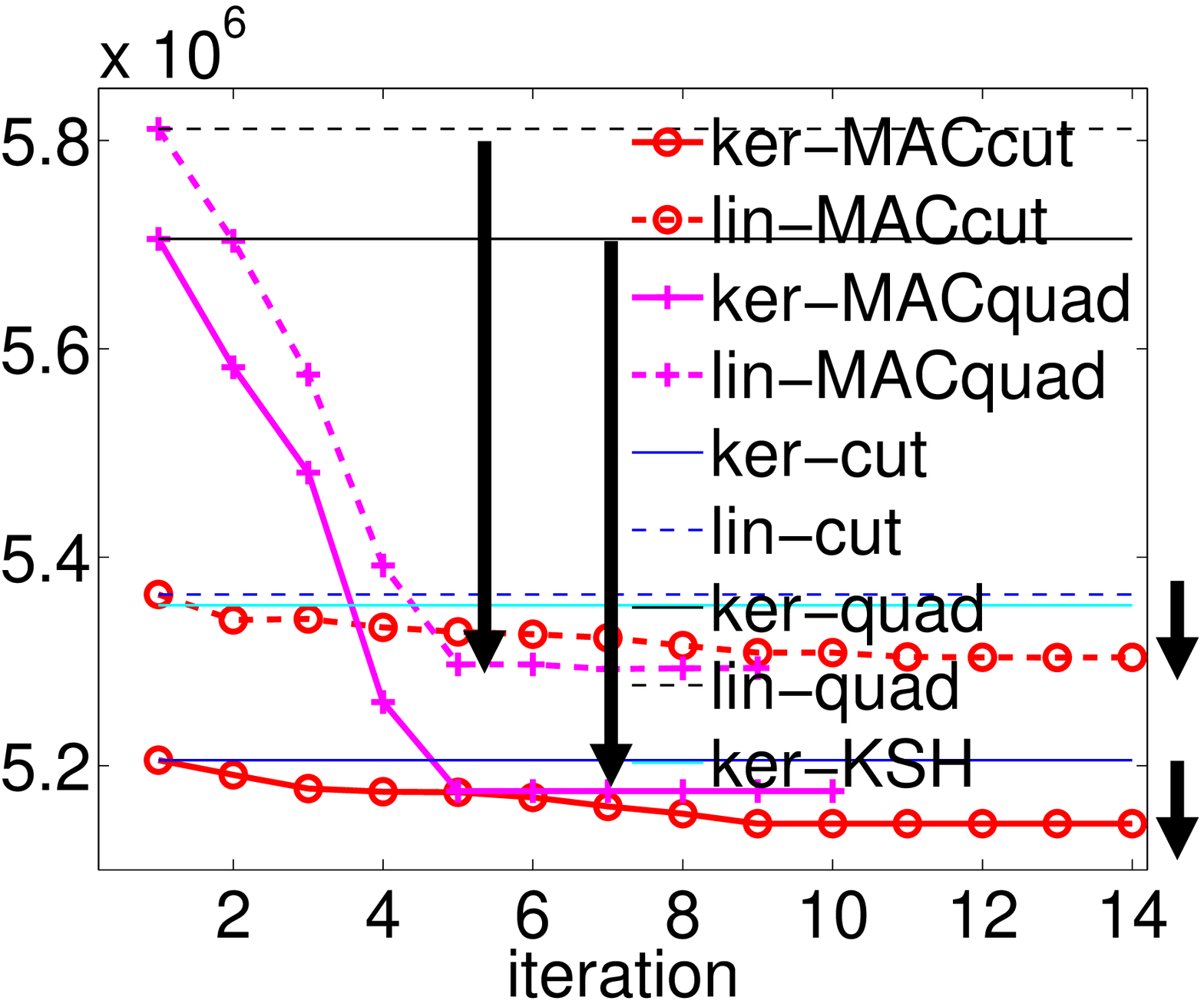} \\
    \hspace{2ex}\rotatebox{90}{\hspace{10ex}precision} &
    \includegraphics[width=0.24\linewidth]{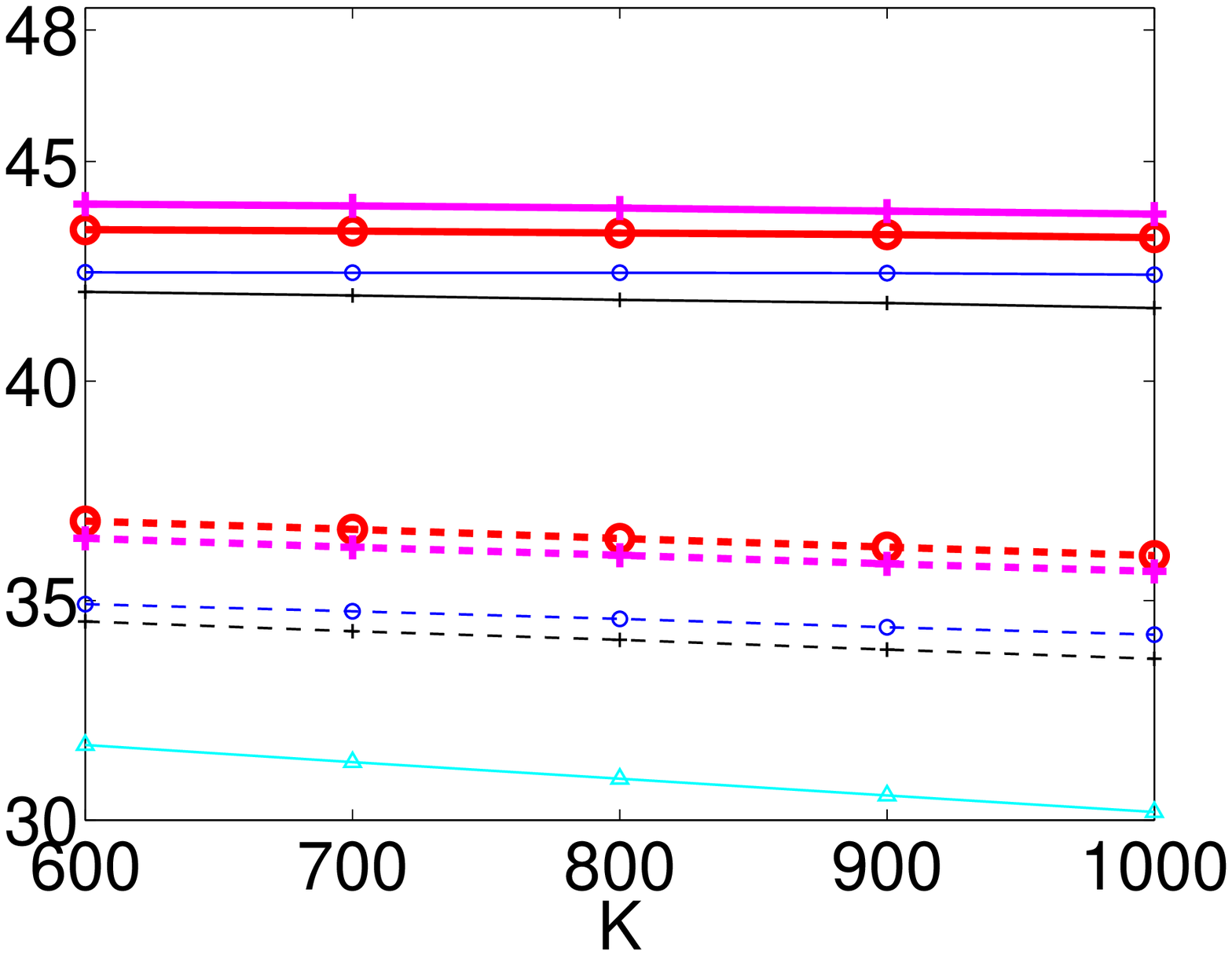} &
    \includegraphics[width=0.240\linewidth]{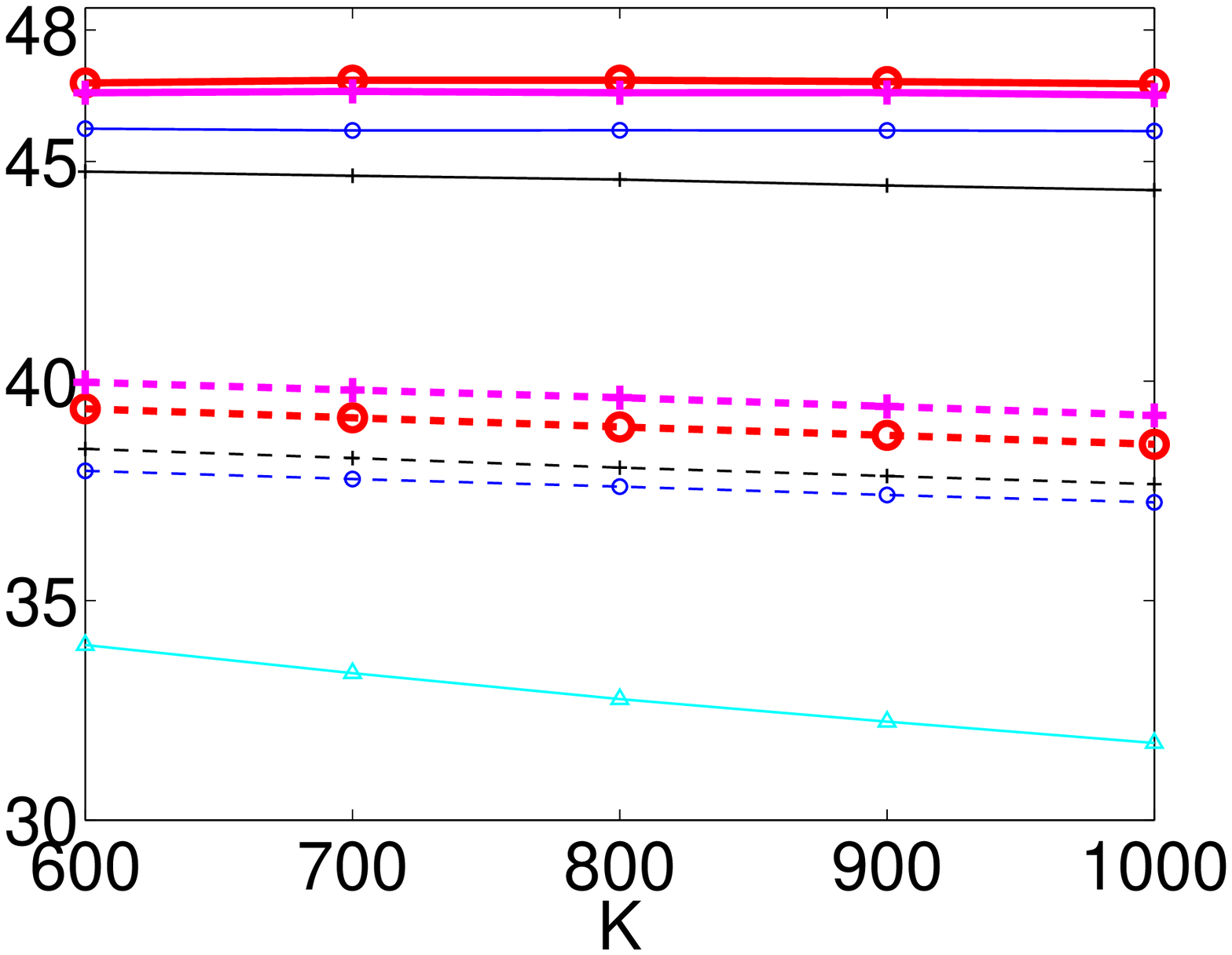} &
    \includegraphics[width=0.240\linewidth]{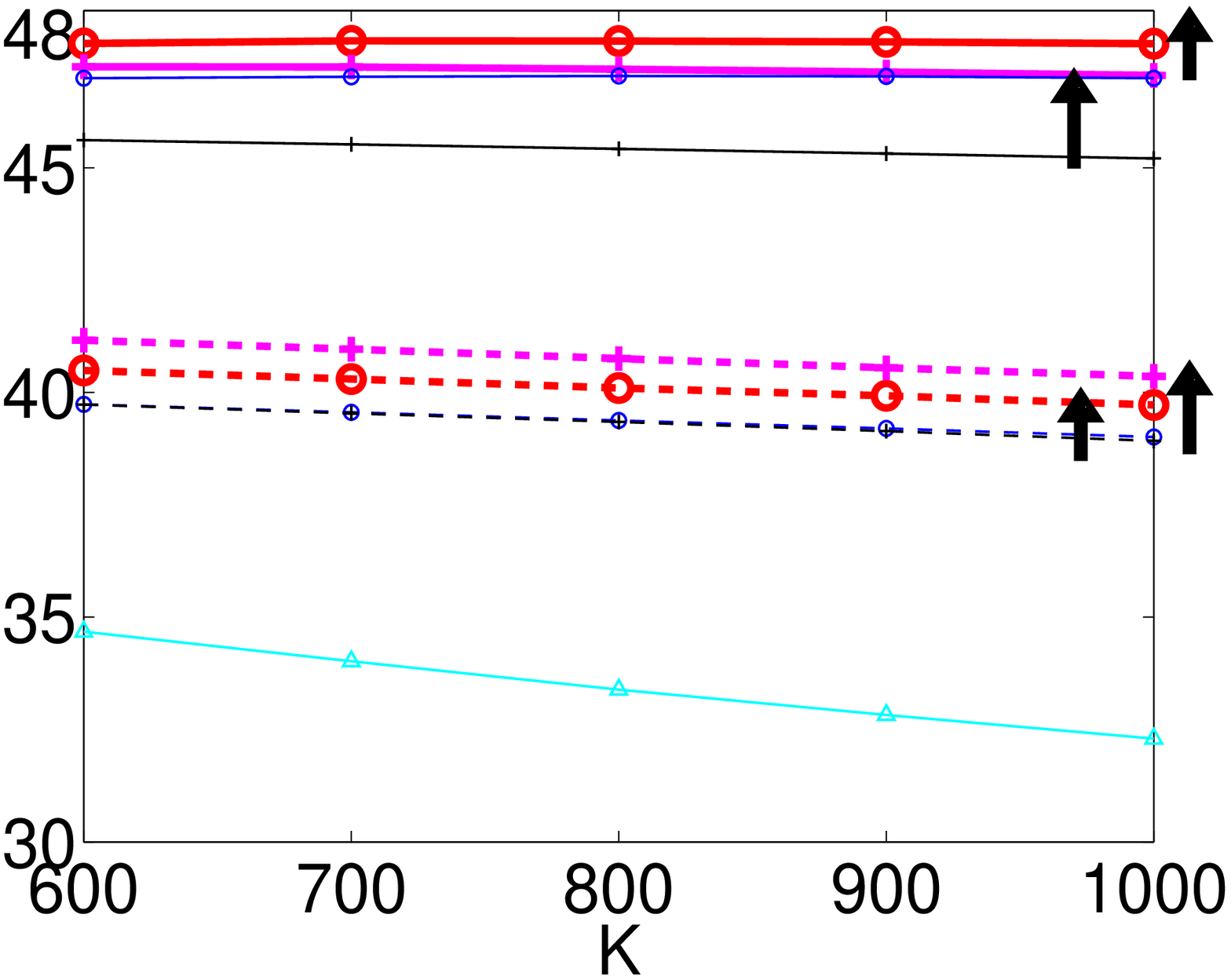} \\
    \hspace{2ex}\rotatebox{90}{\raisebox{3ex}[0pt][0pt]{\makebox[0pt][l]{\hspace{13ex}KSH, $\kappa_+ = 100$, $\kappa_- = 500$, $K = 20\,000$}}\hspace{7ex}precision} &
    \includegraphics[width=0.240\linewidth]{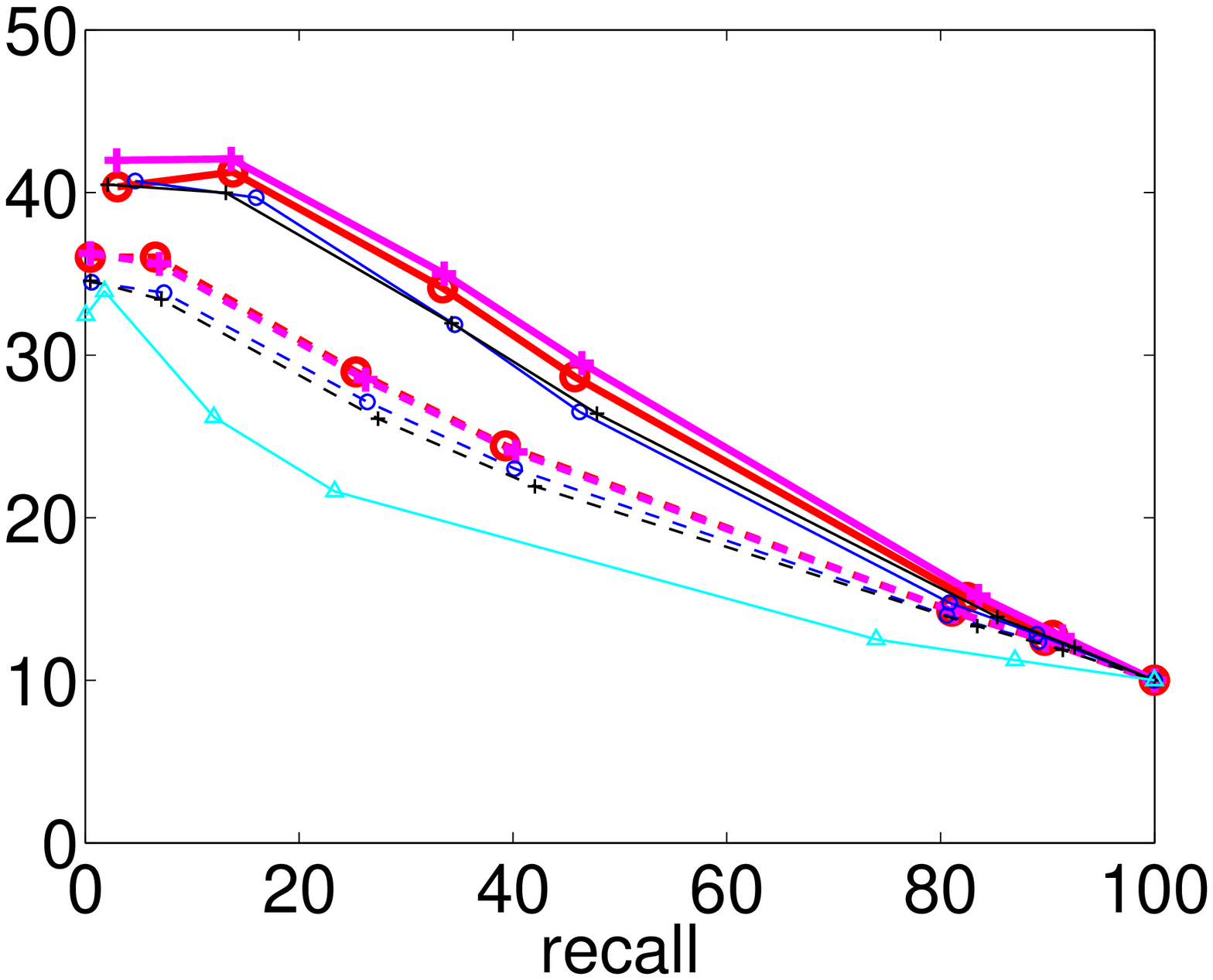} &
    \includegraphics[width=0.240\linewidth]{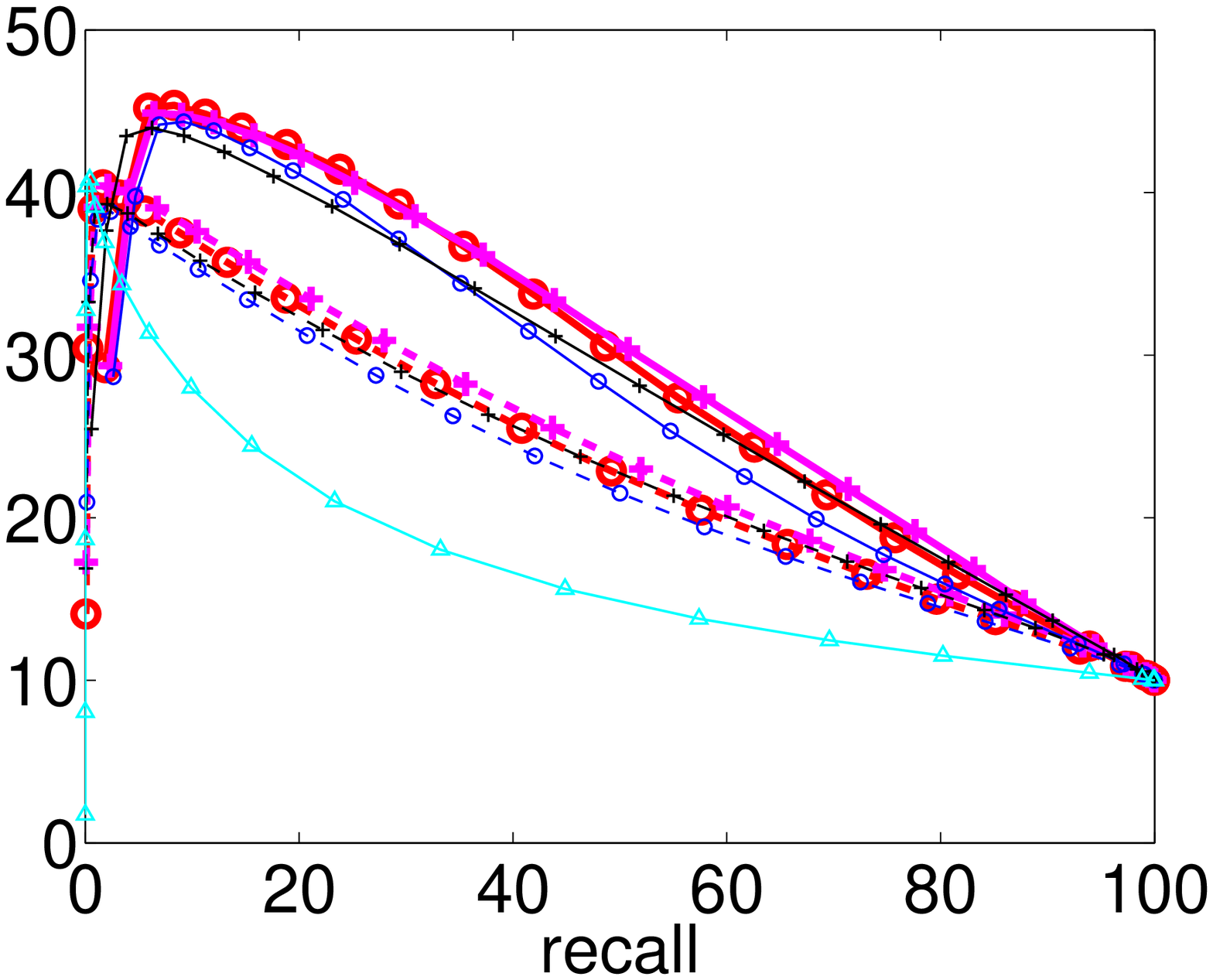} &
    \includegraphics[width=0.240\linewidth]{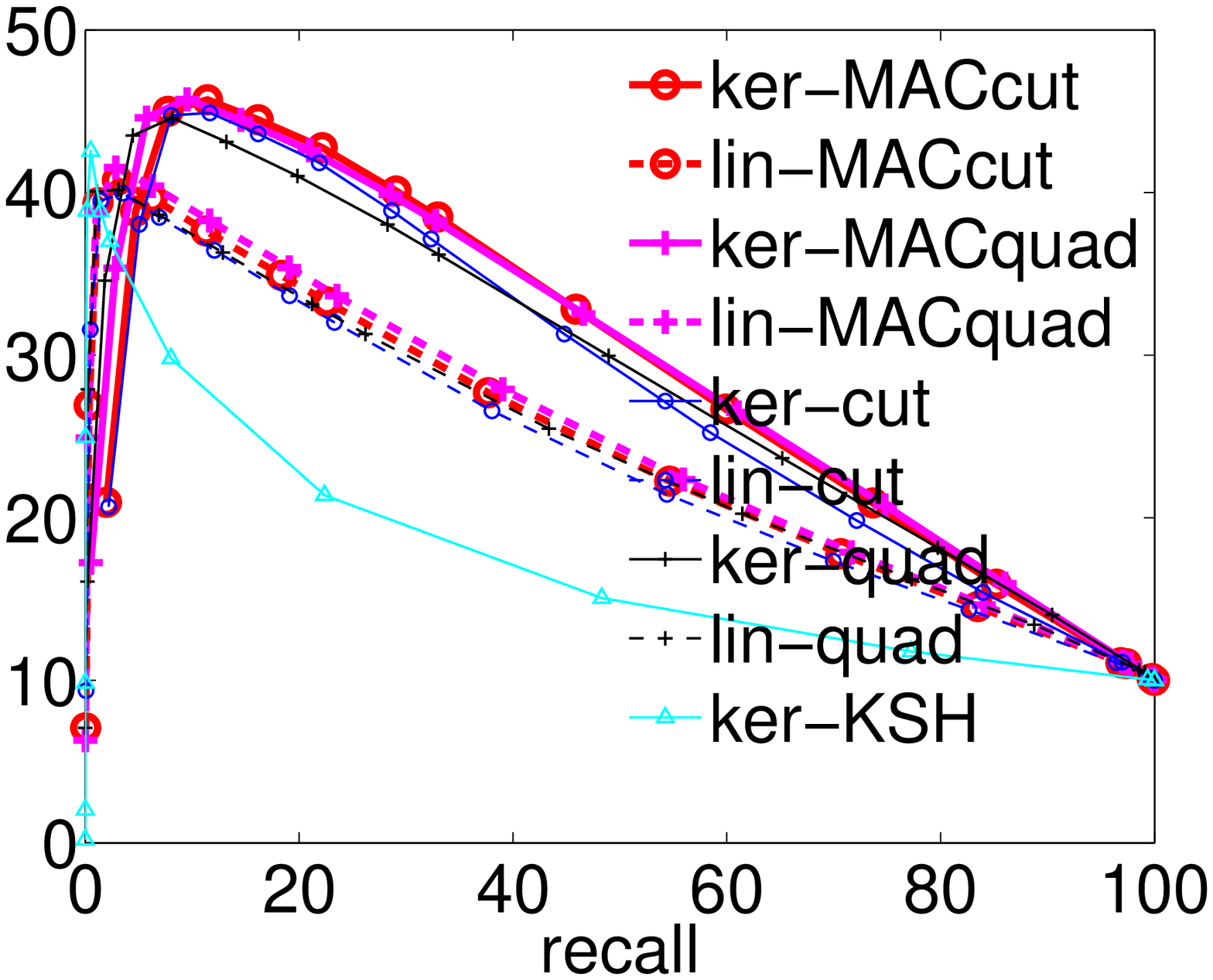}\\
   \hline
    \hspace{2ex}\rotatebox{90}{\hspace{7ex}loss function \calL} &
    \includegraphics[width=0.24\linewidth]{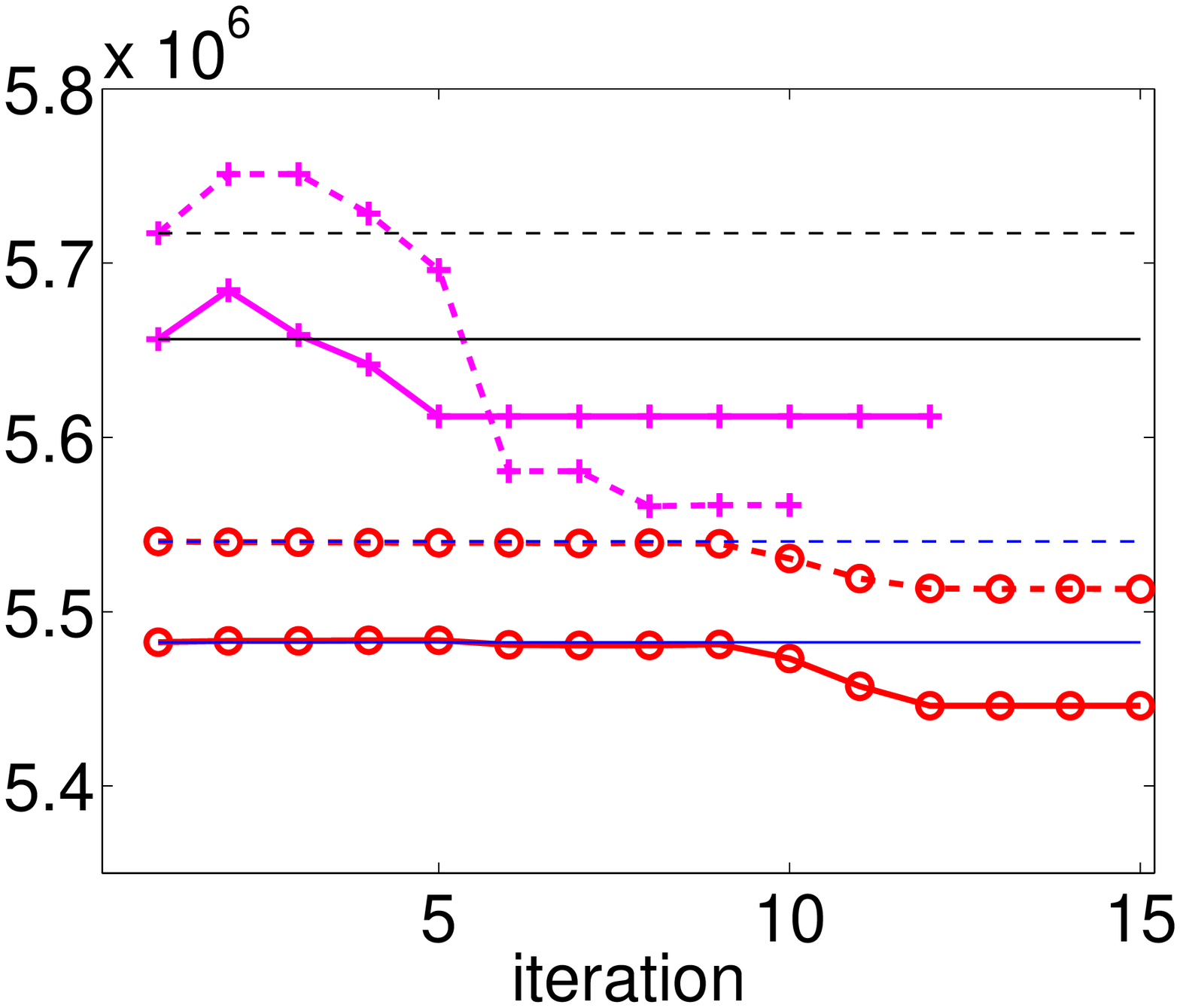} &
    \includegraphics[width=0.24\linewidth]{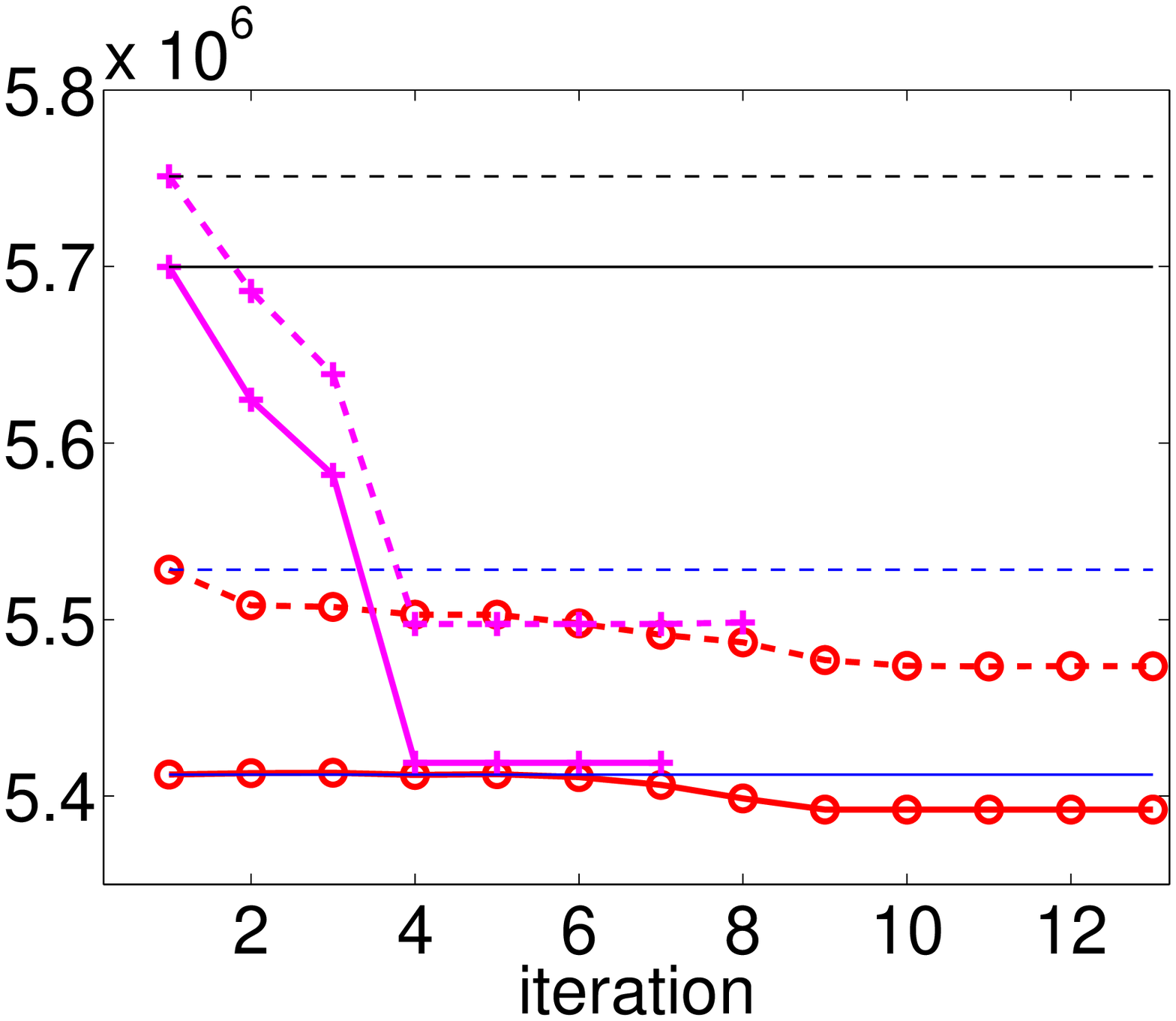} &
    \includegraphics[width=0.24\linewidth]{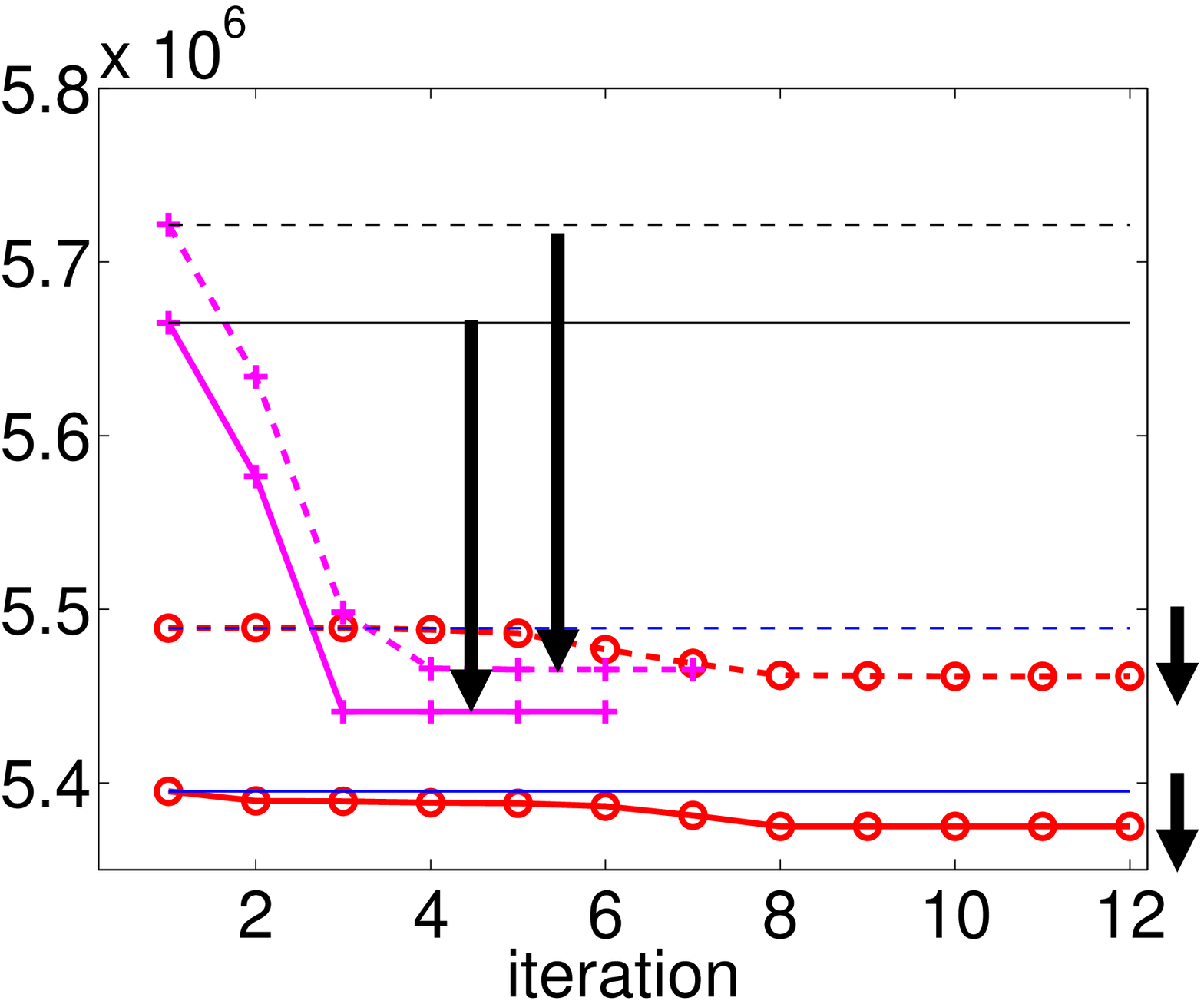} \\
    \hspace{2ex}\rotatebox{90}{\hspace{10ex}precision} &
    \includegraphics[width=0.24\linewidth]{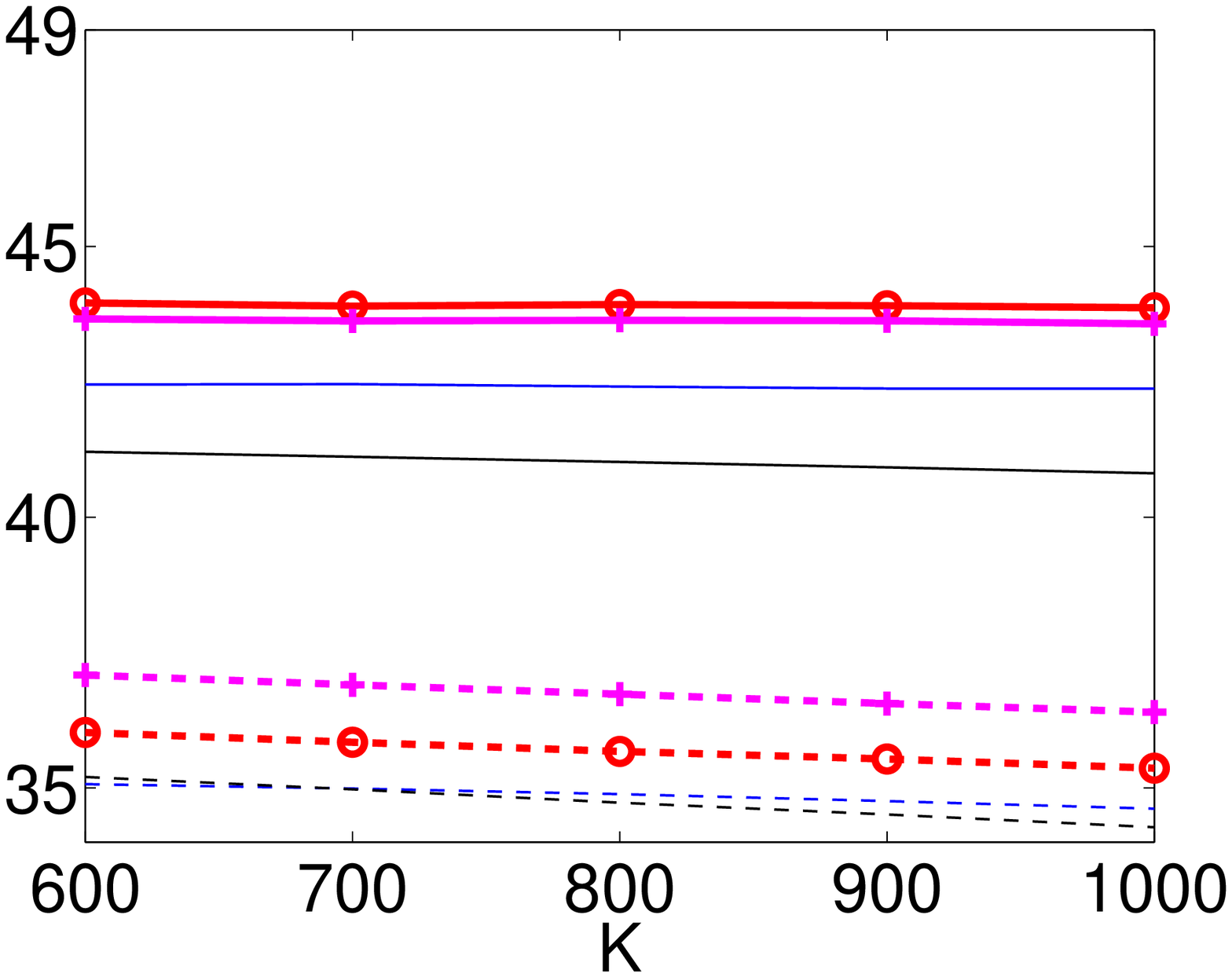} &
    \includegraphics[width=0.24\linewidth]{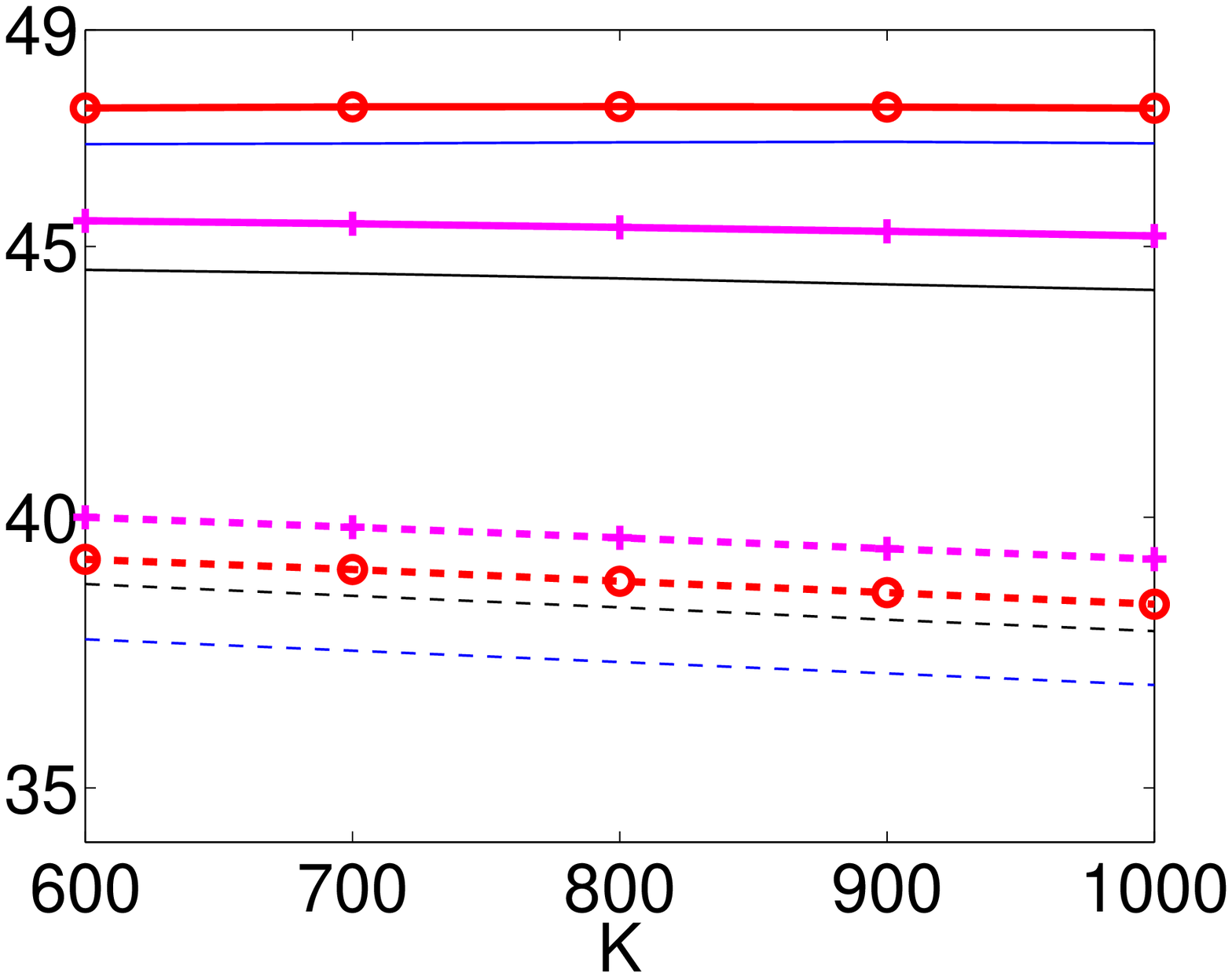} &
    \includegraphics[width=0.24\linewidth]{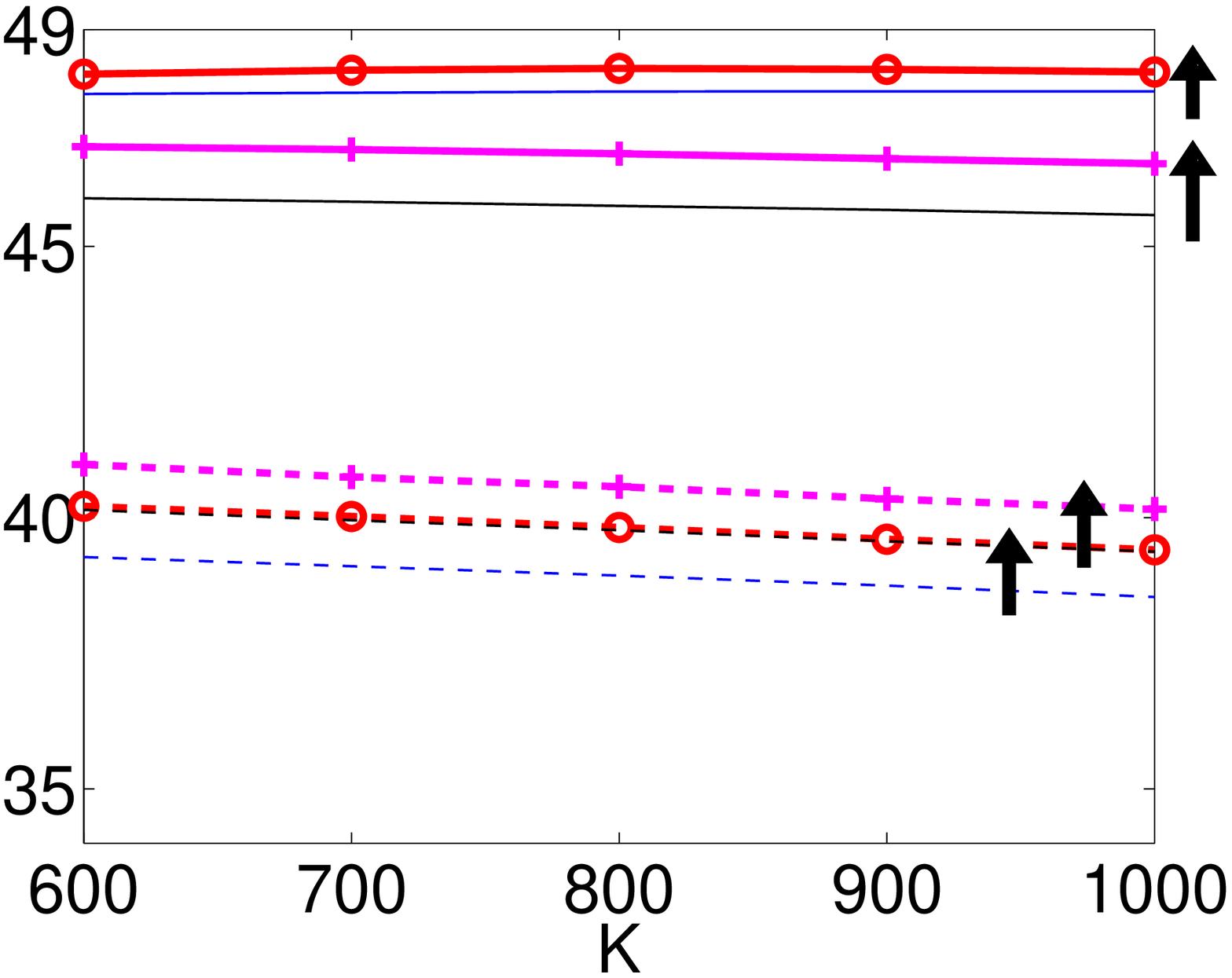} \\
    \hspace{2ex}\rotatebox{90}{\raisebox{3ex}[0pt][0pt]{\makebox[0pt][l]{\hspace{13ex}eSPLH, $\kappa_+ = 100$, $\kappa_- = 500$, $K = 20\,000$}}\hspace{7ex}precision} &
    \includegraphics[width=0.24\linewidth]{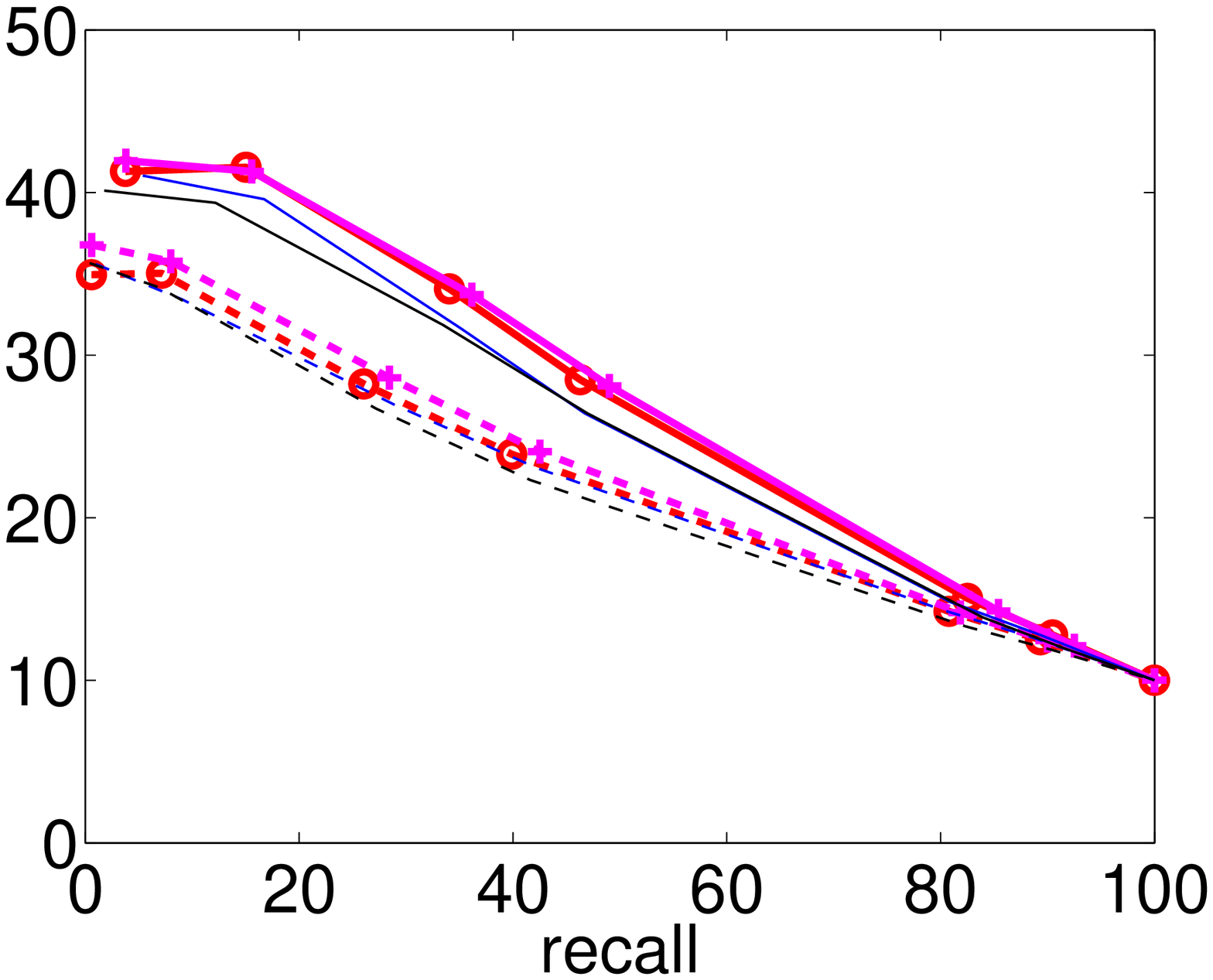} &
    \includegraphics[width=0.24\linewidth]{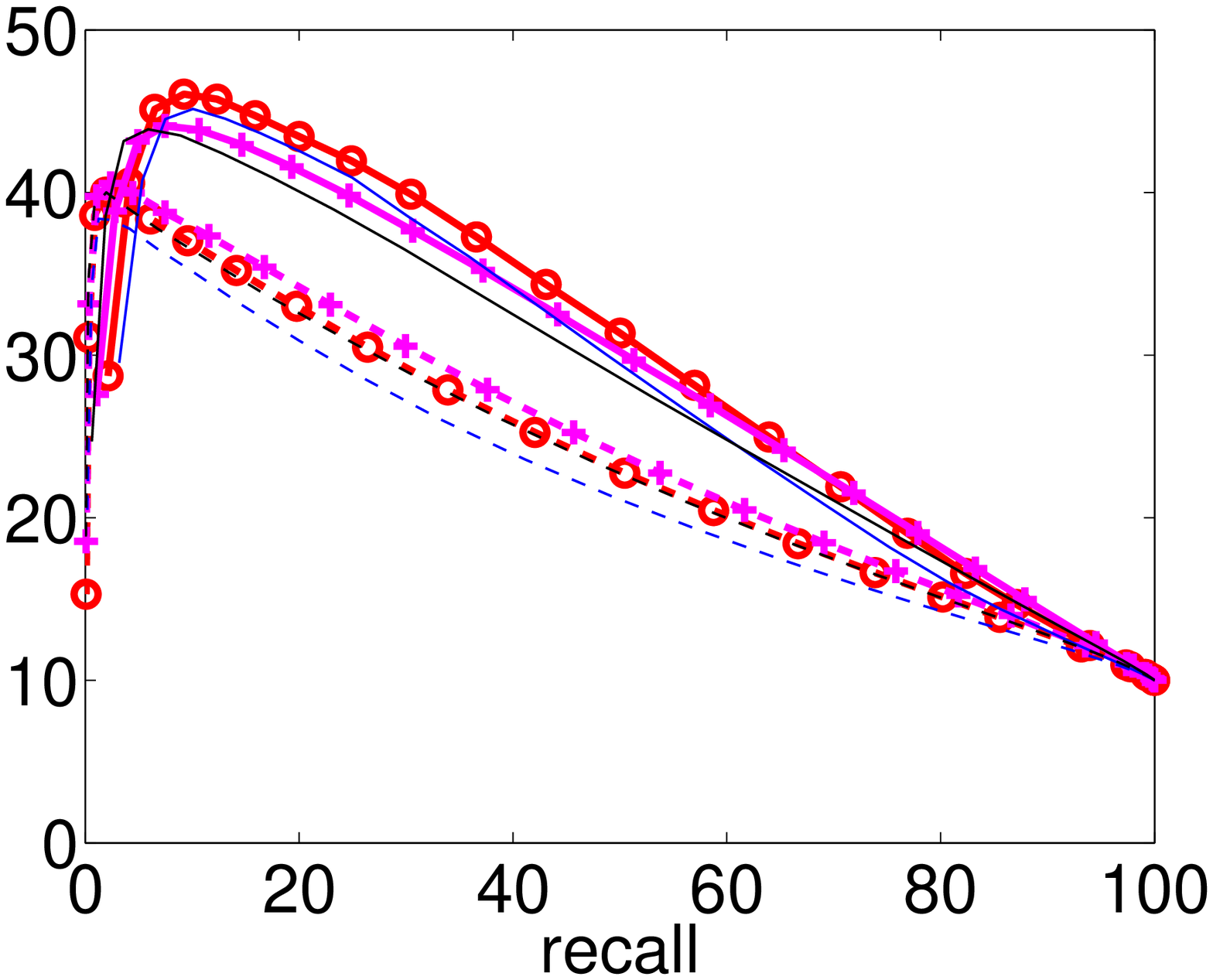} &
    \includegraphics[width=0.24\linewidth]{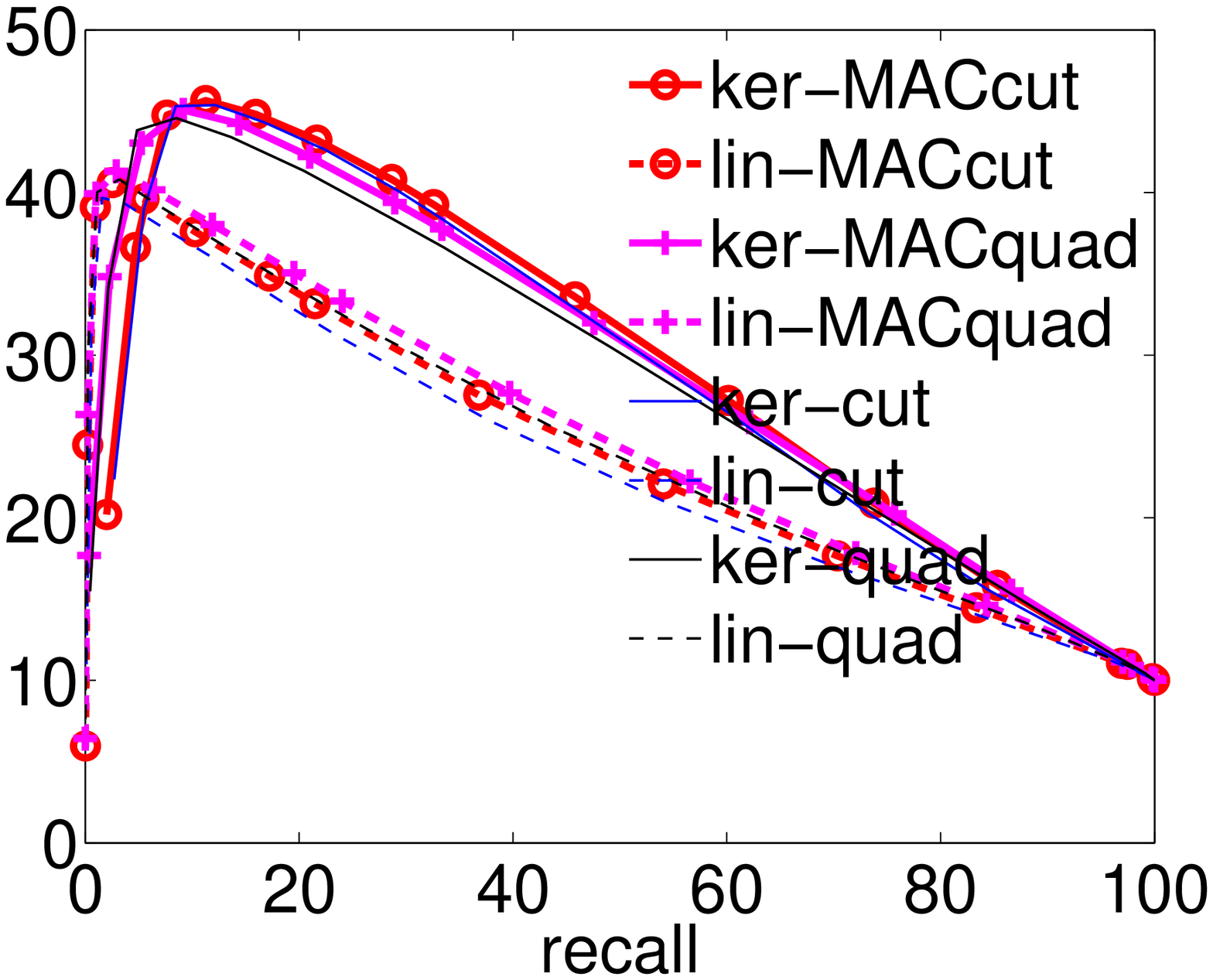}
  \end{tabular}
  \caption{KSH (top panel) and eSPLH (bottom panel) loss functions on CIFAR dataset, using $b=16$ to $48$ bits. The rows in each panel show the value of the loss function \calL, the precision for $k$ retrieved points and the precision/recall (at different Hamming distances).}
  \label{f:sup-nestederr}
\end{figure}

\emph{The goal of this paper is not to introduce a new affinity-based loss or hash function, but to describe a generic framework to construct algorithms that optimize a given combination thereof.} We illustrate its effectiveness here with the CIFAR dataset, with different sizes of retrieved neighbor sets, and using $16$ to $48$ bits. We optimize two affinity-based loss functions (KSH from eq.~\eqref{e:KSH} and eSPLH), and two hash functions (linear and kernel SVM). In all cases, the MAC algorithm achieves a better hash function both in terms of the loss and of the precision/recall. We compare 4 ways of optimizing the loss function: \emph{quad} \citep{Lin_13a}, \emph{cut} \citep{Lin_14b}, \emph{MACquad} and \emph{MACcut}.

For each point $\x_n$ in the training set, we use $\kappa_+ = 100$ positive (similar) and $\kappa_- = 500$ negative (dissimilar) neighbors, chosen at random to have the same or a different label as $\x_n$, respectively. Fig.~\ref{f:sup-nestederr}(top panel) shows the KSH loss function for all the methods (including the original KSH method in \citealp{Liu_12c}) over iterations of the MAC algorithm (KSH, \emph{quad} and \emph{cut} do not iterate), as well as precision and recall. It is clear that \emph{MACcut} (red lines) and \emph{MACquad} (magenta lines) reduce the loss function more than \emph{cut} (blue lines) and \emph{quad} (black lines), respectively, as well as the original KSH algorithm (cyan), in all cases: type of hash function (linear: dashed lines, kernel: solid lines) and number of bits $b = 16$ to $48$. Hence, applying MAC is always beneficial. Reducing the loss nearly always translates into better precision and recall (with a larger gain for linear than for kernel hash functions, usually). The gain of \emph{MACcut}/\emph{MACquad} over \emph{cut}/\emph{quad} is significant, often comparable to the gain obtained by changing from the linear to the kernel hash function within the same algorithm.

We usually find \emph{cut} outperforms \emph{quad} (in agreement with \citealp{Lin_14b}), and correspondingly \emph{MACcut} outperforms \emph{MACquad}. Interestingly, \emph{MACquad} and \emph{MACcut} end up being very similar even though they started very differently. This suggests it is not crucial which of the two methods to use in the MAC \Z\ step, although we still prefer \emph{cut}, because it usually produces somewhat better optima. Finally, fig.~\ref{f:sup-nestederr}(bottom panel) shows the \emph{MACcut} results using the eSPLH loss. All settings are as in the first KSH experiment. As before, \emph{MACcut} outperforms \emph{cut} in both loss function and precision/recall using either a linear or a kernel SVM.

\subsection{Why does MAC learn better hash functions?}

\begin{figure}[t!]
  \centering
  \psfrag{bits}[][]{number of bits $b$}
  \begin{tabular}{@{}l@{\hspace{1em}}c@{\hspace{1em}}c@{}}
    & KSH loss & eSPLH loss \\[-1ex]
    \rotatebox{90}{\hspace{13ex}loss function \calL} &
    \includegraphics[width=0.4\columnwidth]{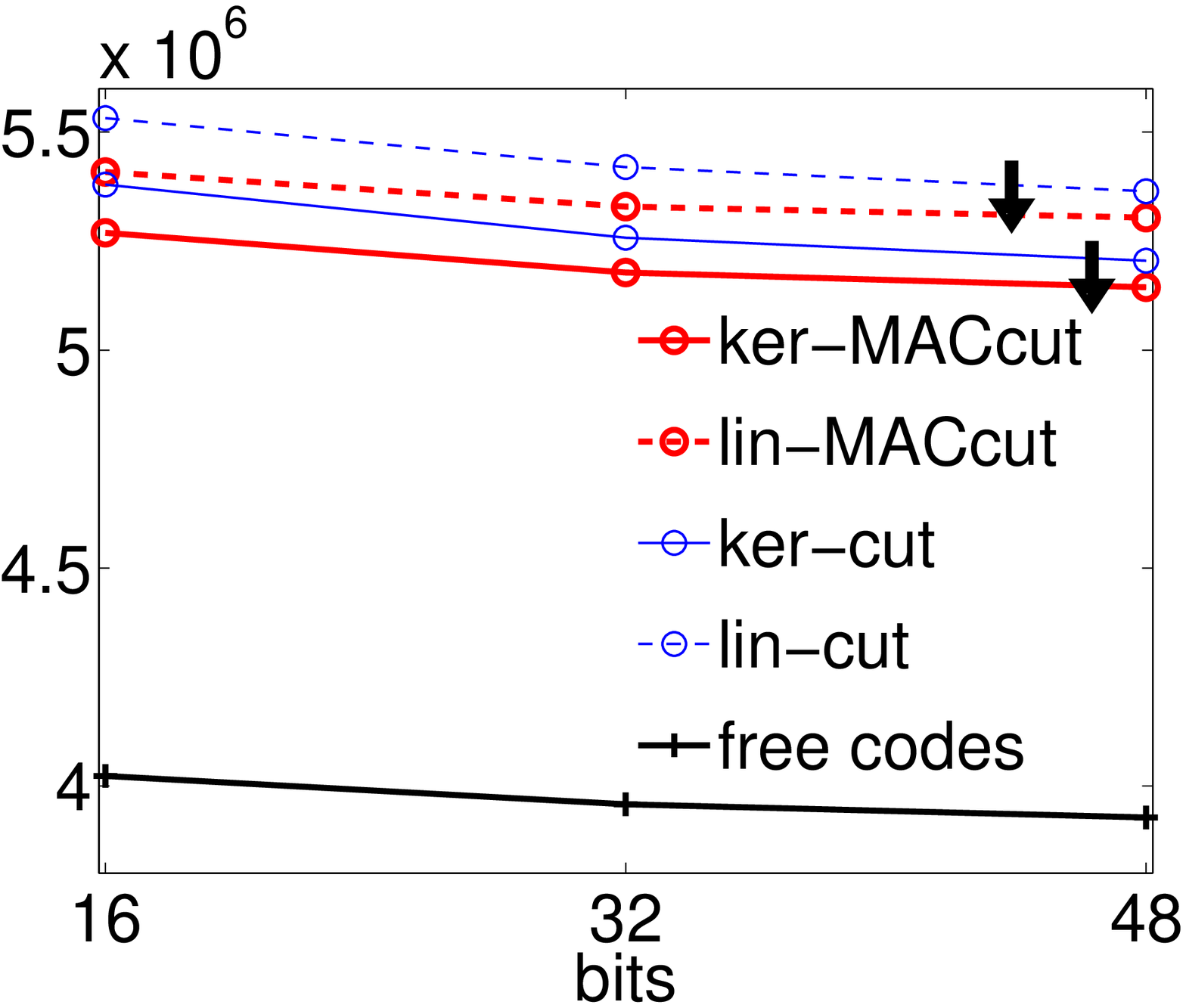} &
    \includegraphics[width=0.4\columnwidth]{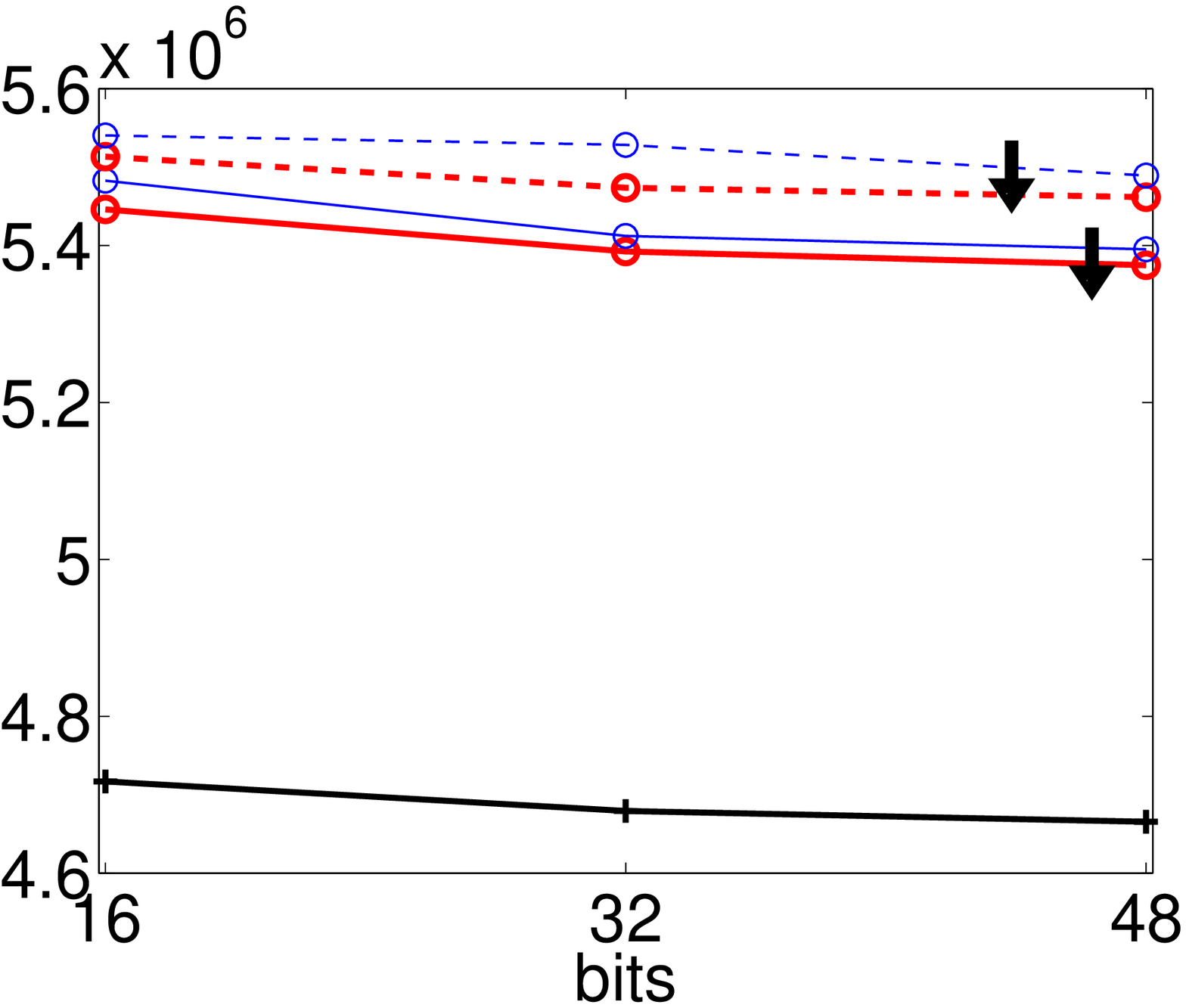}
  \end{tabular}
  \caption{Like fig.~\ref{f:sup-nestederr} but showing the value of the error function $E(\Z)$ of eq.~\eqref{e:objfcnZ} for the ``free'' binary codes, and for the codes produced by the hash functions learned by \emph{cut} (the two-step method) and \emph{MACcut}, with linear and kernel hash functions.}
  \label{f:sup-nestederr-freecodes}
\end{figure}

In both the two-step and MAC approaches, the starting point are the ``free'' binary codes obtained by minimizing the loss over the codes without them being the output of a particular hash function. That is, minimizing~\eqref{e:MAC-constrained} without the ``$\z_n = \h(\x_n)$'' constraints:
\begin{equation}
  \label{e:objfcnZ}
  \min_{\Z}{ E(\Z) = \sum^N_{n=1}{ L(\z_n,\z_m\mathpunct{;}\ y_{nm}) } }, \ \z_1,\dots,\z_N \in \{-1,+1\}^b.
\end{equation}
The resulting free codes try to achieve good precision/recall independently of whether a hash function can actually produce such codes. Constraining the codes to be realizable by a specific family of hash functions (say, linear), means the loss $E(\Z)$ will be larger than for free codes. How difficult is it for a hash function to produce the free codes? Fig.~\ref{f:sup-nestederr-freecodes} plots the loss function for the free codes, the two-step codes from \emph{cut}, and the codes from \emph{MACcut}, for both linear and kernel hash functions in the same experiment as in fig.~\ref{f:sup-nestederr}. It is clear that the free codes have a very low loss $E(\Z)$, which is far from what a kernel function can produce, and even farther from what a linear function can produce. Both of these are relatively smooth functions that cannot represent the presumably complex structure of the free codes. This could be improved by using a very flexible hash function (e.g.\ using a kernel function with many centers), which could better approximate the free codes, but 1) a very flexible function would likely not generalize well, and 2) we require fast hash functions for fast retrieval anyway. Given our linear or kernel hash functions, what the two-step \emph{cut} optimization does is fit the hash function directly to the free codes. This is not guaranteed to find the best hash function in terms of the original problem~\eqref{e:objfcn}, and indeed it produces a pretty suboptimal function. In contrast, MAC gradually optimizes both the codes and the hash function so they eventually match, and finds a better hash function for the original problem (although it is still not guaranteed to find the globally optimal function of problem~\eqref{e:objfcn}, which is NP-complete).

\begin{figure}[b!]
  \centering
  \begin{tabular}{@{}c@{\hspace{7ex}}}
    \psfrag{01}[l][l]{$\{-1,+1\}^{b \times N}$}
    \psfrag{free}[][]{\caja{c}{c}{free binary \\ codes}}
    \psfrag{opt}[l][l]{\caja{c}{c}{codes from optimal \\ hash function}}
    \psfrag{hashfcn}[l][l]{\caja{c}{c}{codes realizable \\ by hash functions}}
    \psfrag{directfit}[l][lB]{\caja{c}{c}{two-step codes}}
    \includegraphics[width=0.60\columnwidth,bb=125 380 462 645,clip]{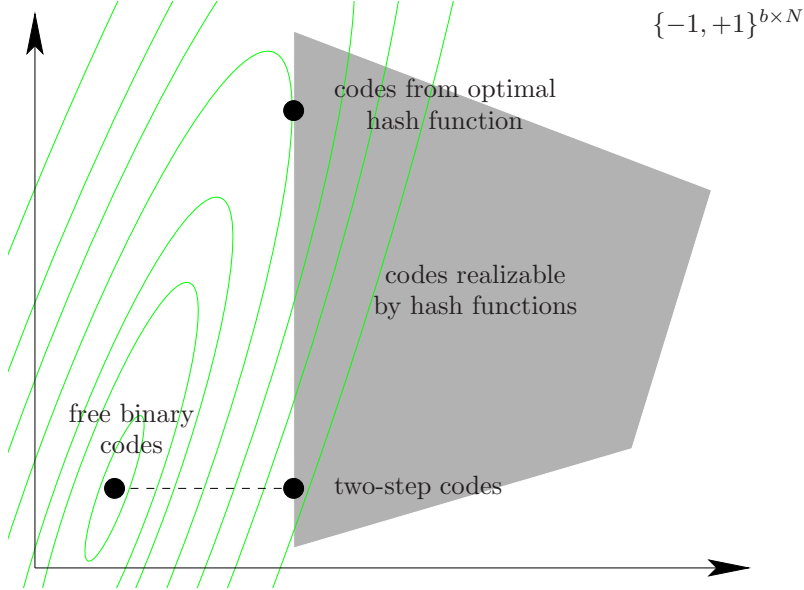}
  \end{tabular}
  \caption{Illustration of free codes, two-step codes and optimal codes realizable by a hash function, in the space $\{-1,+1\}^{b \times N}$.}
  \label{f:free-codes}
\end{figure}

Fig.~\ref{f:free-codes} illustrates this conceptually. It shows the space of all possible binary codes, the contours of $E(\Z)$ (green) and the set of codes that can be produced by (say) linear hash functions \h\ (gray), which is the feasible set $\{\Z \in \{-1,+1\}^{b \times N}\mathpunct{:}\ \Z = \h(\X) \text{ for linear \h}\}$. The two-step codes ``project'' the free codes onto the feasible set, but these are not the codes for the optimal hash function \h.

\subsection{Runtime}
\label{s:runtime}

The runtime per iteration for our $10\,000$-point training sets with $b=48$ bits and $\kappa_+ = 100$ and $\kappa_- = 500$ neighbors in a laptop is 2' for both \emph{MACcut} and \emph{MACquad}. They stop after 10--20 iterations. Each iteration is comparable to a single \emph{cut} or \emph{quad }run, since the \Z\ step dominates the computation. The iterations after the first one are faster because they are warm-started.

\subsection{Comparison with binary hashing methods}
\label{s:comparison}

\begin{figure}[t!]
  \centering
  \psfrag{rerror}[][t]{loss function \calL}
  \psfrag{iteration}[t][]{iterations}
  \psfrag{K}[][]{$k$}
  \psfrag{precision}[][t]{precision}
  \psfrag{recall}[][]{recall}
  \begin{tabular}{@{}l@{}c@{}c@{}c@{}c@{}c@{}}
    & $b=16$ & $b=32$ & $b=48$ & $b=64$\\
    \hspace{2ex}\rotatebox{90}{\hspace{7ex}precision} &
    \includegraphics[width=0.24\linewidth]{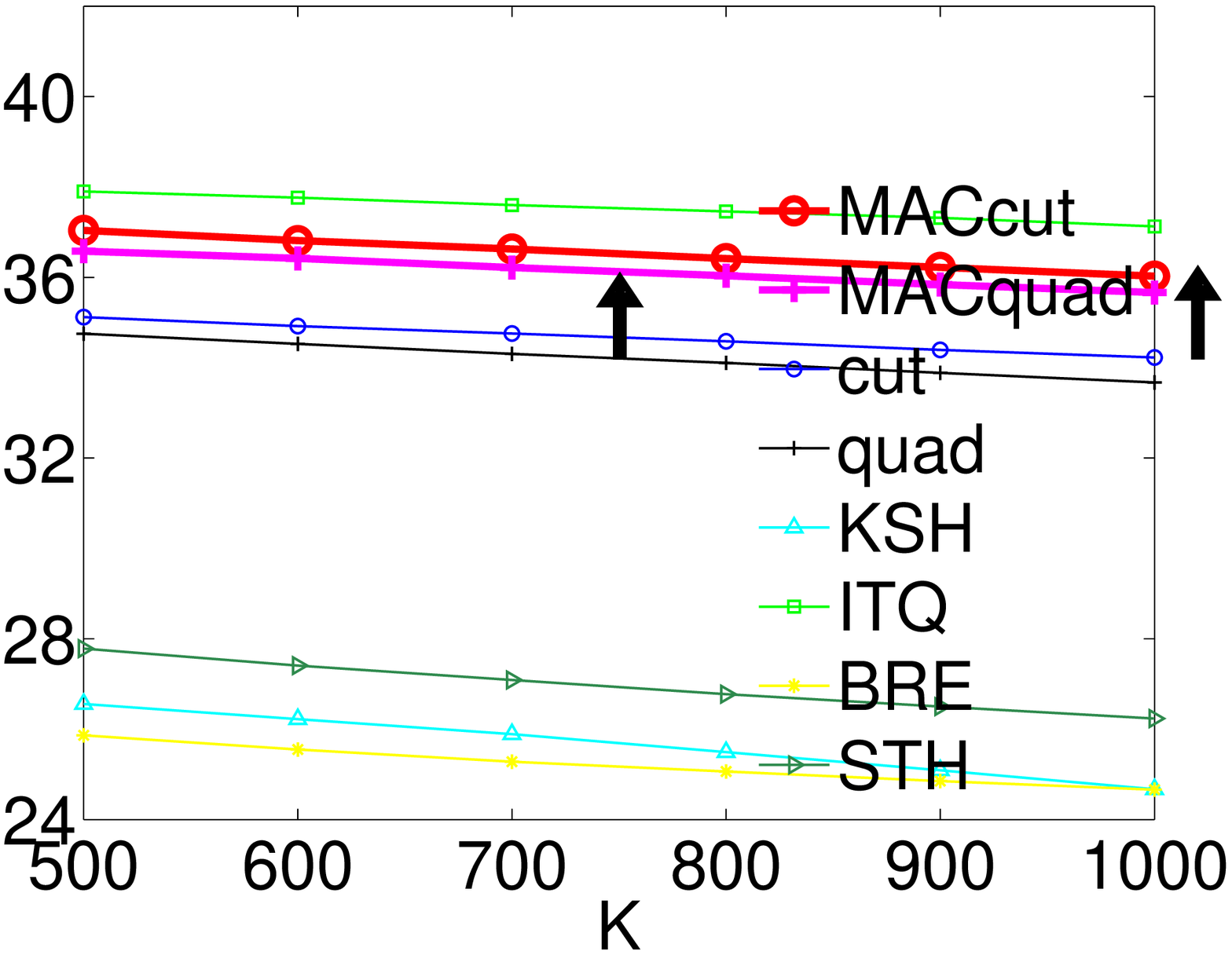} &
    \includegraphics[width=0.24\linewidth]{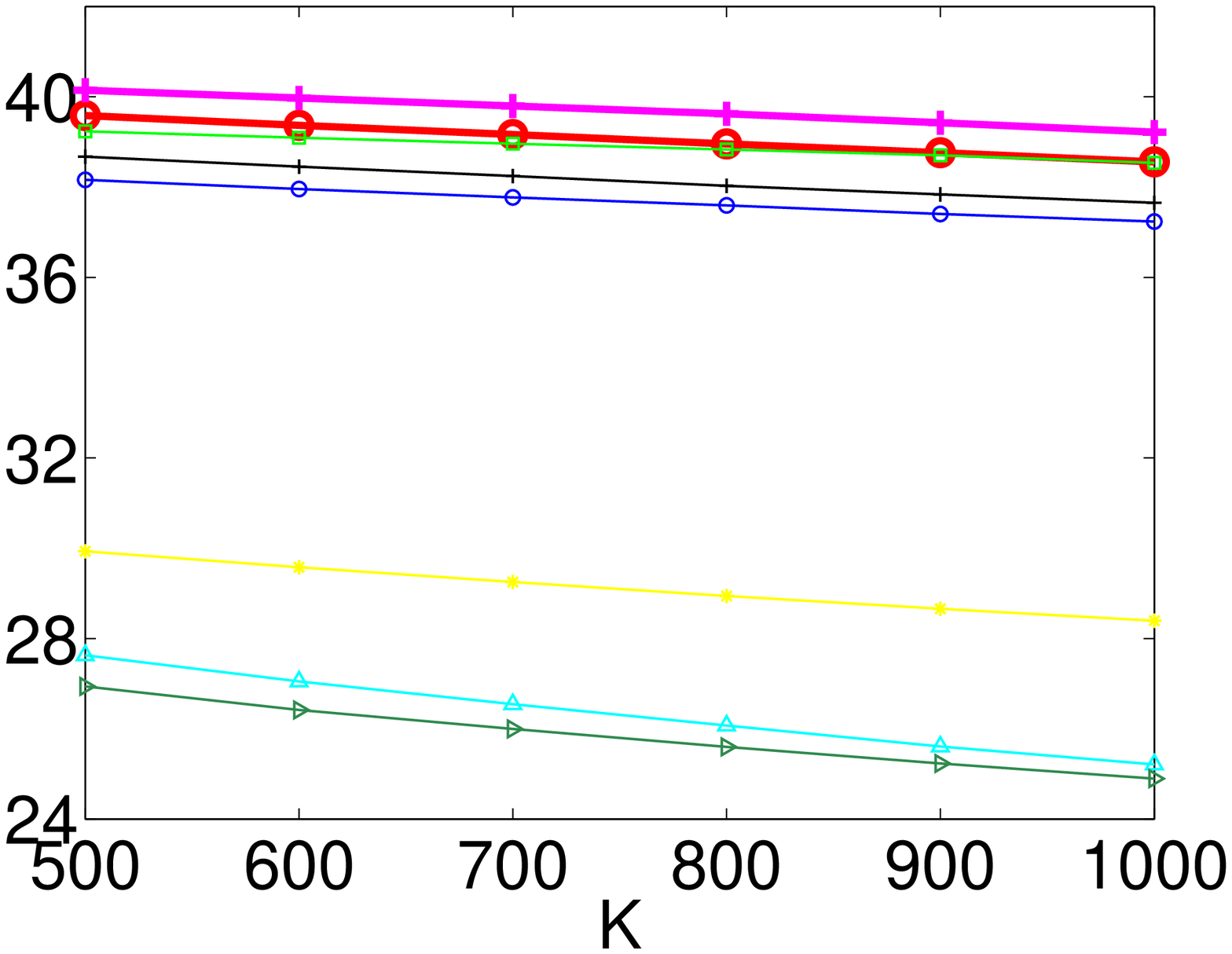} &
    \includegraphics[width=0.24\linewidth]{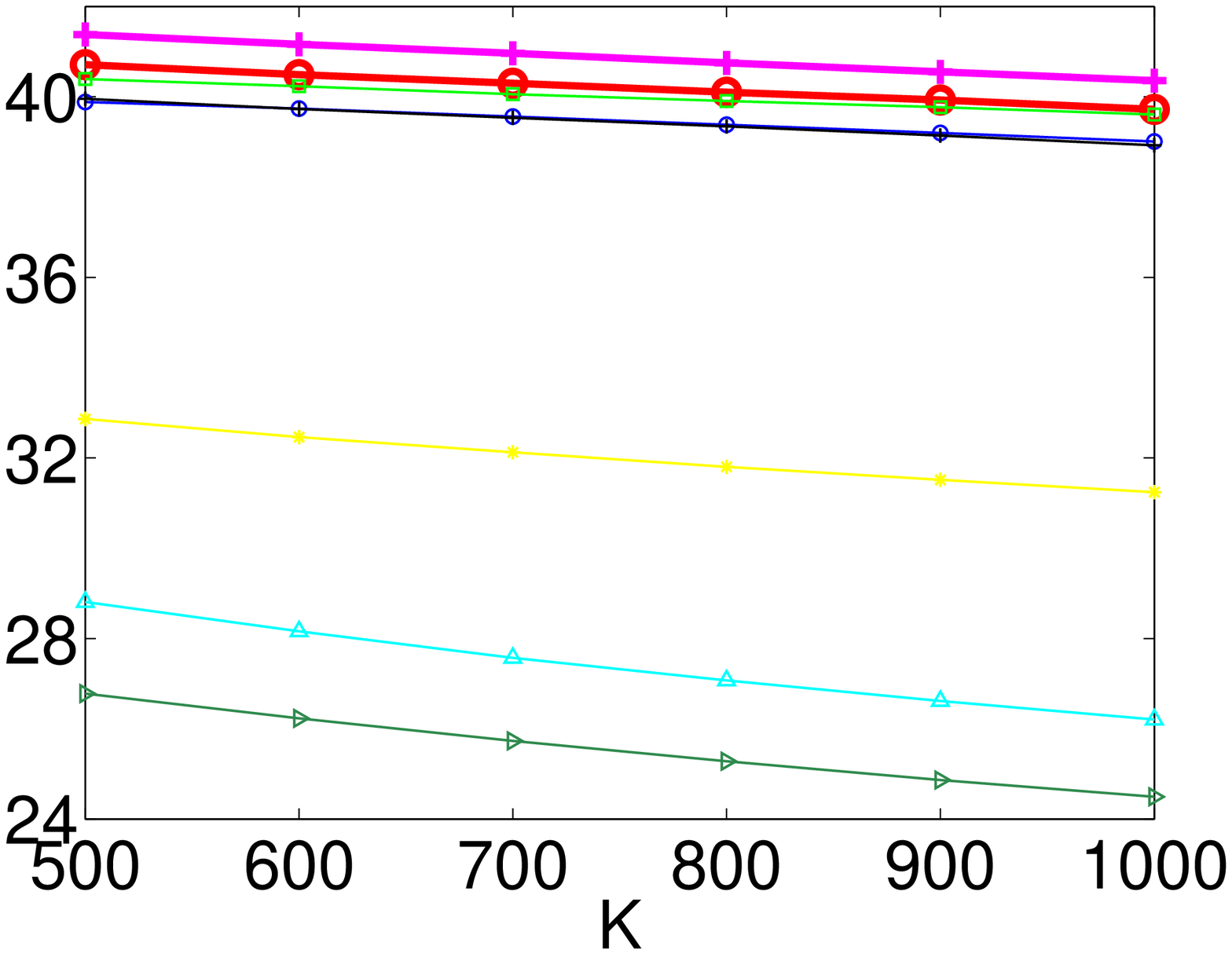} &
    \includegraphics[width=0.24\linewidth]{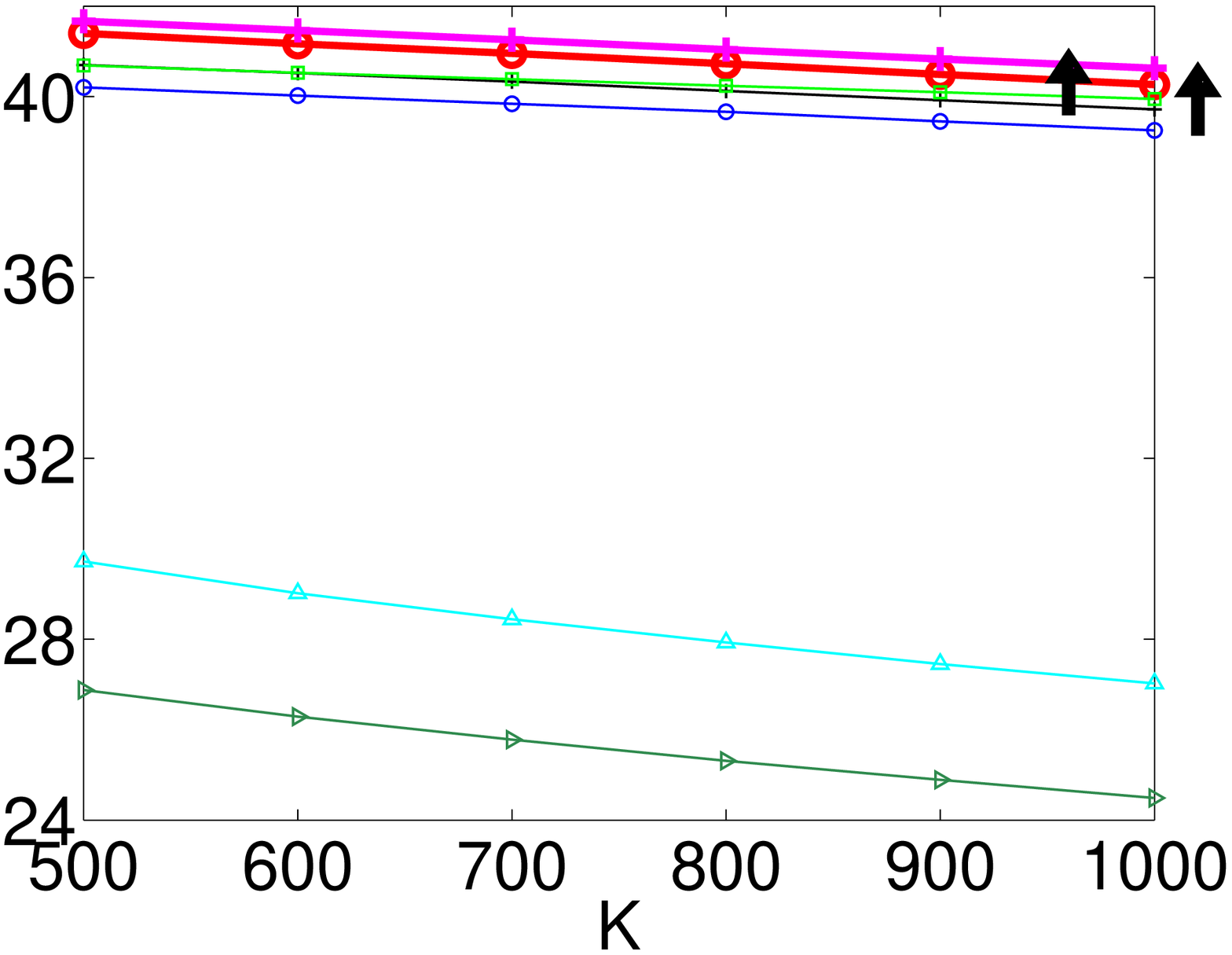} \\[-1ex]
    \hspace{2ex}\rotatebox{90}{\raisebox{3ex}[0pt][0pt]{\makebox[0pt][l]{\hspace{20ex}CIFAR}}\hspace{7ex}precision} &
    \includegraphics[width=0.24\linewidth]{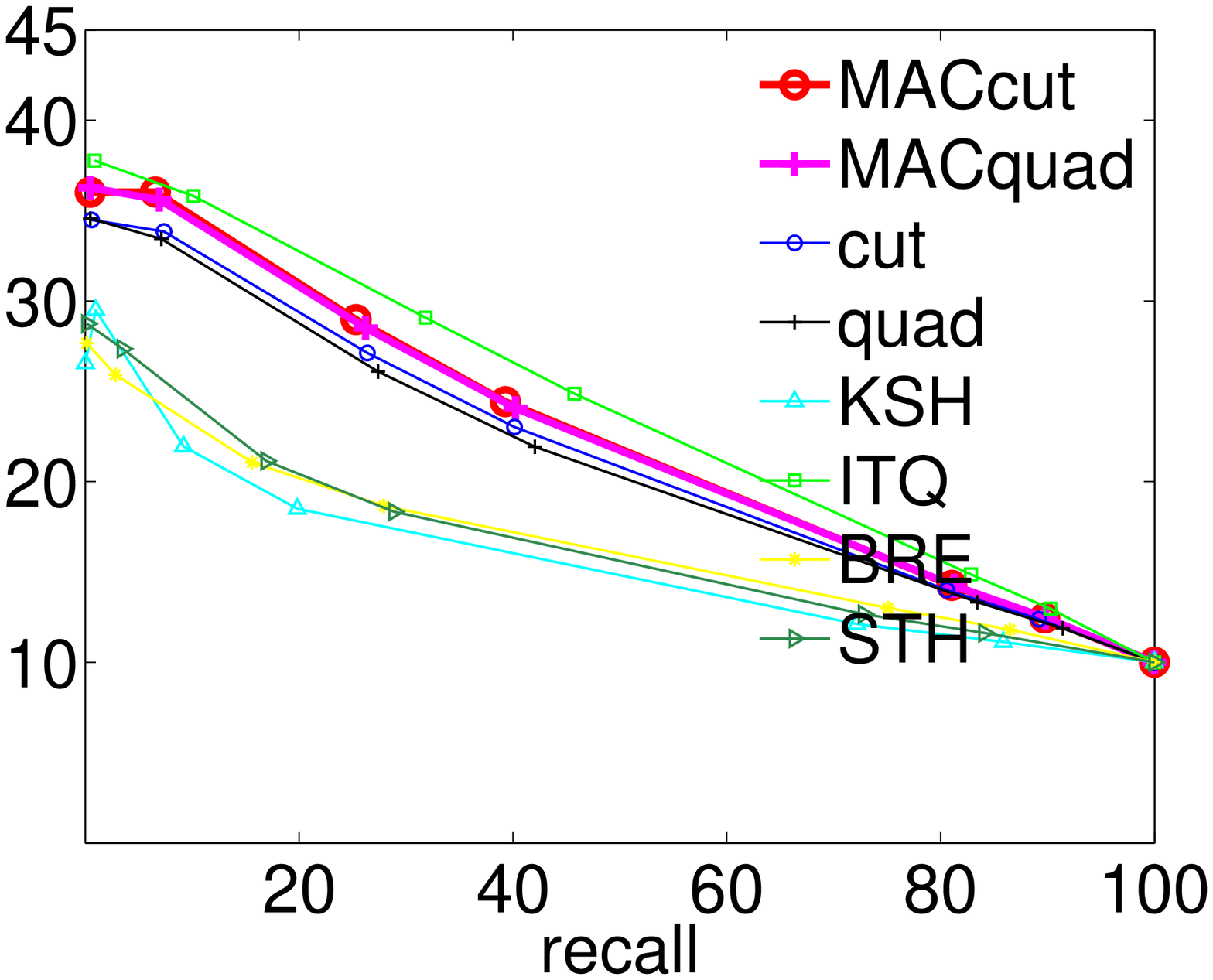} &
    \includegraphics[width=0.24\linewidth]{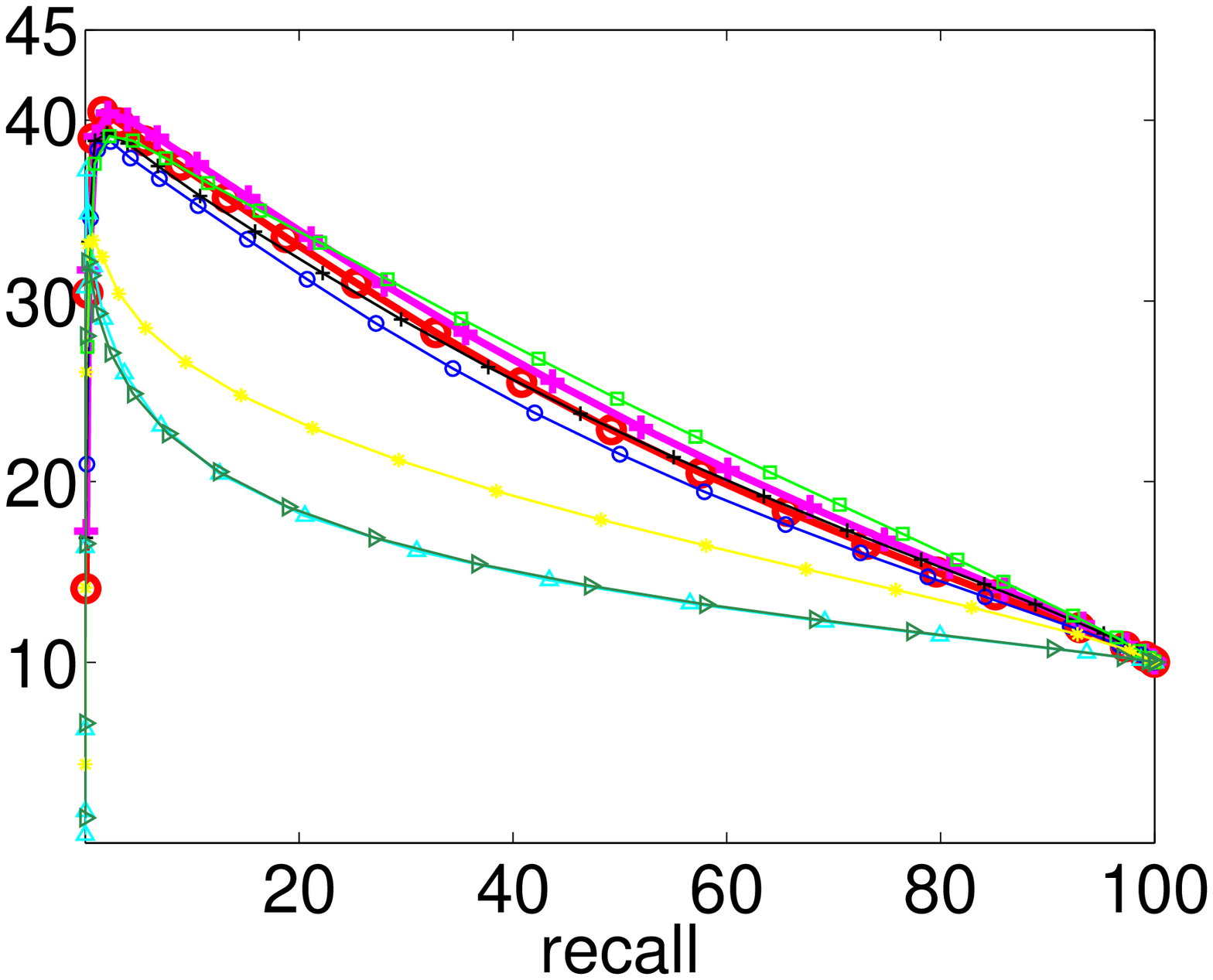} &
    \includegraphics[width=0.24\linewidth]{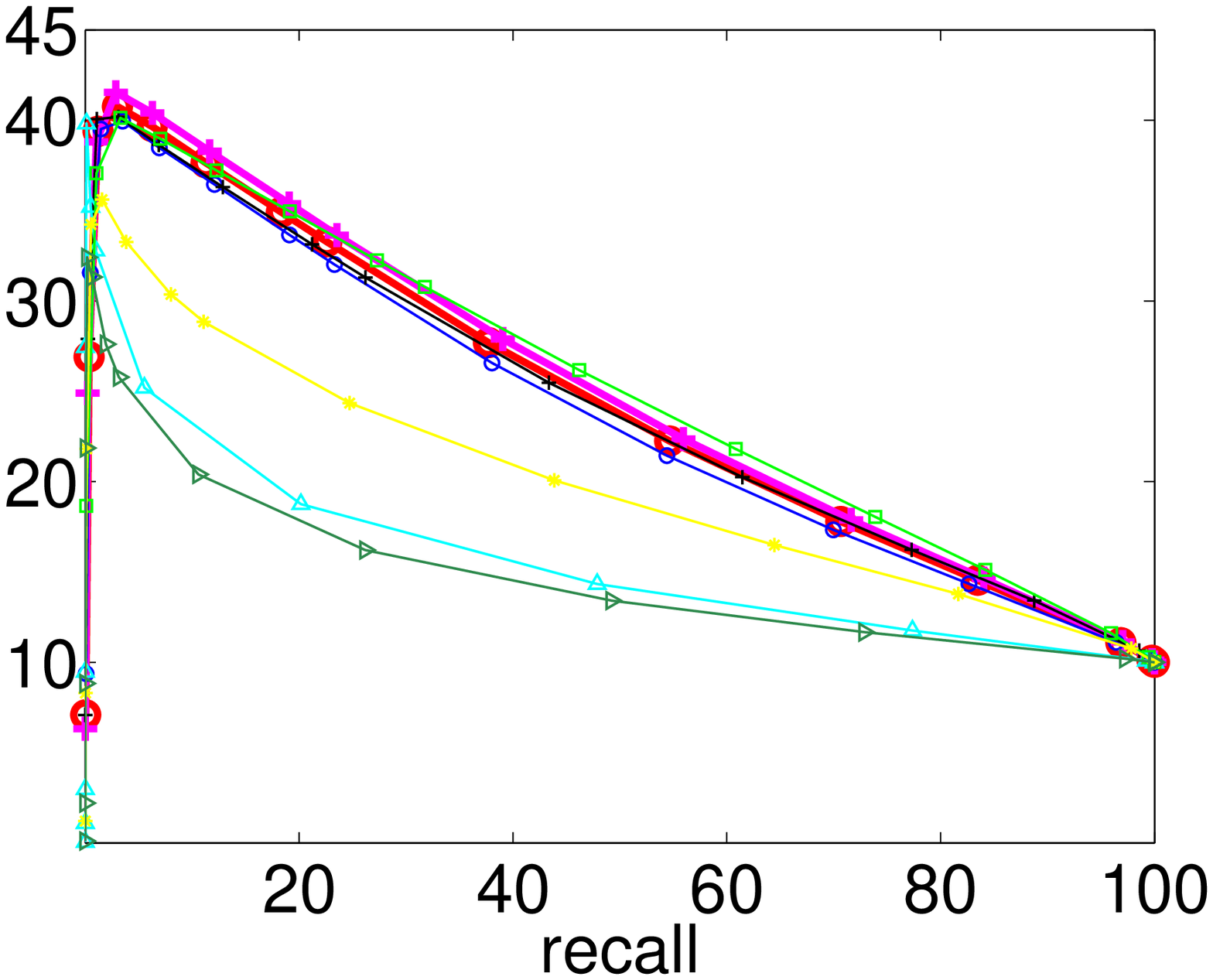} &
    \includegraphics[width=0.24\linewidth]{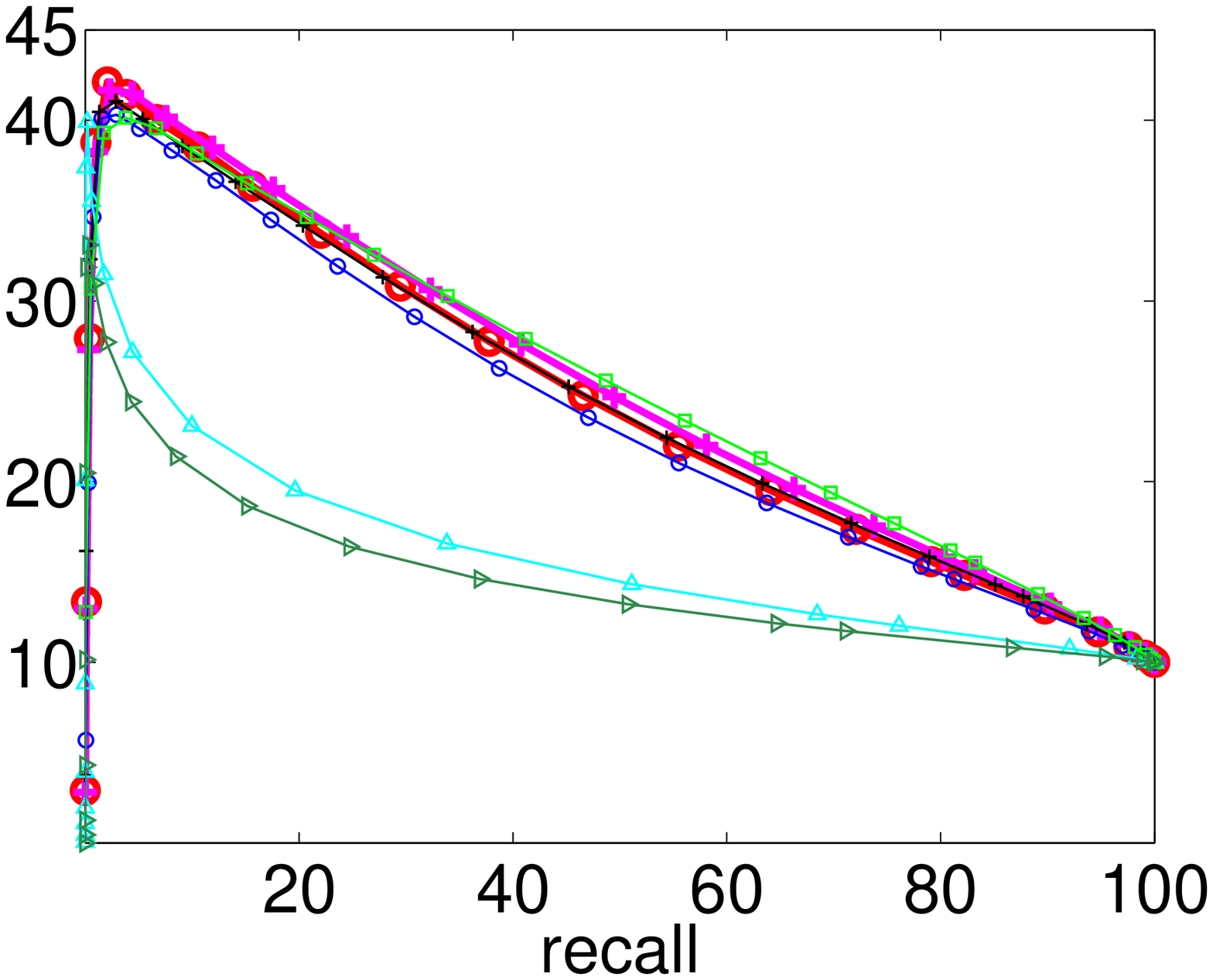}
  \end{tabular} \\[1ex]
  \begin{tabular}{@{}l@{}c@{}c@{}c@{}c@{}c@{}}
    \hspace{2ex}\rotatebox{90}{\hspace{7ex}precision} &
    \includegraphics[width=0.24\linewidth]{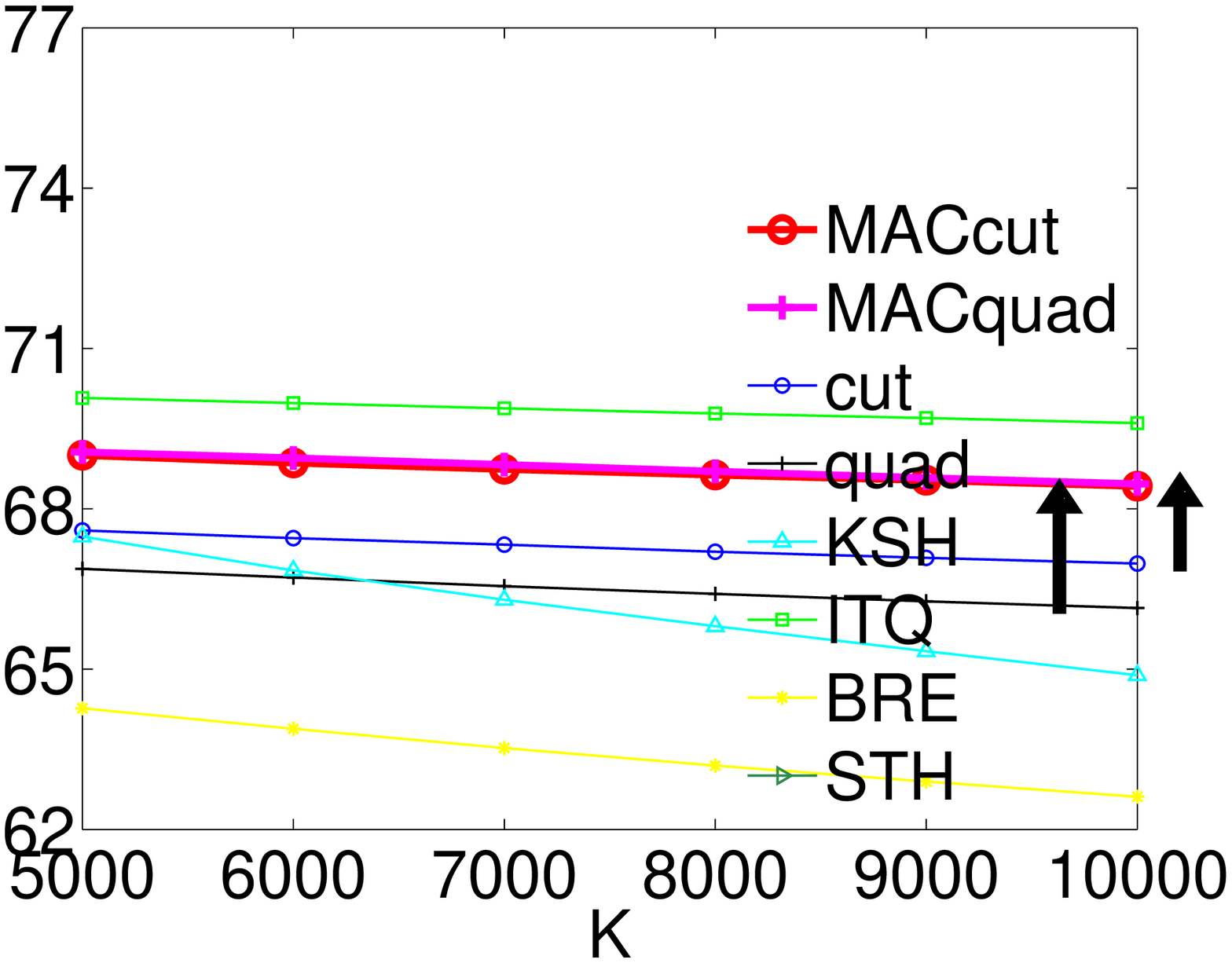} &
    \includegraphics[width=0.24\linewidth]{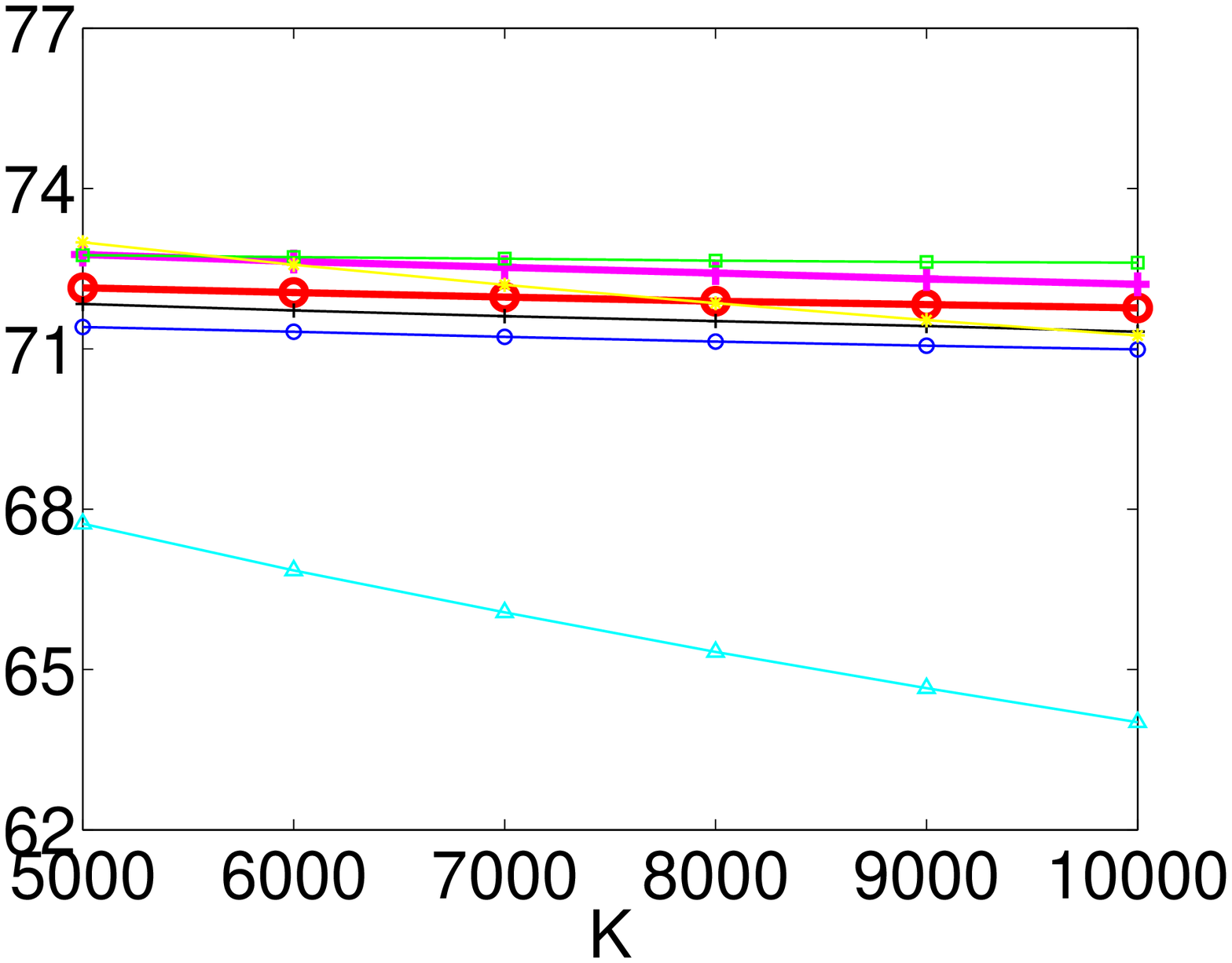} &
    \includegraphics[width=0.24\linewidth]{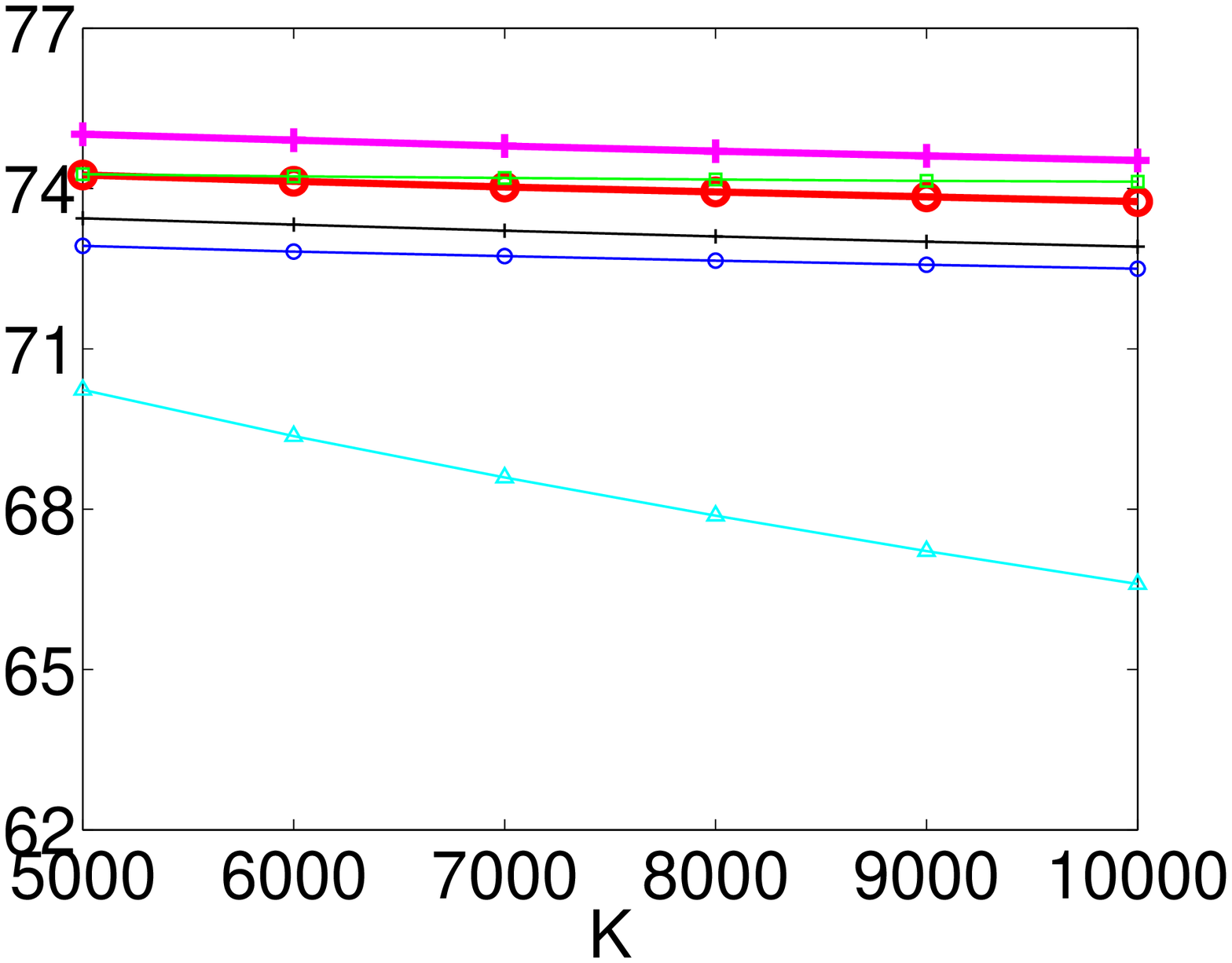} &
    \includegraphics[width=0.24\linewidth]{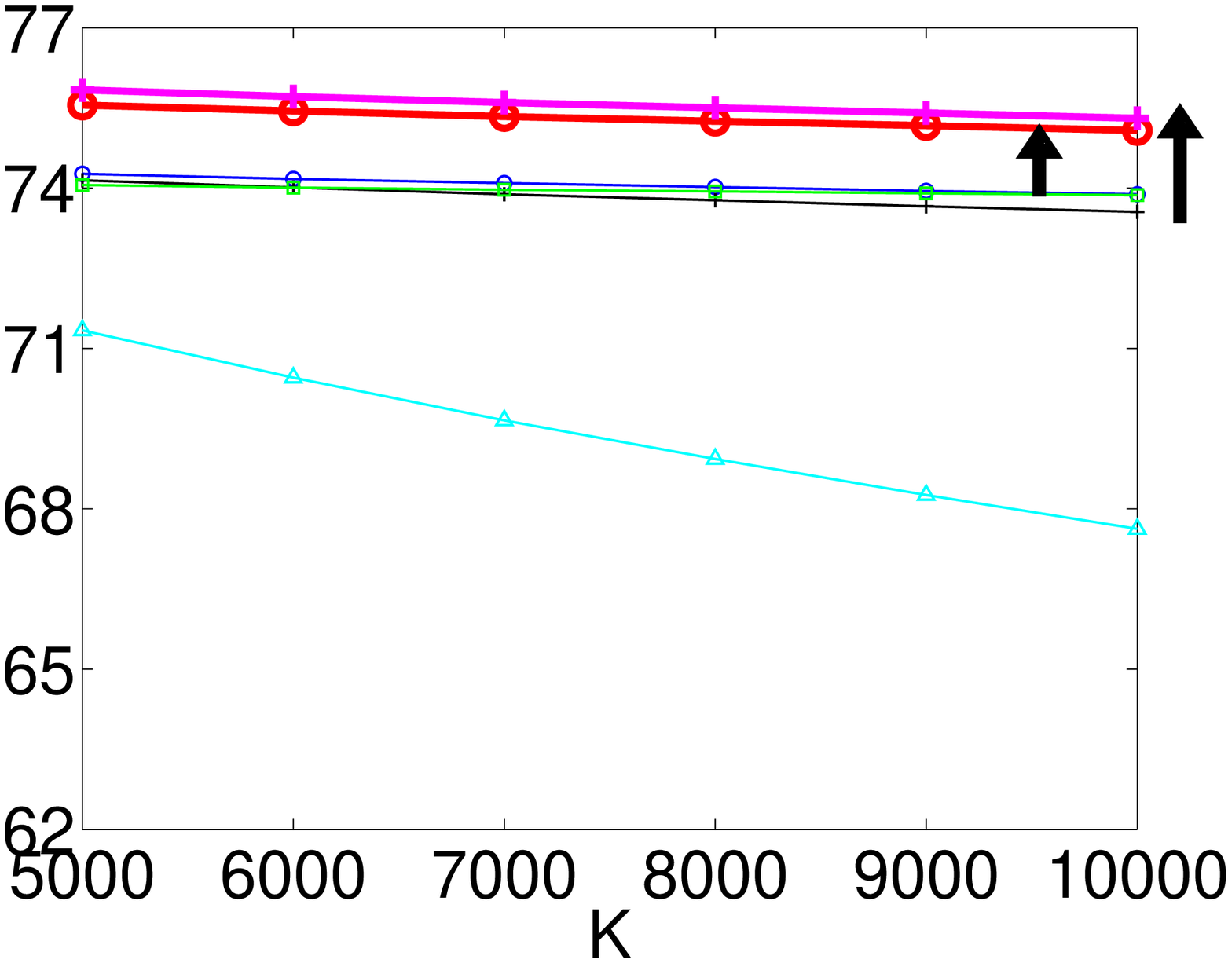} \\[-1ex]
    \hspace{2ex}\rotatebox{90}{\raisebox{3ex}[0pt][0pt]{\makebox[0pt][l]{\hspace{17ex} Infinite MNIST}}\hspace{7ex}precision} &
    \includegraphics[width=0.24\linewidth]{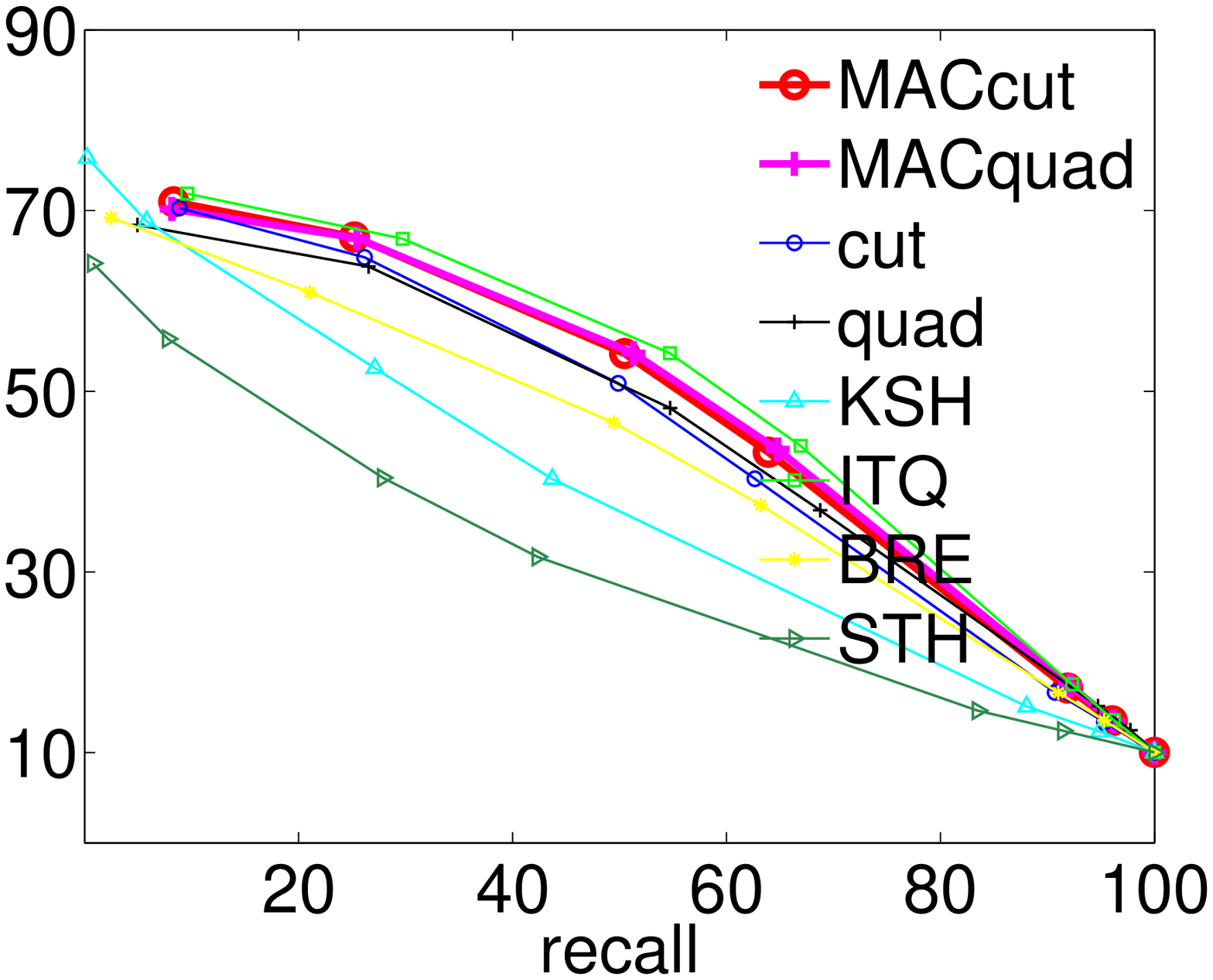} &
    \includegraphics[width=0.24\linewidth]{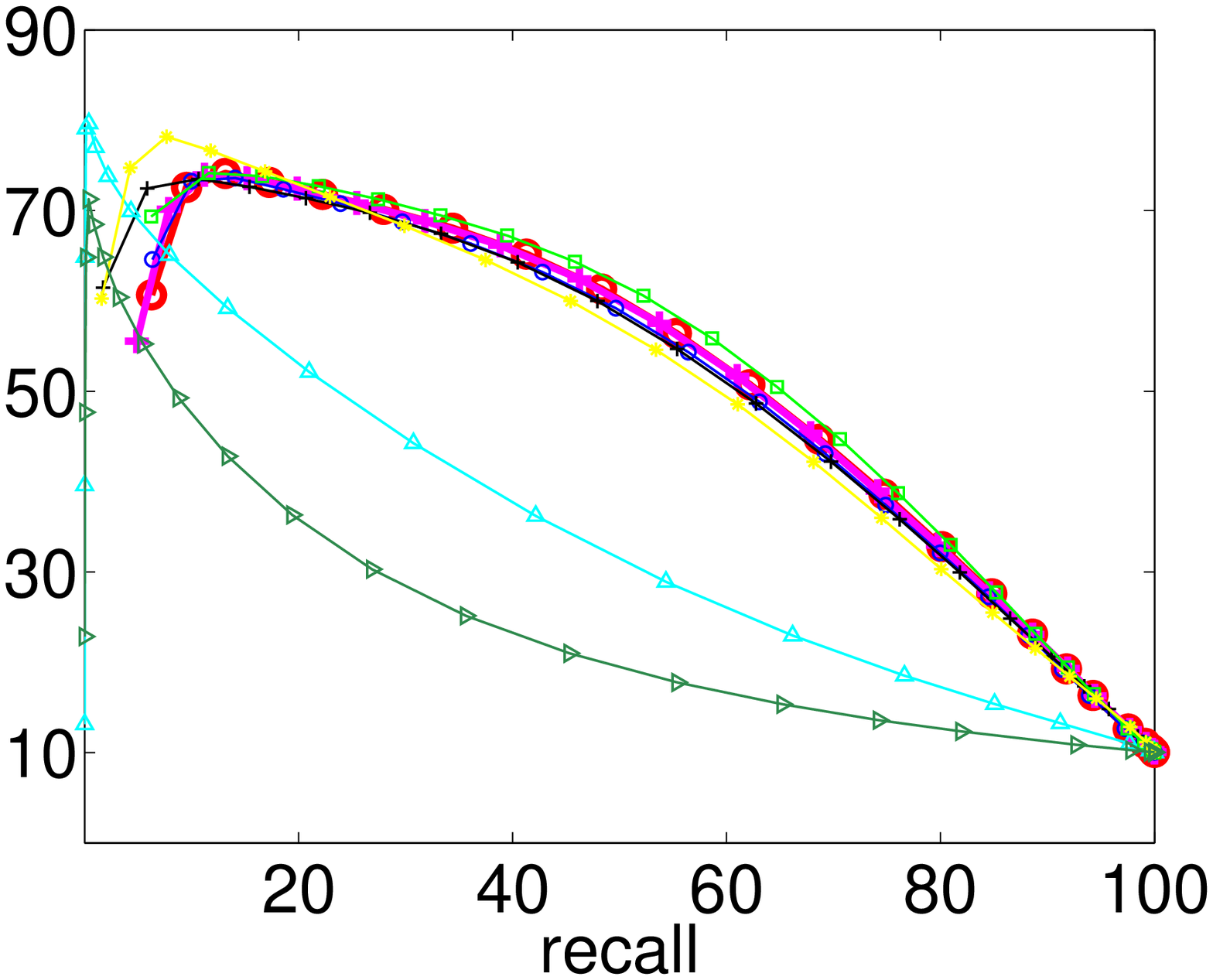} &
    \includegraphics[width=0.24\linewidth]{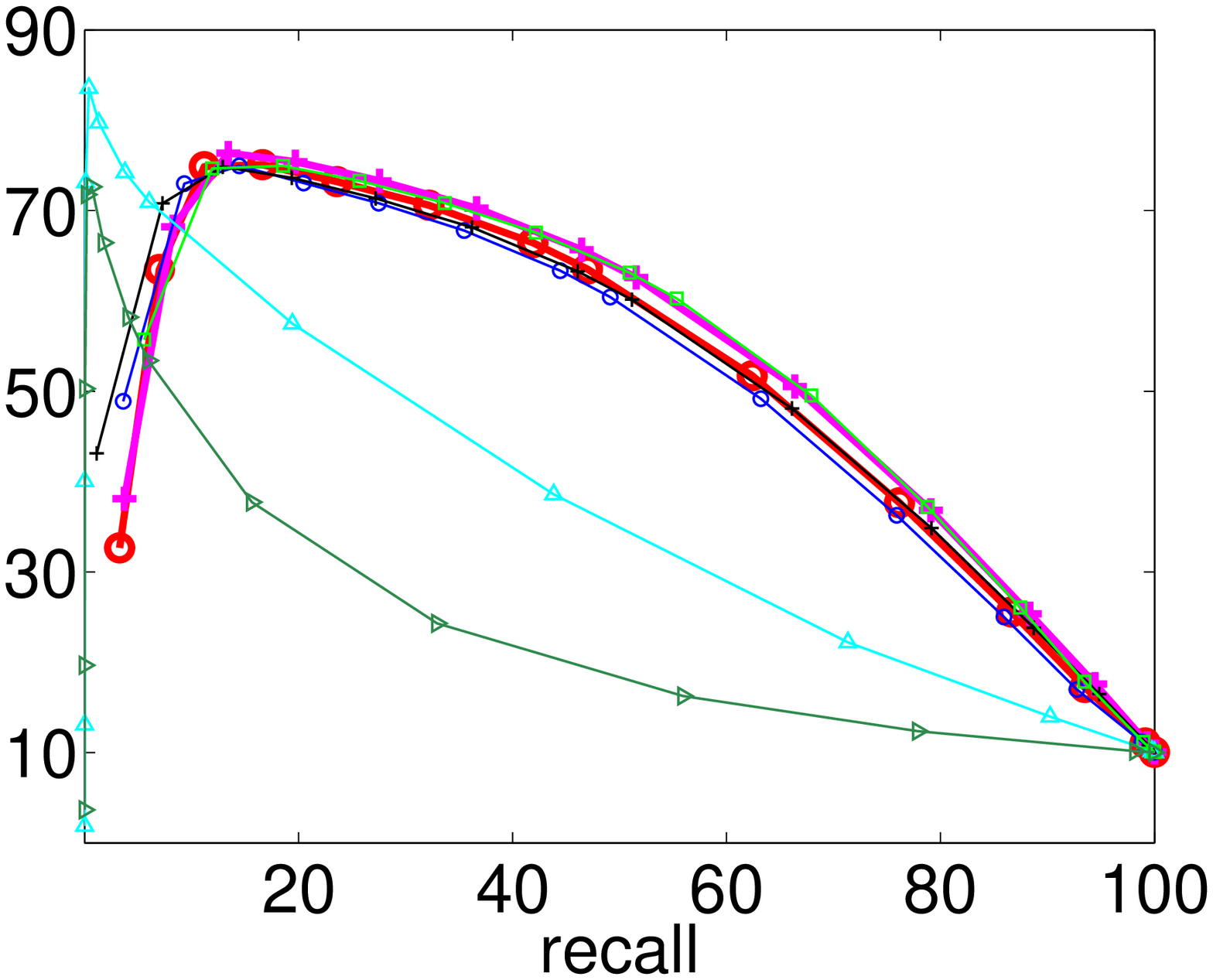} &
    \includegraphics[width=0.24\linewidth]{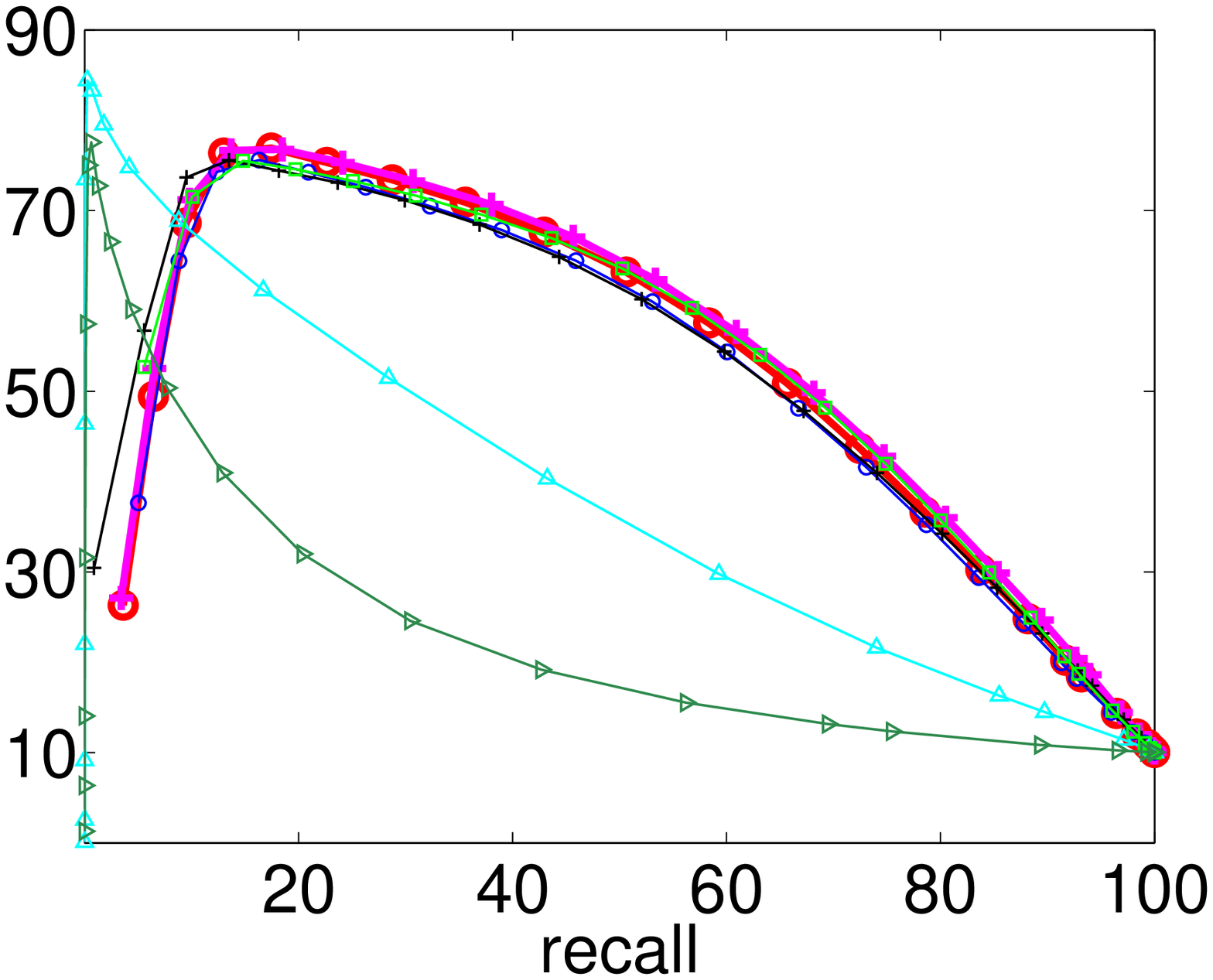}
  \end{tabular}
  \caption{Comparison with binary hashing methods on CIFAR (top panel) and Infinite MNIST (bottom panel), using a linear hash function, using $b=16$ to $64$ bits. The rows in each panel show the precision for $k$ retrieved points, for a range of $k$, and the precision/recall at different Hamming distances.}
  \label{f:sup-comparison}
\end{figure}

Fig.~\ref{f:sup-comparison} shows results on CIFAR and Infinite MNIST. We create affinities $y_{nm}$ for all methods using the dataset labels as before, with $\kappa_+ = 100$ similar neighbors and $\kappa_- = 500$ dissimilar neighbors. We compare \emph{MACquad} and \emph{MACcut} with Two-Step Hashing (\emph{quad}) \citep{Lin_13a}, FastHash (\emph{cut}) \citep{Lin_14b}, Hashing with Kernels (KSH) \citep{Liu_12c}, Iterative Quantization (ITQ) \citep{Gong_13a}, Binary Reconstructive Embeddings (BRE) \citep{KulisDarrel09a} and Self-Taught Hashing (STH) \citep{Zhang_10e}. \emph{MACquad}, \emph{MACcut}, \emph{quad} and \emph{cut} all use the KSH loss function~\eqref{e:KSH}. The results show that \emph{MACcut} (and \emph{MACquad}) generally outperform all other methods, often by a large margin, in nearly all situations (dataset, number of bits, size of retrieved set). In particular, \emph{MACcut} and \emph{MACquad} are the only ones to beat ITQ, as long as one uses sufficiently many bits.

\begin{figure}[t!]
  \centering
  \psfrag{rerror}[][t]{loss function \calL}
  \psfrag{iteration}[t][]{iterations}
  \psfrag{K}[][]{$k$}
  \psfrag{precision}[][t]{precision}
  \psfrag{recall}[][]{recall}
  \begin{tabular}{@{}l@{}c@{}c@{}c@{}c@{}c@{}}
    & $b=16$ & $b=32$ & $b=48$ & $b=64$\\
    \hspace{2ex}\rotatebox{90}{\hspace{7ex}precision} &
    \includegraphics[width=0.24\linewidth]{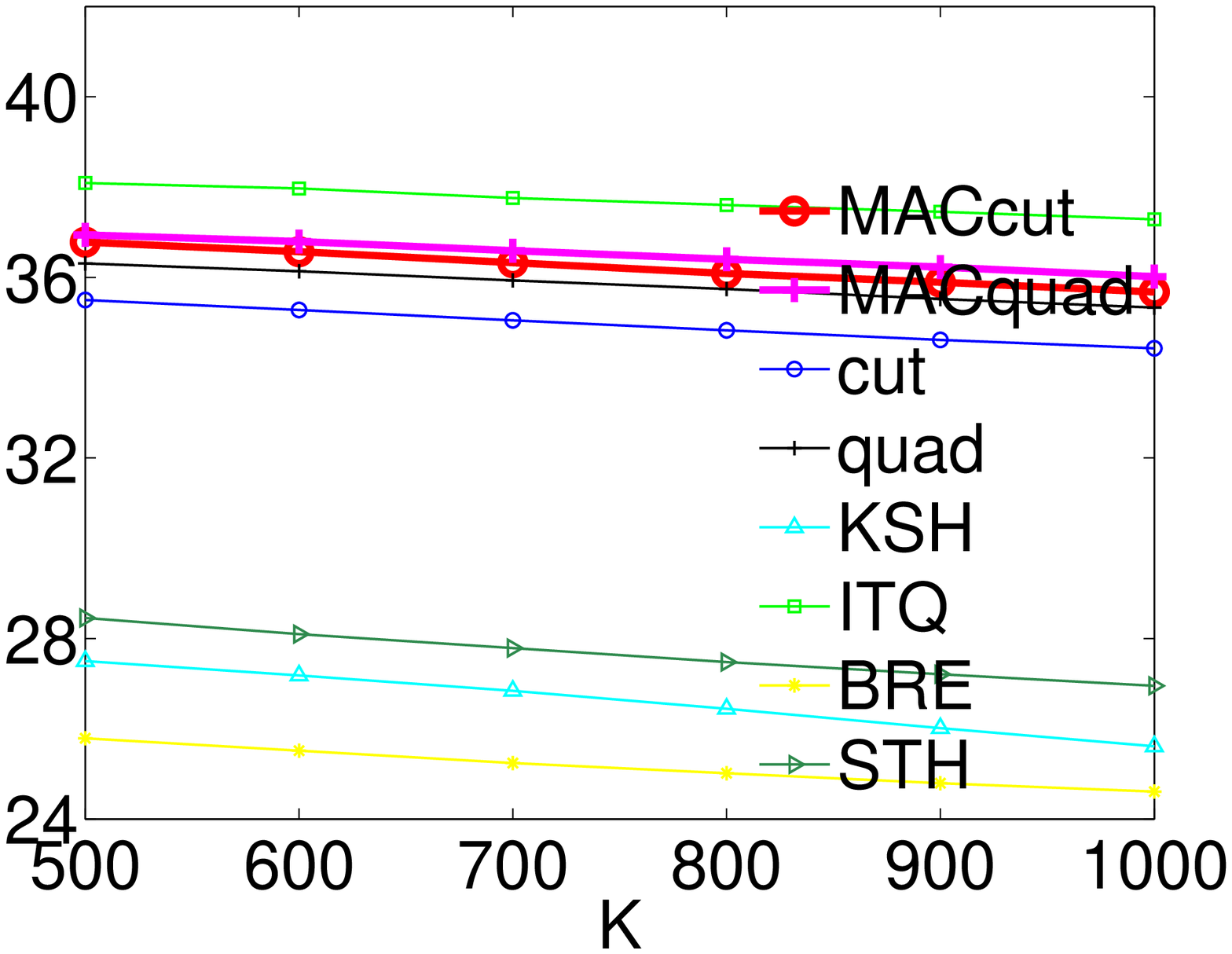} &
    \includegraphics[width=0.24\linewidth]{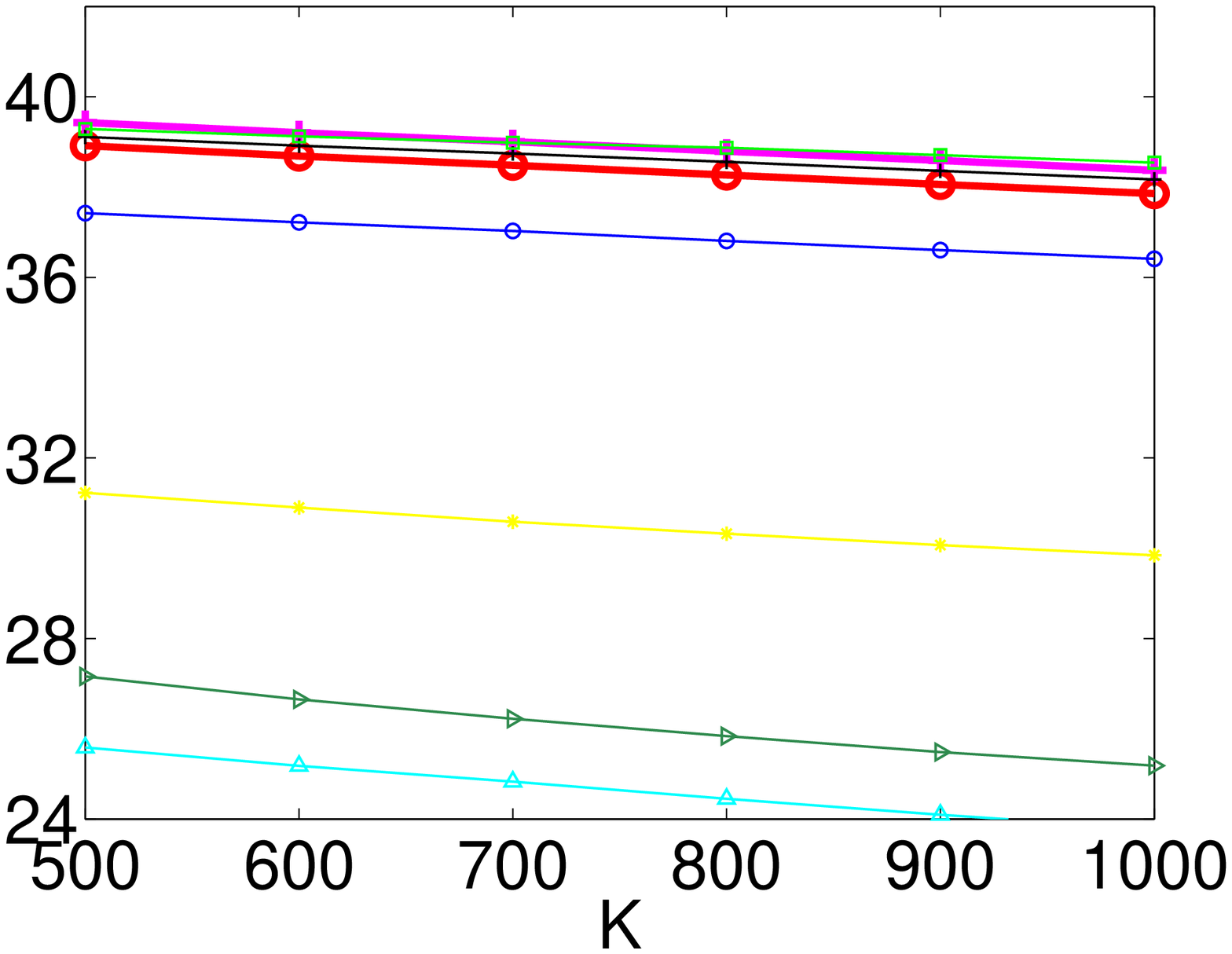} &
    \includegraphics[width=0.24\linewidth]{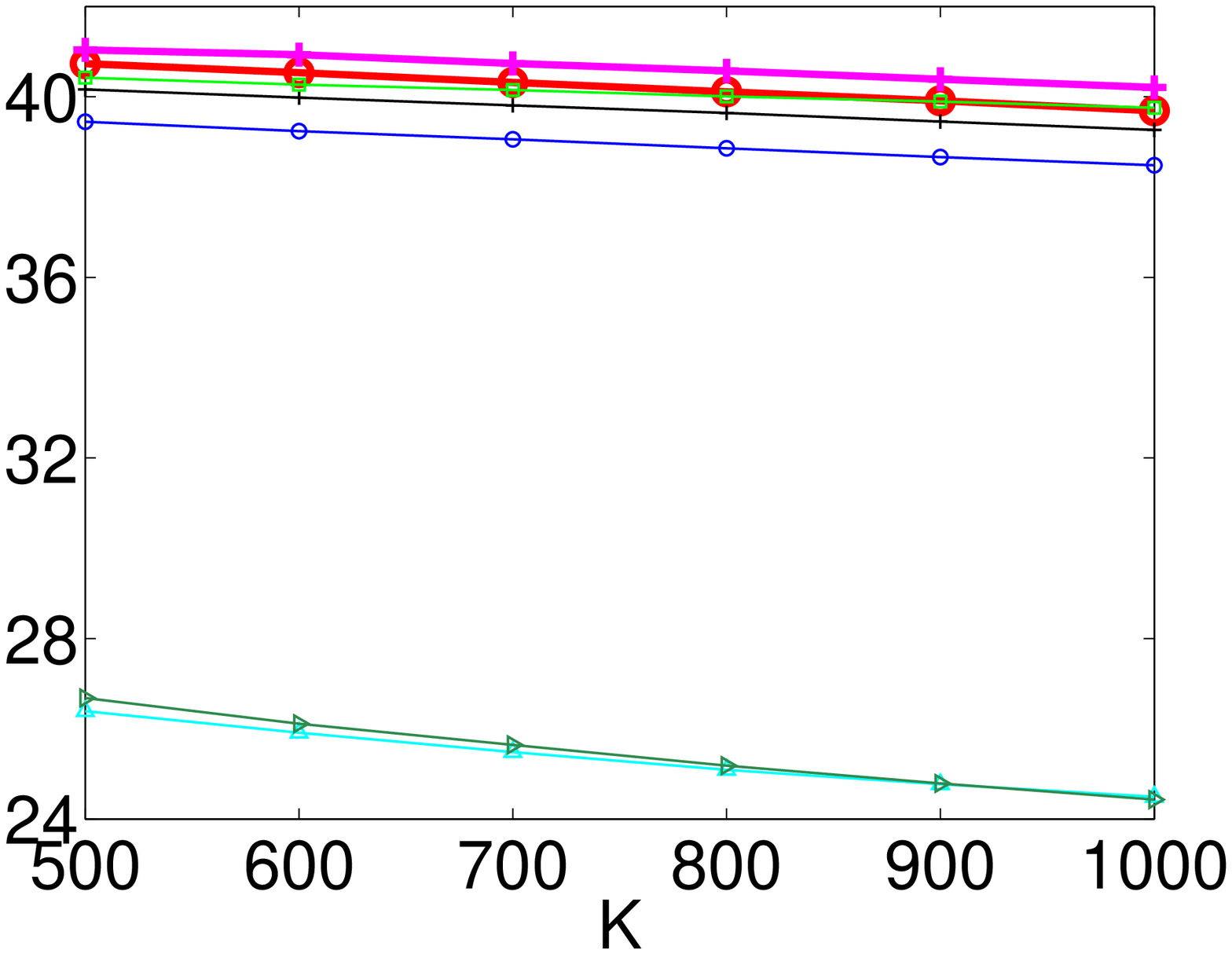} &
    \includegraphics[width=0.24\linewidth]{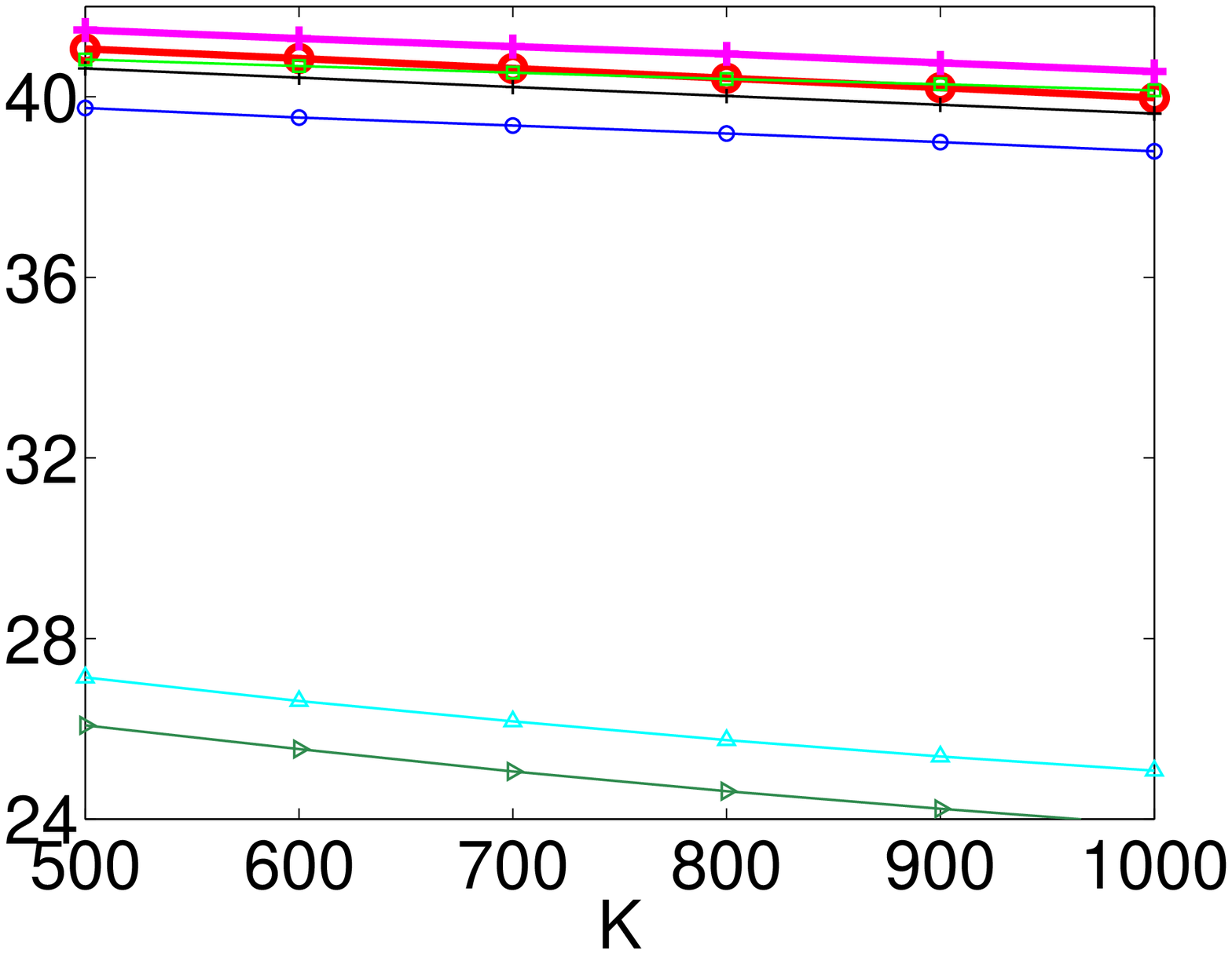} \\[-1ex]
    \hspace{2ex}\rotatebox{90}{\raisebox{3ex}[0pt][0pt]{\makebox[0pt][l]{\hspace{20ex}CIFAR}}\hspace{7ex}precision} &
    \includegraphics[width=0.24\linewidth]{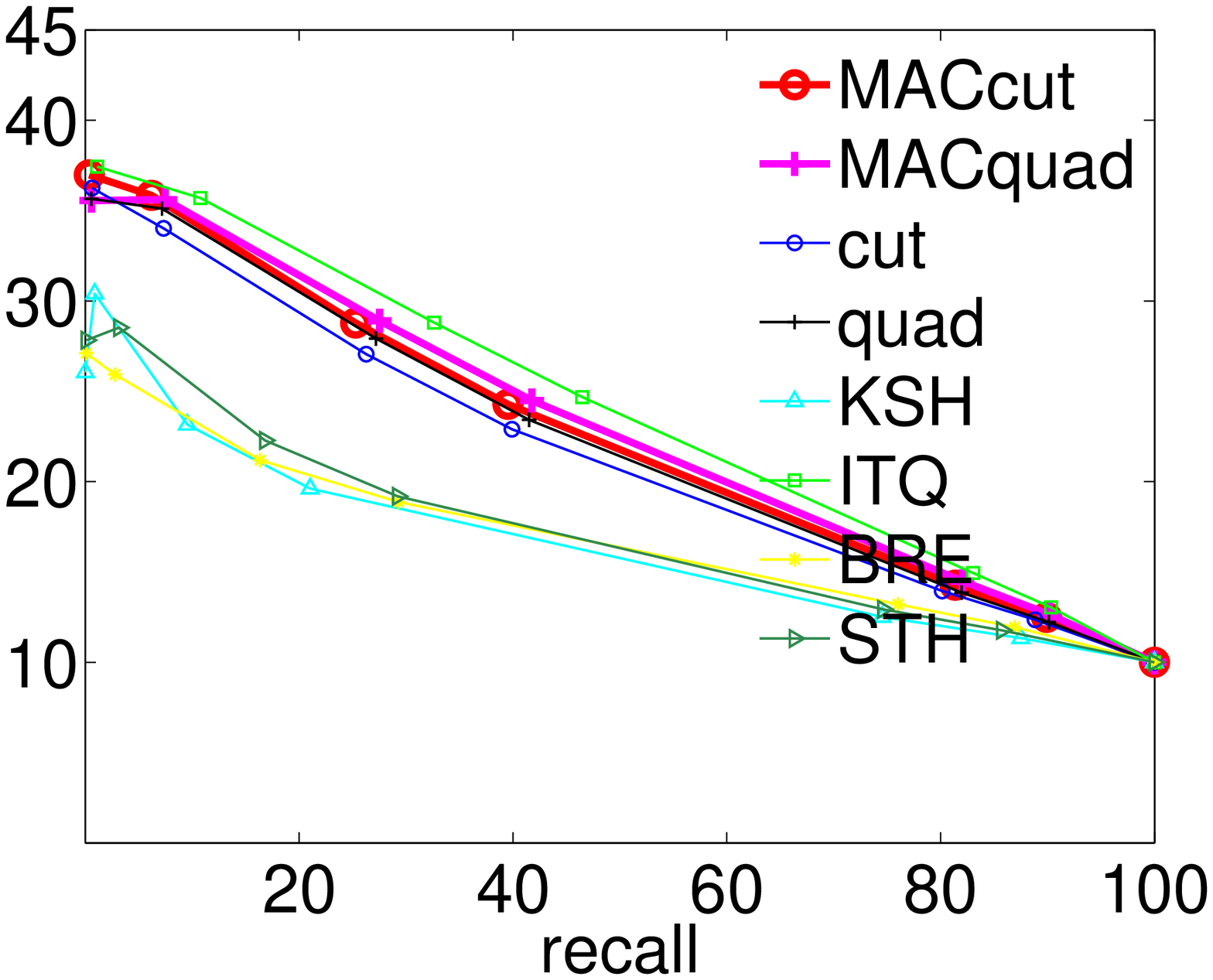} &
    \includegraphics[width=0.24\linewidth]{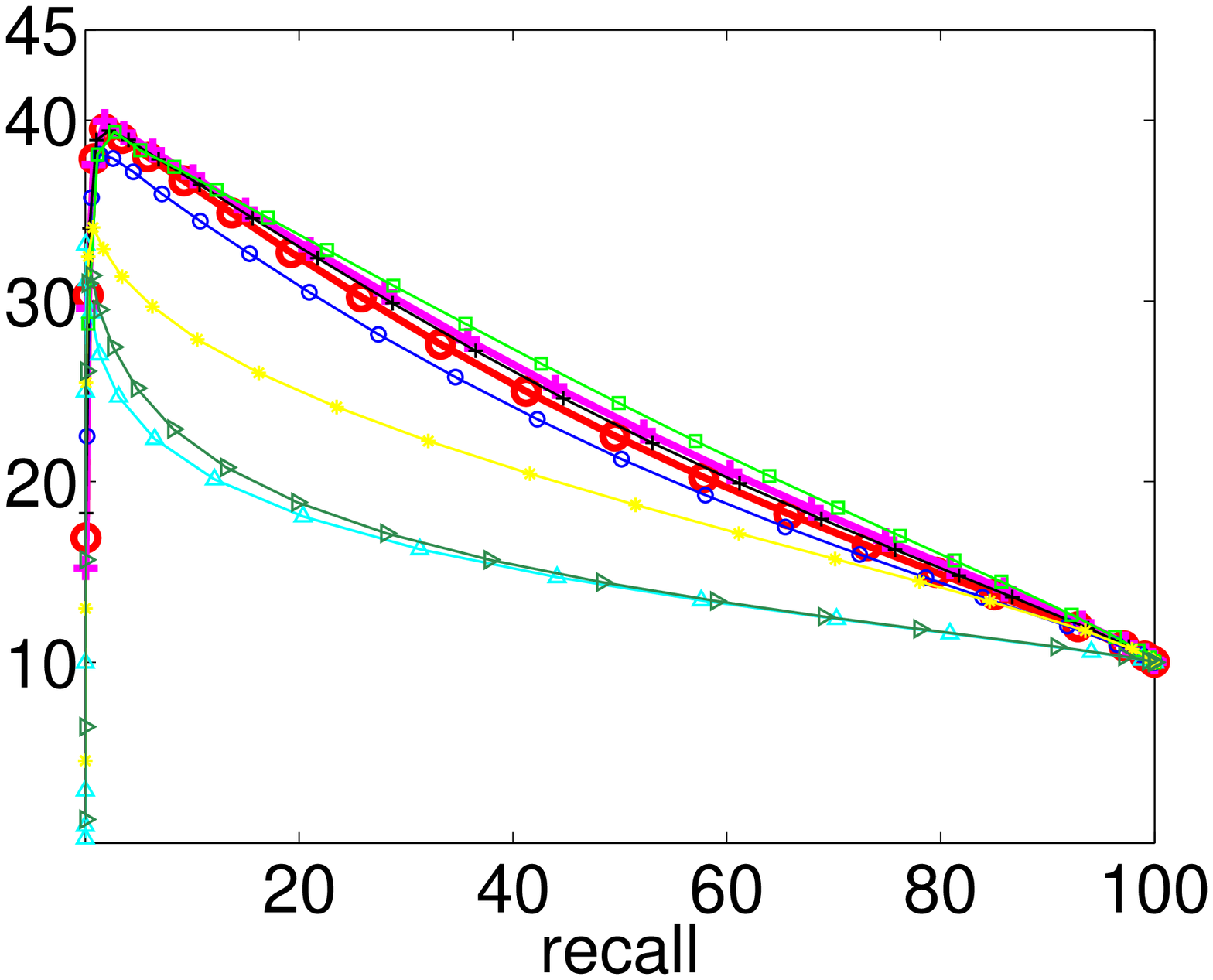} &
    \includegraphics[width=0.24\linewidth]{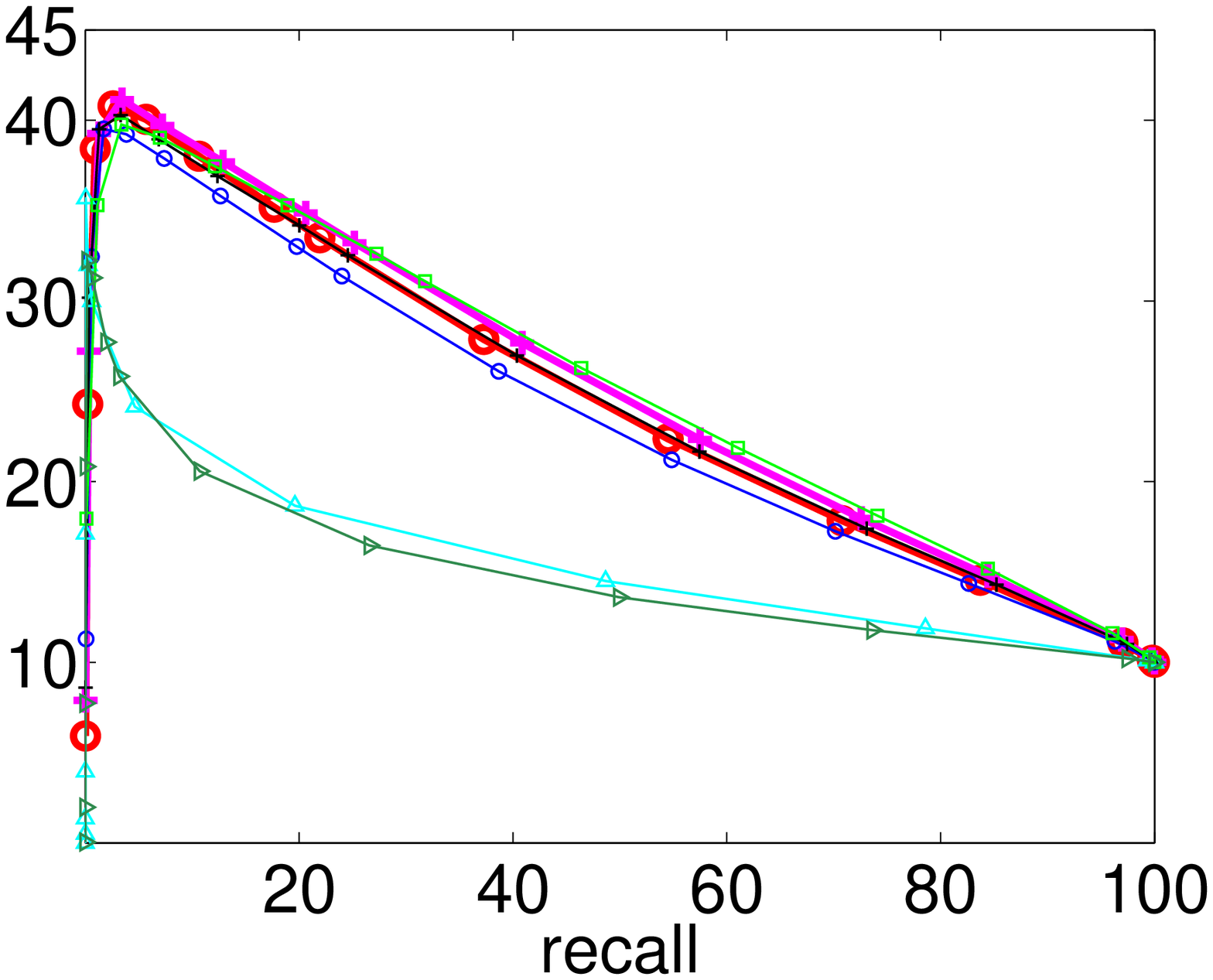} &
    \includegraphics[width=0.24\linewidth]{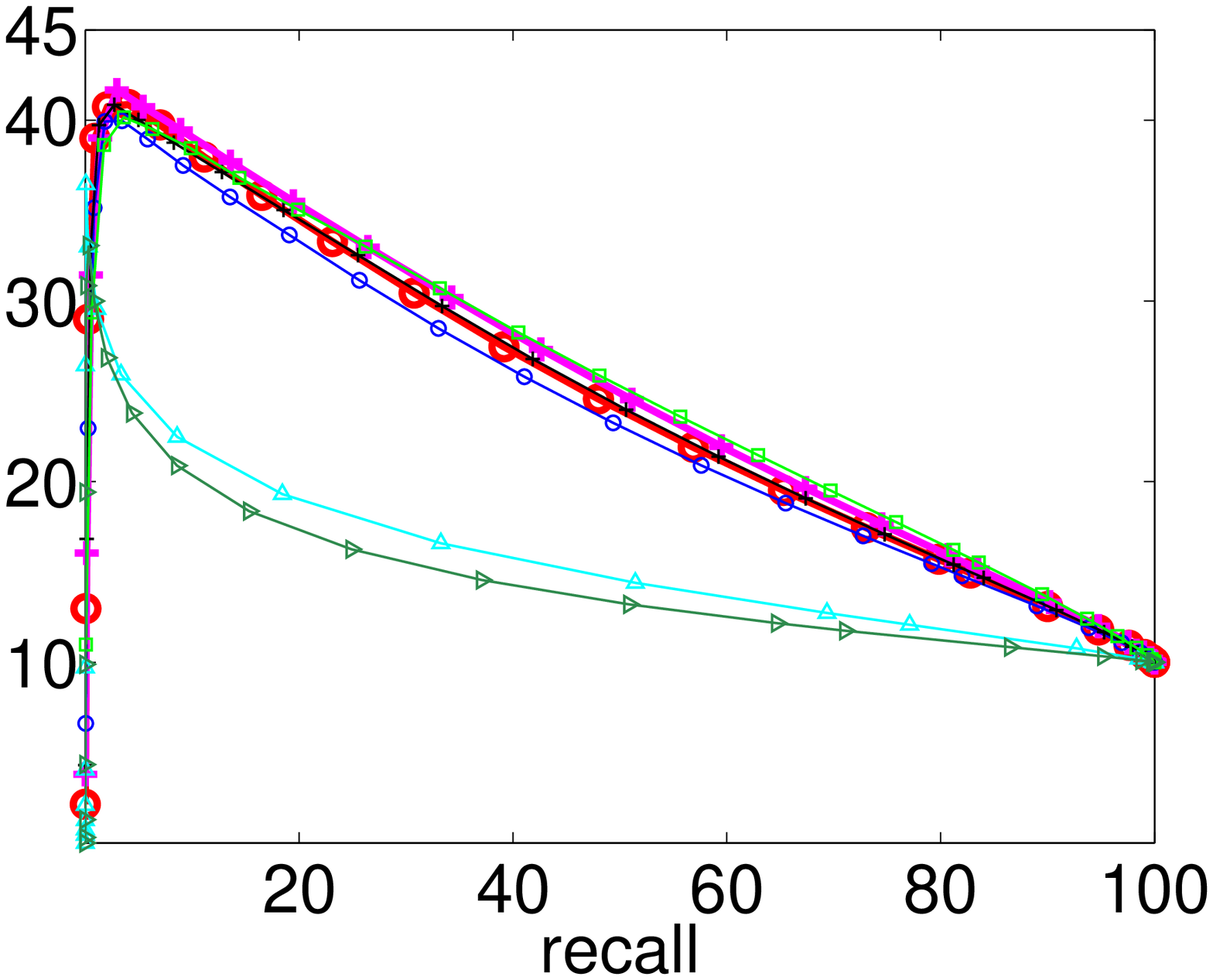}
  \end{tabular}
  \caption{As in fig.~\ref{f:sup-comparison} but using the cosine similarity instead of the Euclidean distance to find neighbors (i.e., all the points are centered and normalized before training and testing), on CIFAR.}
  \label{f:sup-cosine}
\end{figure}

\section{Discussion}
\label{s:discussion}

\paragraph{Two-step approaches vs the MAC algorithm for affinity-based loss functions}

The two-step approach of Two-Step Hashing \citep{Lin_13a} and FastHash \citep{Lin_14b} is a significant advance in finding good codes for binary hashing, but it also causes a maladjustment between the codes and the hash function, since the codes were learned without knowledge of what hash function would use them. Ignoring the interaction between the loss and the hash function limits the quality of the results. For example, a linear hash function will have a harder time than a nonlinear one at learning such codes. In our algorithm, this tradeoff is enforced gradually (as $\mu$ increases) in the \Z\ step as a regularization term (eq.~\eqref{e:MAC-QP}): it finds the best codes according to the loss function, but makes sure they are close to being realizable by the current hash function. Our experiments demonstrate that significant, consistent gains are achieved in both the loss function value and the precision/recall in image retrieval over the two-step approach.

A similar, well-known situation arises in feature selection for classification \citep{KohaviJohn98a}. The best combination of classifier and features will result from jointly minimizing the classification error with respect to both classifier and features (the ``wrapper'' approach), rather than first selecting features according to some criterion and then using them to learn a particular classifier (the ``filter'' approach). From this point of view, the two-step approaches of \citep{Lin_13a,Lin_14b} are filter approaches that first optimize the loss function over the codes \Z\ (equivalently, optimize $\calL_P$ with $\mu = 0$) and then fit the hash function \h\ to those codes. Any such filter approach is then equivalent to optimizing $\calL_P$ over $(\Z,\h)$ for $\mu\rightarrow 0^+$.

The method of auxiliary coordinates algorithmically decouples (within each iteration) the two elements that make up a binary hashing model: the hash function and the loss function. Both elements act in combination to produce a function that maps input patterns to binary codes so that they represent neighborhood in input space, but they play distinct roles. The hash function role is to map input patterns to binary codes. The loss function role is to assign binary codes to input patterns in order to preserve neighborhood relations, regardless of how easy it is for a mapping to produce such binary codes. By itself, the loss function would produce a nonparametric hash function for the training set with the form of a table of (image,code) pairs. However, the hash function and the loss function cannot act independently, because the objective function depends on both. The optimal combination of hash and loss is difficult to obtain, because of the nonlinear and discrete nature of the objective. Several previous optimization attempts for binary hashing first find codes that optimize the loss, and then fit a hash function to them, thus imposing a strict, suboptimal separation between loss function and hash function. In MAC, both elements are decoupled within each iteration, while still optimizing the correct objective: the step over the hash function does not involve the loss, and the step over the codes does not involve the hash function, but both are iterated. The connection between both steps occurs through the auxiliary coordinates, which are the binary codes themselves. The penalty regularizes the loss so that its optimal codes are progressively closer to what a hash function from the given class (e.g.\ linear) can achieve.

What is the best type of hash function to use? The answer to this is not unique, as it depends on application-specific factors: quality of the codes produced (to retrieve the correct images), time to compute the codes on high-dimensional data (since, after all, the reason to use binary hashing is to speed up retrieval), ease of implementation within a given hardware architecture and software libraries, etc. Our MAC framework facilitates considerably this choice, because training different types of hash functions simply involves reusing an existing classification algorithm within the \h\ step, with no changes to the \Z\ step.

In terms of runtime, the resulting MAC algorithm is not much slower than the two-step approach; it is comparable to iterating the latter a few times. Besides, since all iterations except the first are warm-started, the average cost of one iteration is lower than for the two-step approach.

Finally, note that the method of auxiliary coordinates can be used also to learn an out-of-sample mapping for a \emph{continuous embedding} \citep{CarreirVladym15a}, such as the elastic embedding \citep{Carreir10a} or $t$-SNE \citep{MaatenHinton08a}---rather than to learn hash functions for a discrete embedding, as is our case in binary hashing. The resulting MAC algorithm optimizes over the out-of-sample mapping and the auxiliary coordinates (which are the data points' low-dimensional projections), by alternating two steps. One step optimizes the out-of-sample mapping that projects high-dimensional points to the continuous, latent space, given the auxiliary coordinates \Z. This is a regression problem, while in binary hashing this is a classification problem (per hash function). The other step optimizes the auxiliary coordinates \Z\ given the mapping and is a regularized continuous embedding problem. Both steps can be solved using existing algorithms. In particular, solving the \Z\ step can be done efficiently with large datasets by using $N$-body methods and efficient optimization techniques \citep{Carreir10a,VladymCarreir12a,Maaten13a,Yang_13a,VladymCarreir14a}. In binary hashing the \Z\ step is a combinatorial optimization and, at present, far more challenging to solve. However, with continuous embeddings one must drive the penalty parameter $\mu$ to infinity for the constraints to be satisfied and so the solution follows a continuous path over $\mu \in \bbR$, while with binary hashing the solution follows a discretized, piecewise path which terminates at a finite value of $\mu$.

\paragraph{Binary autoencoder vs affinity-based loss, trained with MAC}

The method of auxiliary coordinates has also been applied in the context of binary hashing to a different objective function, the \emph{binary autoencoder (BA)} \citep{CarreirRaziper15a}:
\begin{equation}
  \label{e:BA}
  E_{\text{BA}}(\h,\f) = \sum^N_{n=1}{ \norm{\x_n - \f(\h(\x_n))}^2 }
\end{equation}
where \h\ is the hash function, or encoder (which outputs binary values), and \f\ is a decoder. (ITQ, \citealp{Gong_13a}, can be seen as a suboptimal way to optimize this.) As with the affinity-based loss function, the MAC algorithm alternates between fitting the hash function (and the decoder) given the codes, and optimizing over the codes. However, \emph{in the binary autoencoder the optimization over the codes decouples over every data point} (since the objective function involves one term per data point). This has an important computational advantage in the \Z\ step: rather than having to solve one large optimization problem $\{\z_1,\dots,\z_N\}$ over $Nb$ binary variables, it has to solve $N$ small optimization problems $\{\z_1\},\dots,\{\z_N\}$ each over $b$ variables, which is much faster and easier to solve (since $b$ is relatively small in practice), and to parallelize. Also, the BA objective does not require any neighborhood information (e.g.\ the affinity between pairs of neighbors) and scales linearly with the dataset. Computing these affinity values, or even finding pairs of neighbors in the first place, is computationally costly. For these reasons, the BA can scale to training on larger datasets than affinity-based loss functions.

The BA objective function does have the disadvantage of being less directly related to the goals that are desirable from an information retrieval point of view, such as precision and recall. Neighborhood relations are only indirectly preserved by autoencoders \citep{CarreirRaziper15a}, whose direct aim is to reconstruct its inputs and thus to learn the data manifold (imperfectly, because of the binary projection layer). Affinity-based loss functions of the form~\eqref{e:objfcn} allow the user to specify more complex neighborhood relations, for example based on class labels, which may significantly differ from the actual distances in image feature space. Still, finding more efficient and scalable optimization methods for binary embeddings (in the \Z\ step of the MAC algorithm), that are able to handle larger numbers of training and neighbor points, would improve the quality of the loss function. This is an important topic of future research.

\section{Conclusion}
\label{s:concl}

We have proposed a general framework for optimizing binary hashing using affinity-based loss functions. It improves over previous, two-step approaches based on learning binary codes first and then learning the hash function. Instead, it optimizes jointly over the binary codes and the hash function in alternation, so that the binary codes eventually match the hash function, resulting in a better local optimum of the affinity-based loss. This was possible by introducing auxiliary variables that conditionally decouple the codes from the hash function, and gradually enforcing the corresponding constraints. Our framework makes it easy to design an optimization algorithm for a new choice of loss function or hash function: one simply reuses existing software that optimizes each in isolation. The resulting algorithm is not much slower than the suboptimal two-step approach---it is comparable to iterating the latter a few times---and well worth the improvement in precision/recall.

The step over the hash function is essentially a solved problem if using a classifier, since this can be learned in an accurate and scalable way using machine learning techniques. The most difficult and time-consuming part in our approach is the optimization over the binary codes, which is NP-complete and involves many binary variables and terms in the objective. Although some techniques exist \citep{Lin_13a,Lin_14b} that produce practical results, designing algorithms that reliably find good local optima and scale to large training sets is an important topic of future research.

Another direction for future work involves learning more sophisticated hash functions that go beyond mapping image features onto output binary codes using simple classifiers such as SVMs. This is possible because the optimization over the hash function parameters is confined to the \h\ step and takes the form of a supervised classification problem, so we can apply an array of techniques from machine learning and computer vision. For example, it may be possible to learn image features that work better with hashing than standard features such as SIFT, or to learn transformations of the input to which the binary codes should be invariant, such as translation, rotation or alignment.

\appendix

\section{Additional experiments}
\label{s:expts-additional}

\subsection{Unsupervised dataset}

Although affinity-based hashing is intended to work with supervised datasets, it can also be used with unsupervised ones, and our MAC approach applies just as well. We use the SIFT1M dataset \citep{Jegou_11a}, which contains $N = 1\,000\,000$ training high-resolution color images and $10\,000$ test images, each represented by $D=128$ SIFT features. The experiments and conclusions are generally the same as with supervised datasets, with small differences in the settings of the experiments. In order to construct an affinity-based objective function, we define neighbors as follows. For each point in the training set we use the $\kappa_+ = 100$ nearest neighbors as positive (similar) neighbors, and $\kappa_- = 500$ points chosen randomly among the remaining points as negative (dissimilar) neighbors. We report precision and precision/recall for the test set queries using as ground truth (set of true neighbors in original space) the $K$ nearest neighbors in unsupervised datasets, and all the training points with the same label in supervised datasets.

\begin{figure}[p]
  \centering
  \psfrag{rerror}[][t]{loss function \calL}
  \psfrag{iteration}[t][]{iterations}
  \psfrag{K}[t][]{$k$}
  \psfrag{precision}[][t]{precision}
  \psfrag{recall}[][b]{recall}
  \begin{tabular}{@{}l@{}c@{}c@{}c@{}c@{}}
    & $b=8$ & $b=16$ & $b=24$ & $b=32$\\
    \hspace{2ex}\rotatebox{90}{\hspace{4ex}loss function \calL} &
    \includegraphics[width=0.240\linewidth]{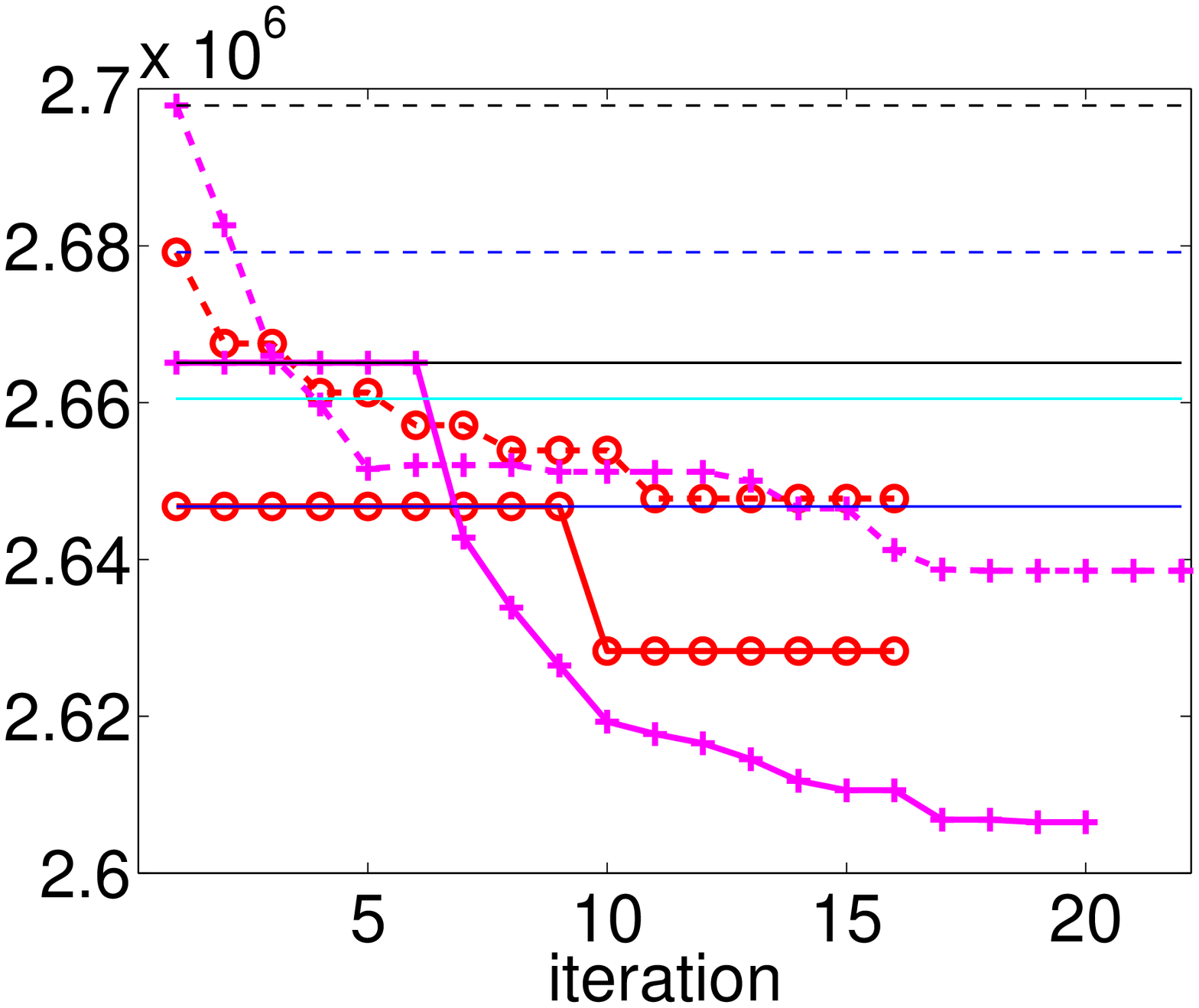} &
    \includegraphics[width=0.240\linewidth]{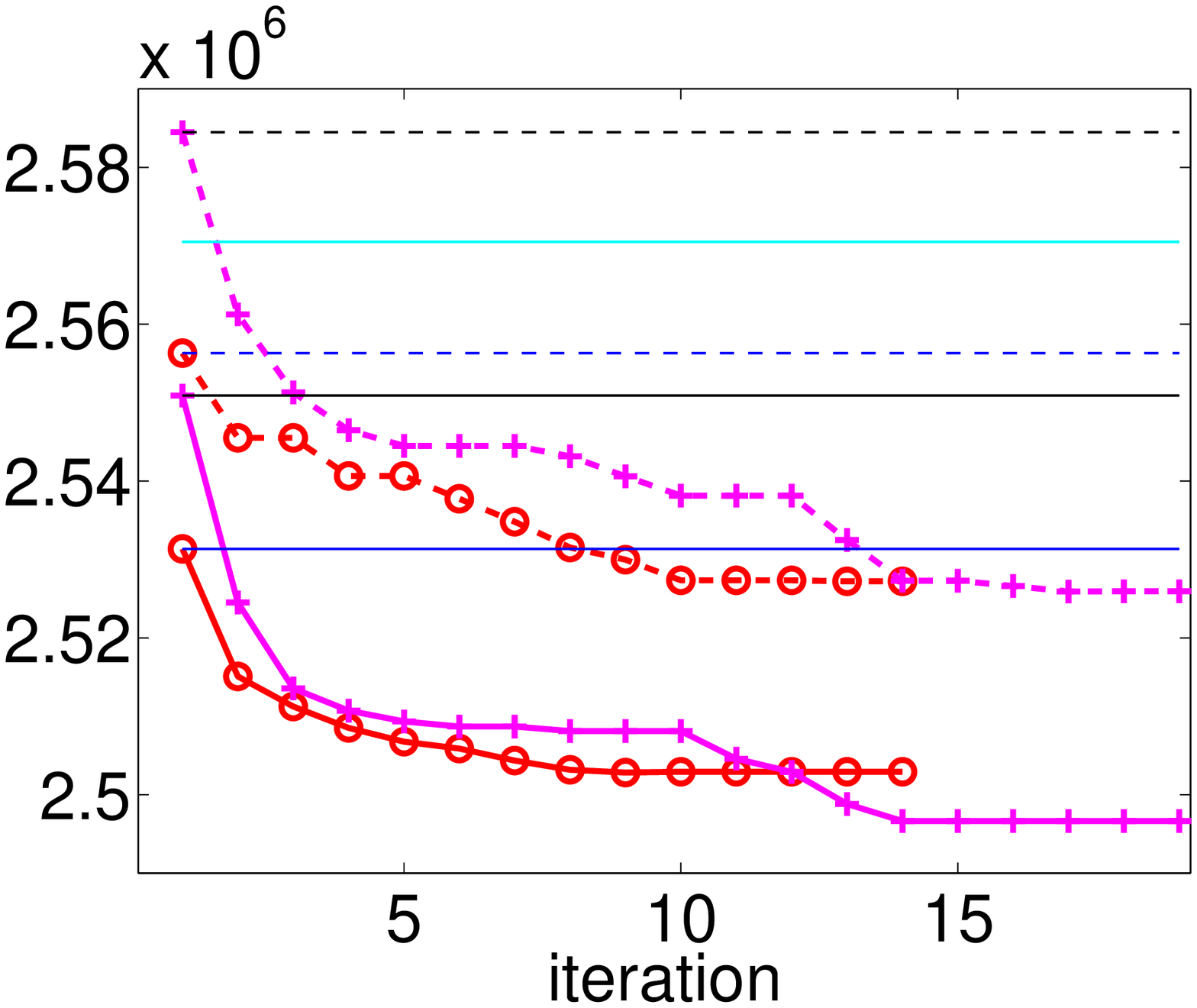} &
    \includegraphics[width=0.240\linewidth]{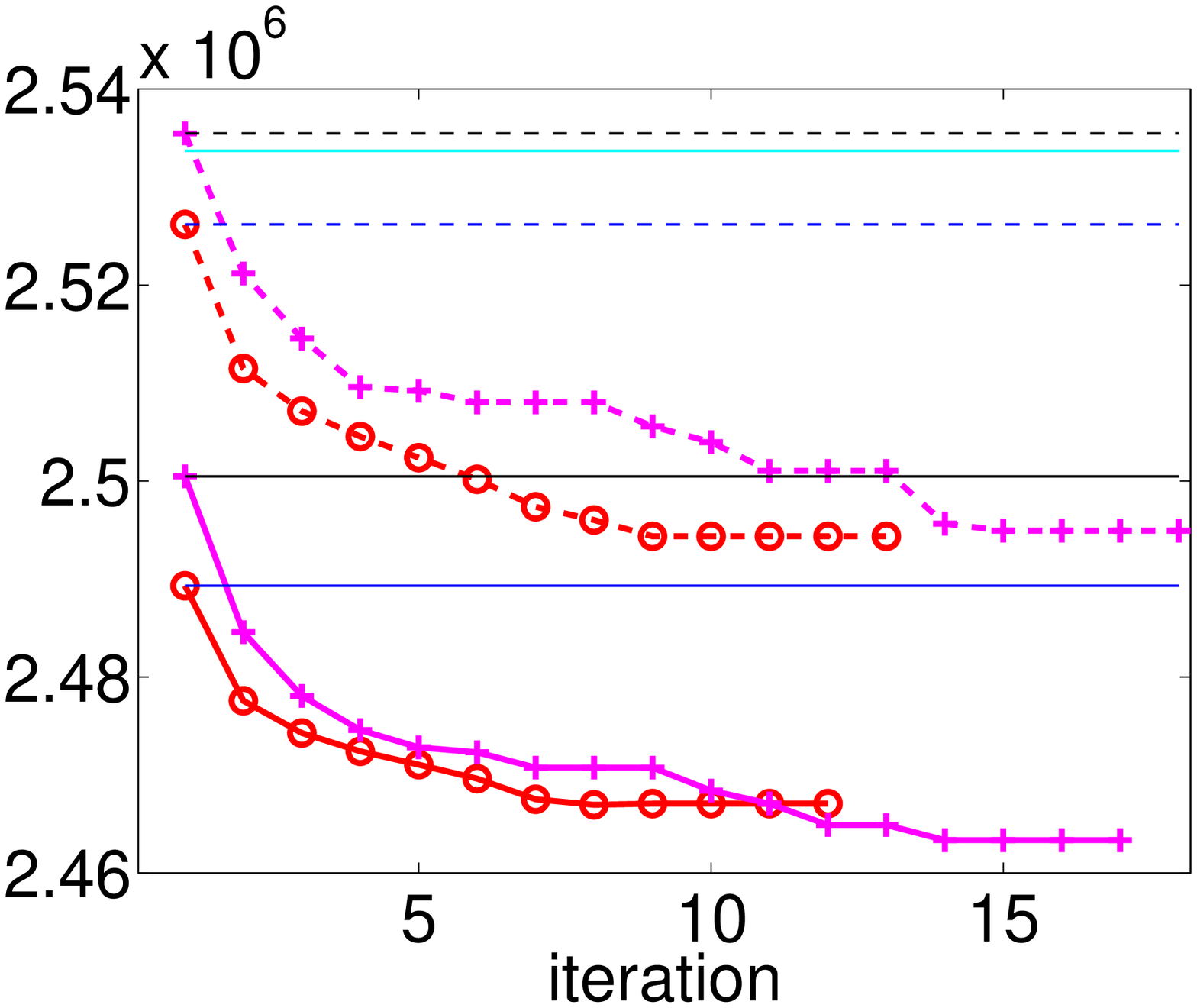} &
    \includegraphics[width=0.240\linewidth]{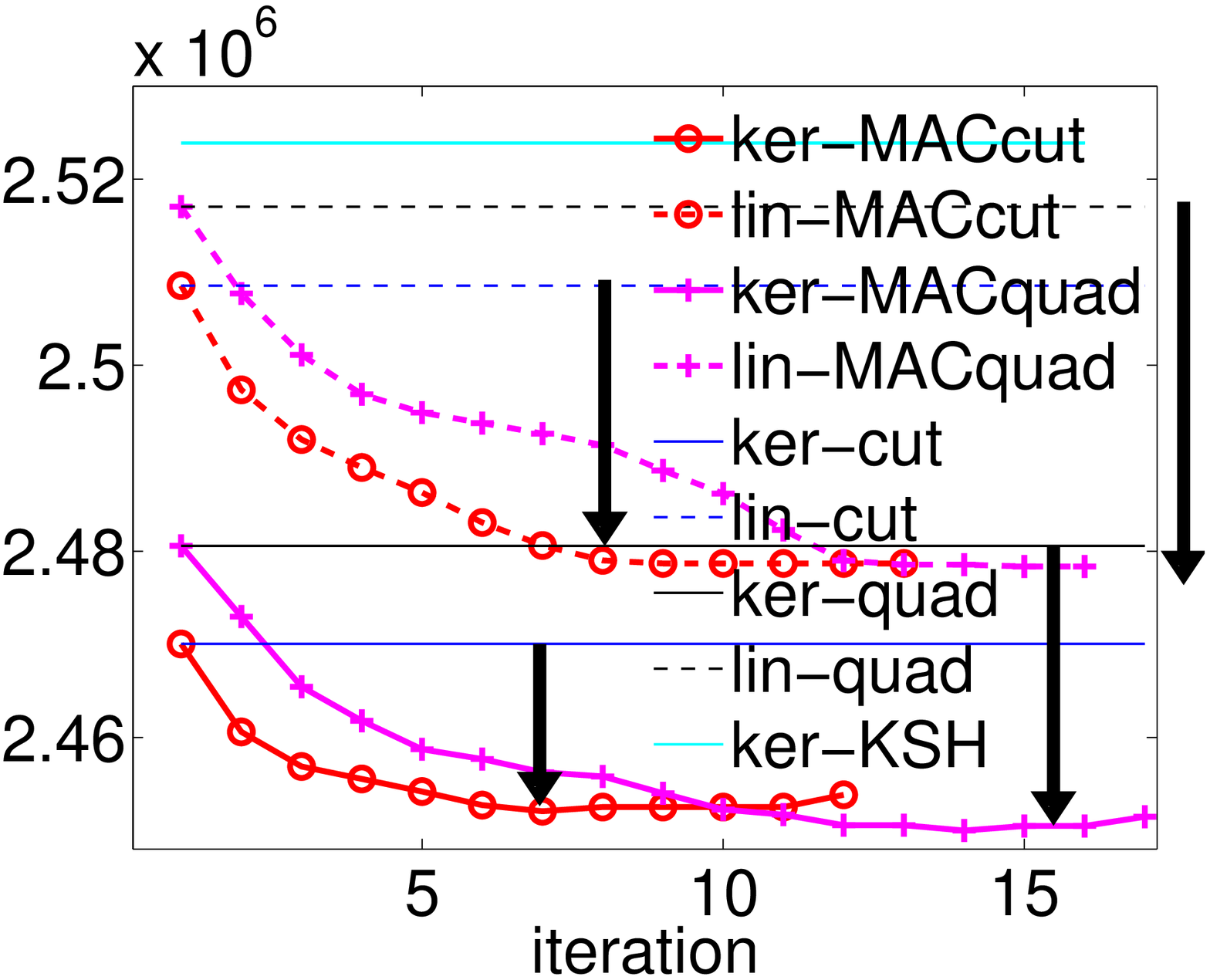} \\
    \hspace{2ex}\rotatebox{90}{\hspace{7ex}precision} &
    \includegraphics[width=0.240\linewidth]{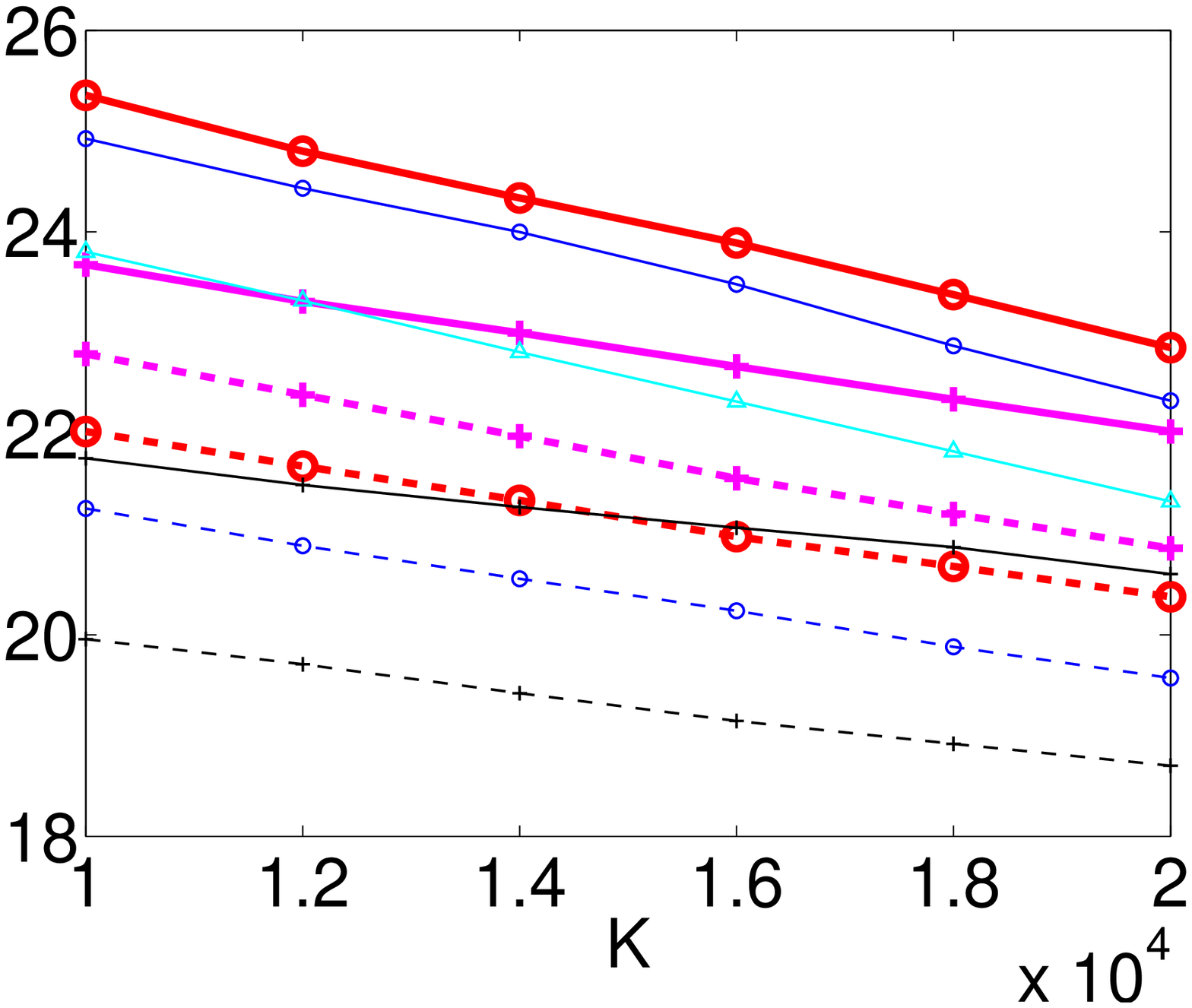} &
    \includegraphics[width=0.240\linewidth]{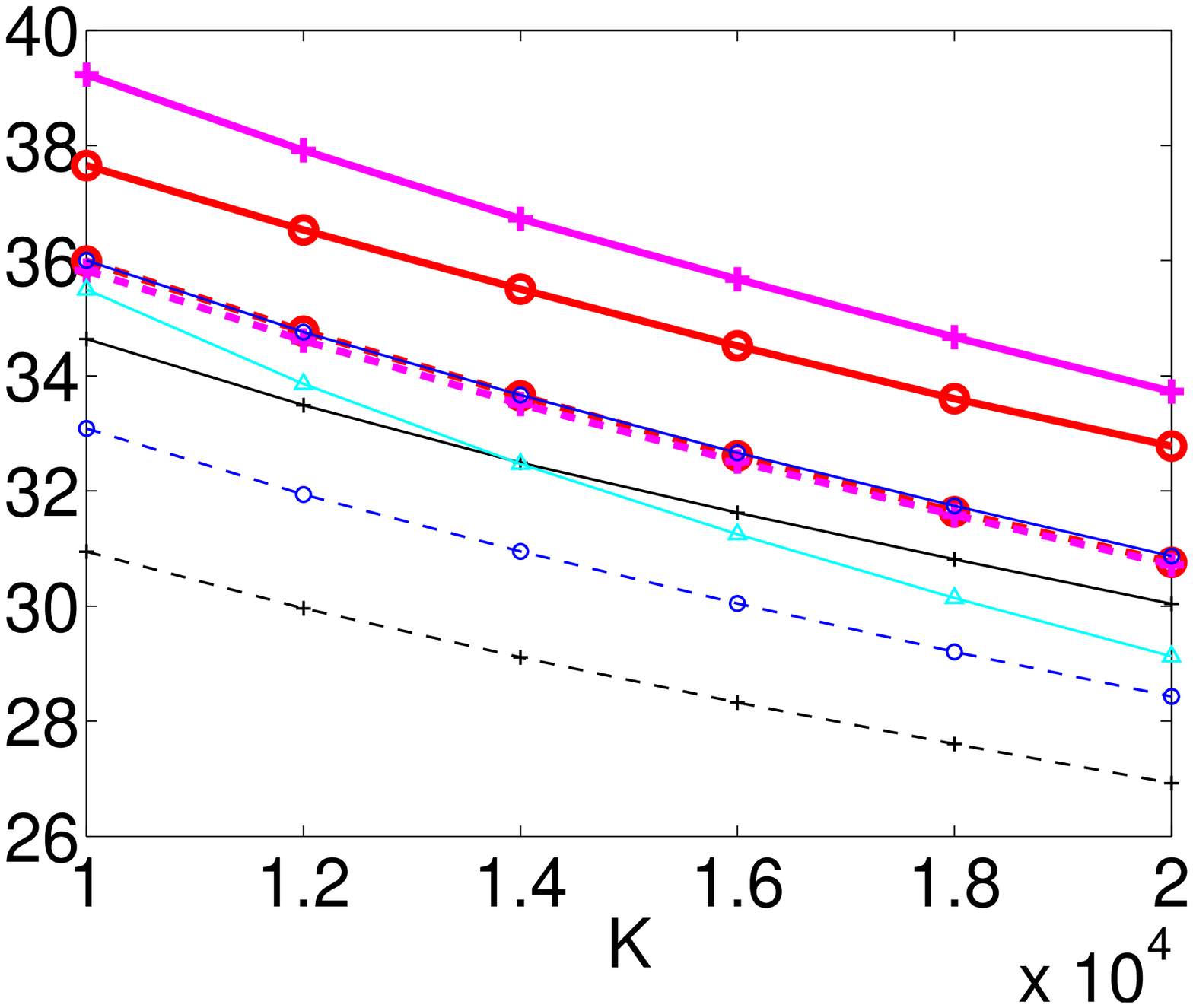} &
    \includegraphics[width=0.240\linewidth]{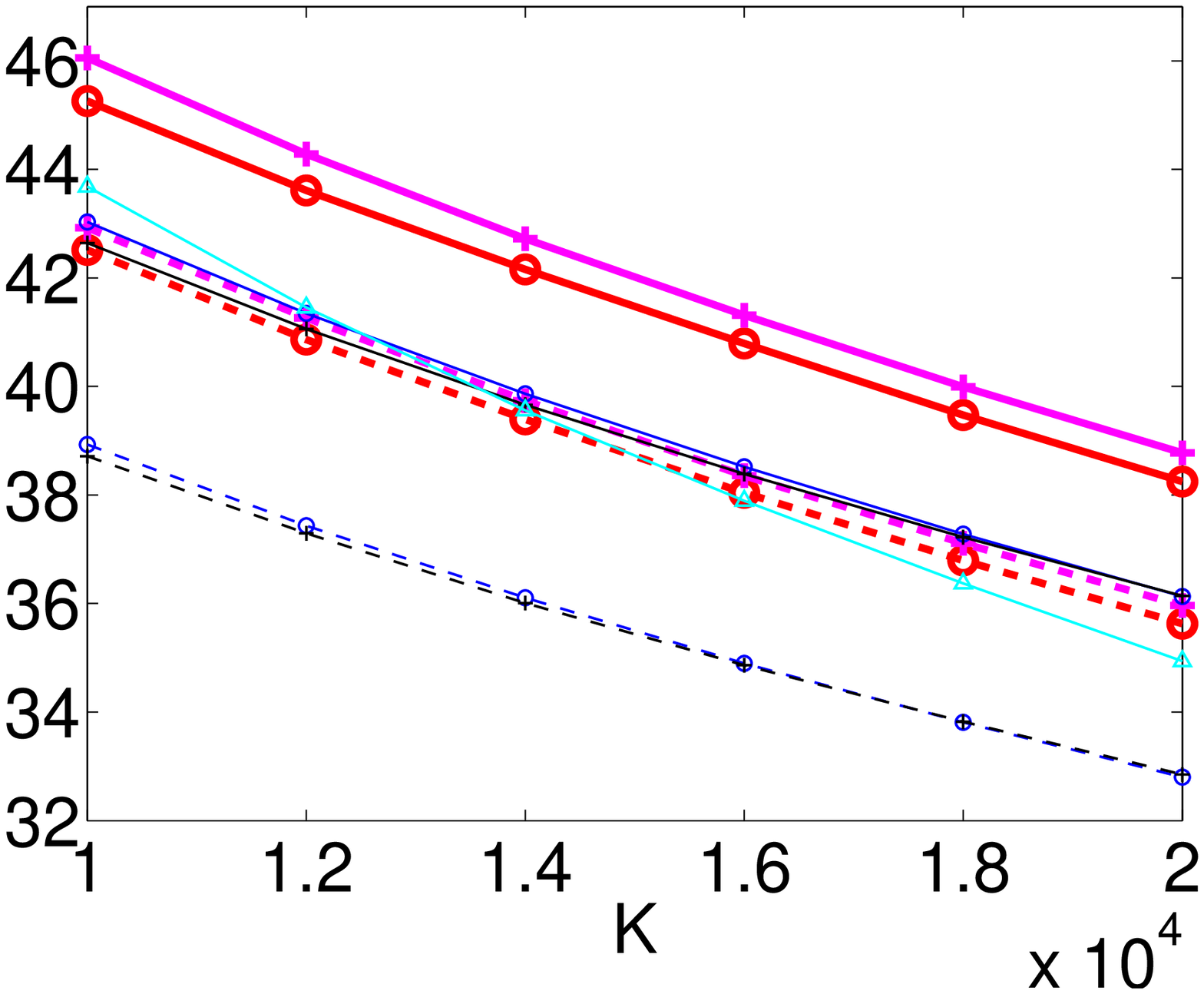} &
    \includegraphics[width=0.240\linewidth]{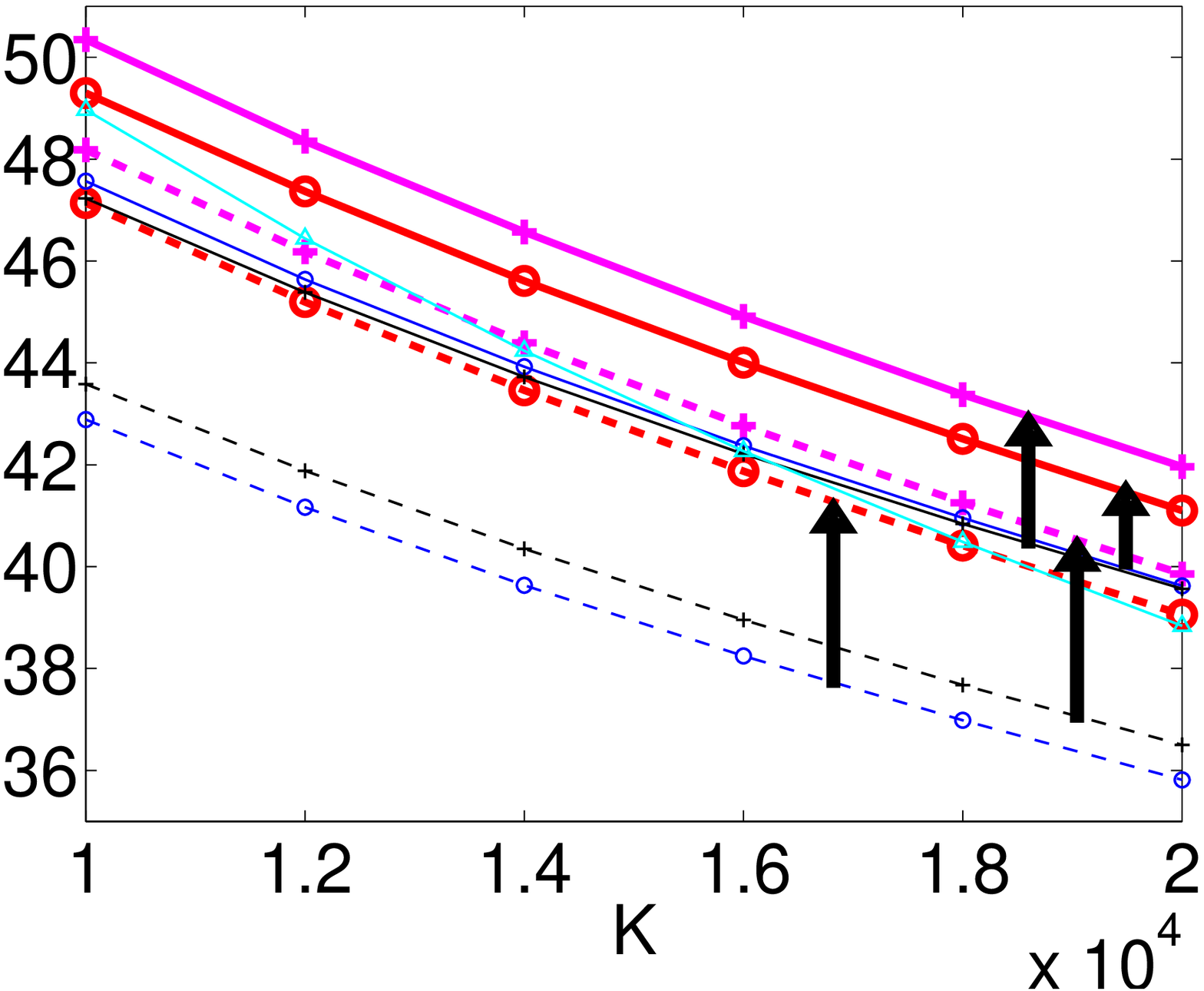} \\
    \hspace{2ex}\rotatebox{90}{\raisebox{3ex}[0pt][0pt]{\makebox[0pt][l]{\hspace{13ex}KSH, $\kappa_+ = 100$, $\kappa_- = 500$, $K = 20\,000$}}\hspace{7ex}precision} &
    \includegraphics[width=0.240\linewidth]{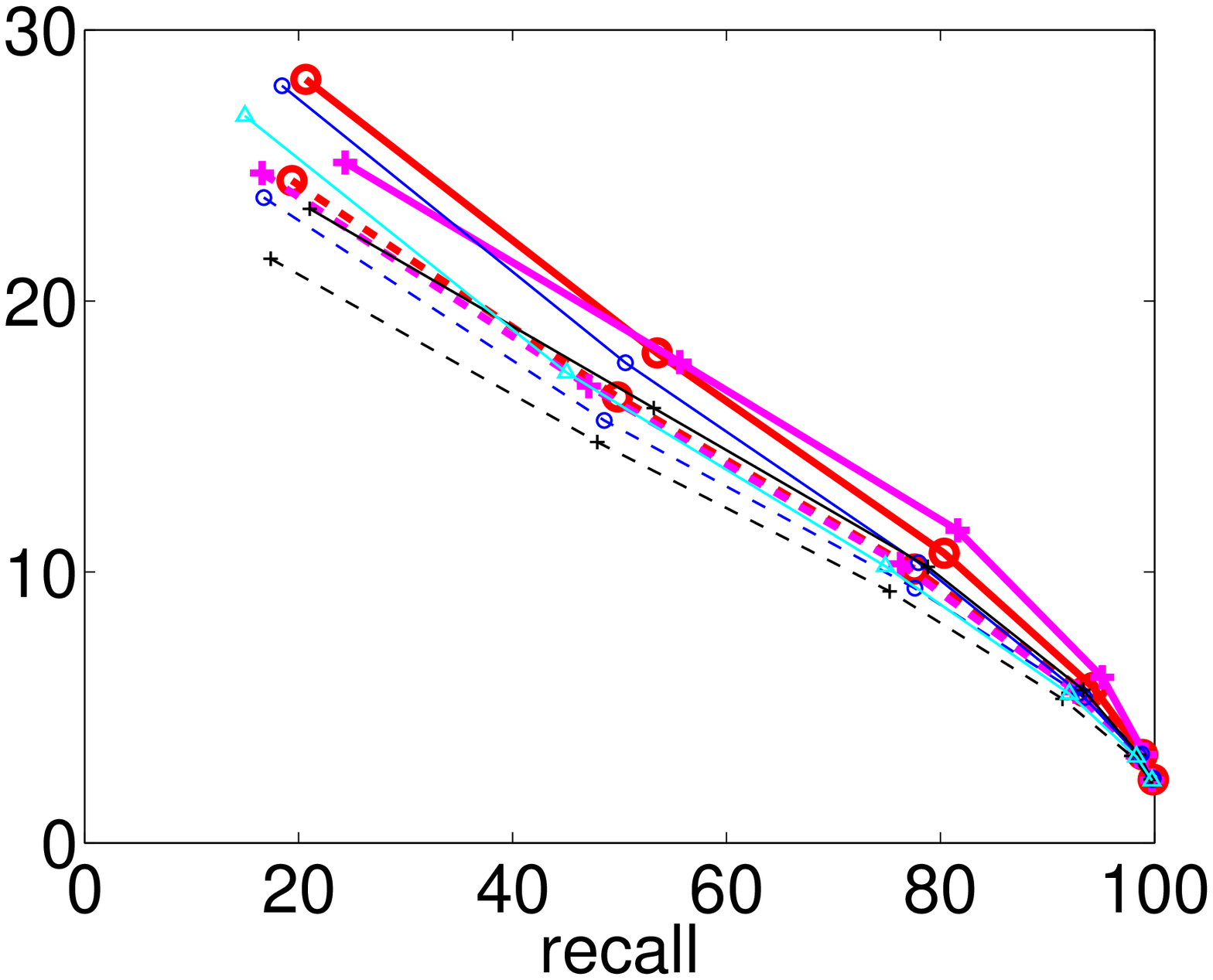} &
    \includegraphics[width=0.240\linewidth]{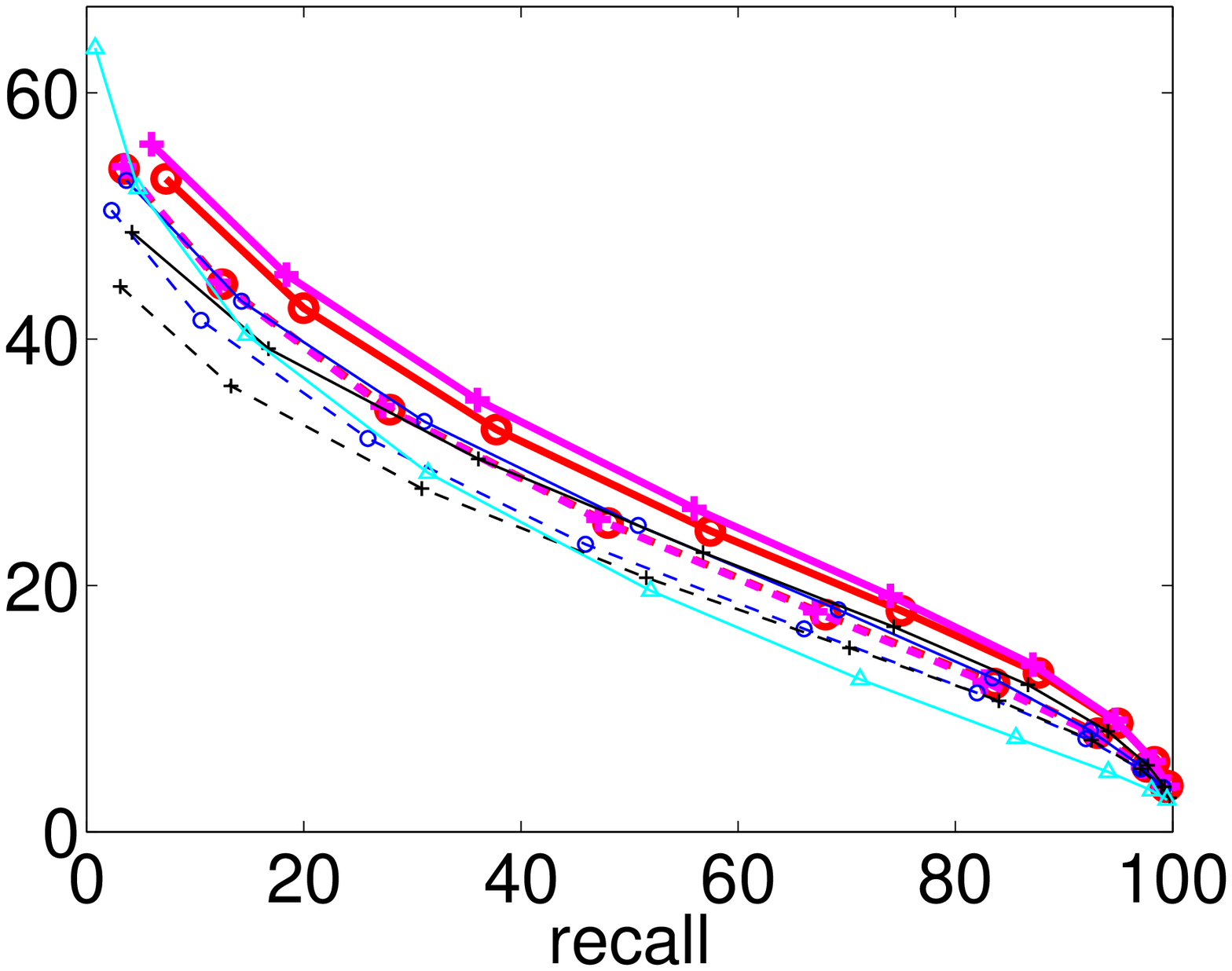} &
    \includegraphics[width=0.240\linewidth]{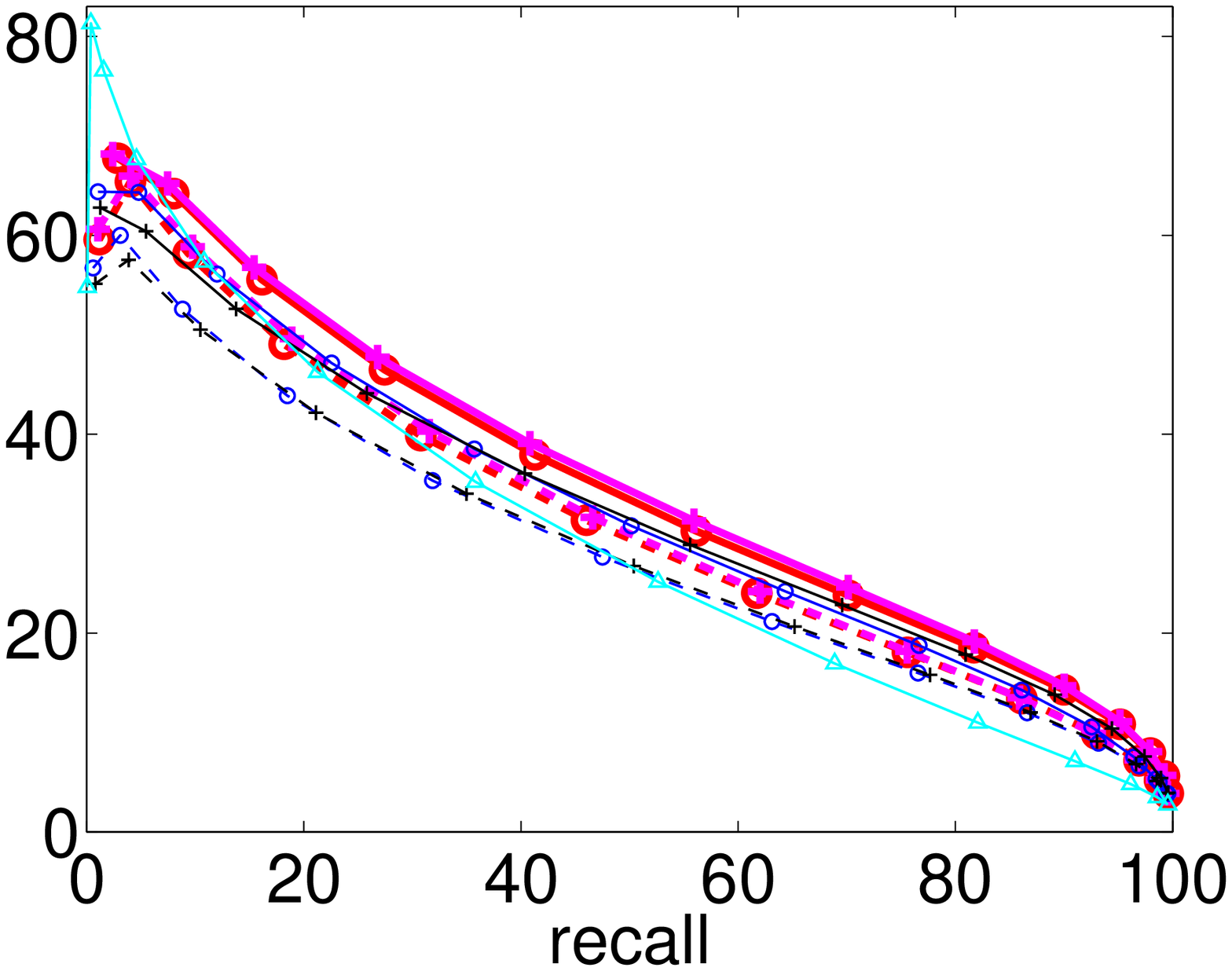} &
    \includegraphics[width=0.240\linewidth]{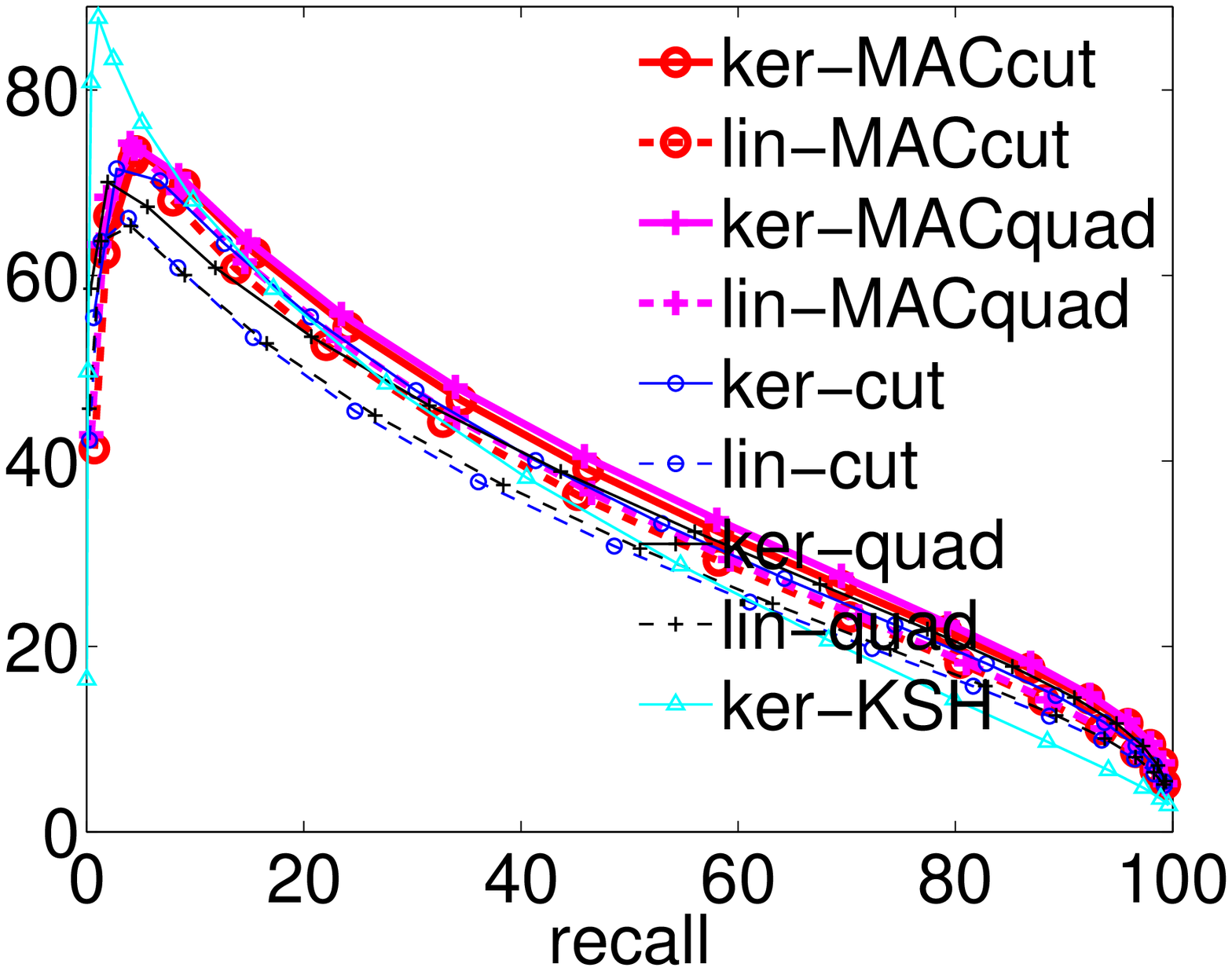} \\
    \hline
    \hspace{2ex}\rotatebox{90}{\hspace{4ex}loss function \calL} &
    \includegraphics[width=0.240\linewidth]{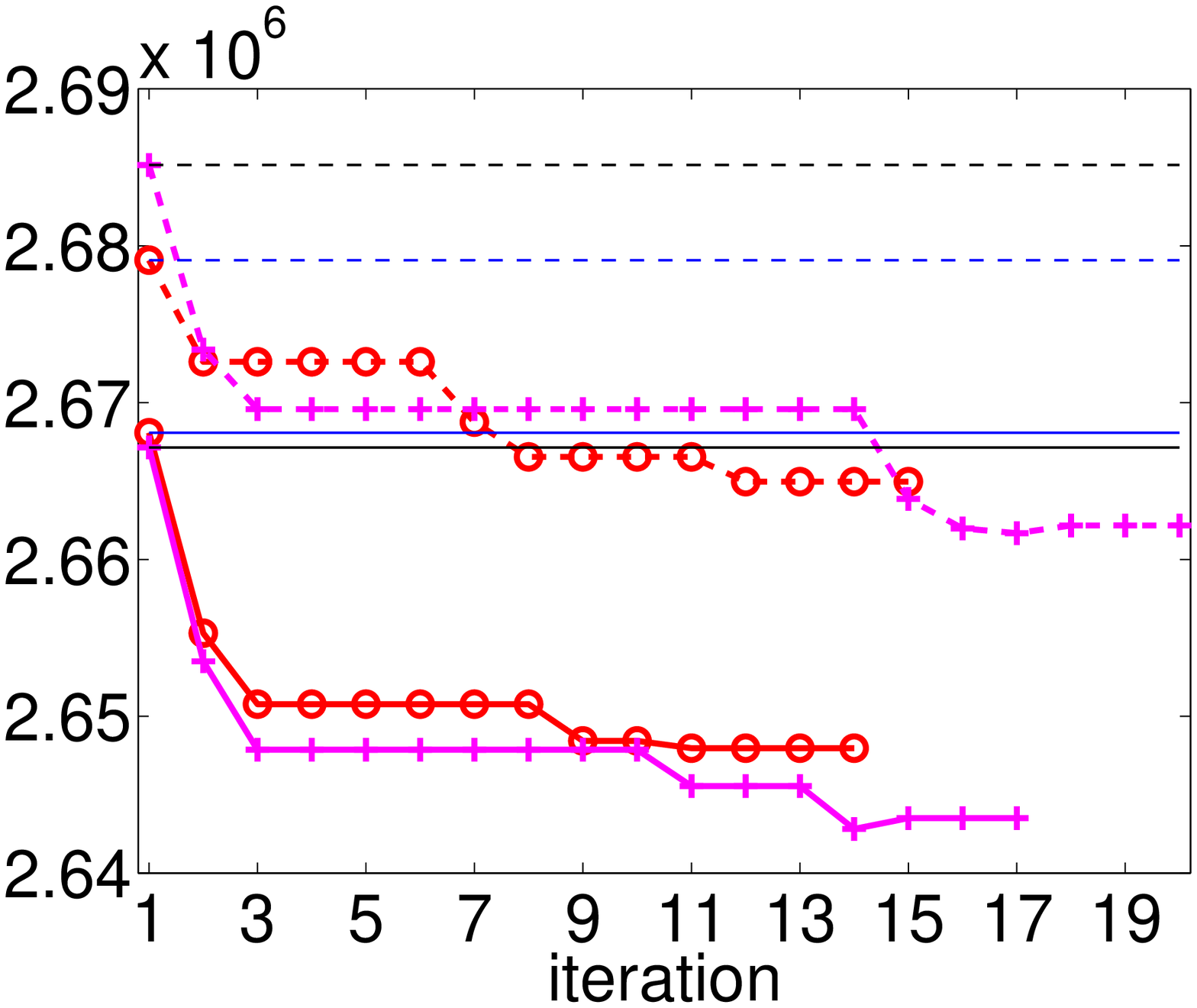} &
    \includegraphics[width=0.240\linewidth]{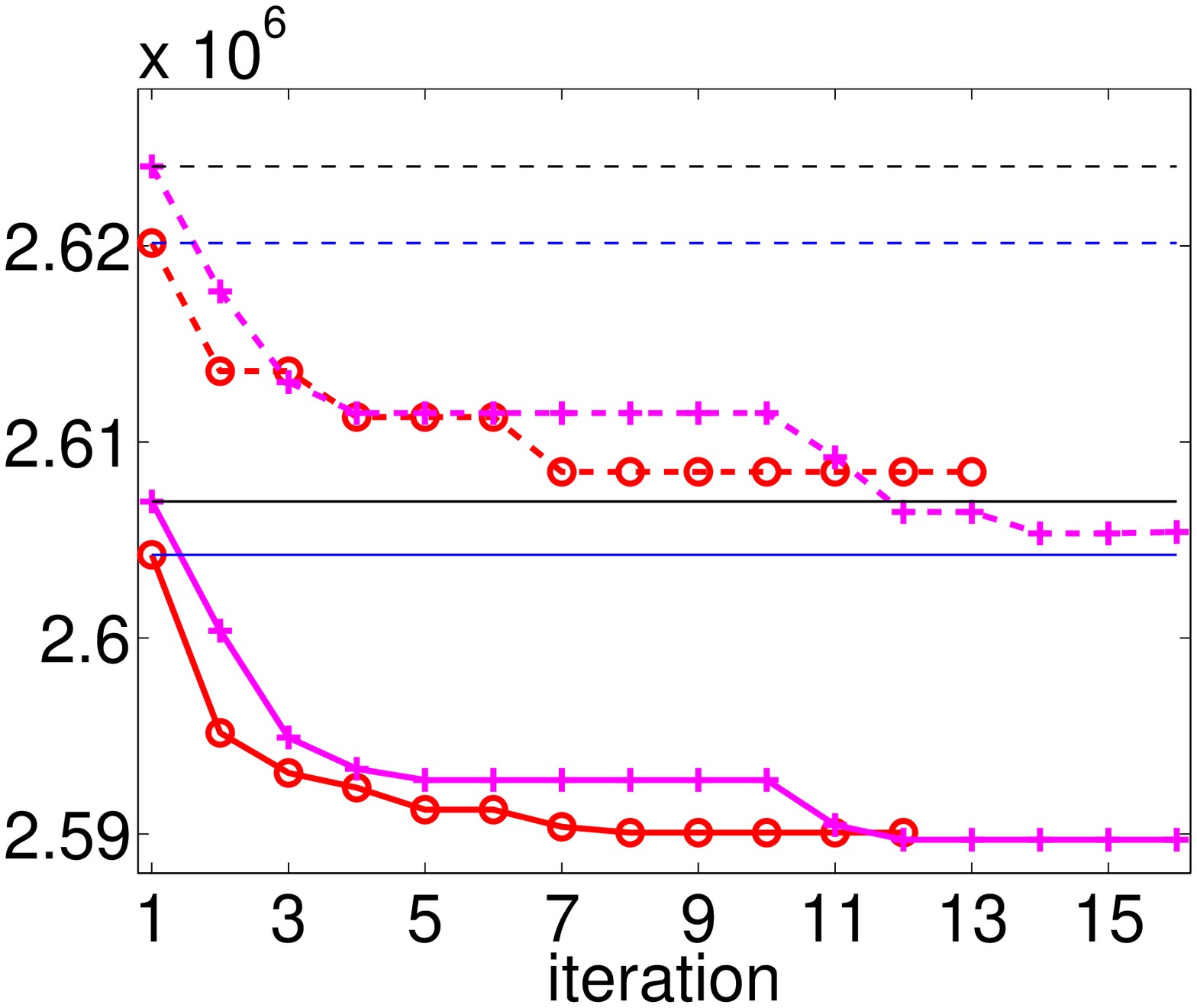} &
    \includegraphics[width=0.240\linewidth]{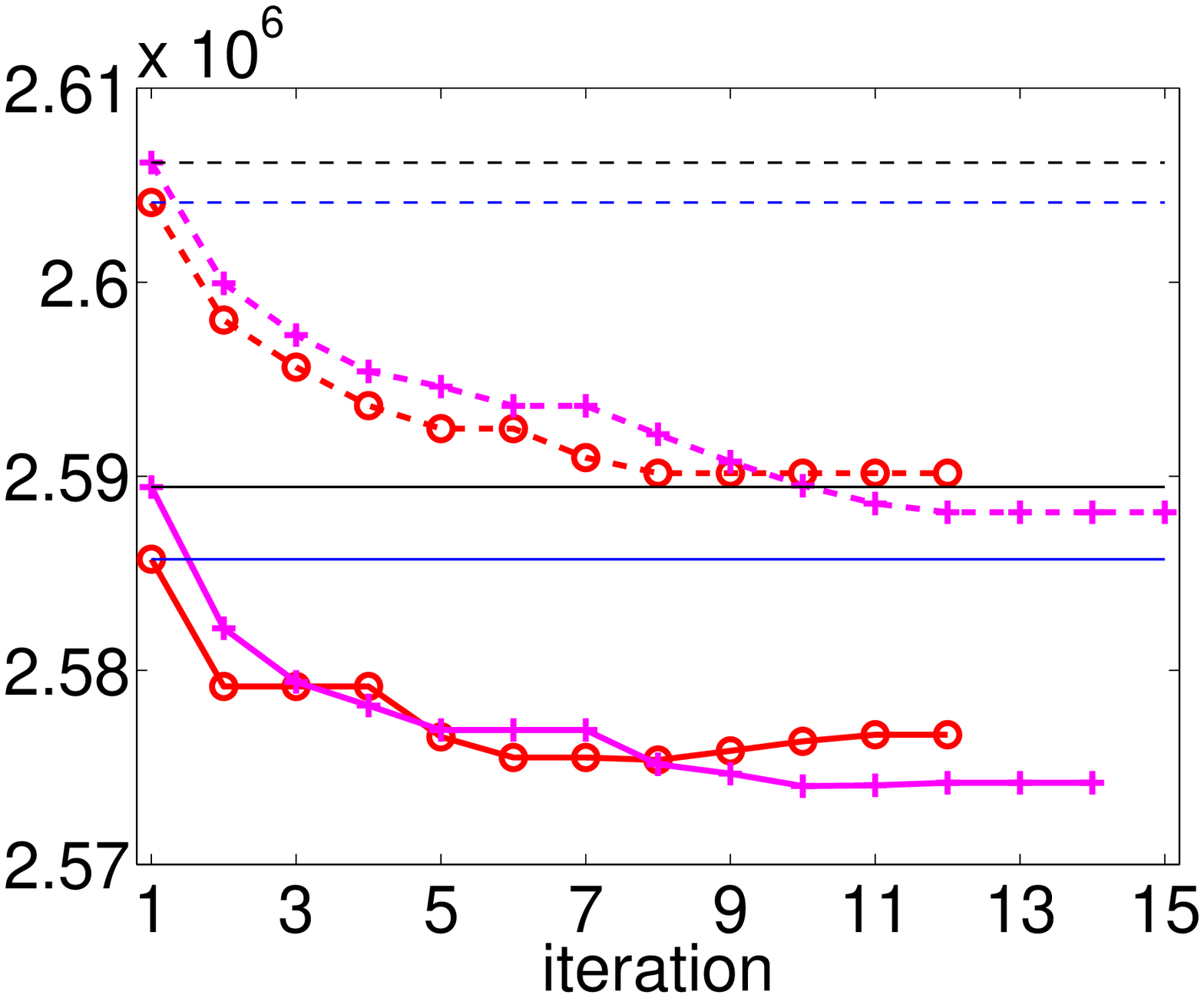} &
    \includegraphics[width=0.240\linewidth]{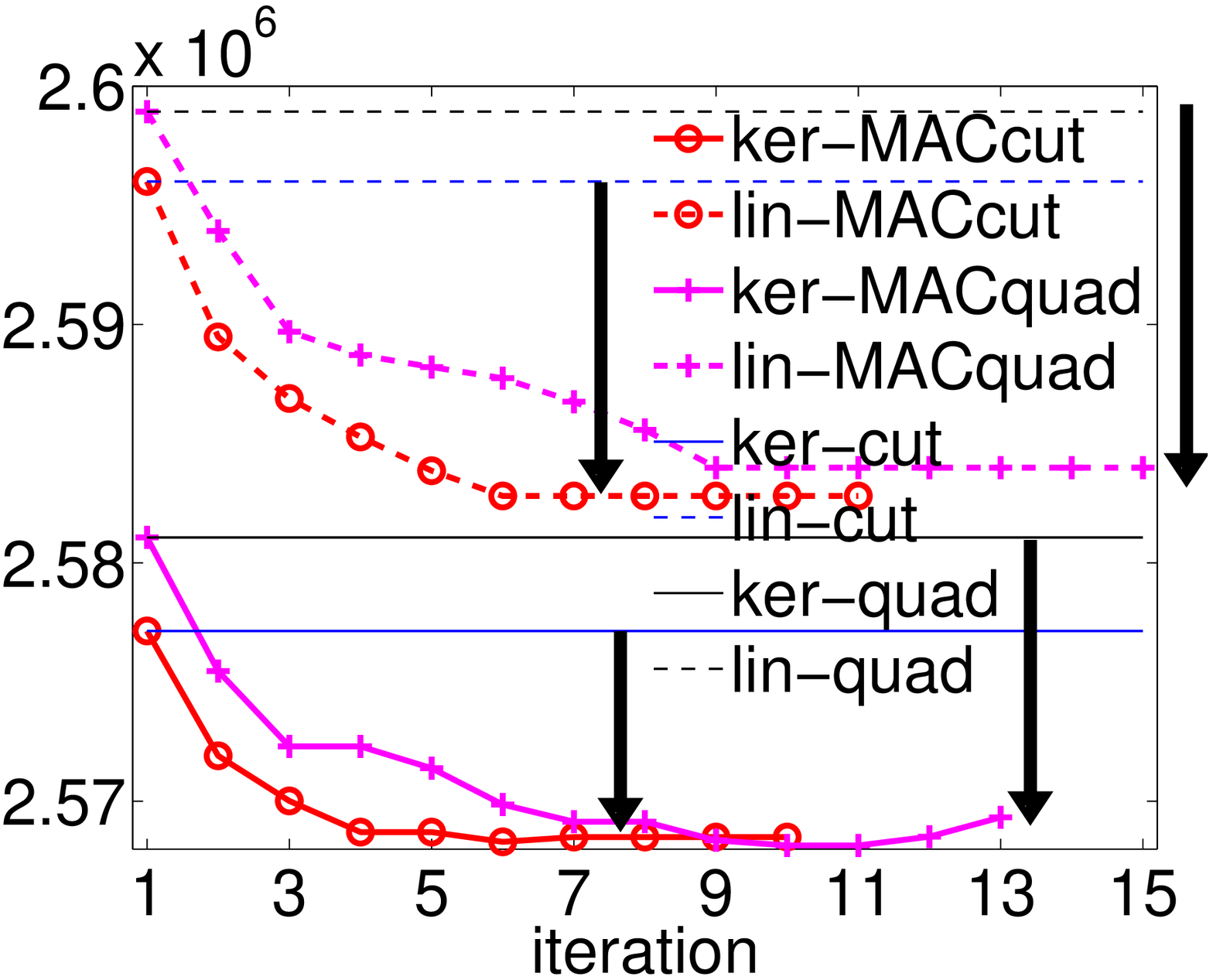} \\
    \hspace{2ex}\rotatebox{90}{\hspace{7ex}precision} &
    \includegraphics[width=0.240\linewidth]{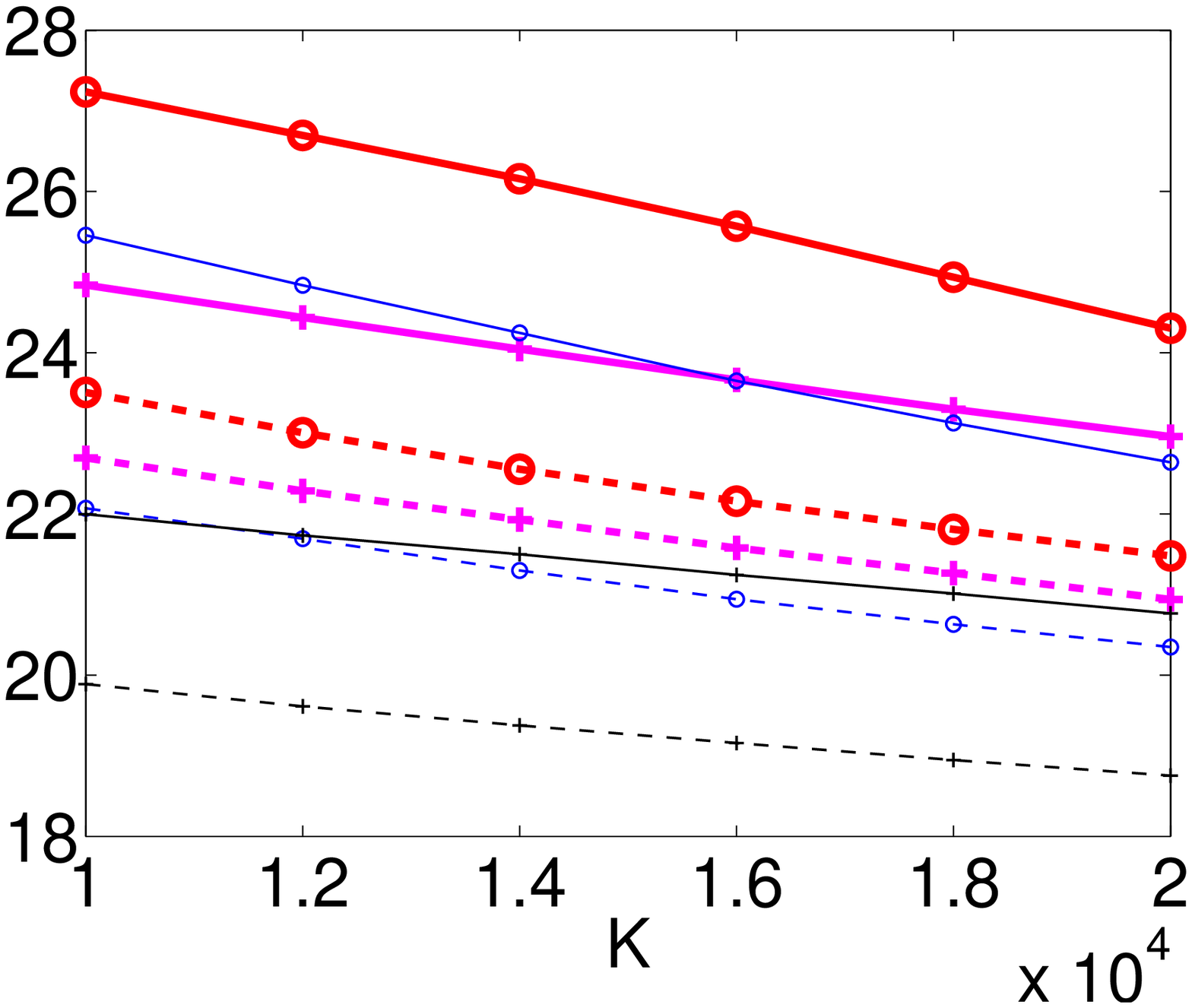} &
    \includegraphics[width=0.240\linewidth]{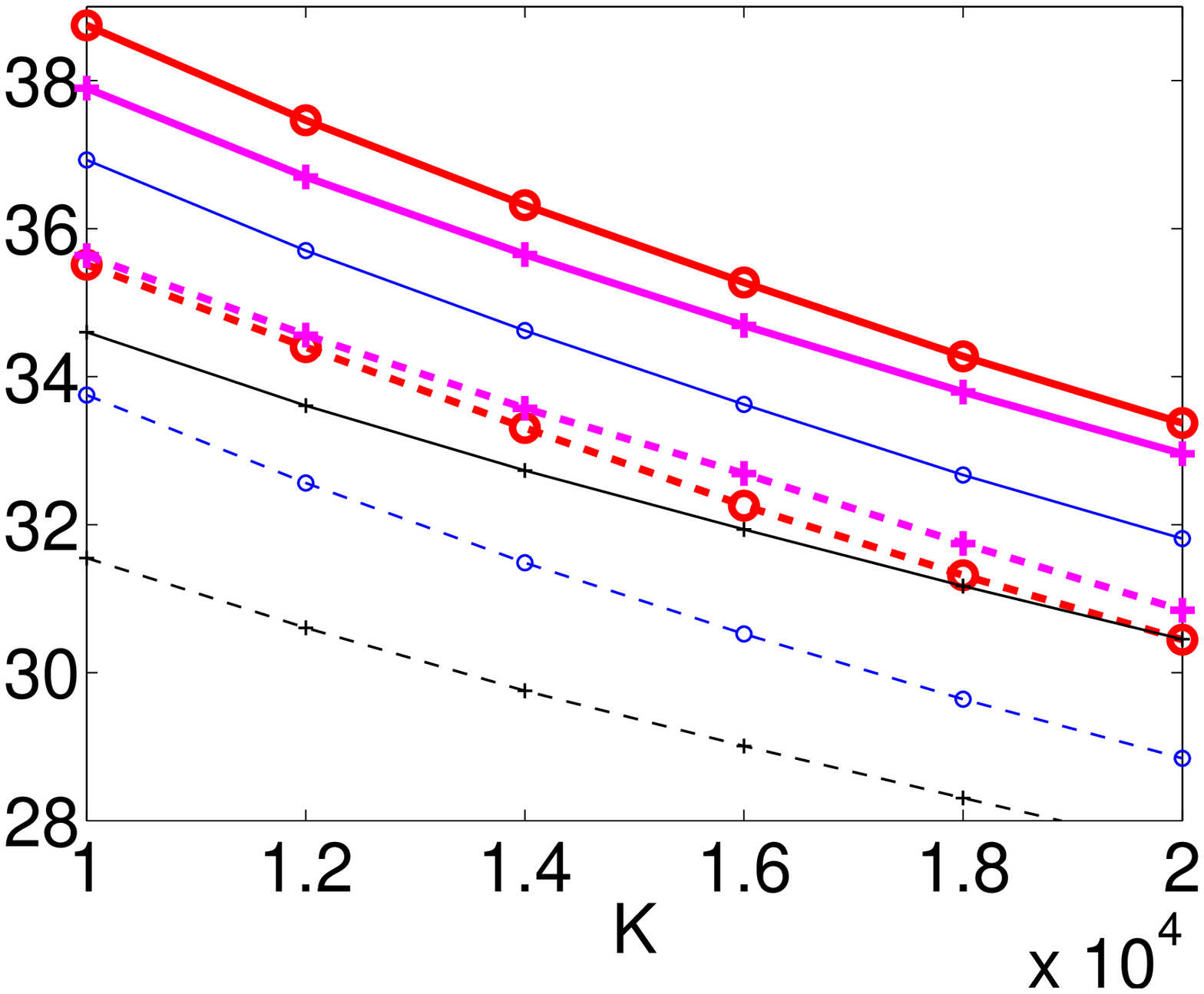} &
    \includegraphics[width=0.240\linewidth]{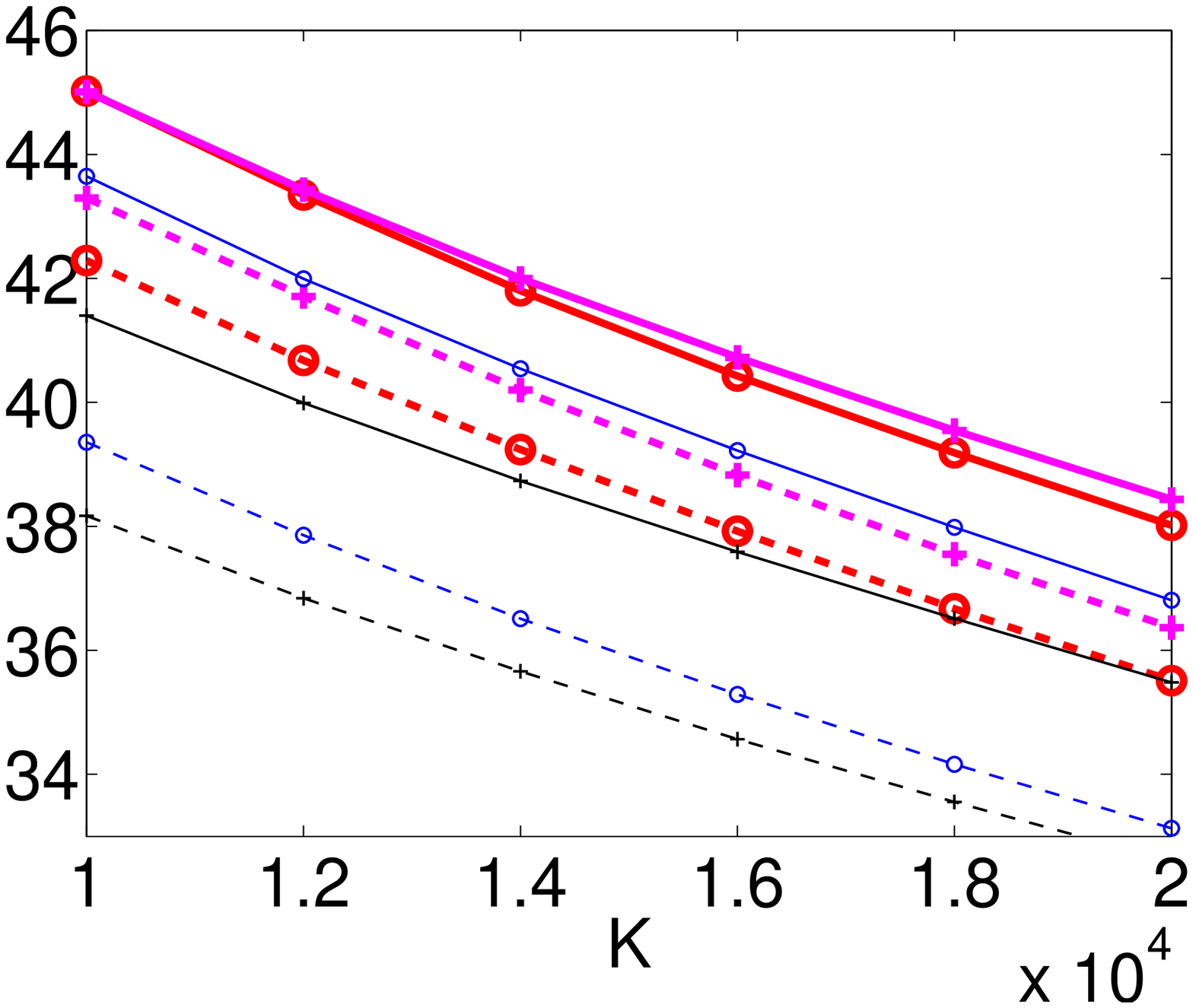} &
    \includegraphics[width=0.240\linewidth]{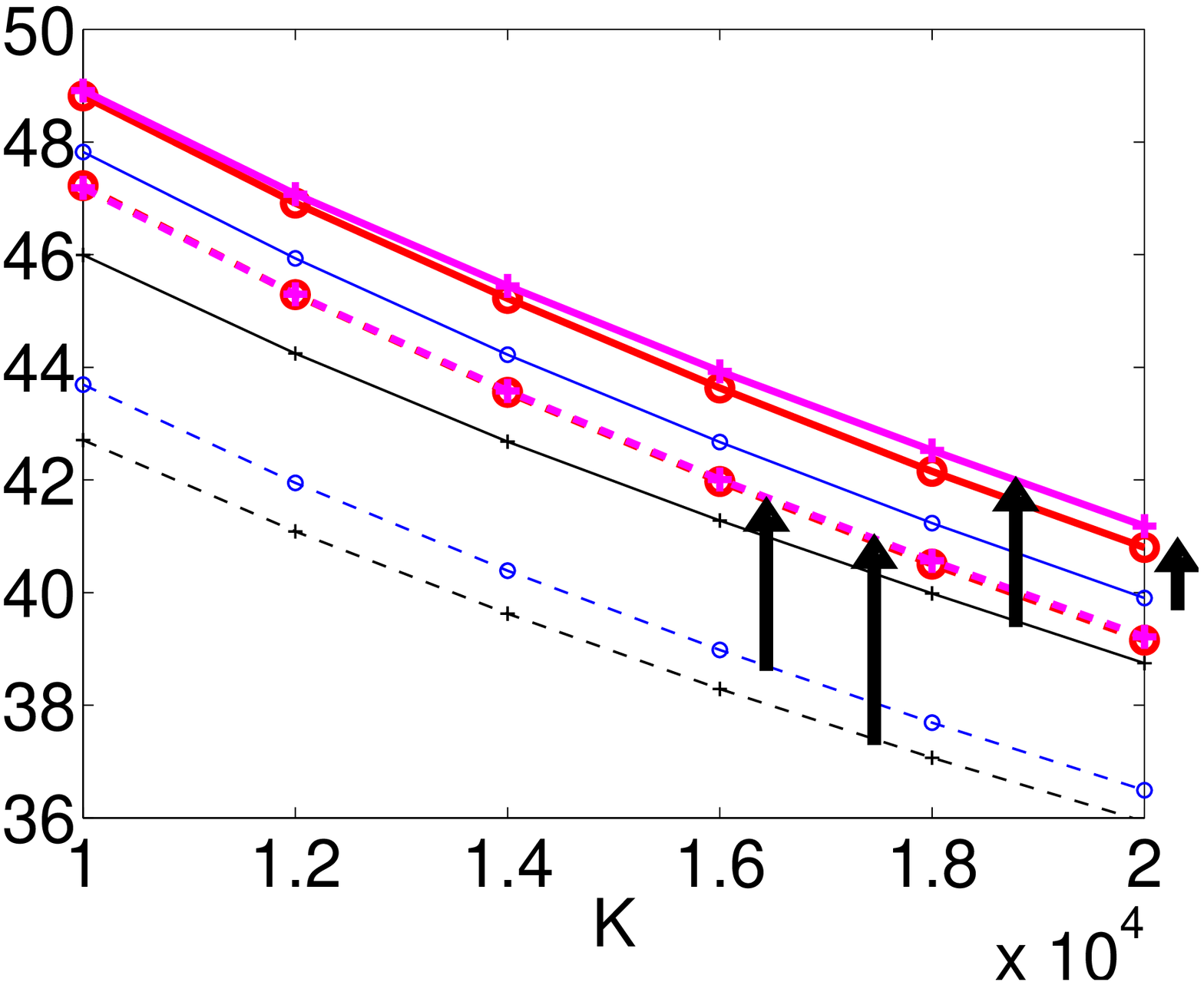} \\
    \hspace{2ex}\rotatebox{90}{\raisebox{3ex}[0pt][0pt]{\makebox[0pt][l]{\hspace{13ex}eSPLH, $\kappa_+ = 100$, $\kappa_- = 500$, $K = 20\,000$}}\hspace{7ex}precision} &
    \includegraphics[width=0.240\linewidth]{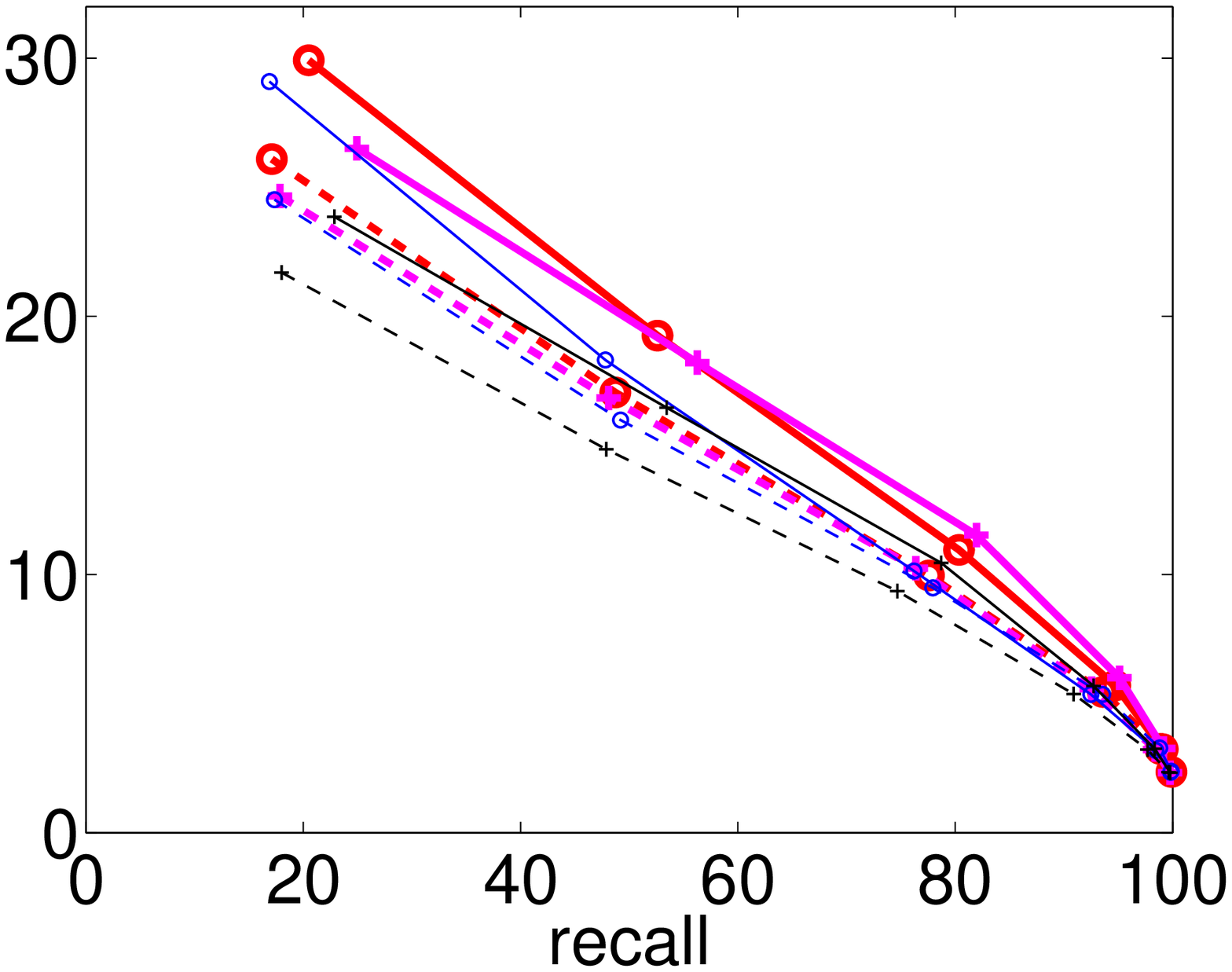} &
    \includegraphics[width=0.240\linewidth]{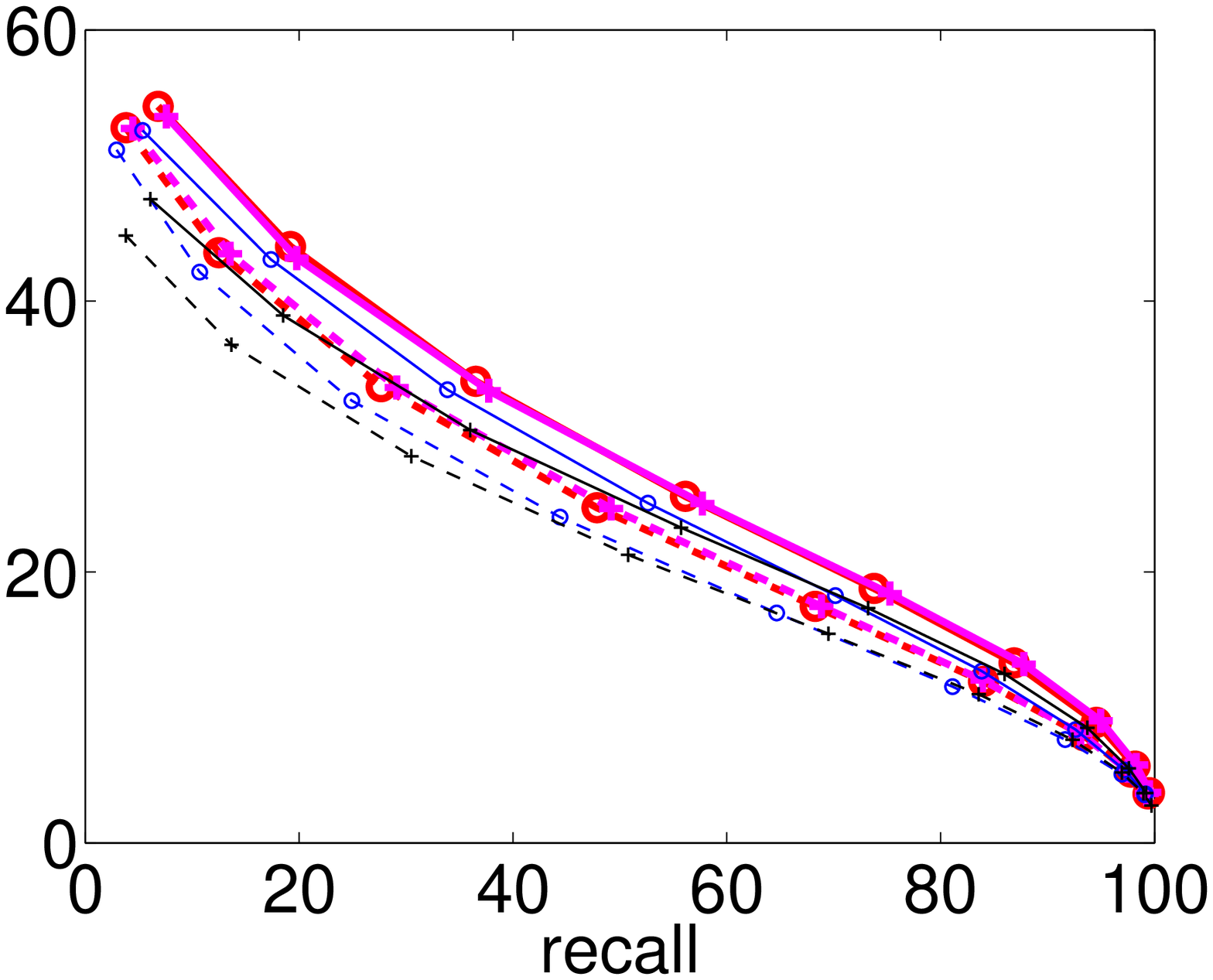} &
    \includegraphics[width=0.240\linewidth]{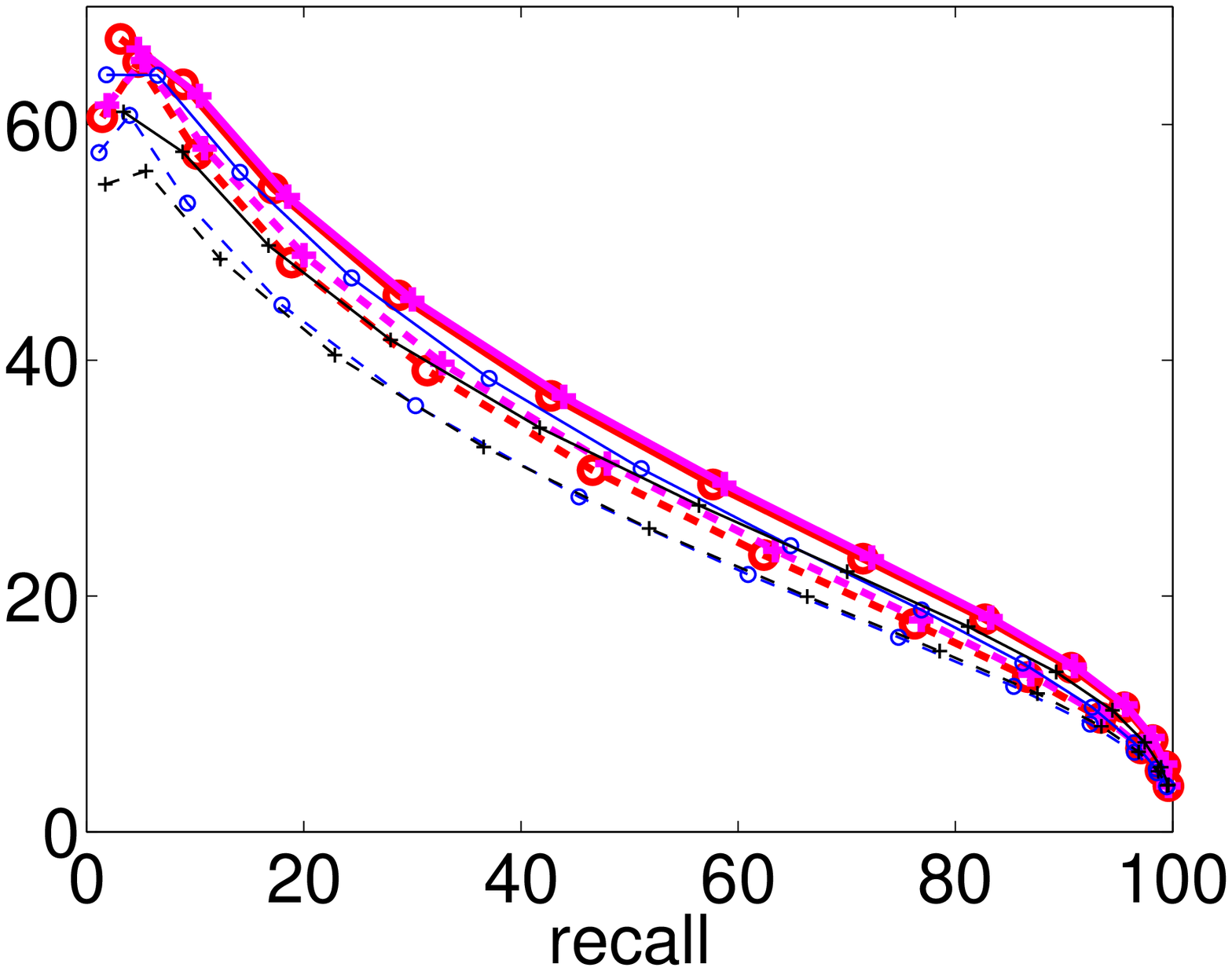} &
    \includegraphics[width=0.240\linewidth]{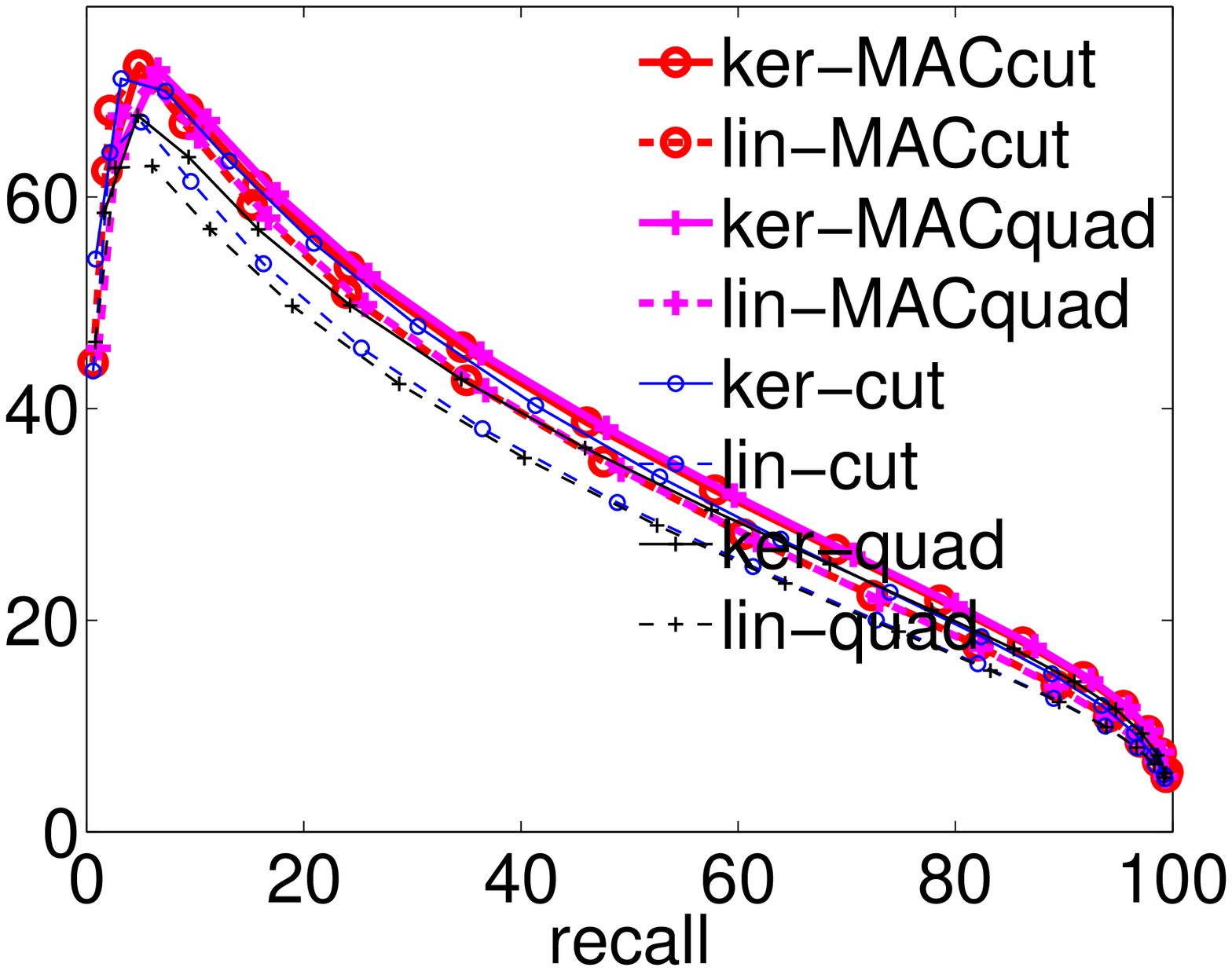}
  \end{tabular}
  \caption{Like fig.~\ref{f:sup-nestederr} but on SIFT1M dataset, for the KSH (top panel) and eSPLH (bottom panel) loss functions. The rows show the value of the loss function \calL, the precision (for a number of retrieved points $k$) and the precision/recall (at different Hamming distances), using $b=8$ to $32$ bits.}
  \label{f:unsup-nestederr}
\end{figure}

\begin{figure}[h]
  \centering
  \psfrag{rerror}[][t]{loss function \calL}
  \psfrag{iteration}[t][]{iterations}
  \psfrag{K}[t][]{$k$}
  \psfrag{precision}[][t]{precision}
  \psfrag{recall}[][b]{recall}
  \begin{tabular}{@{}l@{}c@{}c@{}c@{}c@{}}
    & $b=8$ & $b=16$ & $b=24$ & $b=32$\\
    \hspace{2ex}\rotatebox{90}{\hspace{4ex}loss function \calL} &
    \includegraphics[width=0.240\linewidth]{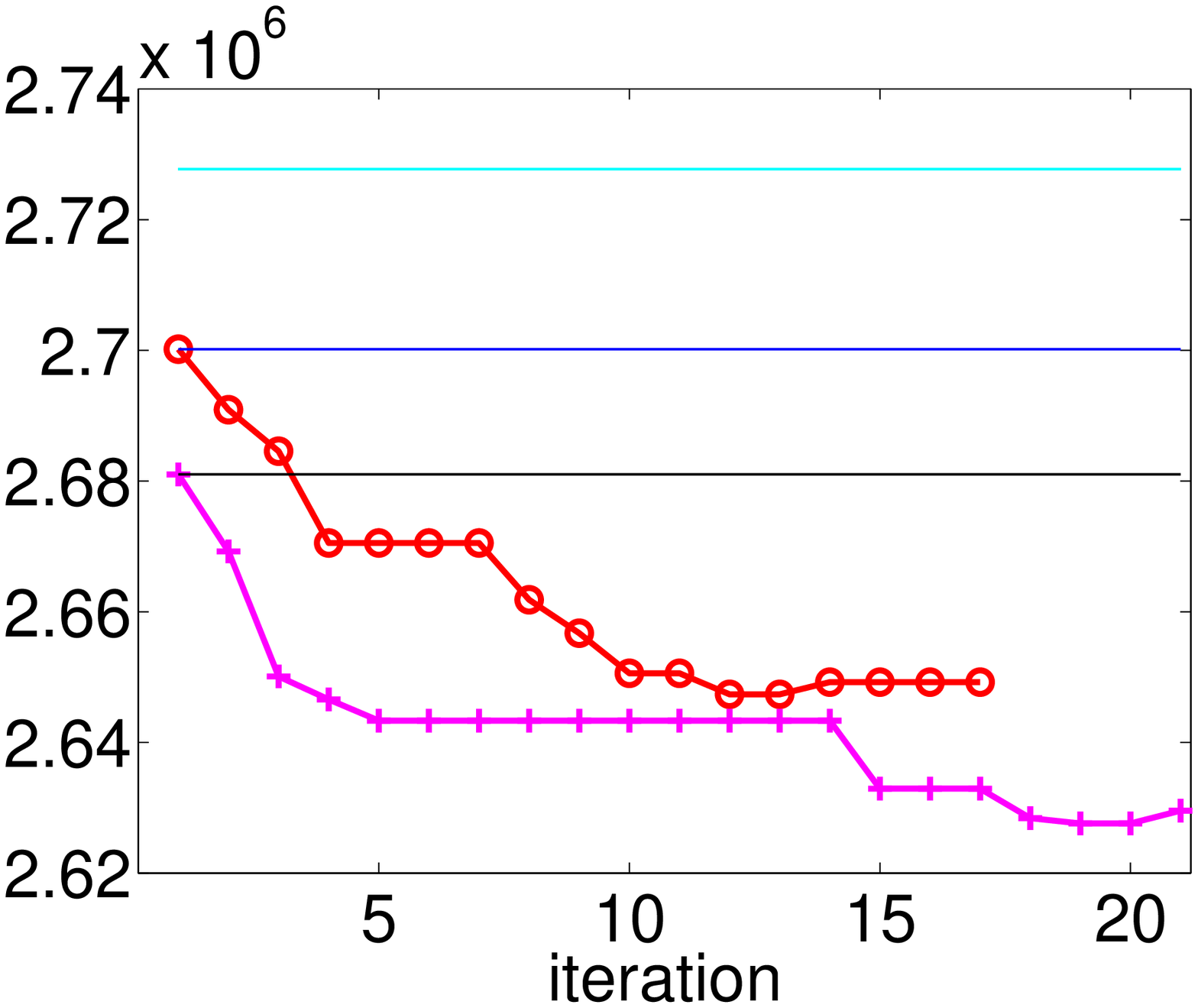} &
    \includegraphics[width=0.240\linewidth]{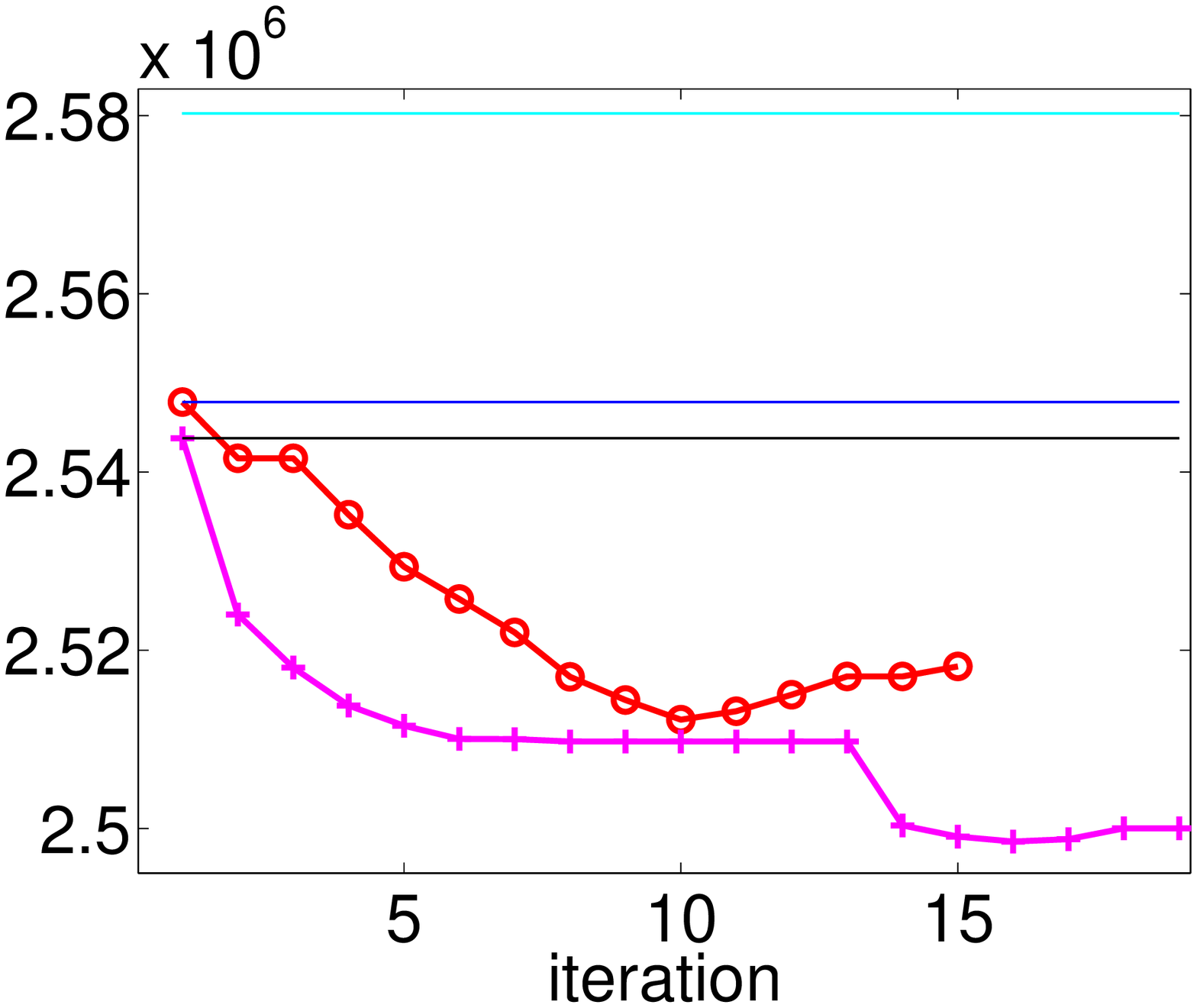} &
    \includegraphics[width=0.240\linewidth]{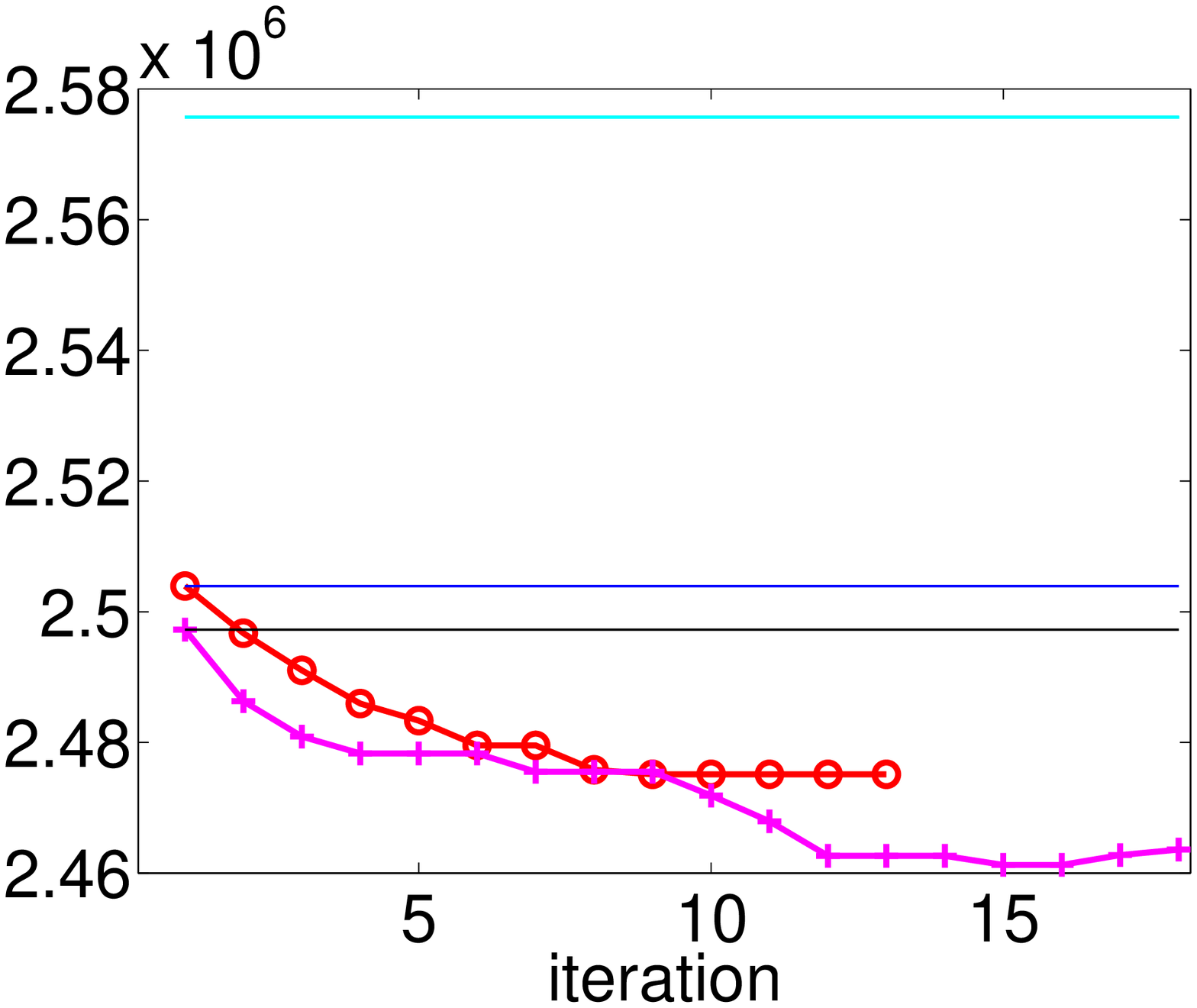} &
    \includegraphics[width=0.240\linewidth]{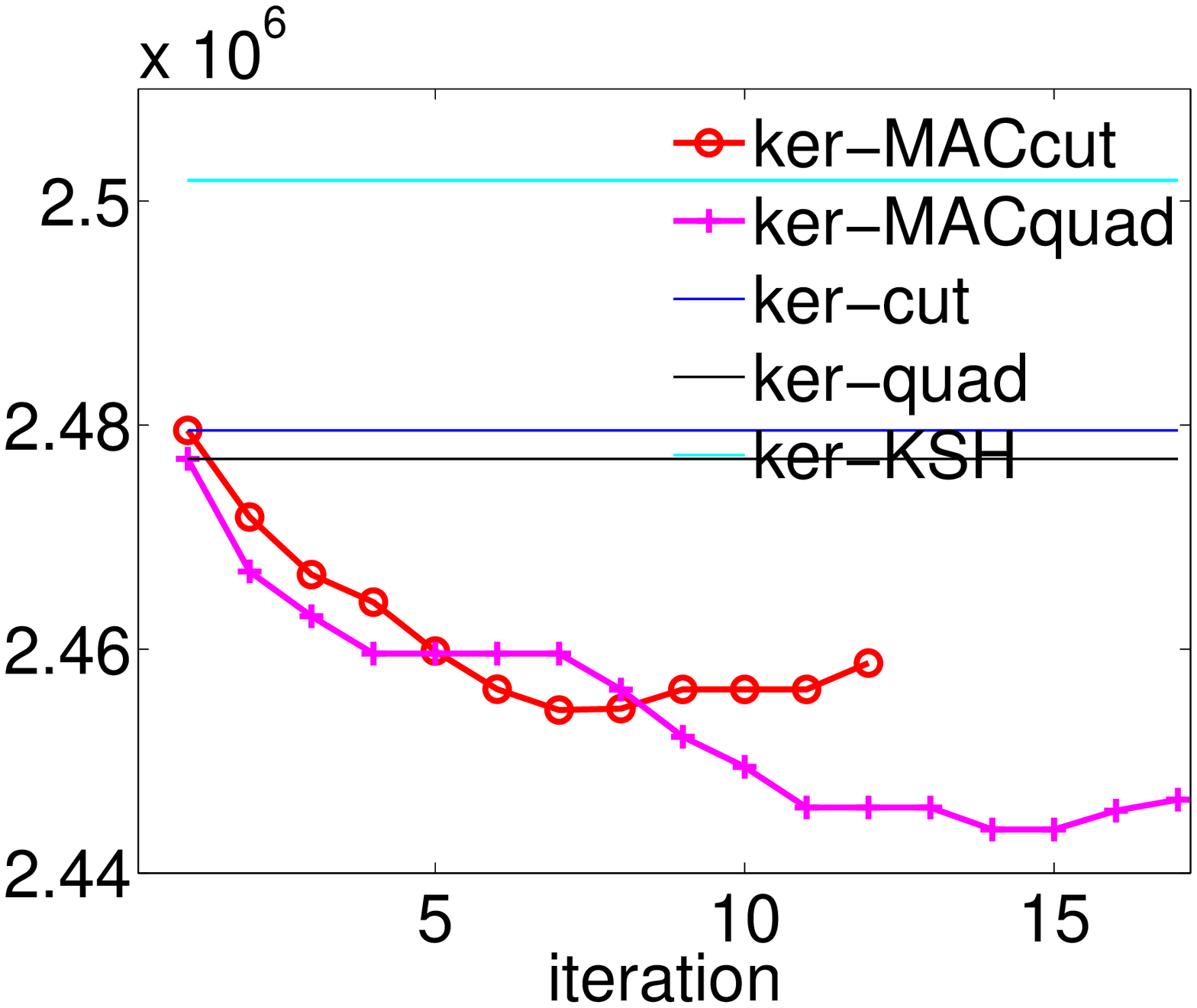} \\
    \hspace{2ex}\rotatebox{90}{\hspace{7ex}precision} &
    \includegraphics[width=0.240\linewidth]{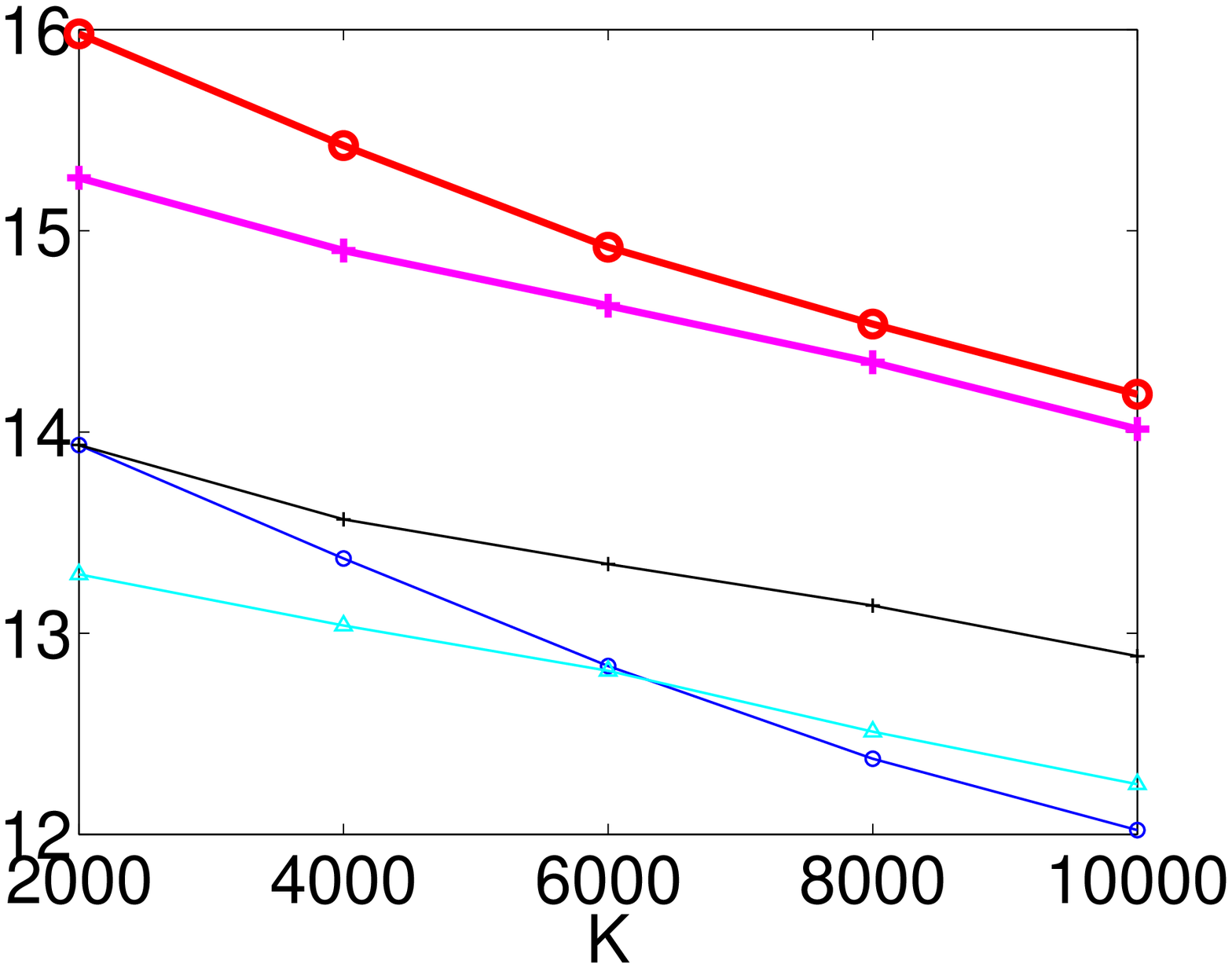} &
    \includegraphics[width=0.240\linewidth]{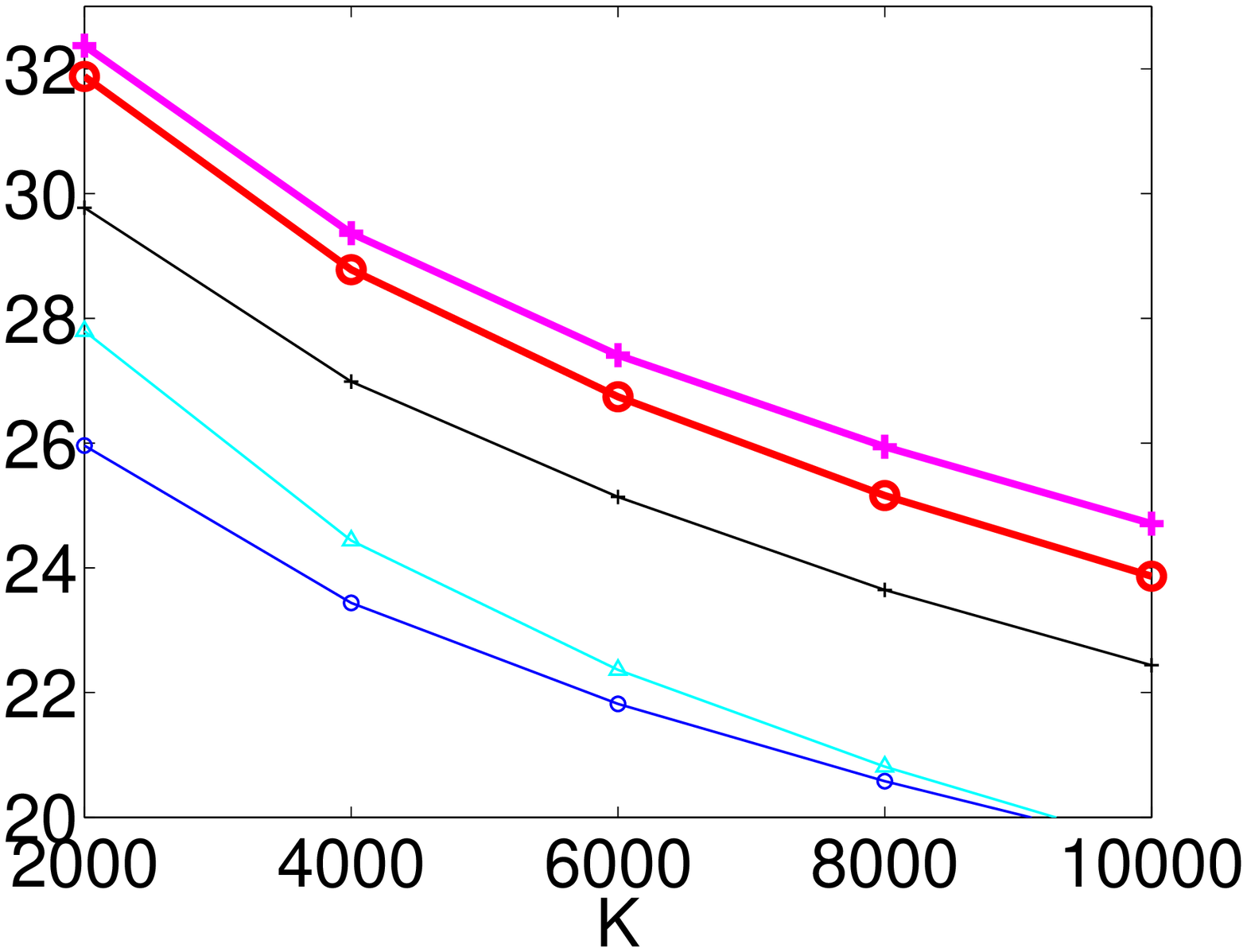} &
    \includegraphics[width=0.240\linewidth]{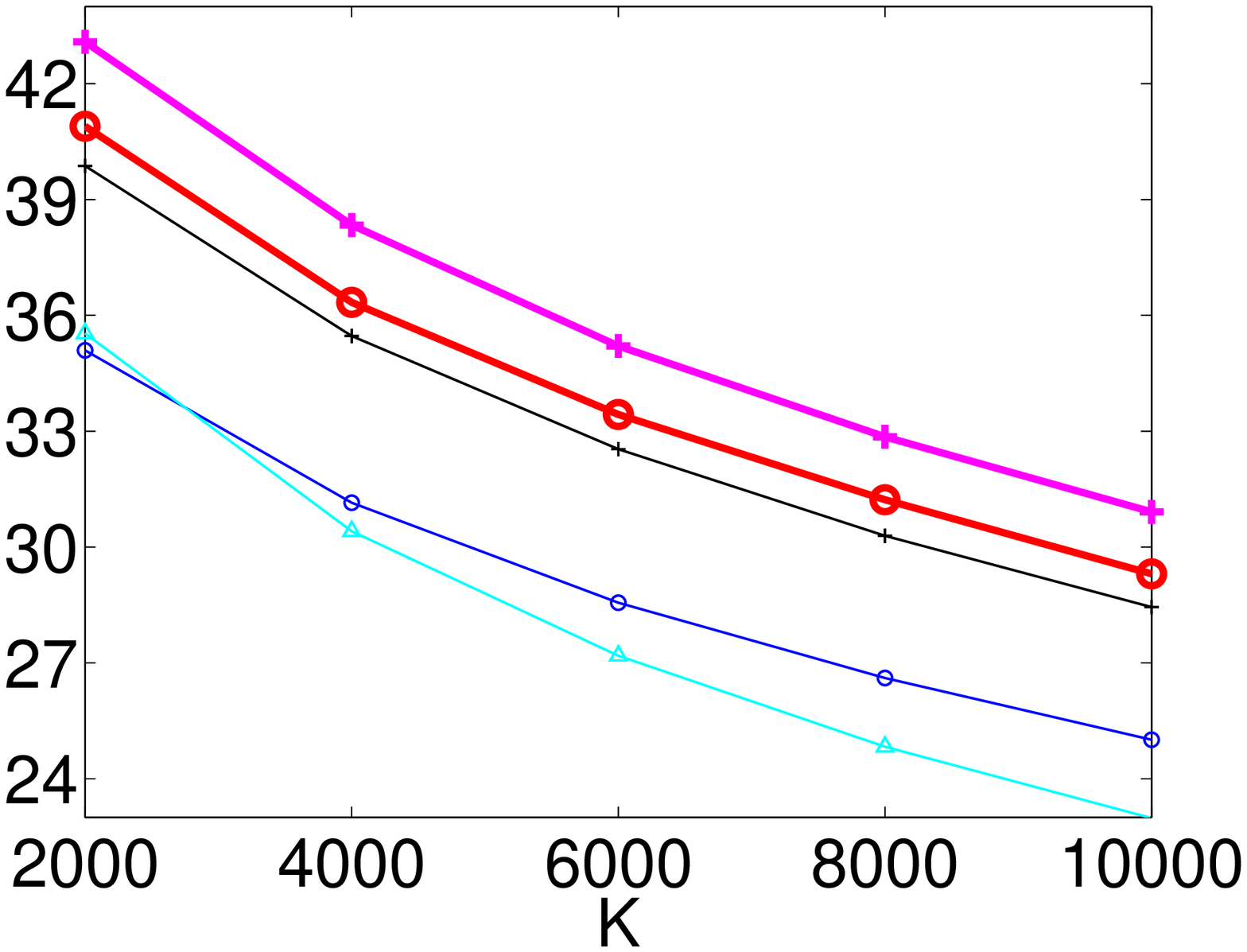} &
    \includegraphics[width=0.240\linewidth]{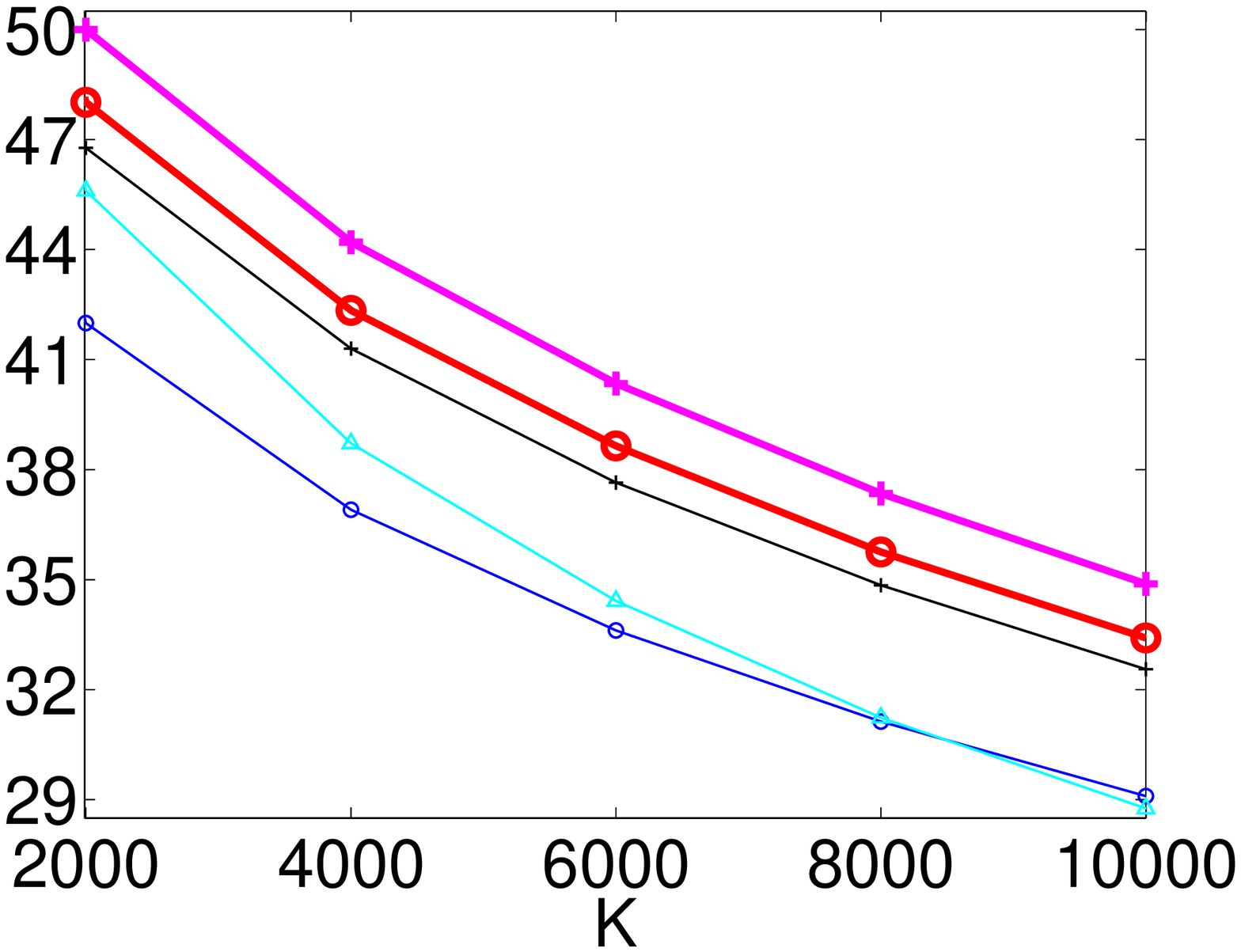} \\
    \hspace{2ex}\rotatebox{90}{\raisebox{3ex}[0pt][0pt]{\makebox[0pt][l]{\hspace{13ex}KSH, $\kappa_+ = 50$, $\kappa_- = 1\,000$, $K = 10\,000$}}\hspace{7ex}precision} &
    \includegraphics[width=0.240\linewidth]{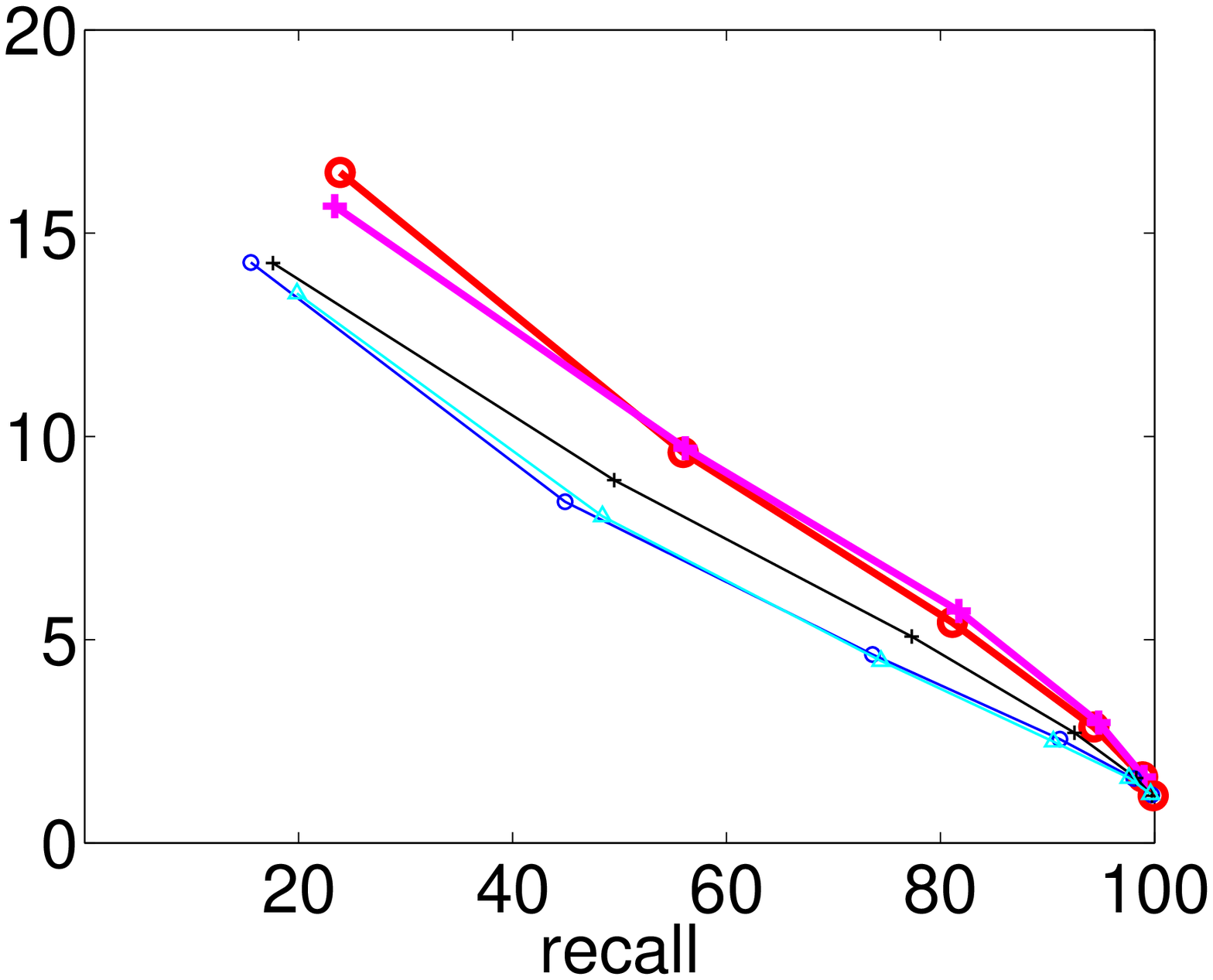} &
    \includegraphics[width=0.240\linewidth]{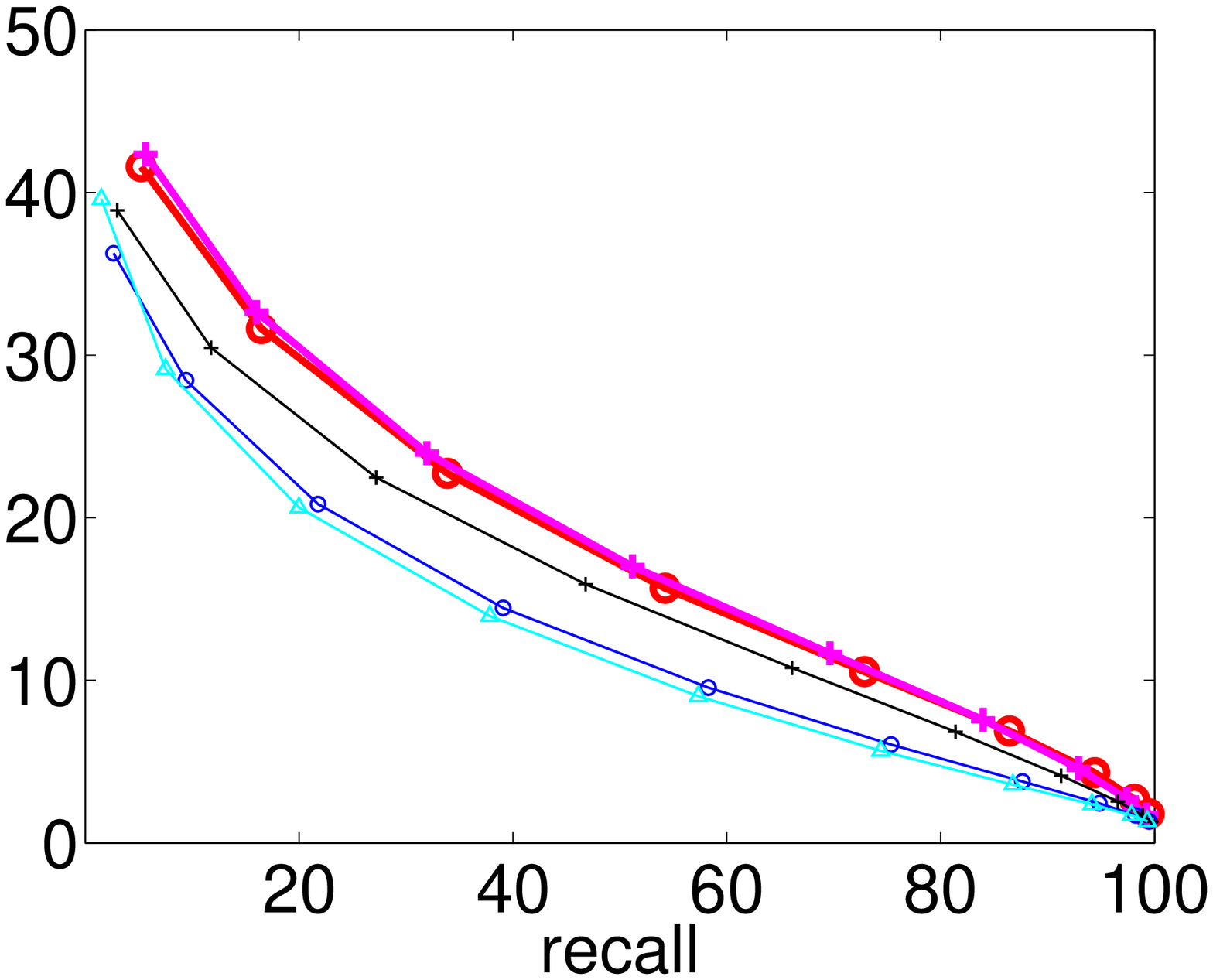} &
    \includegraphics[width=0.240\linewidth]{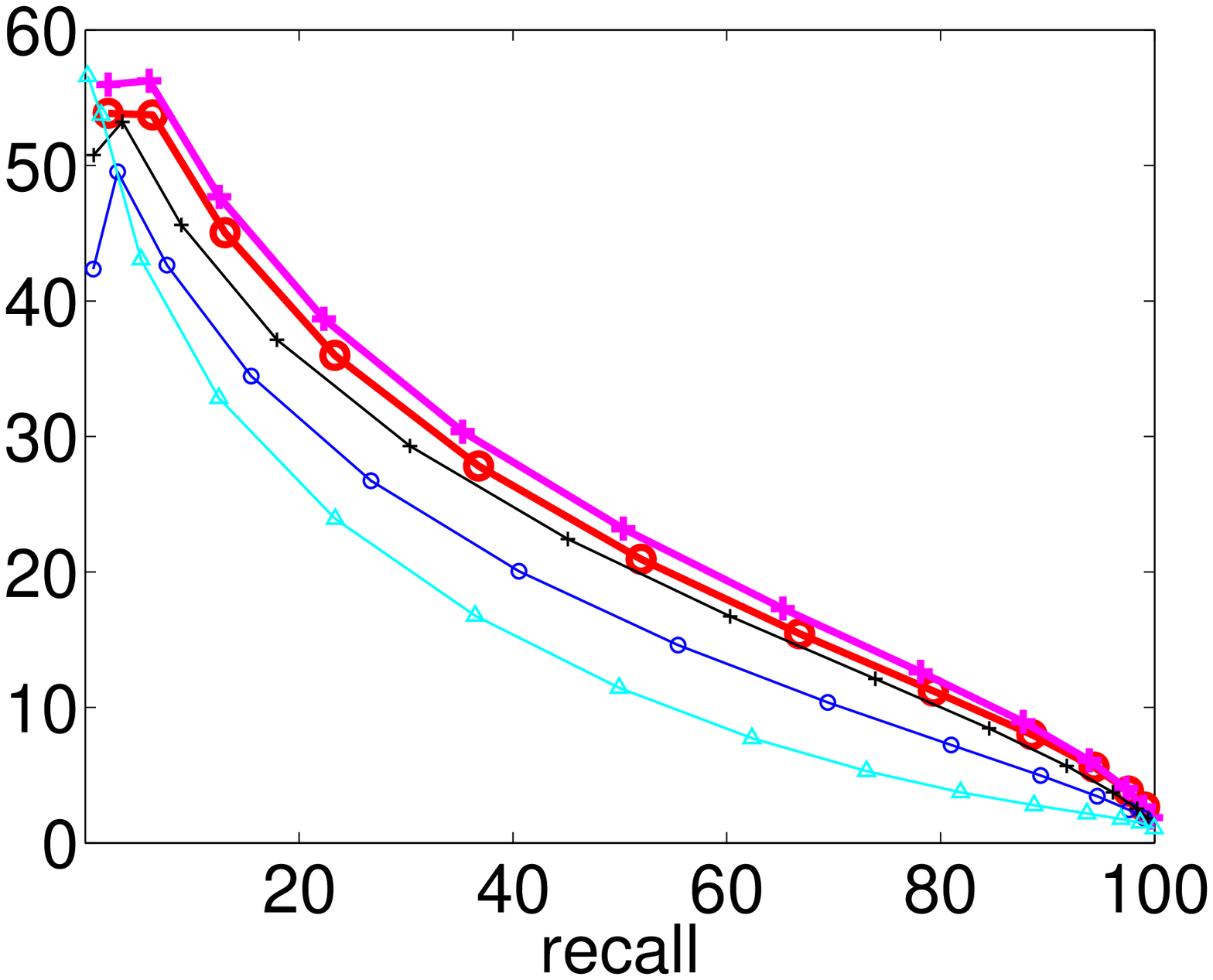} &
    \includegraphics[width=0.240\linewidth]{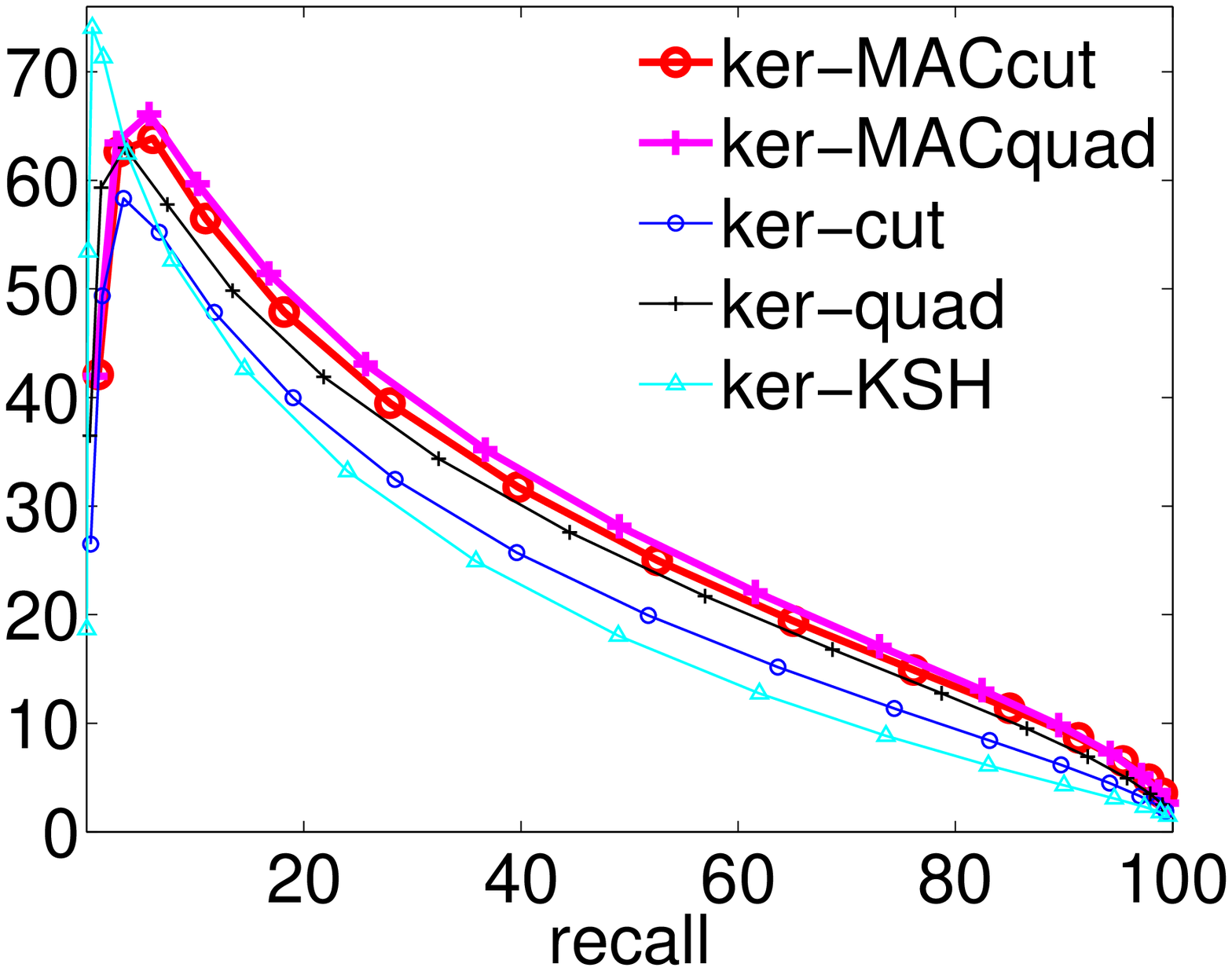}
  \end{tabular}
  \caption{Like the top panel of fig.~\ref{f:unsup-nestederr} (KSH loss) but with $\kappa_+ = 50$, $\kappa_- = 1\,000$, $K = 10\,000$.}
  \label{f:unsup-nestederr1}
\end{figure}

Fig.~\ref{f:unsup-nestederr} shows results using KSH and eSLPH loss functions, respectively, with different sizes of retrieved neighbor sets and using $8$ to $32$ bits. As with the supervised datasets, it is clear that the MAC algorithm finds better optima and that \emph{MACcut} is generally better than \emph{MACquad}. Fig.~\ref{f:unsup-nestederr1} shows one case, using $\kappa_+ = 50$, $\kappa_- = 1\,000$ and $K = 10\,000$ (1\% of the base set), where \emph{quad} outperforms \emph{cut} and correspondingly \emph{MACquad} outperforms \emph{MACcut}, although both MAC results are very close, particularly in precision and recall.

\begin{figure}[t!]
  \centering
  \psfrag{rerror}[][t]{loss function \calL}
  \psfrag{iteration}[t][]{iterations}
  \psfrag{K}[][]{$k$}
  \psfrag{precision}[][t]{precision}
  \psfrag{recall}[][]{recall}
  \begin{tabular}{@{}l@{}c@{}c@{}c@{}c@{}}
    & $b=8$ & $b=16$ & $b=24$ & $b=32$ \\
    \hspace{2ex}\rotatebox{90}{\hspace{7ex}precision} &
    \includegraphics[width=0.240\linewidth]{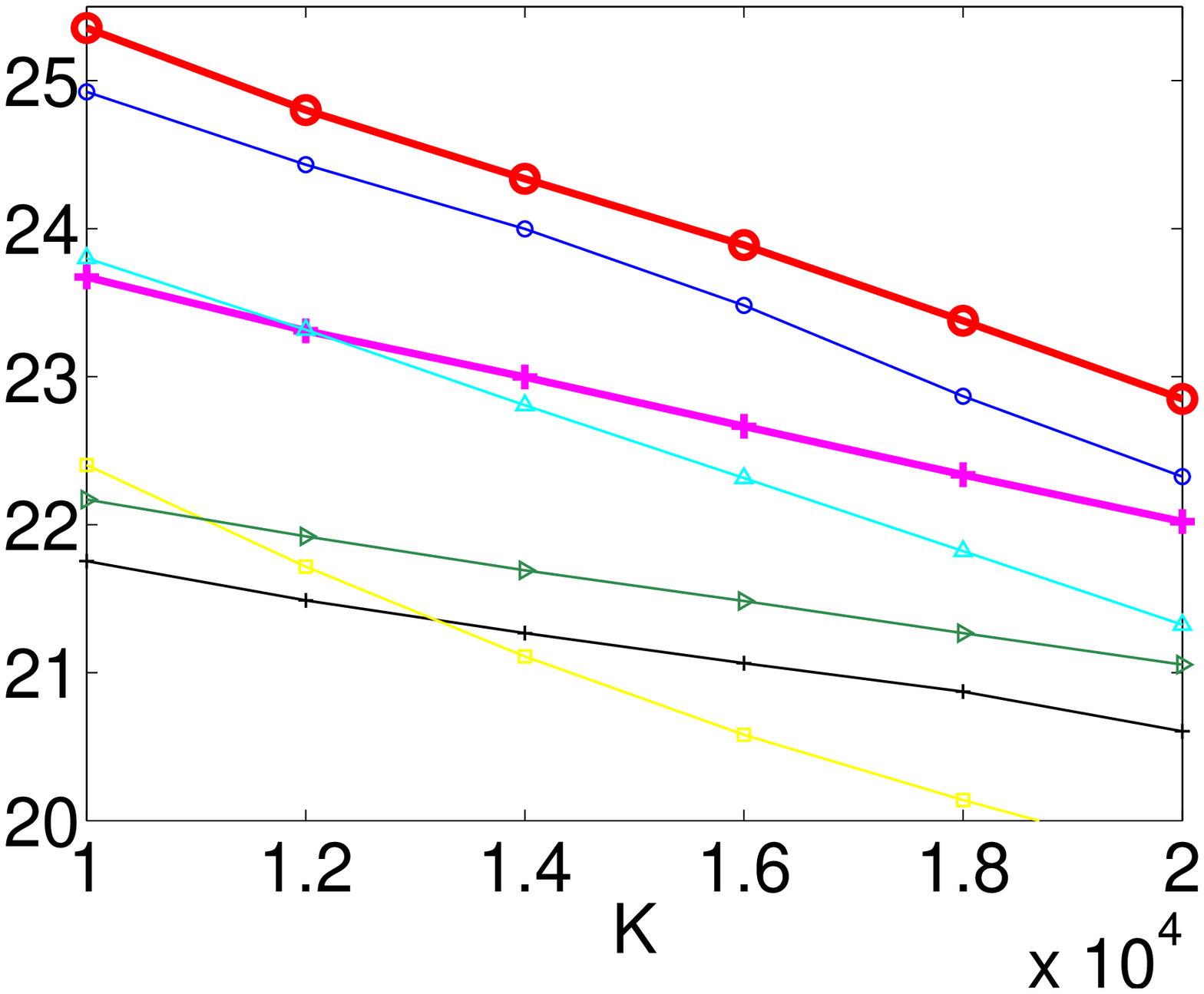} &
    \includegraphics[width=0.240\linewidth]{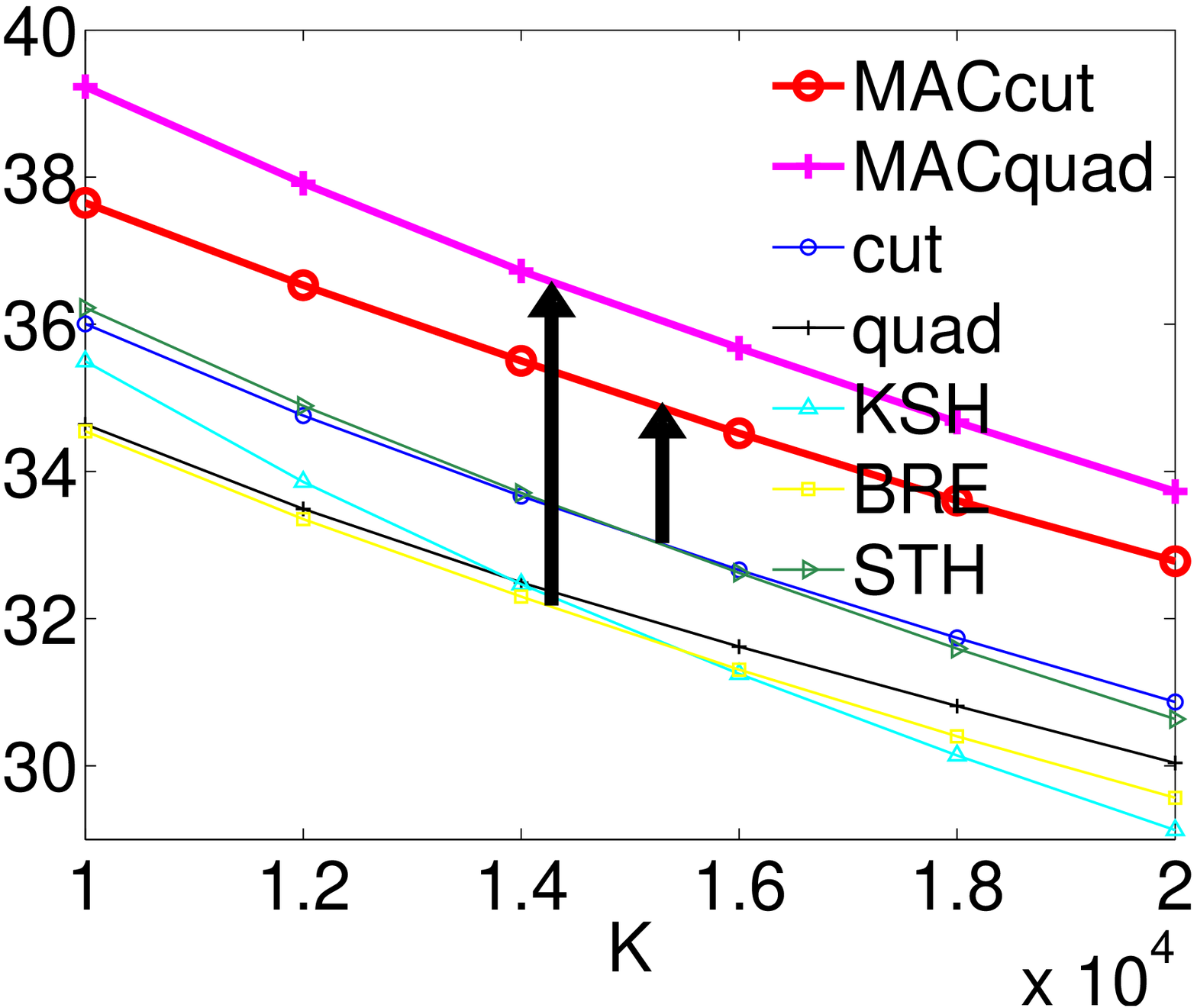} &
    \includegraphics[width=0.240\linewidth]{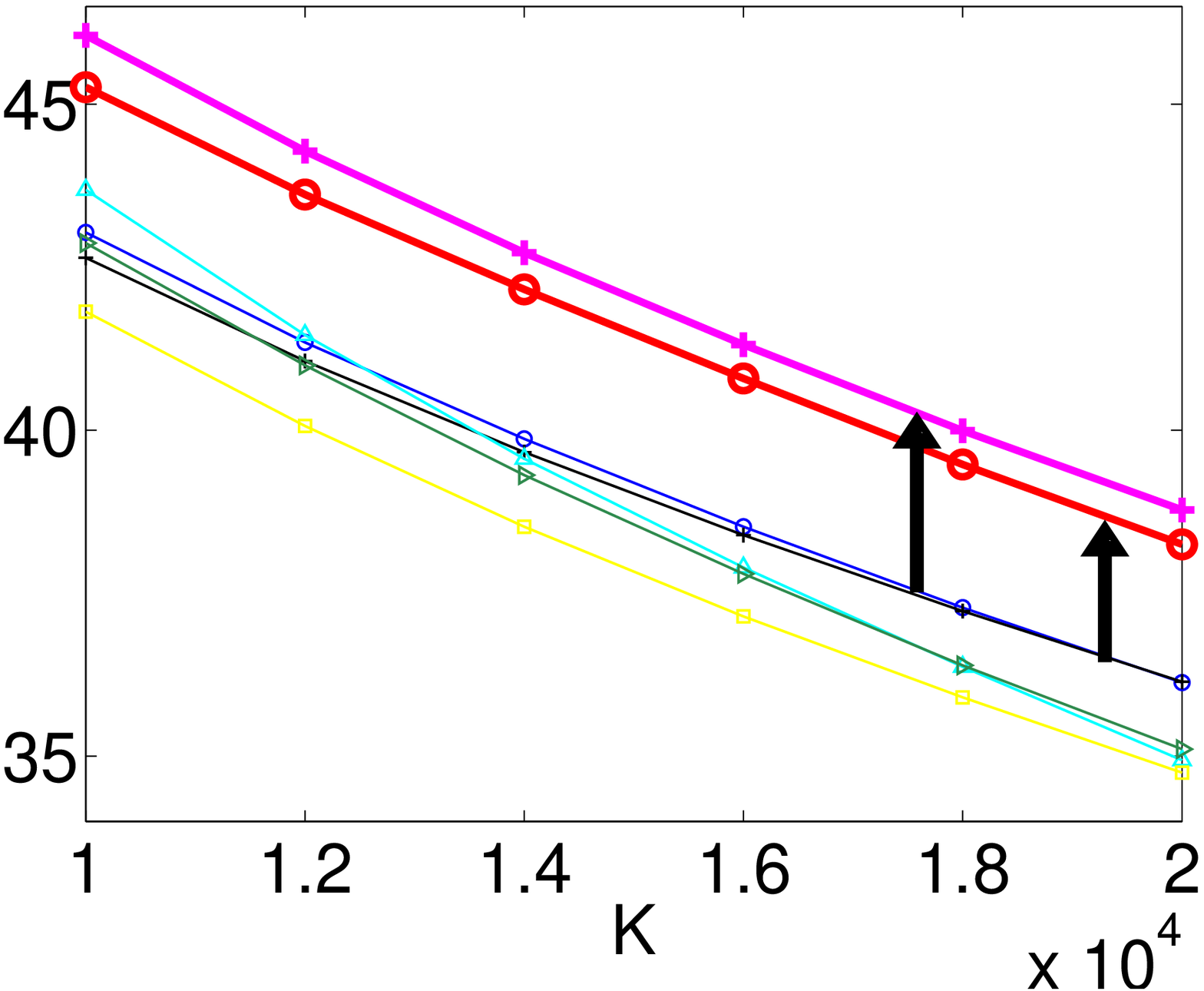} &
    \includegraphics[width=0.240\linewidth]{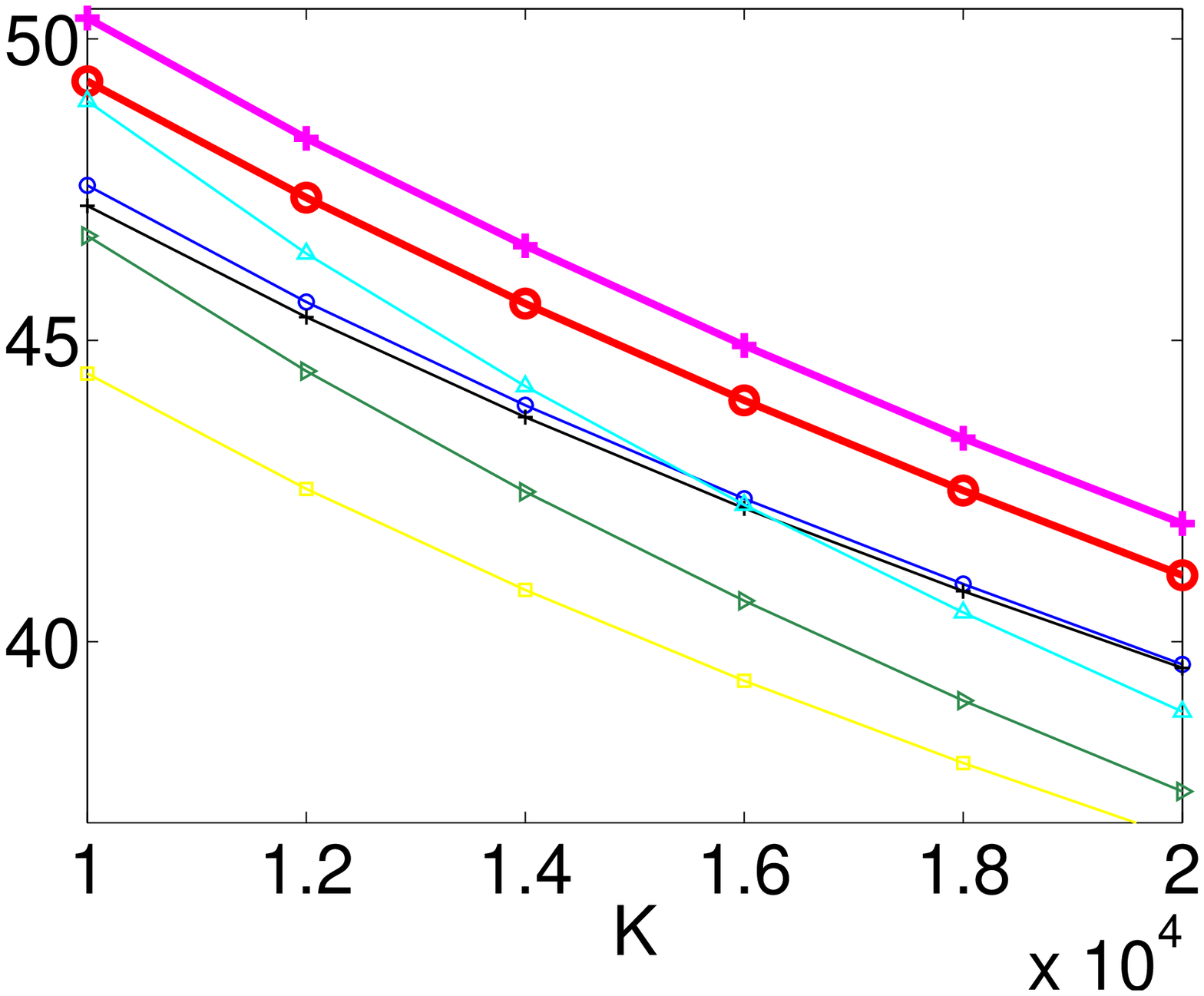} \\[-1ex]
    \hspace{2ex}\rotatebox{90}{\raisebox{3ex}[0pt][0pt]{\makebox[0pt][l]{\hspace{-1ex}SIFT1M (using distance-based pseudolabels)}}\hspace{7ex}precision} &
    \includegraphics[width=0.240\linewidth]{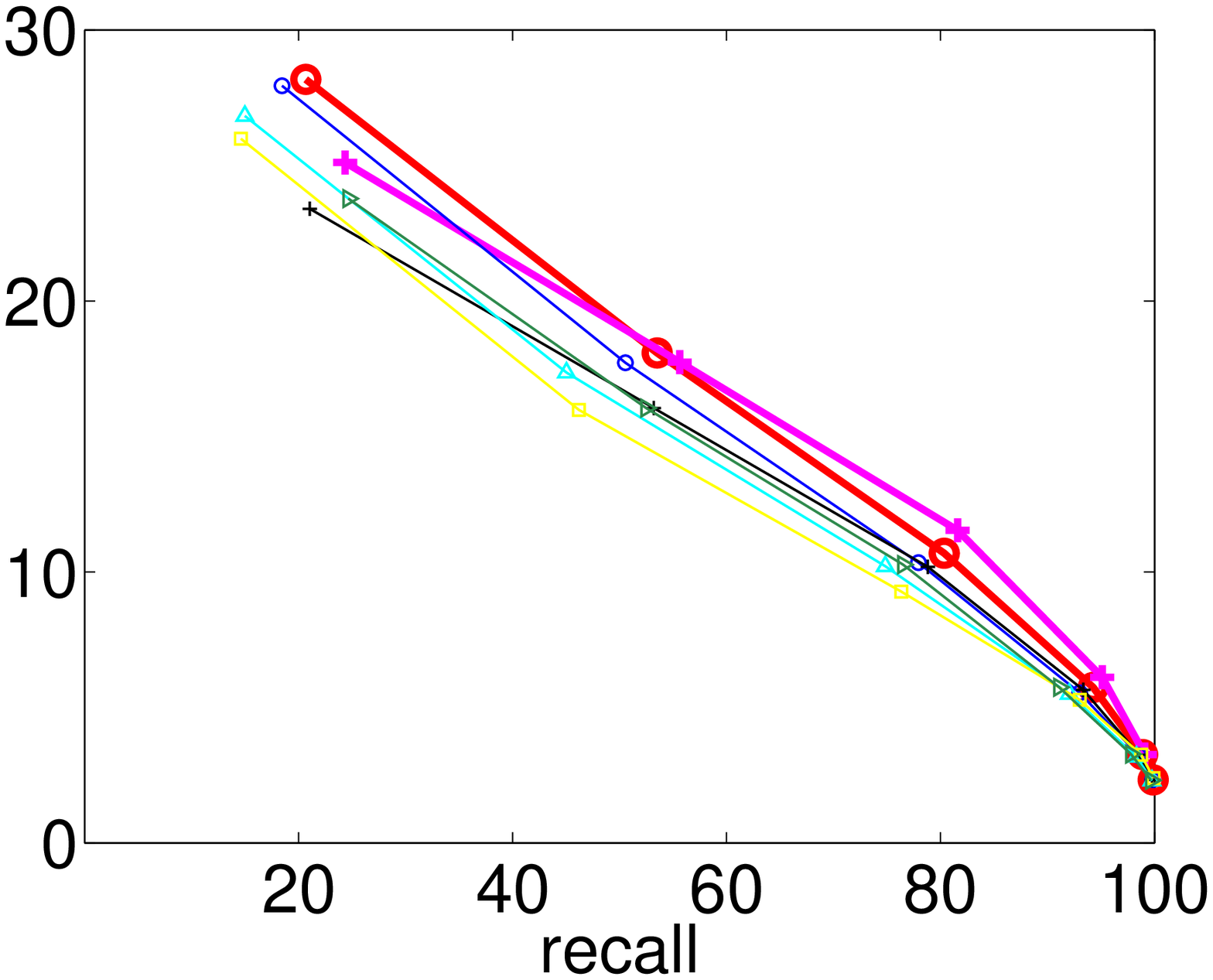} &
    \includegraphics[width=0.240\linewidth]{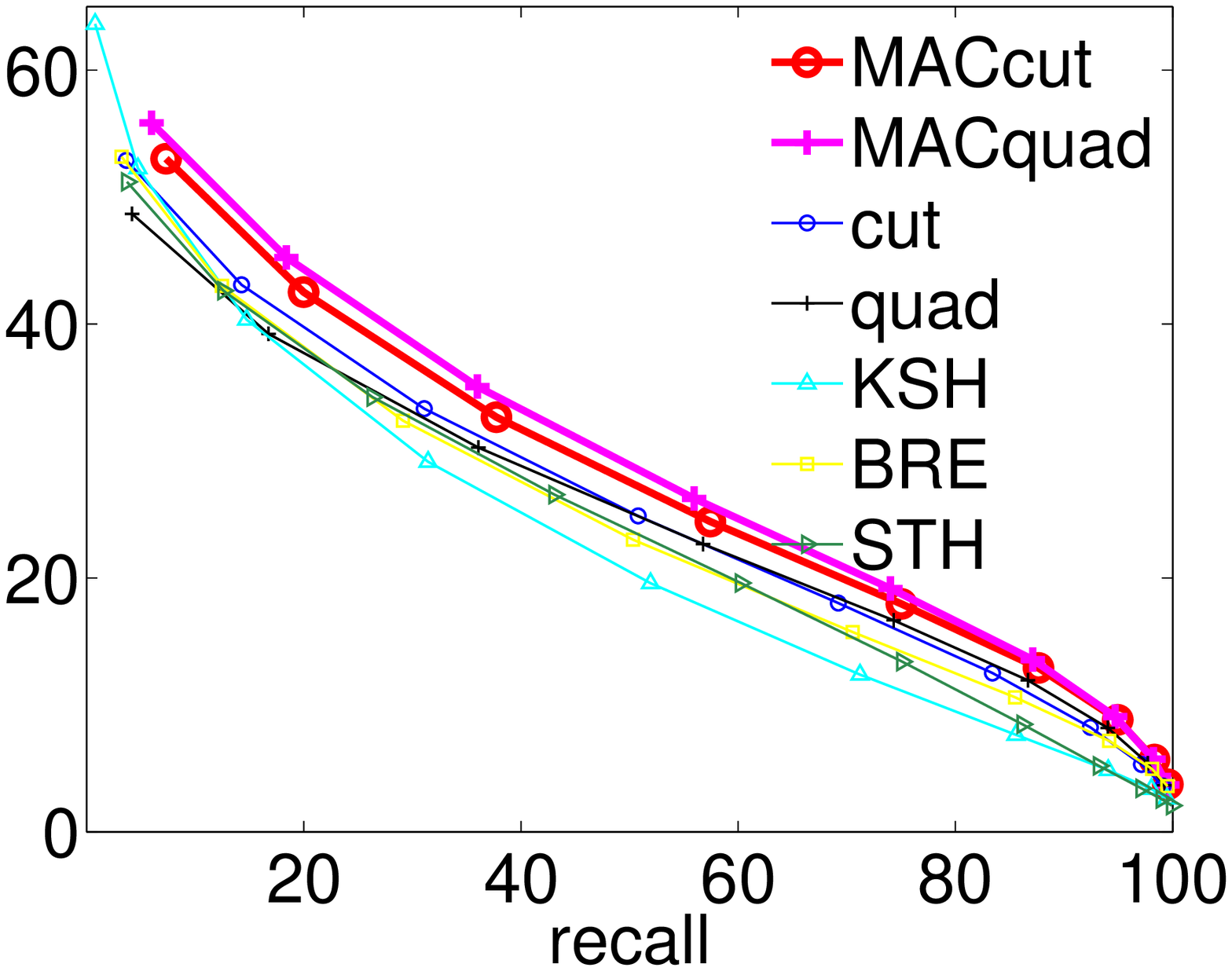} &
    \includegraphics[width=0.240\linewidth]{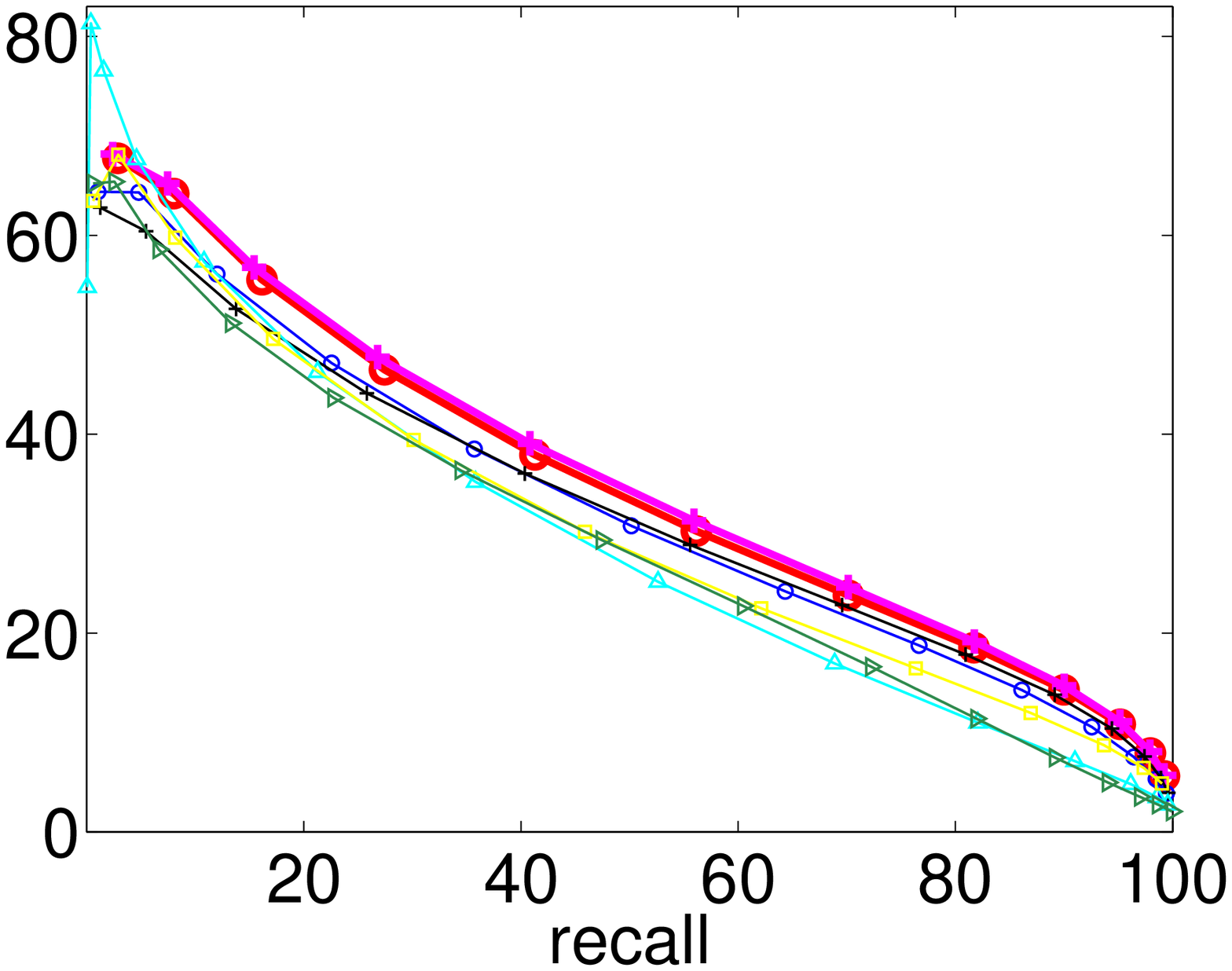} &
    \includegraphics[width=0.240\linewidth]{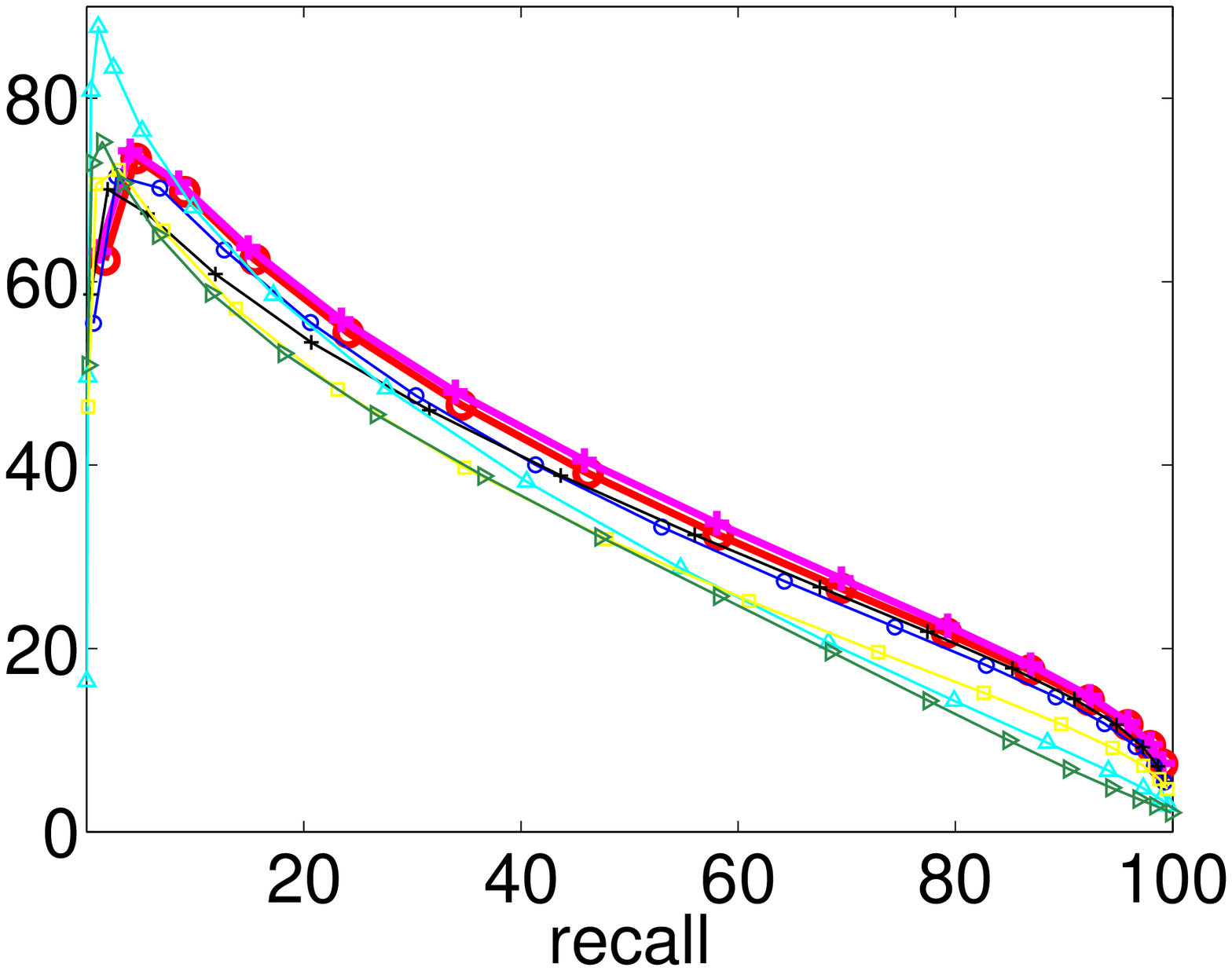}
  \end{tabular} \\[1ex]
  \begin{tabular}{@{}l@{}c@{}c@{}c@{}c@{}}
    \hspace{2ex}\rotatebox{90}{\hspace{7ex}precision} &
    \includegraphics[width=0.240\linewidth]{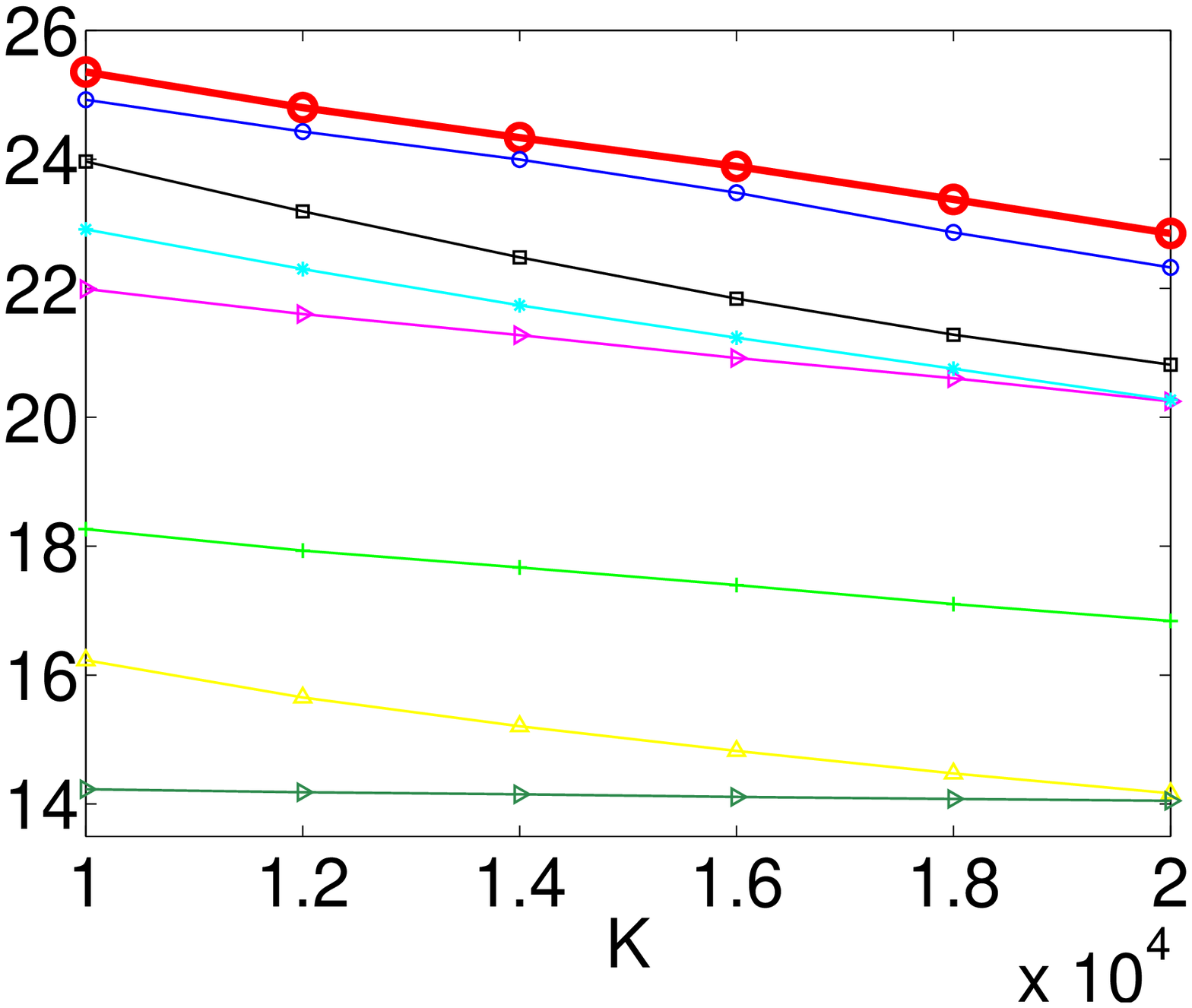} &
    \includegraphics[width=0.240\linewidth]{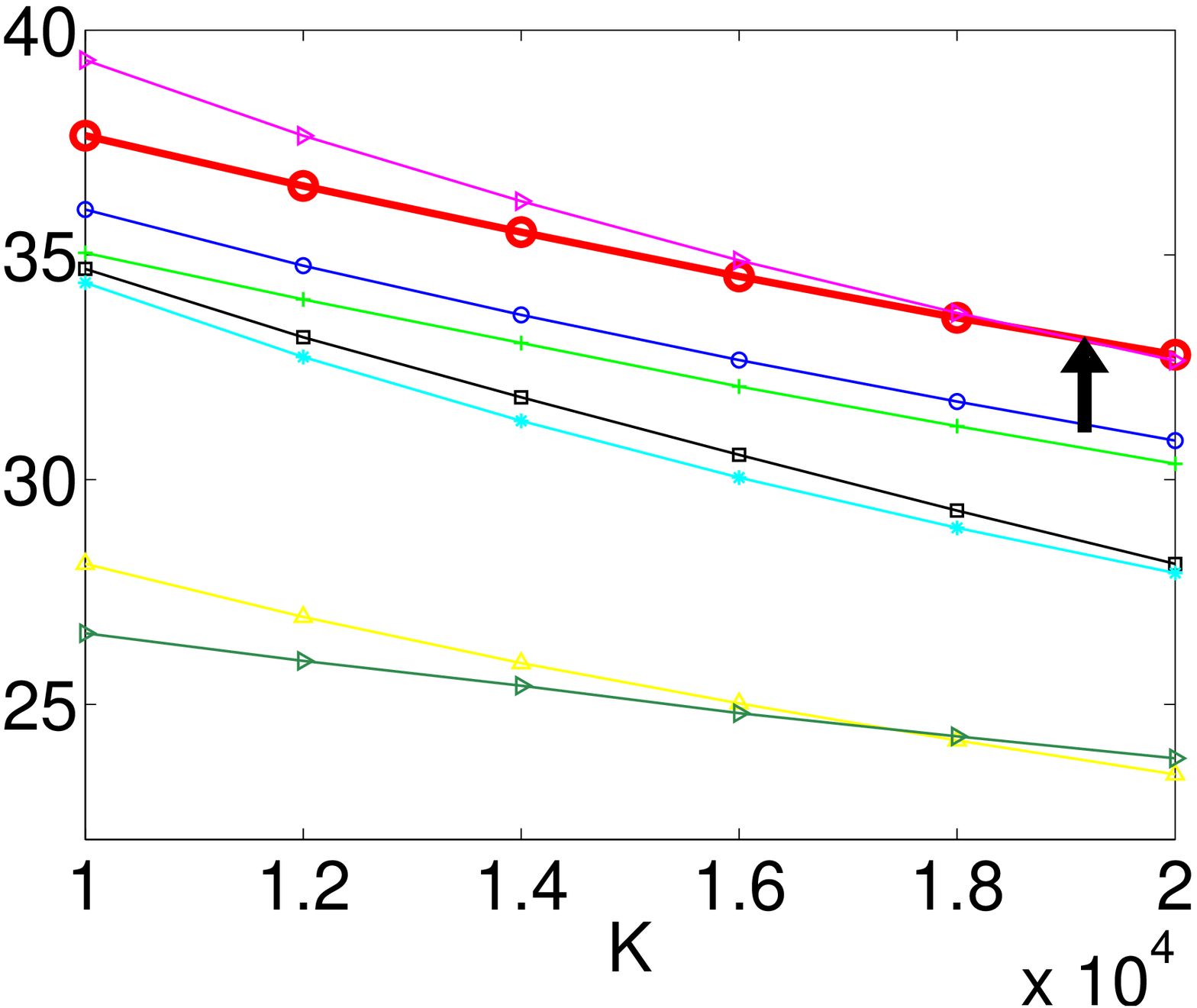} &
    \includegraphics[width=0.240\linewidth]{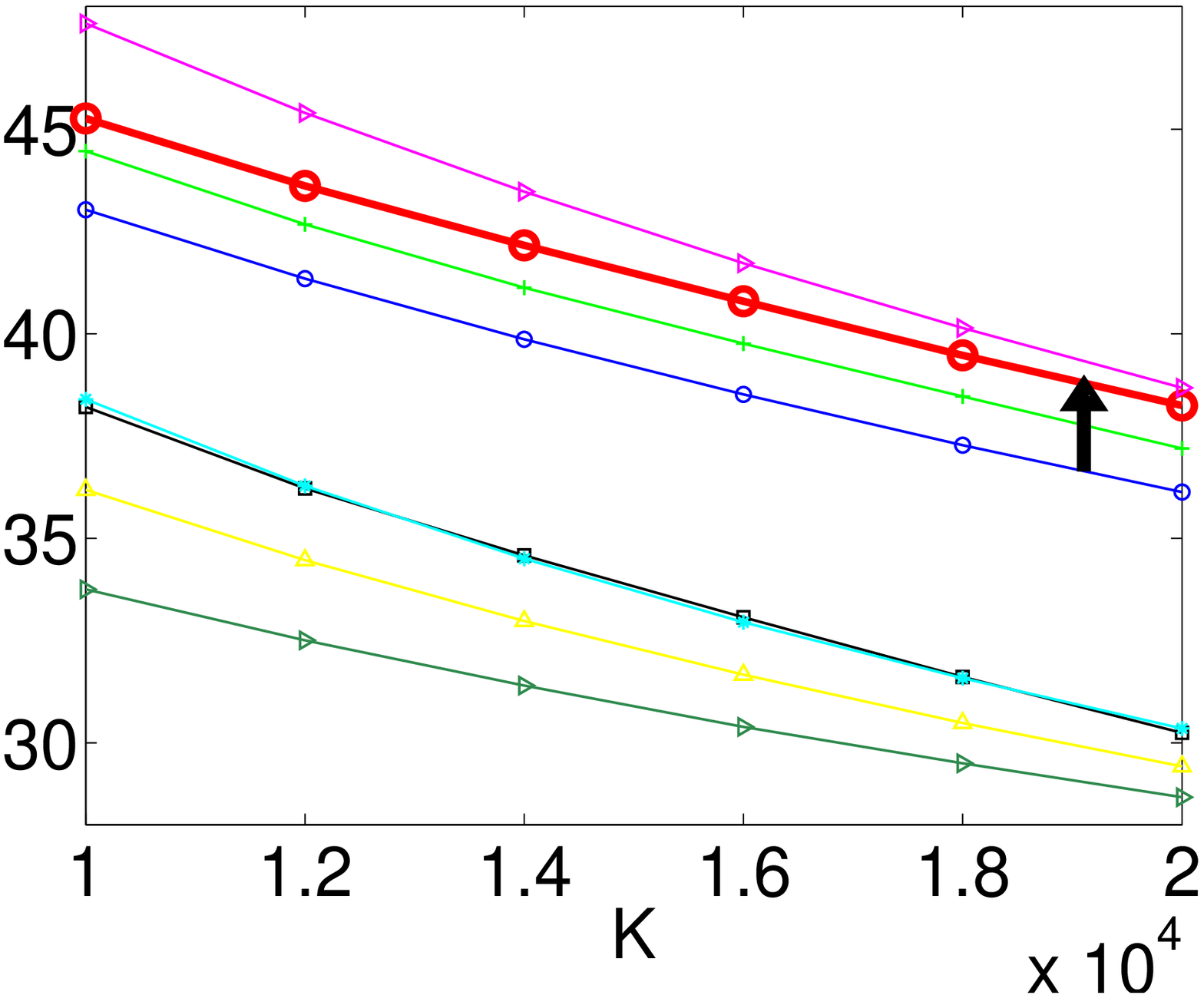} &
    \includegraphics[width=0.240\linewidth]{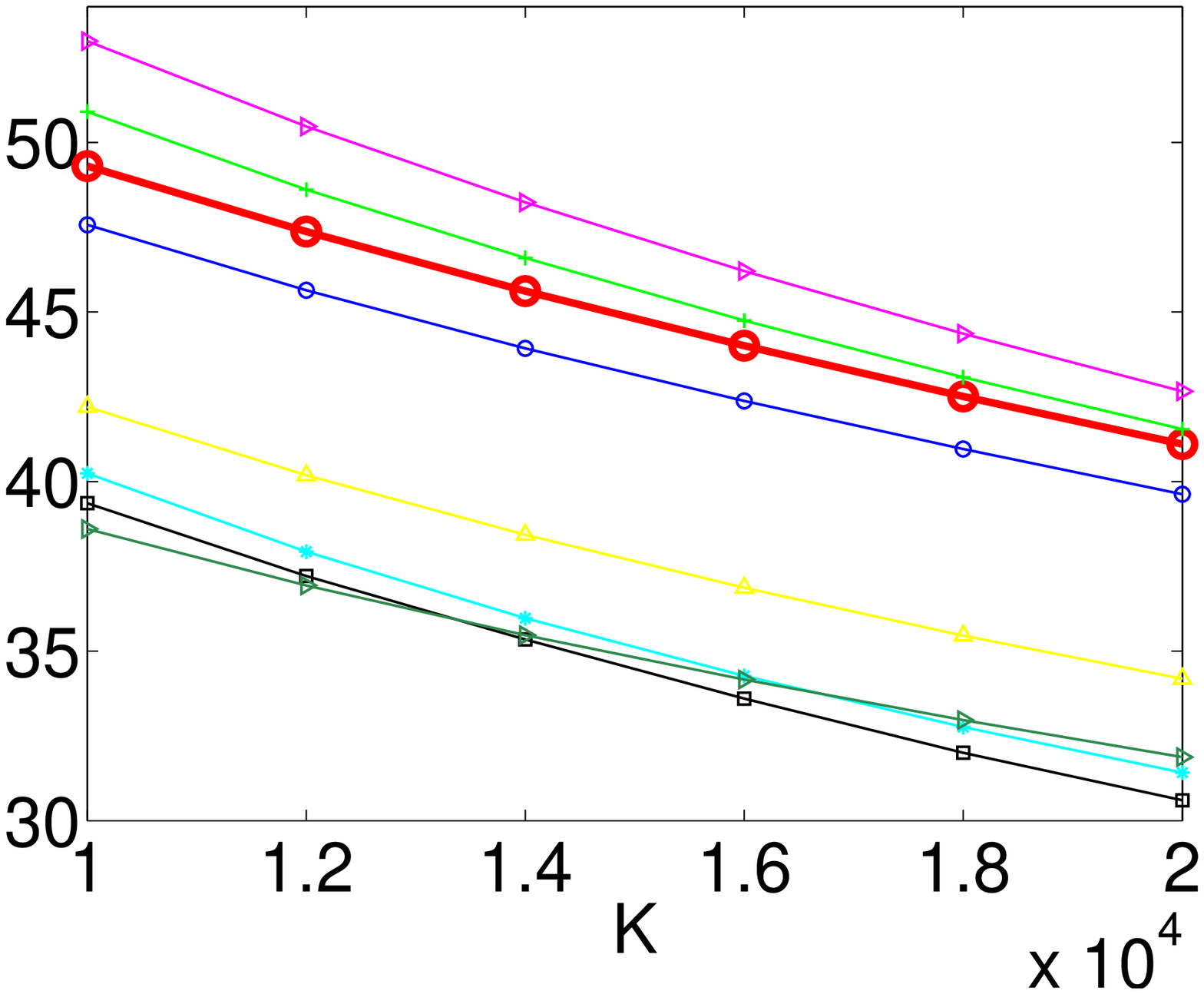} \\[-1ex]
    \hspace{2ex}\rotatebox{90}{\raisebox{3ex}[0pt][0pt]{\makebox[0pt][l]{\hspace{5ex}SIFT1M (unsupervised, no labels)}}\hspace{7ex}precision} &
    \includegraphics[width=0.240\linewidth]{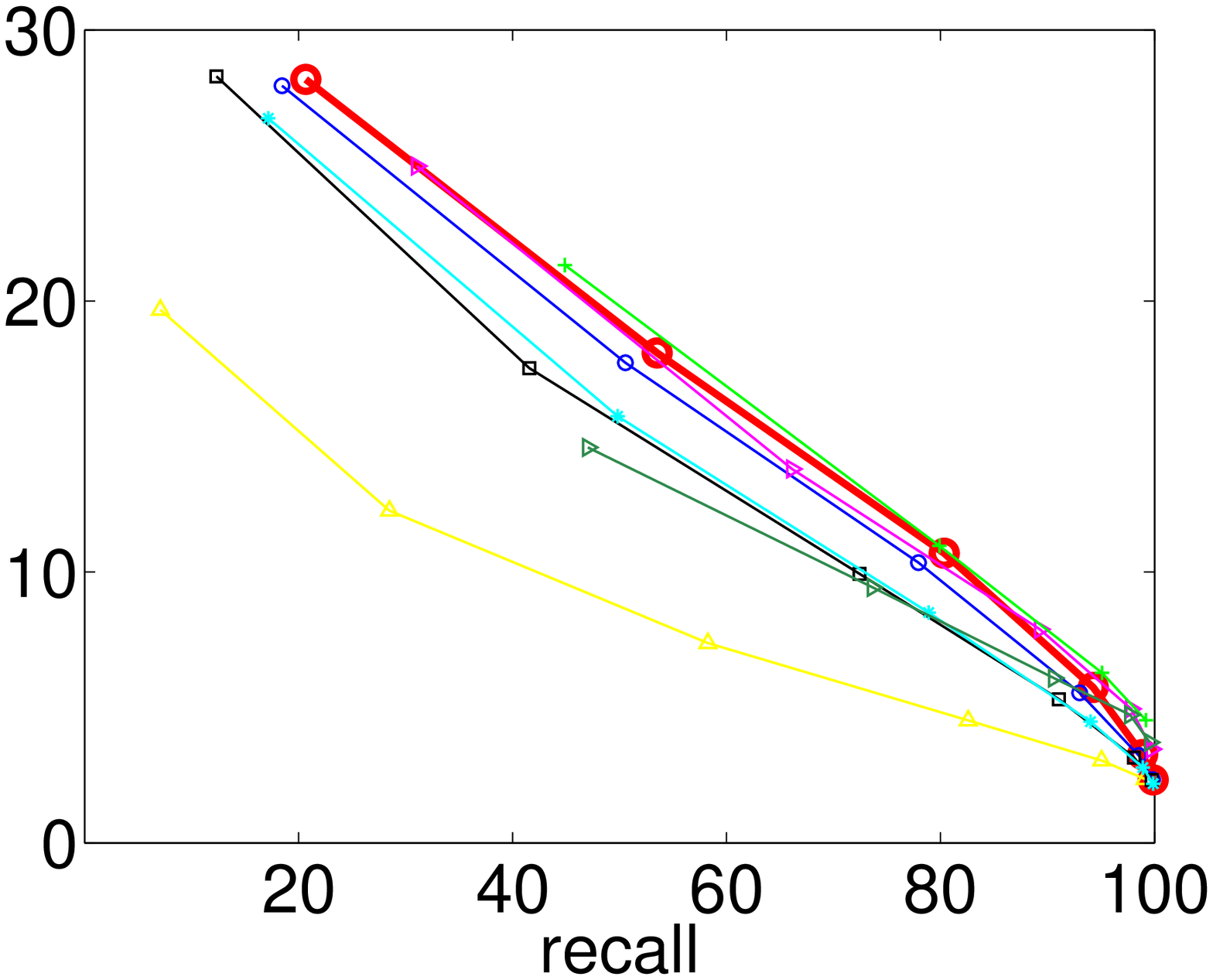} &
    \includegraphics[width=0.240\linewidth]{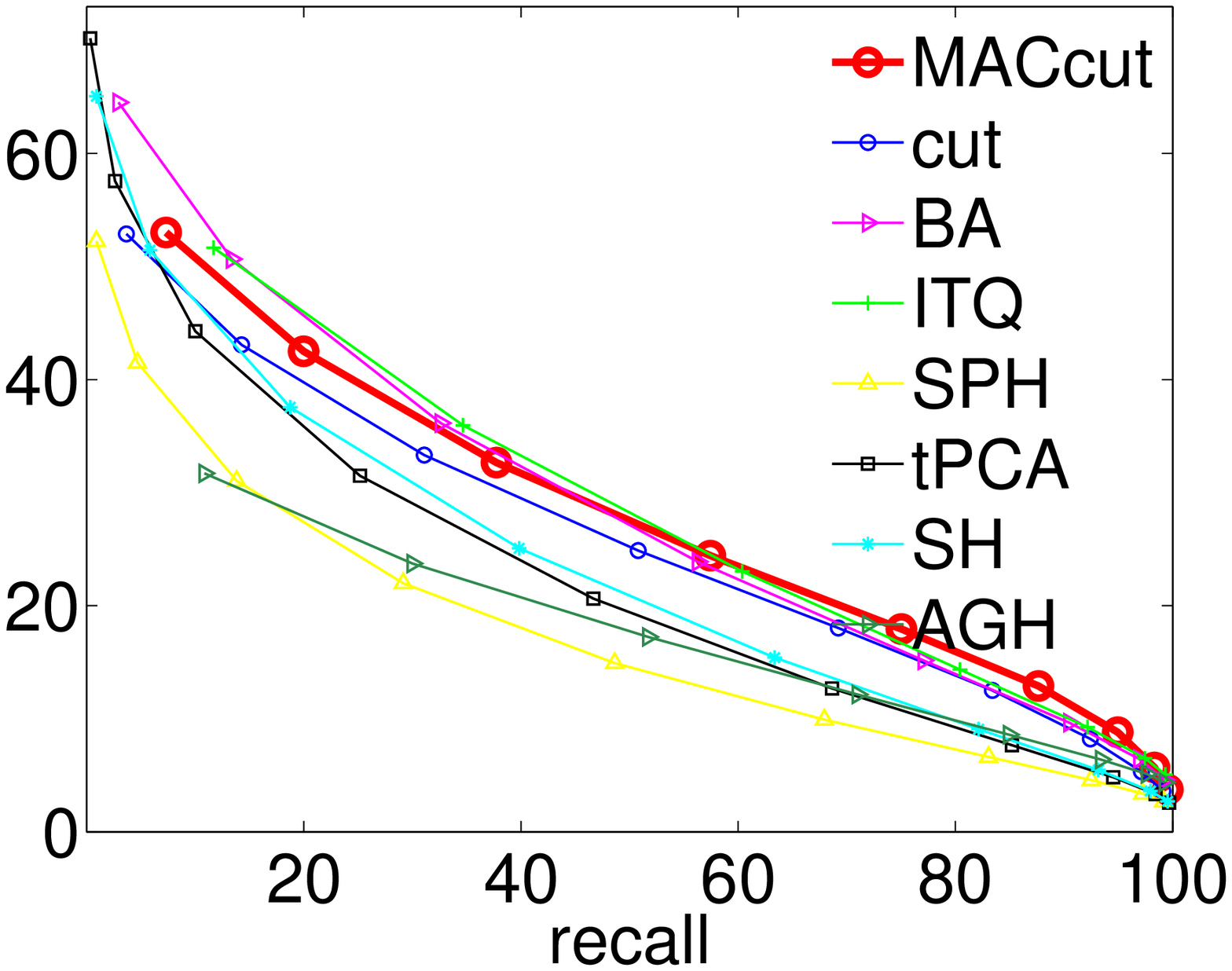} &
    \includegraphics[width=0.240\linewidth]{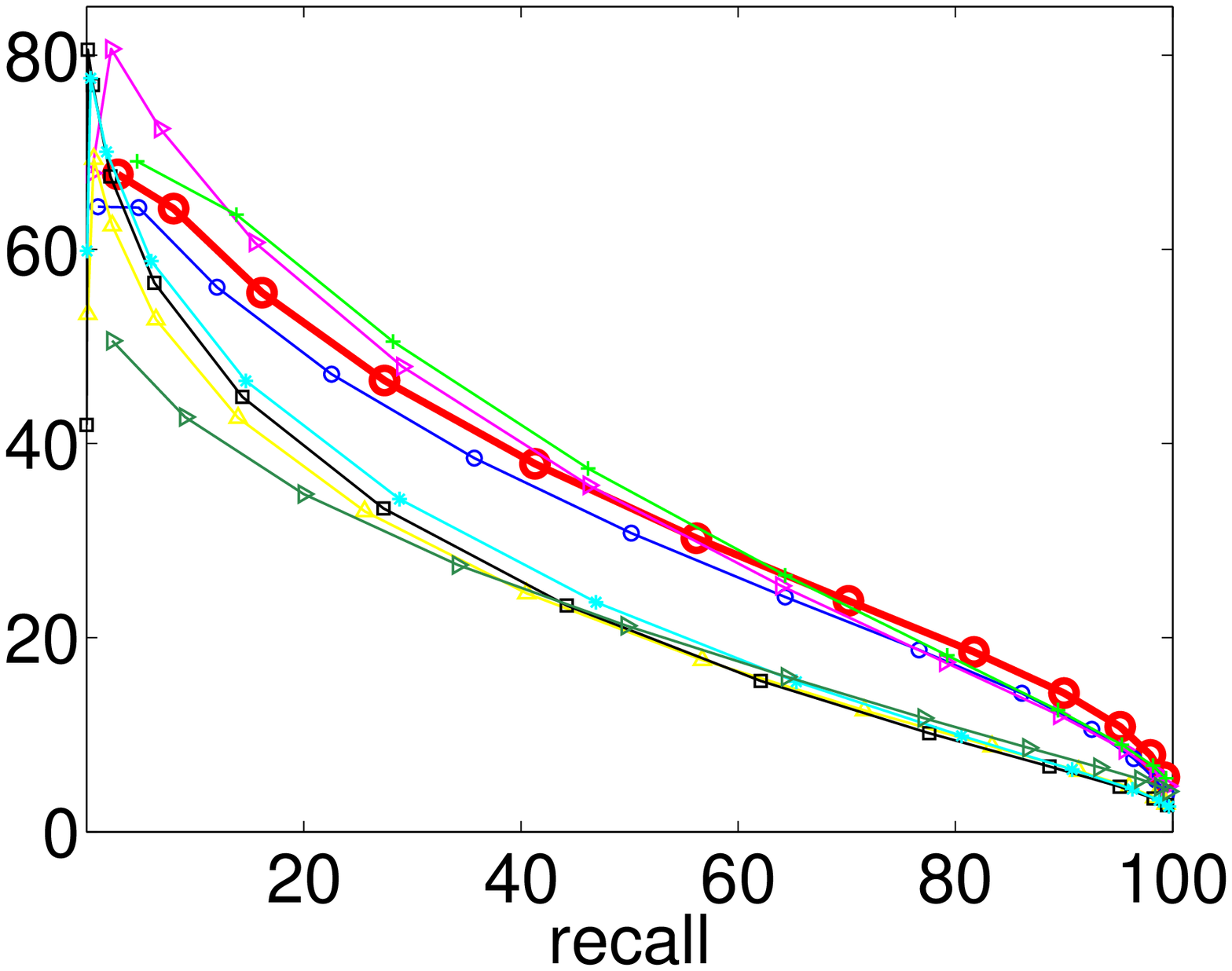} &
    \includegraphics[width=0.240\linewidth]{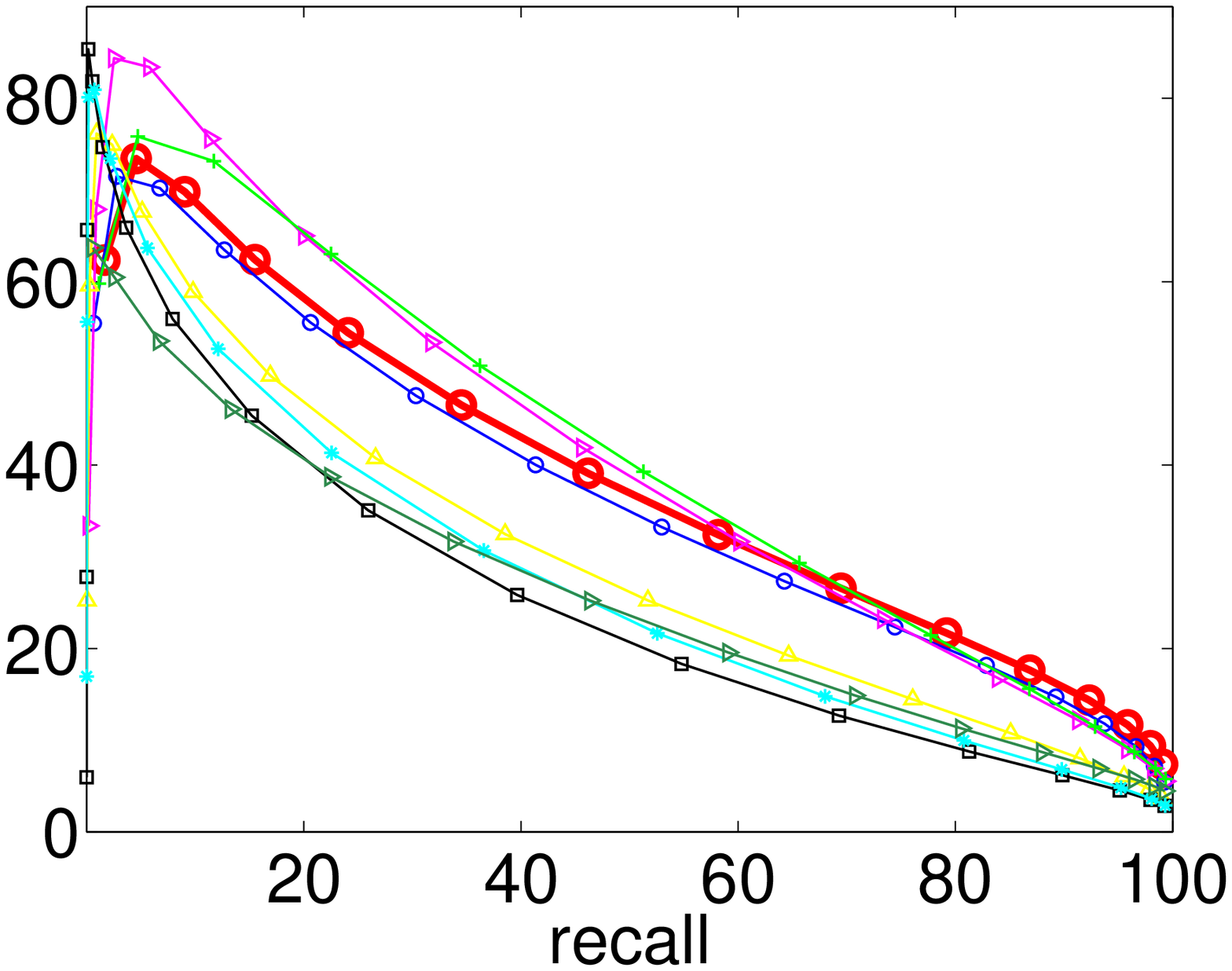}
  \end{tabular}
  \caption{Comparison with binary hashing methods on SIFT1M using pseudolabels (top panel) and without labels (bottom panel). The rows in each panel show the precision (for a range of retrieved points $k$) and the precision/recall (at different Hamming distances), using $b=8$ to $32$ bits.}
  \label{f:unsup-comparison}
\end{figure}

Fig.~\ref{f:unsup-comparison} shows results comparing with binary hashing methods. All methods are trained on a subset of $5\,000$ points. We consider two types of methods. In the first type, we create pseudolabels for each point and then apply supervised methods as in CIFAR (in particular, \emph{cut}/\emph{quad} and \emph{MACcut}/\emph{MACquad}, using the KSH loss function). The pseudolabels $y_{nm}$ for each training point $\x_n$ are obtained by declaring as similar points its $\kappa_+ = 100$ true nearest neighbors and as dissimilar points a random subset of $\kappa_- = 500$ points among the remaining points. In the second type, we use purely unsupervised methods (not based on similar/dissimilar affinities): thresholded PCA (tPCA), Iterative Quantization (ITQ) \citep{Gong_13a}, Binary Autoencoder (BA) \citep{CarreirRaziper15a}, Spectral Hashing (SH) \citep{Weiss_09a}, AnchorGraph Hashing (AGH) \citep{Liu_11a}, and Spherical Hashing (SPH) \citep{Heo_12a}. The results are again in general agreement with the conclusions in the main paper.

\begin{figure}[b!]
  \centering
  \psfrag{bits}[][]{$b$}
    \psfrag{binary}[l][l]{binary codes}
  \begin{tabular}{@{}l@{\hspace{1em}}c@{\hspace{1em}}c@{}}
    & KSH & eSPLH\\
    \rotatebox{90}{\hspace{7ex}Objective function value} &
    \includegraphics[width=0.4\linewidth]{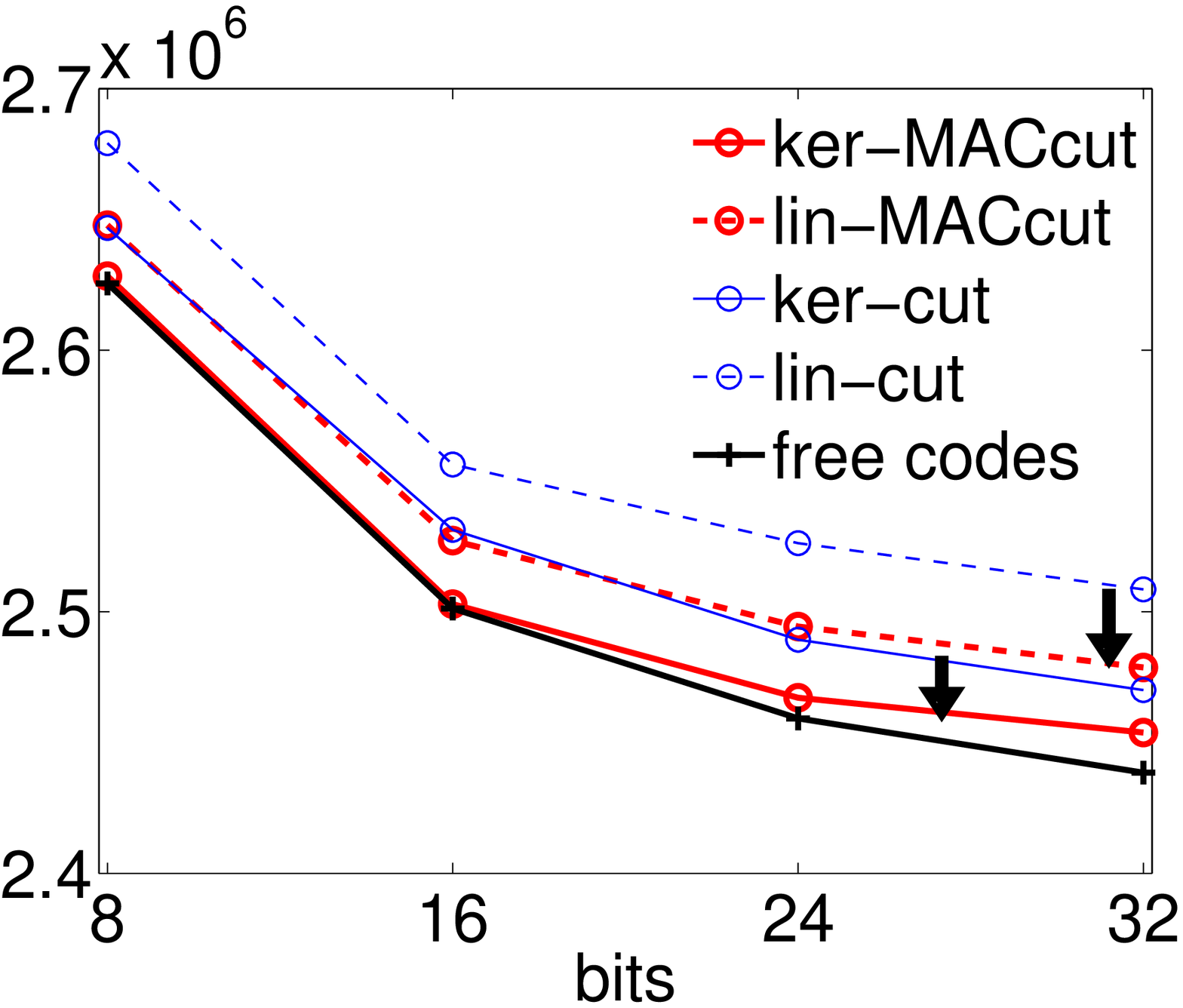} &
    \includegraphics[width=0.4\linewidth]{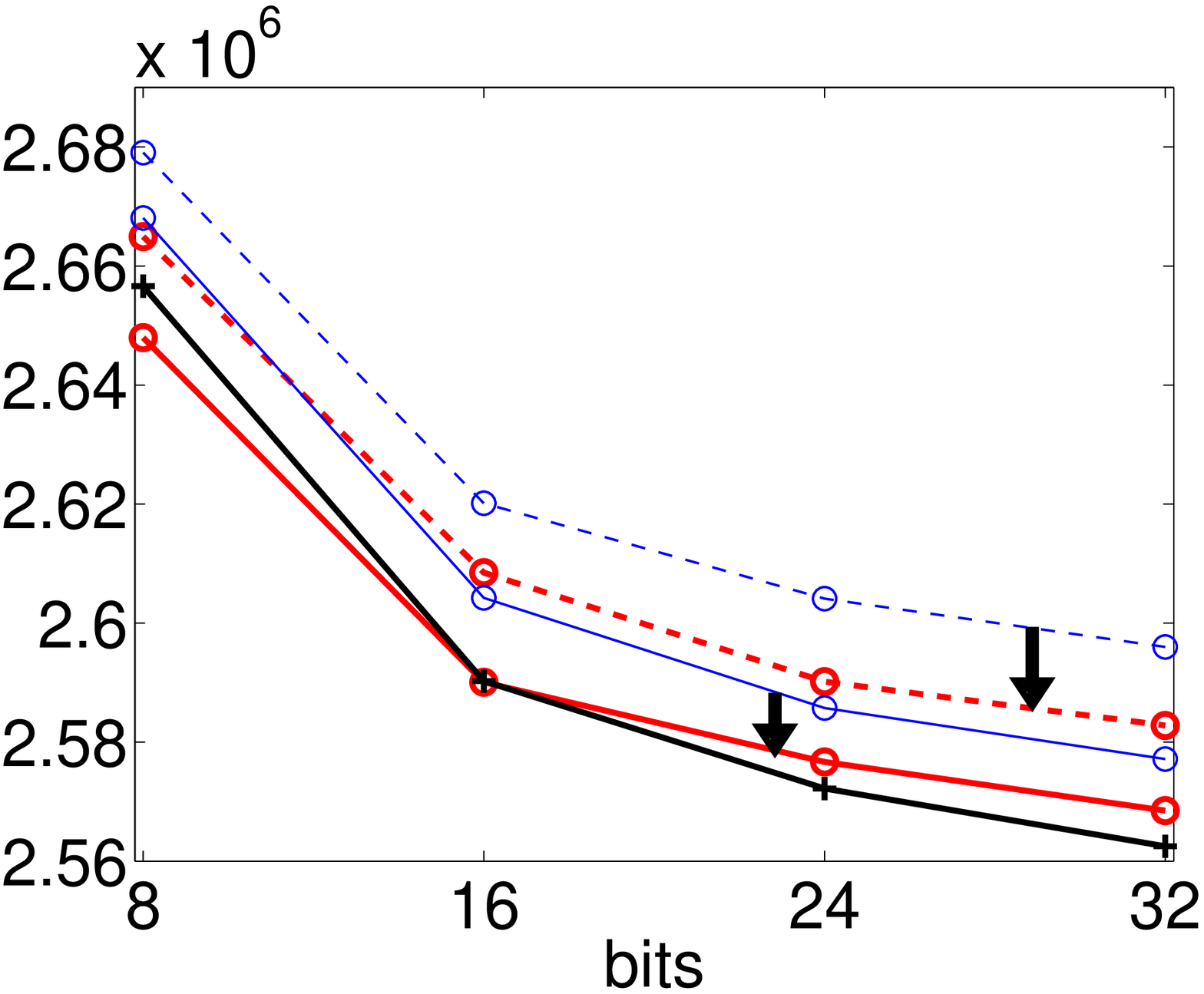}
  \end{tabular}
  \caption{Like fig.~\ref{f:sup-nestederr-freecodes} but for the SIFT1M dataset.}
\label{f:unsup-free}
\end{figure}

\begin{figure}[t!]
  \centering
  \psfrag{rerror}[][t]{loss function \calL}
  \psfrag{iteration}[t][]{iterations}
  \psfrag{K}[][]{$k$}
  \psfrag{precision}[][t]{precision}
  \psfrag{recall}[][]{recall}
  \begin{tabular}{@{}l@{}c@{}c@{}c@{}c@{}}
    & $b=8$ & $b=16$ & $b=24$ & $b=32$ \\
    \hspace{2ex}\rotatebox{90}{\hspace{7ex}precision} &
    \includegraphics[width=0.240\linewidth]{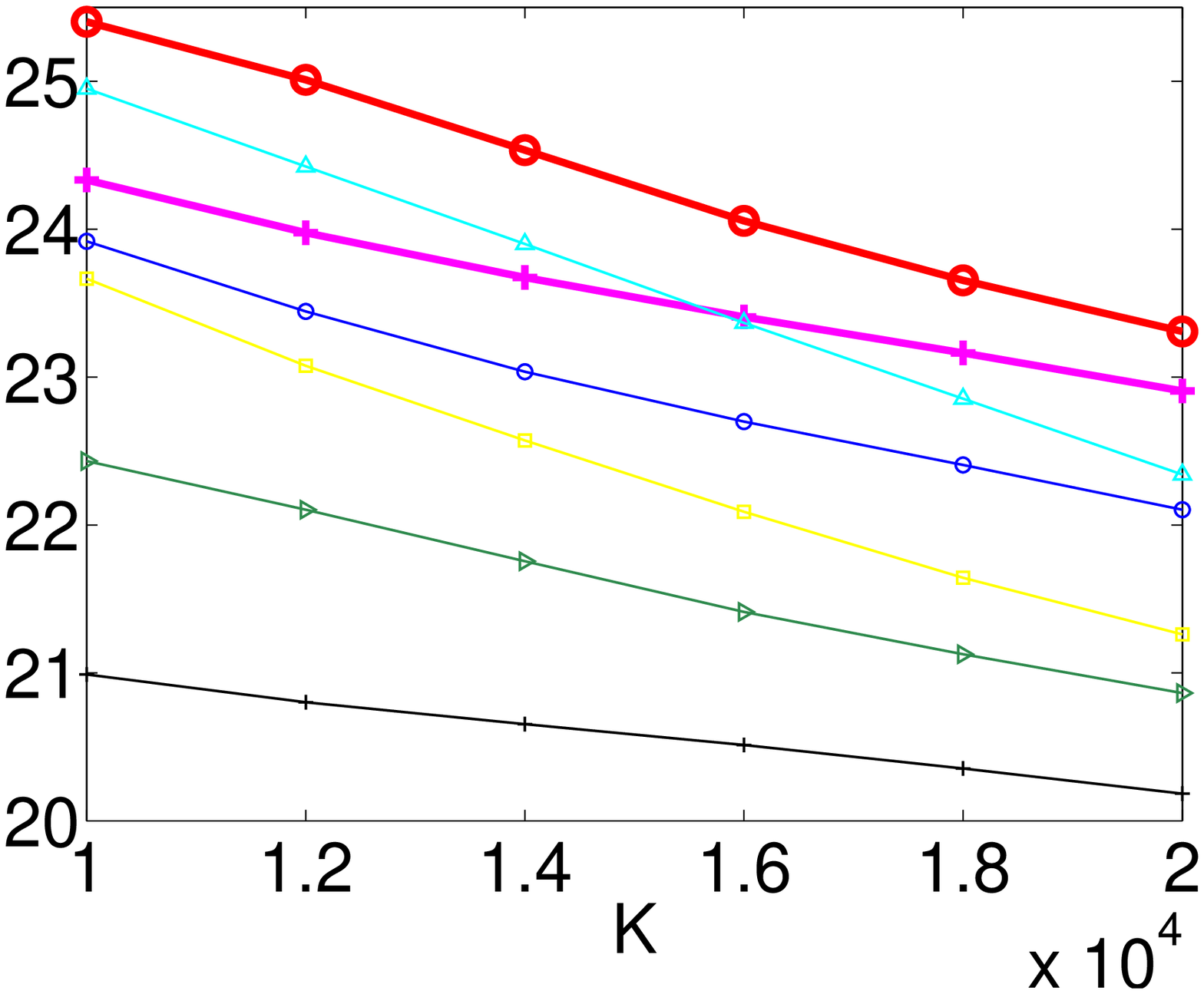} &
    \includegraphics[width=0.240\linewidth]{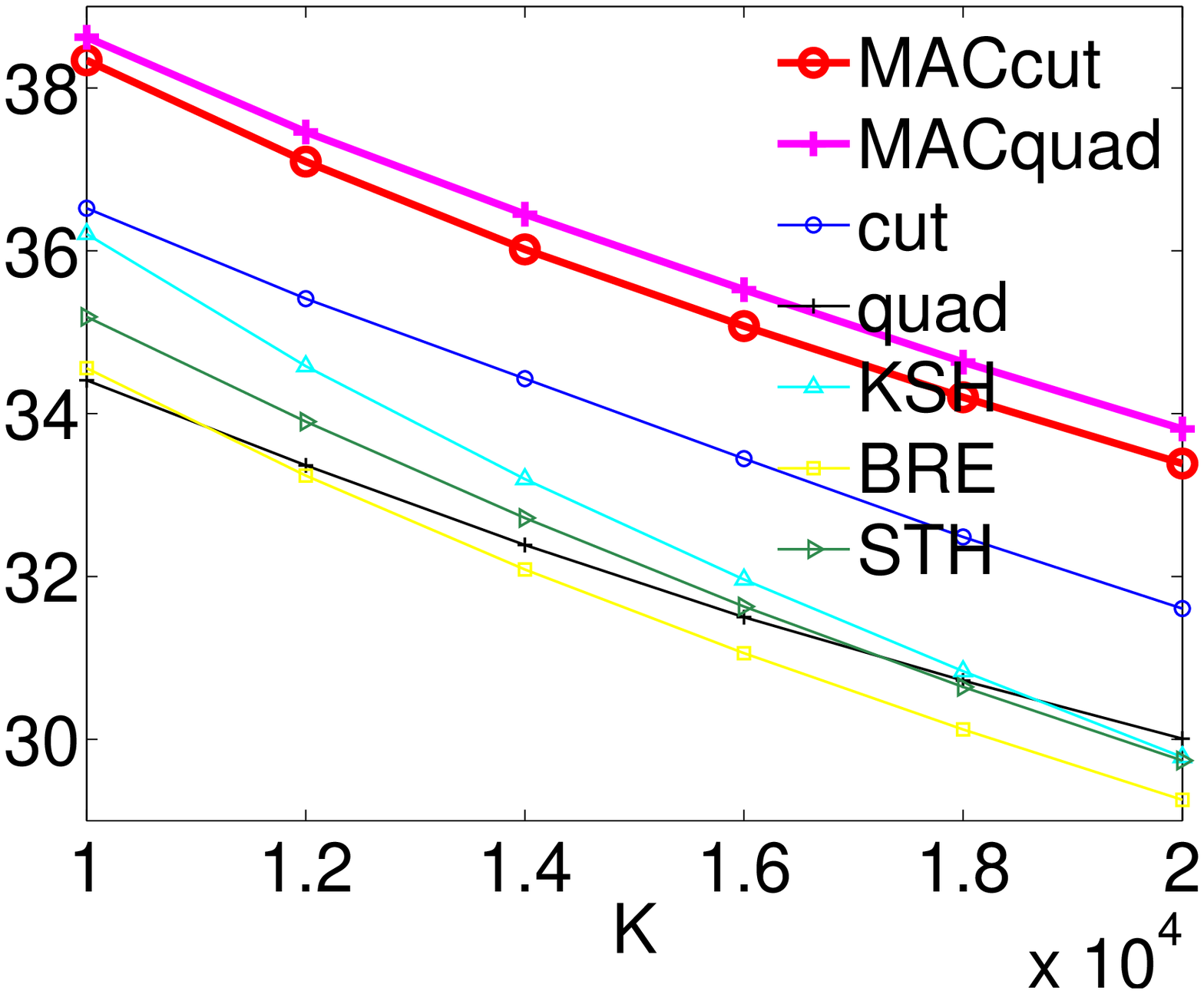} &
    \includegraphics[width=0.240\linewidth]{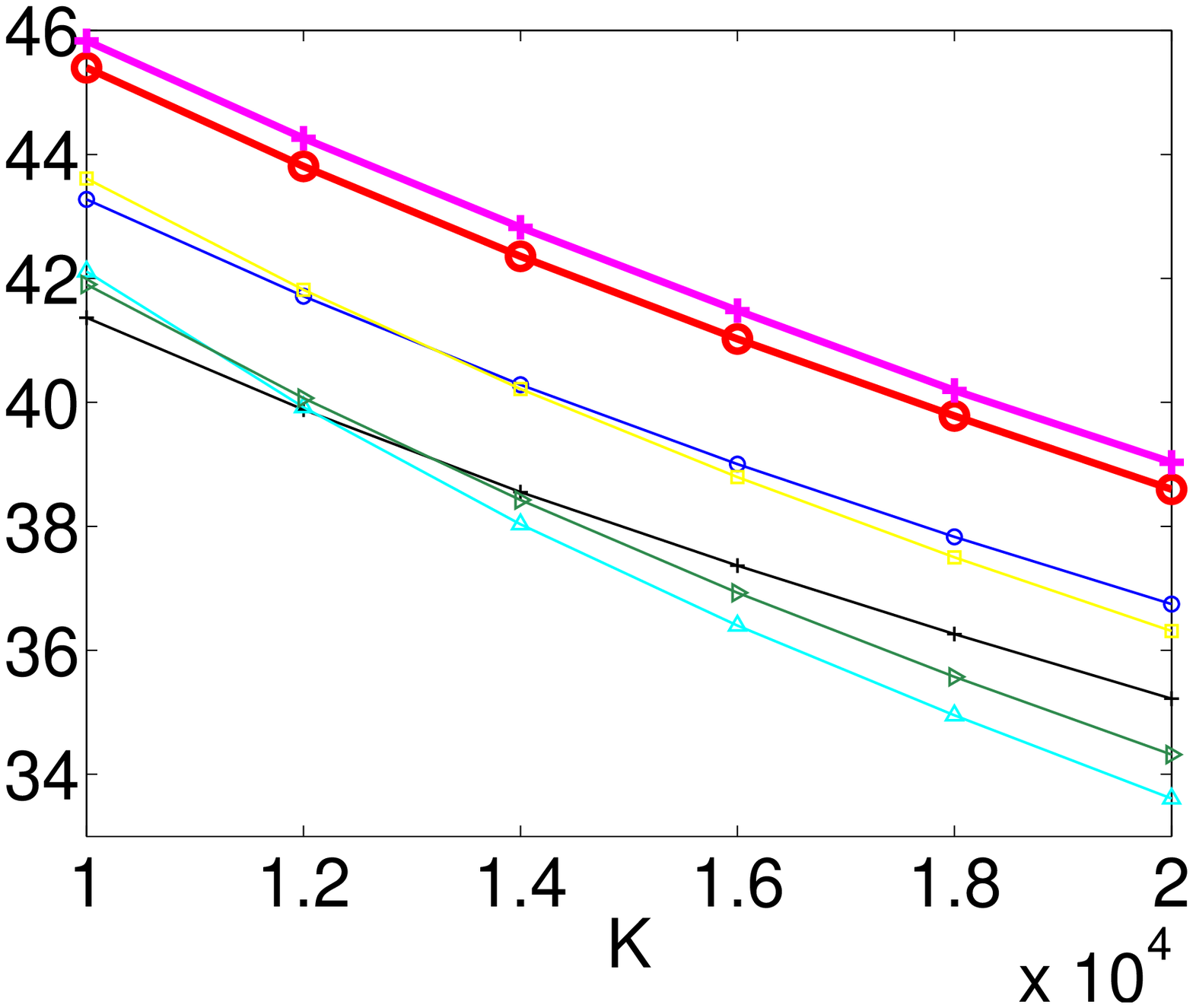} &
    \includegraphics[width=0.240\linewidth]{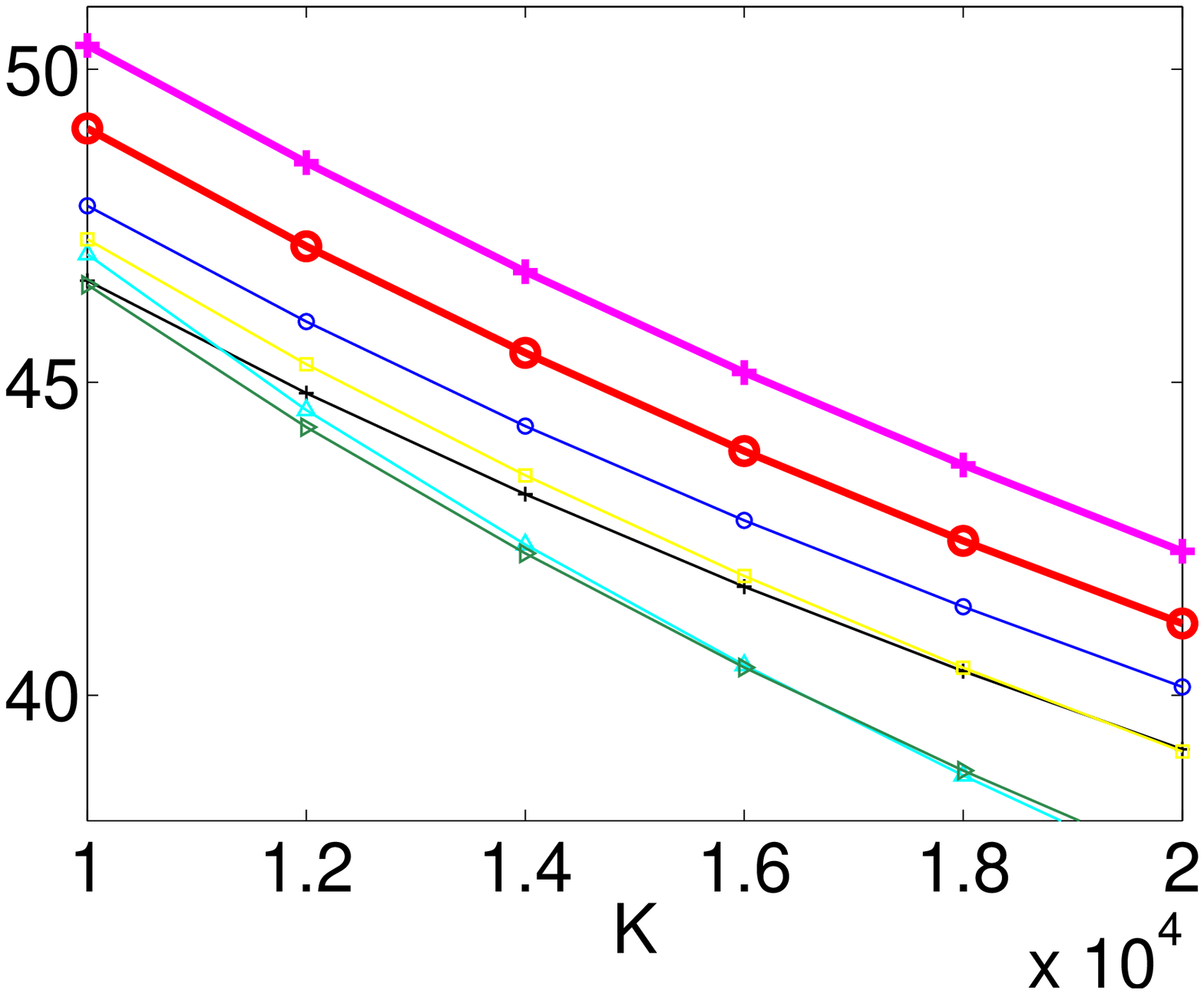} \\[-1ex]
    \hspace{2ex}\rotatebox{90}{\raisebox{3ex}[0pt][0pt]{\makebox[0pt][l]{\hspace{-1ex}SIFT1M (using distance-based pseudolabels)}}\hspace{7ex}precision} &
    \includegraphics[width=0.240\linewidth]{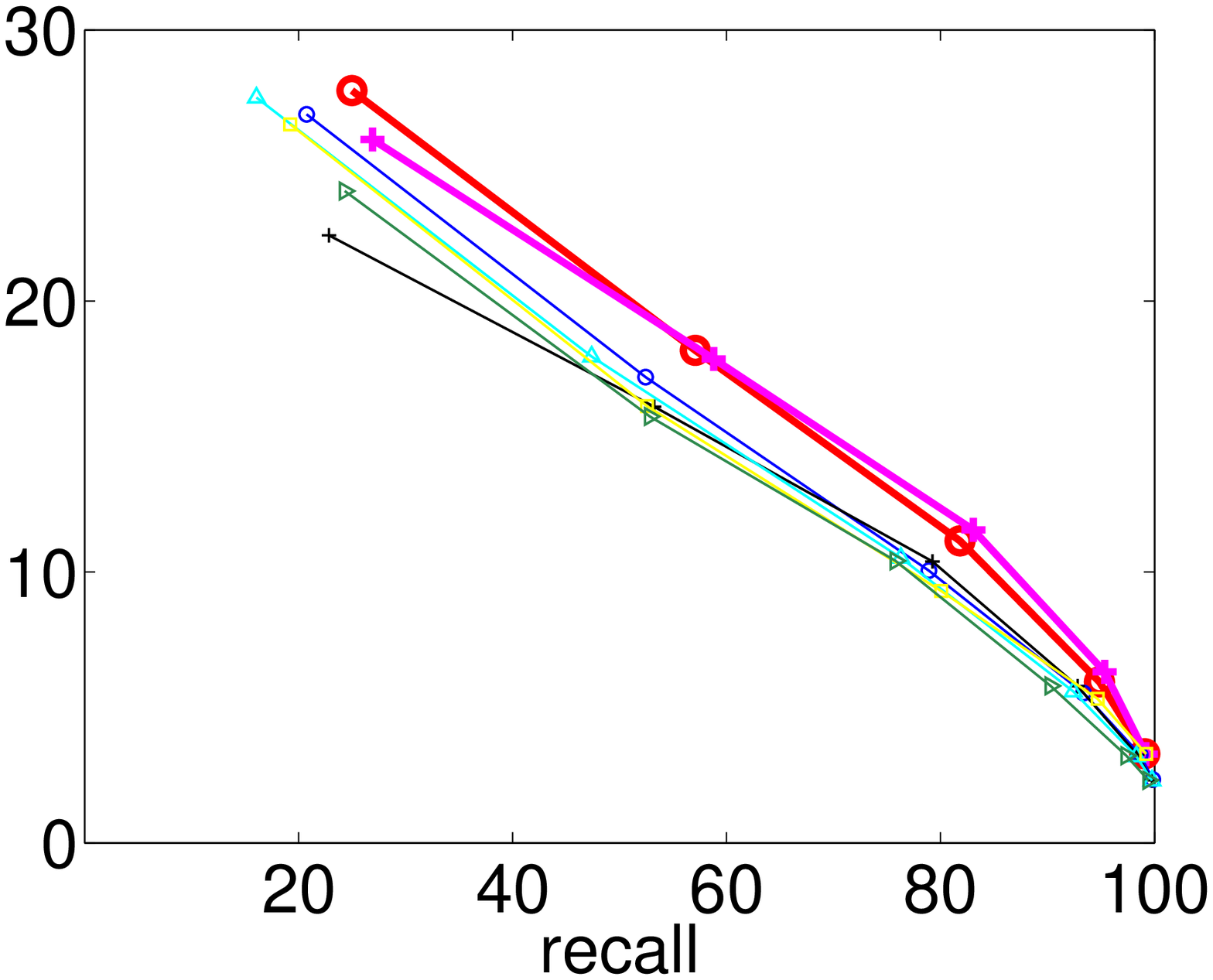} &
    \includegraphics[width=0.240\linewidth]{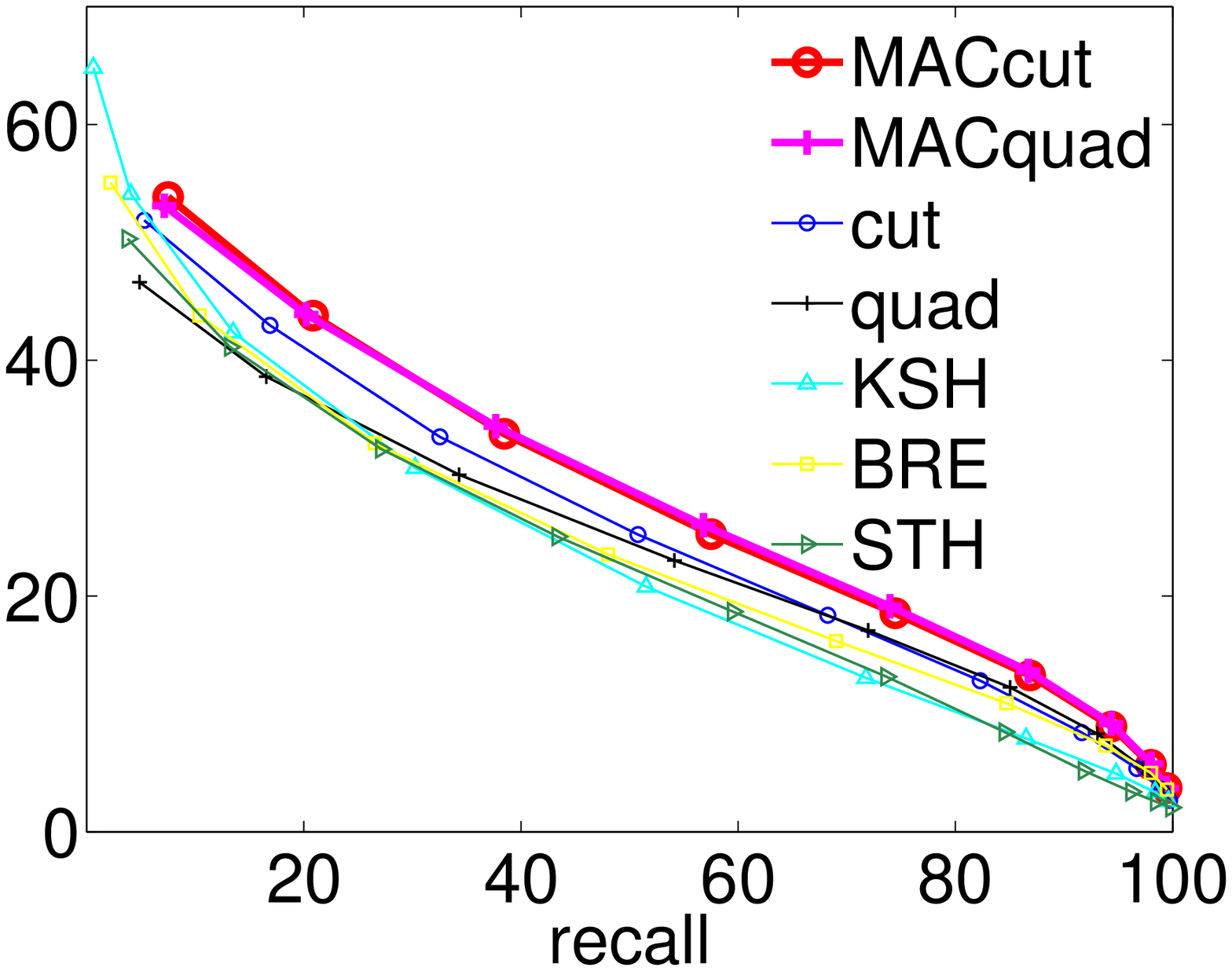} &
    \includegraphics[width=0.240\linewidth]{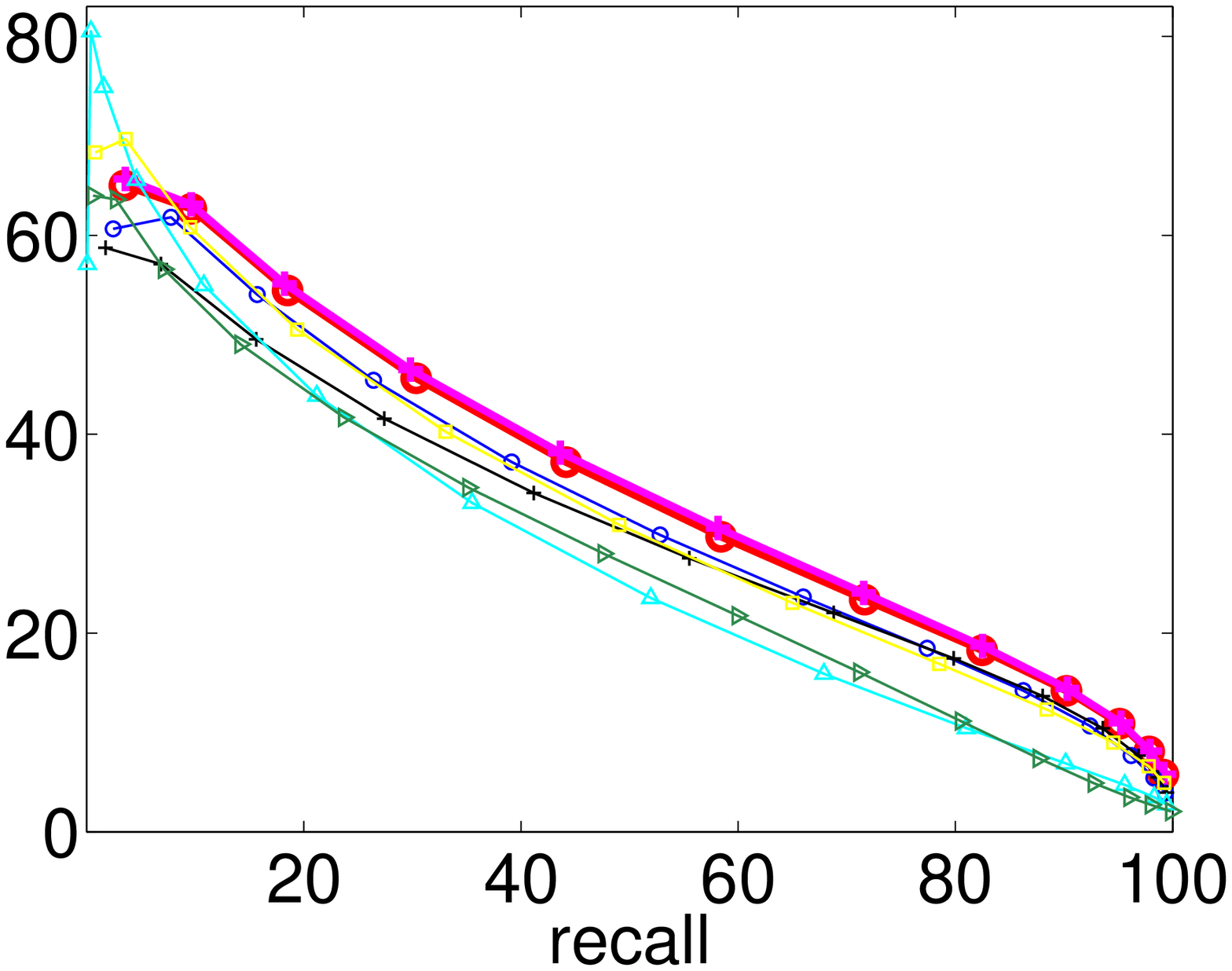} &
    \includegraphics[width=0.240\linewidth]{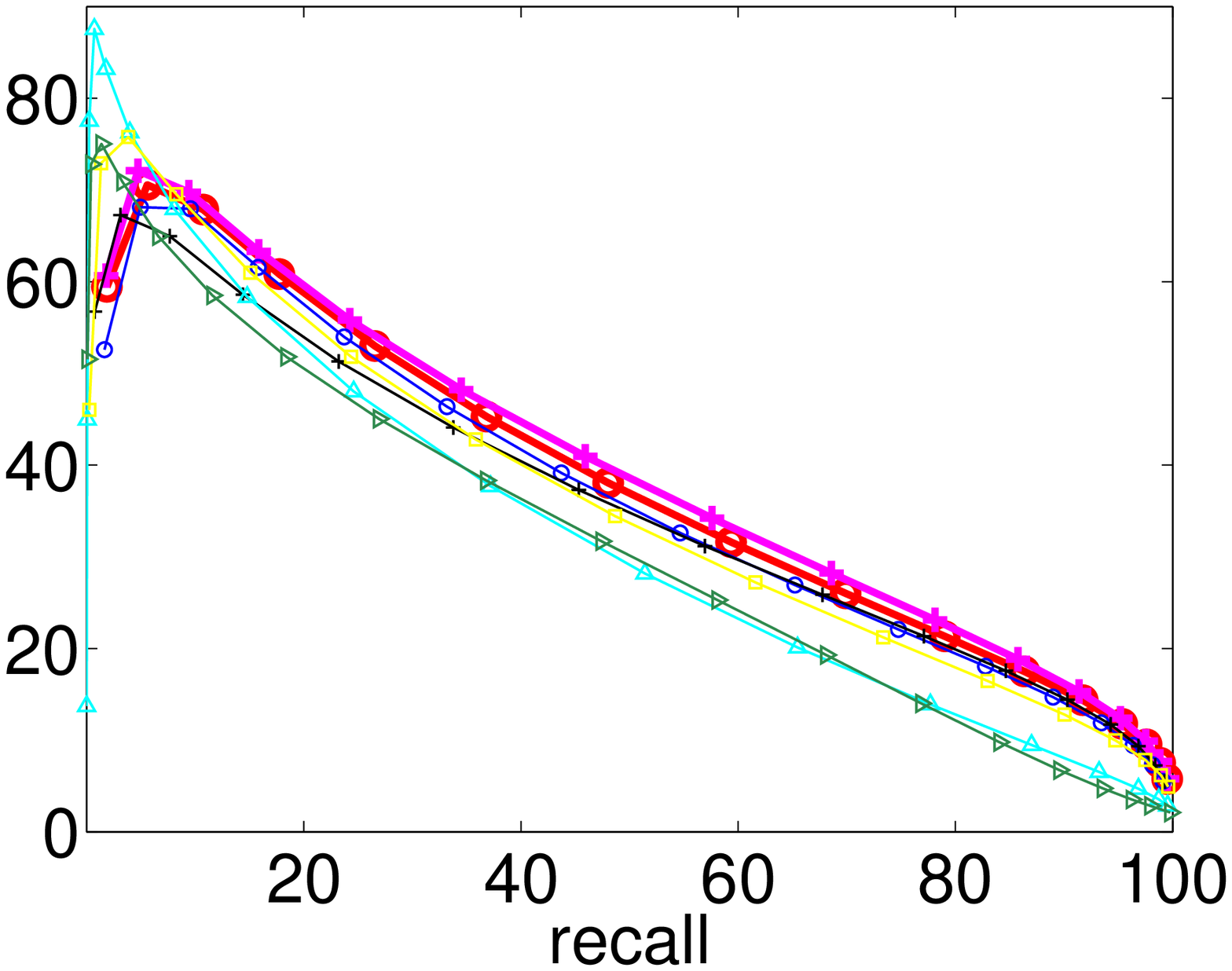}
  \end{tabular} \\[1ex]
  \begin{tabular}{@{}l@{}c@{}c@{}c@{}c@{}}
    \hspace{2ex}\rotatebox{90}{\hspace{7ex}precision} &
    \includegraphics[width=0.240\linewidth]{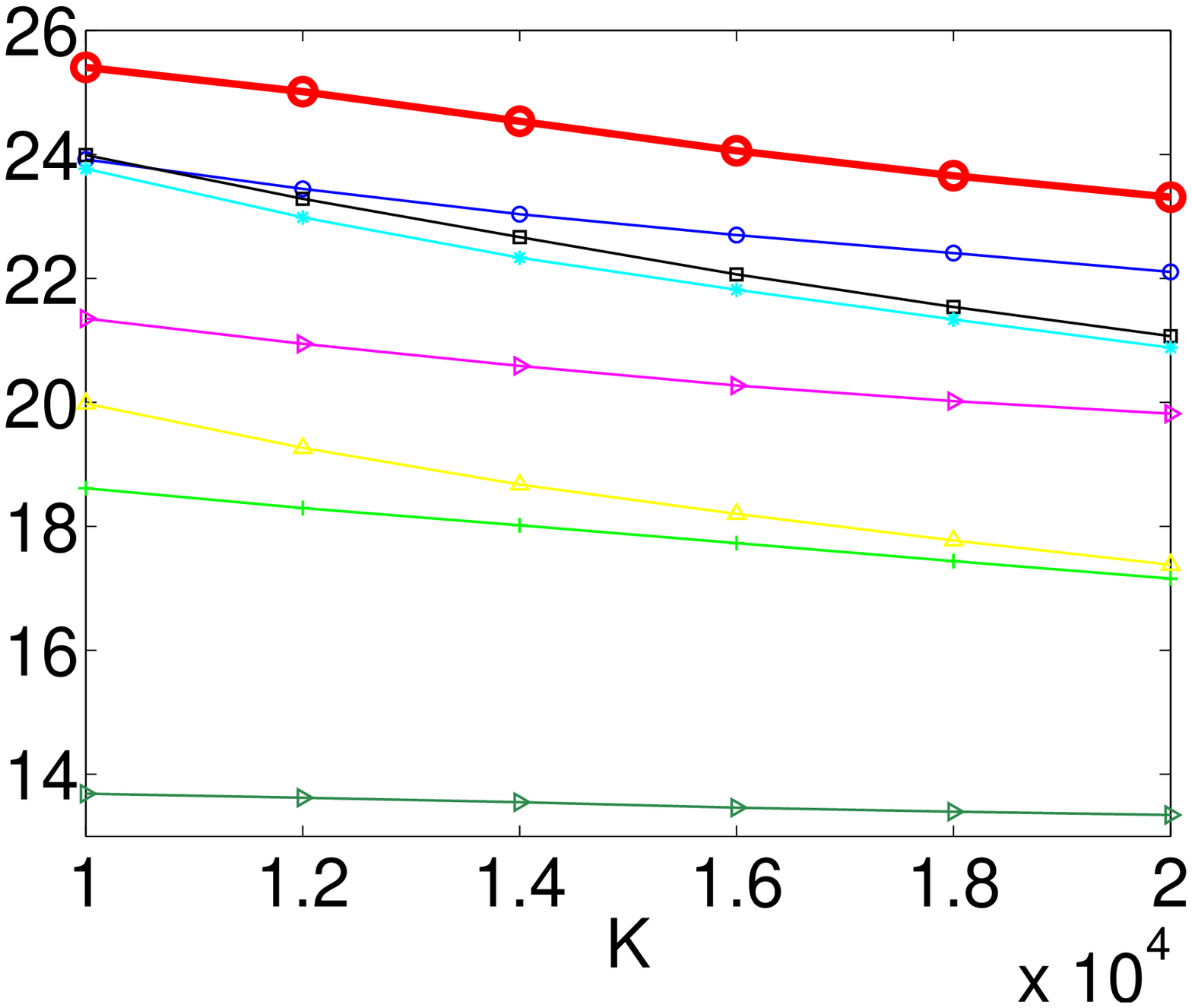} &
    \includegraphics[width=0.240\linewidth]{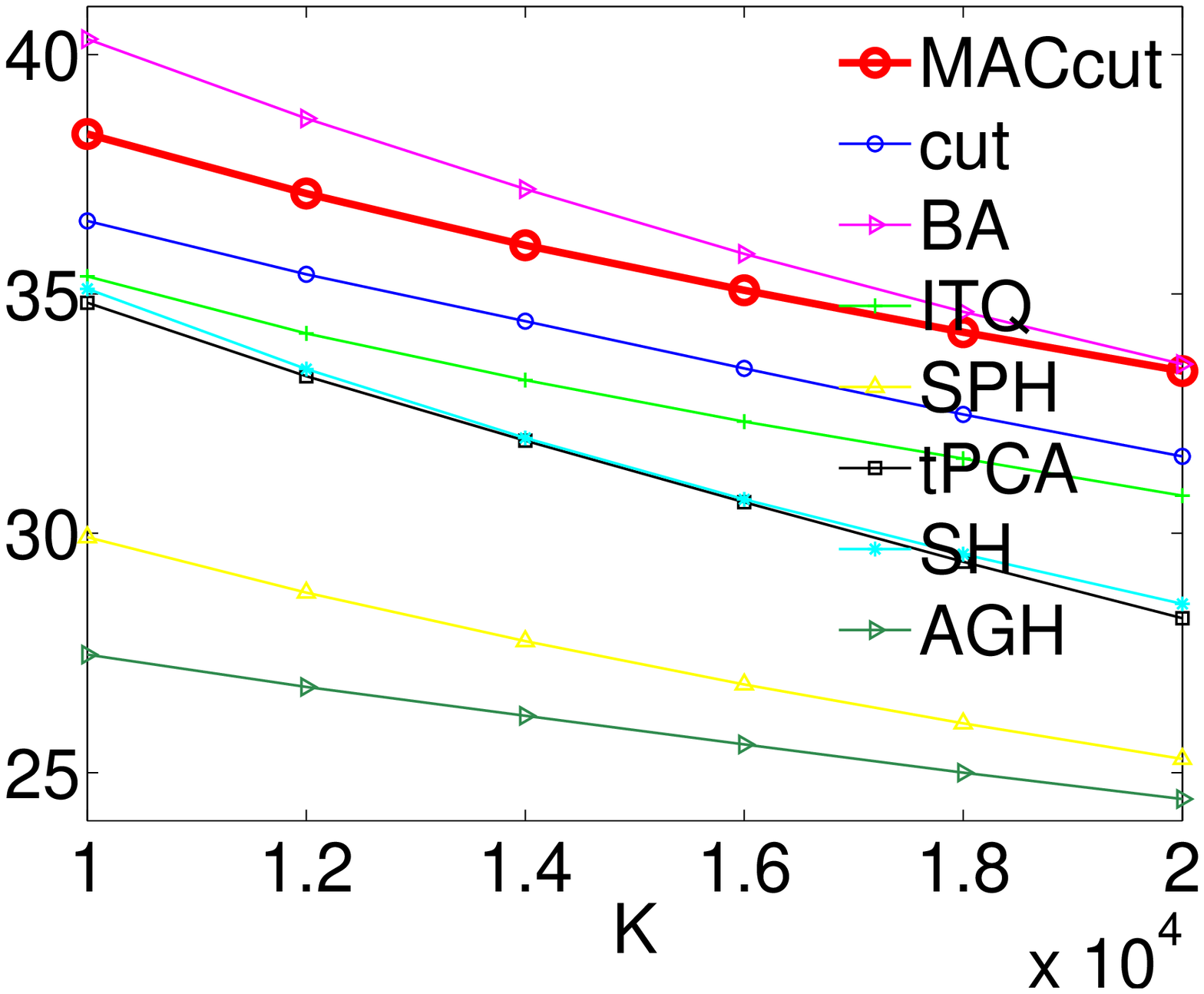} &
    \includegraphics[width=0.240\linewidth]{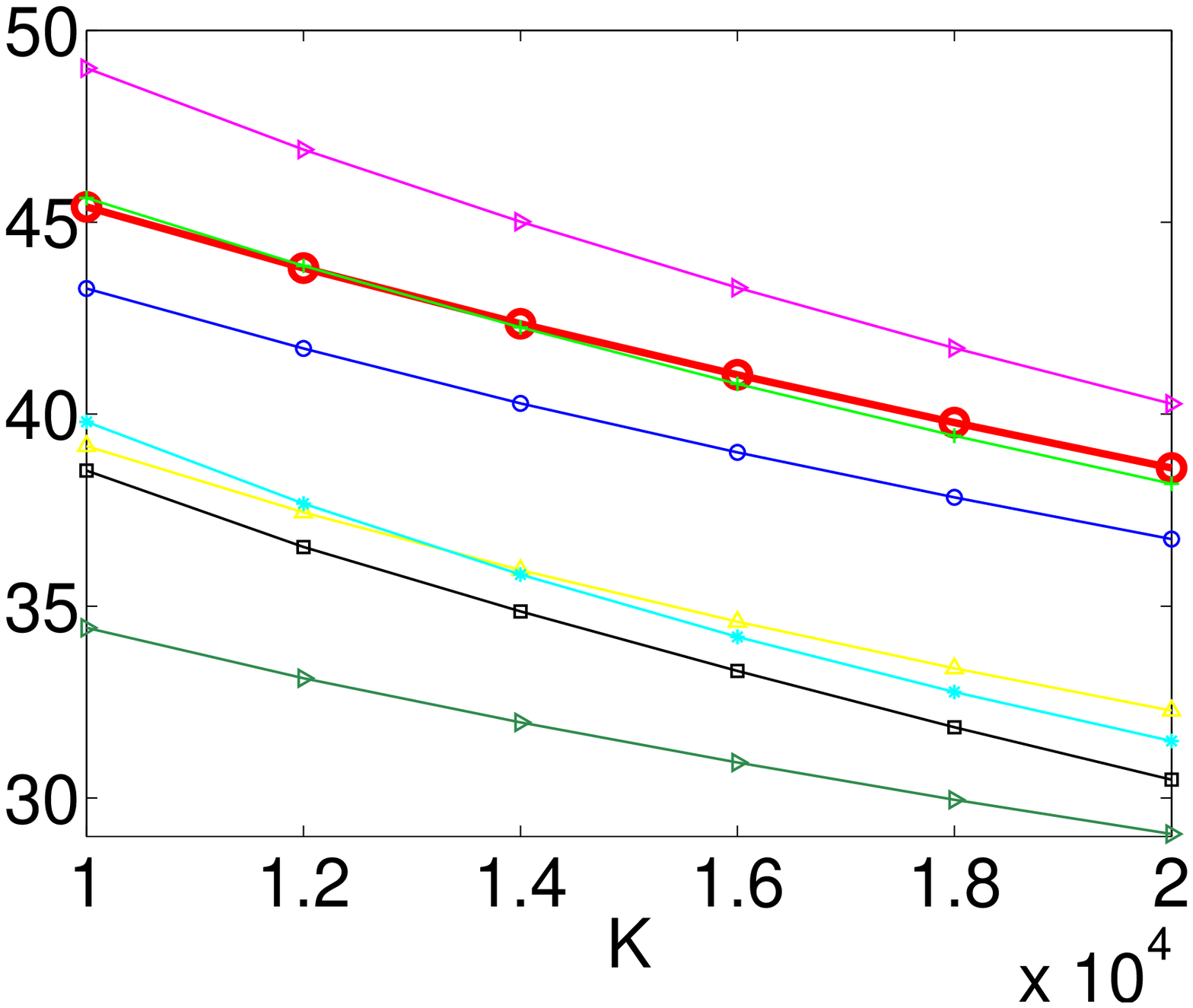} &
    \includegraphics[width=0.240\linewidth]{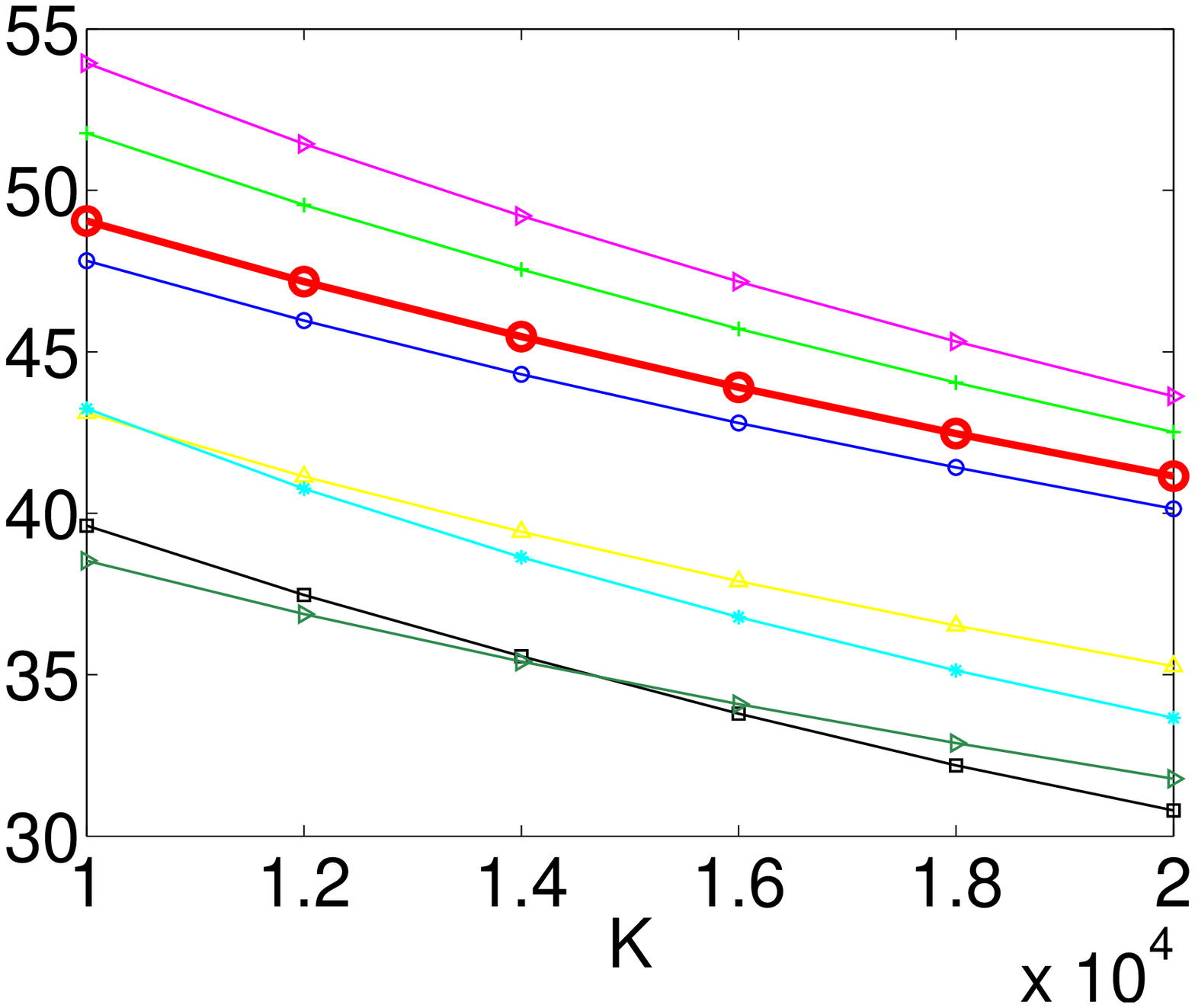} \\[-1ex]
    \hspace{2ex}\rotatebox{90}{\raisebox{3ex}[0pt][0pt]{\makebox[0pt][l]{\hspace{5ex}SIFT1M (unsupervised, no labels)}}\hspace{7ex}precision} &
    \includegraphics[width=0.240\linewidth]{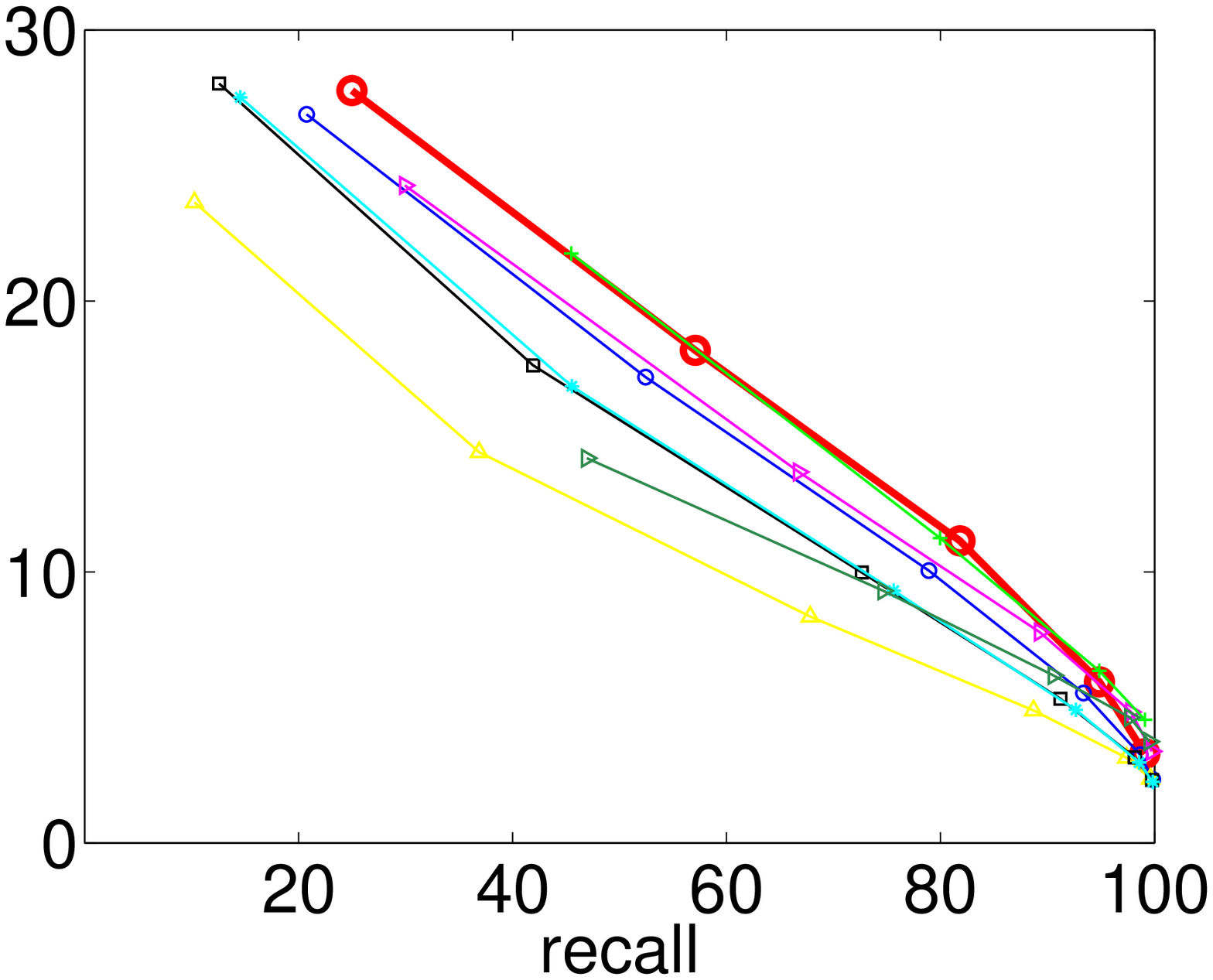} &
    \includegraphics[width=0.240\linewidth]{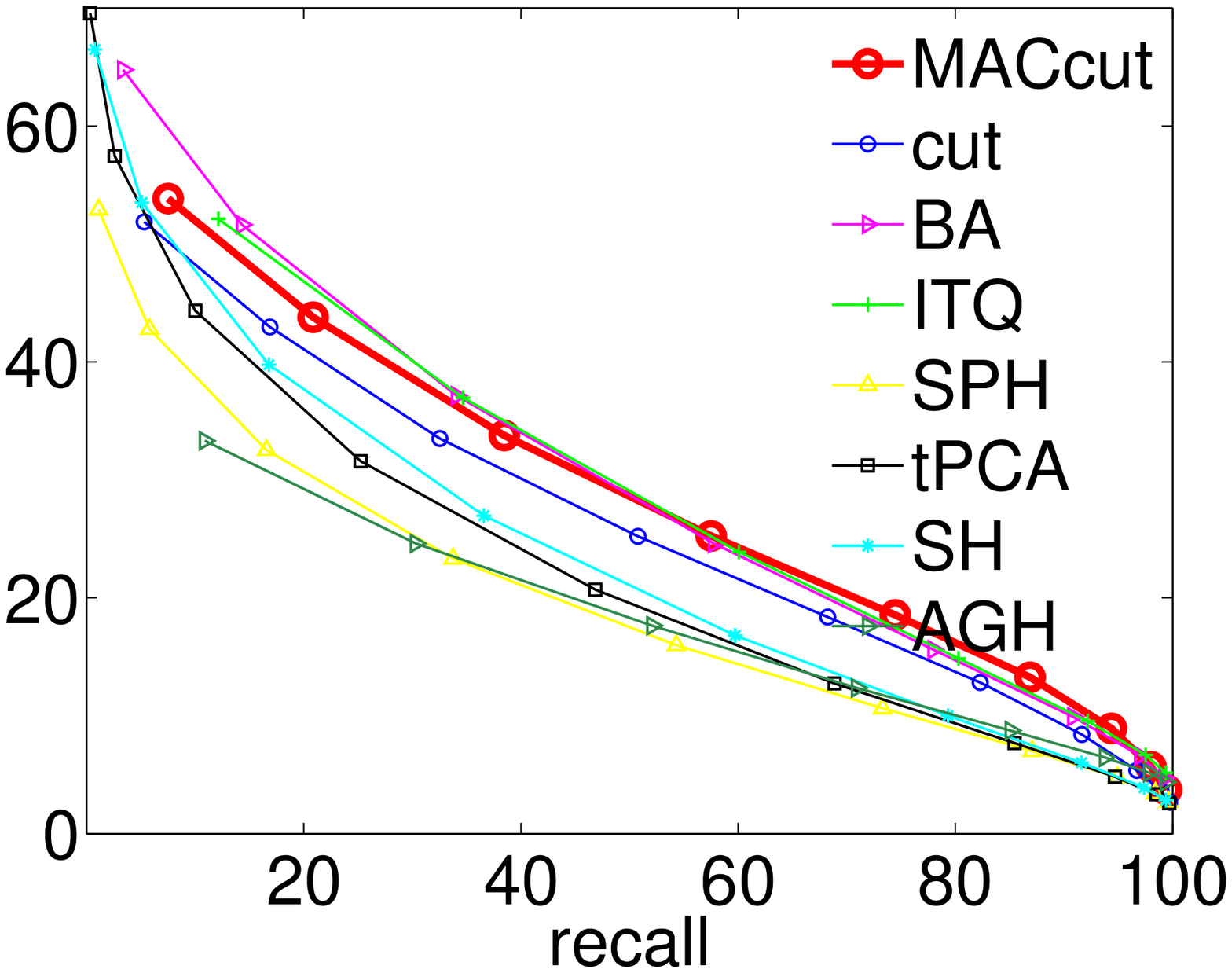} &
    \includegraphics[width=0.240\linewidth]{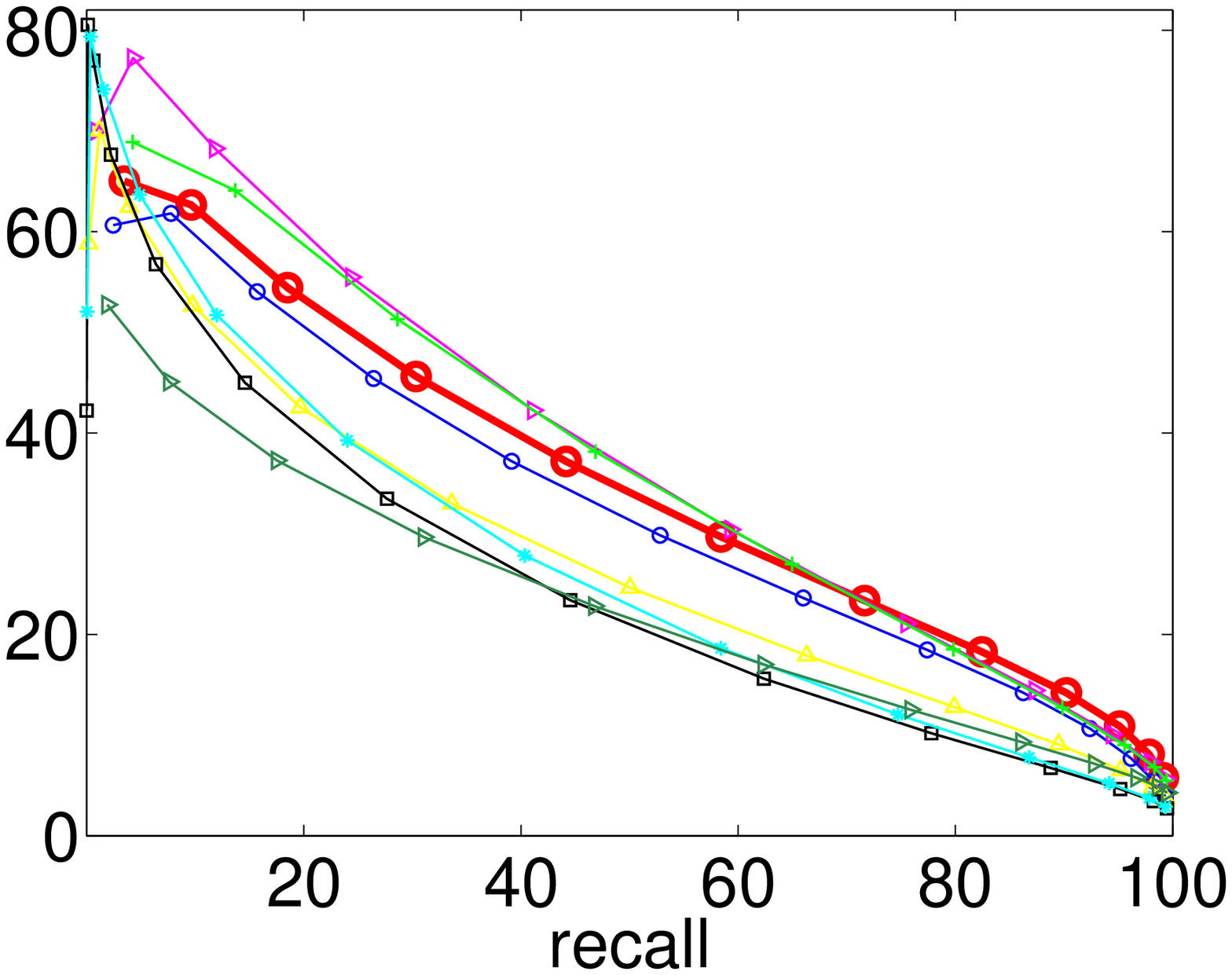} &
    \includegraphics[width=0.240\linewidth]{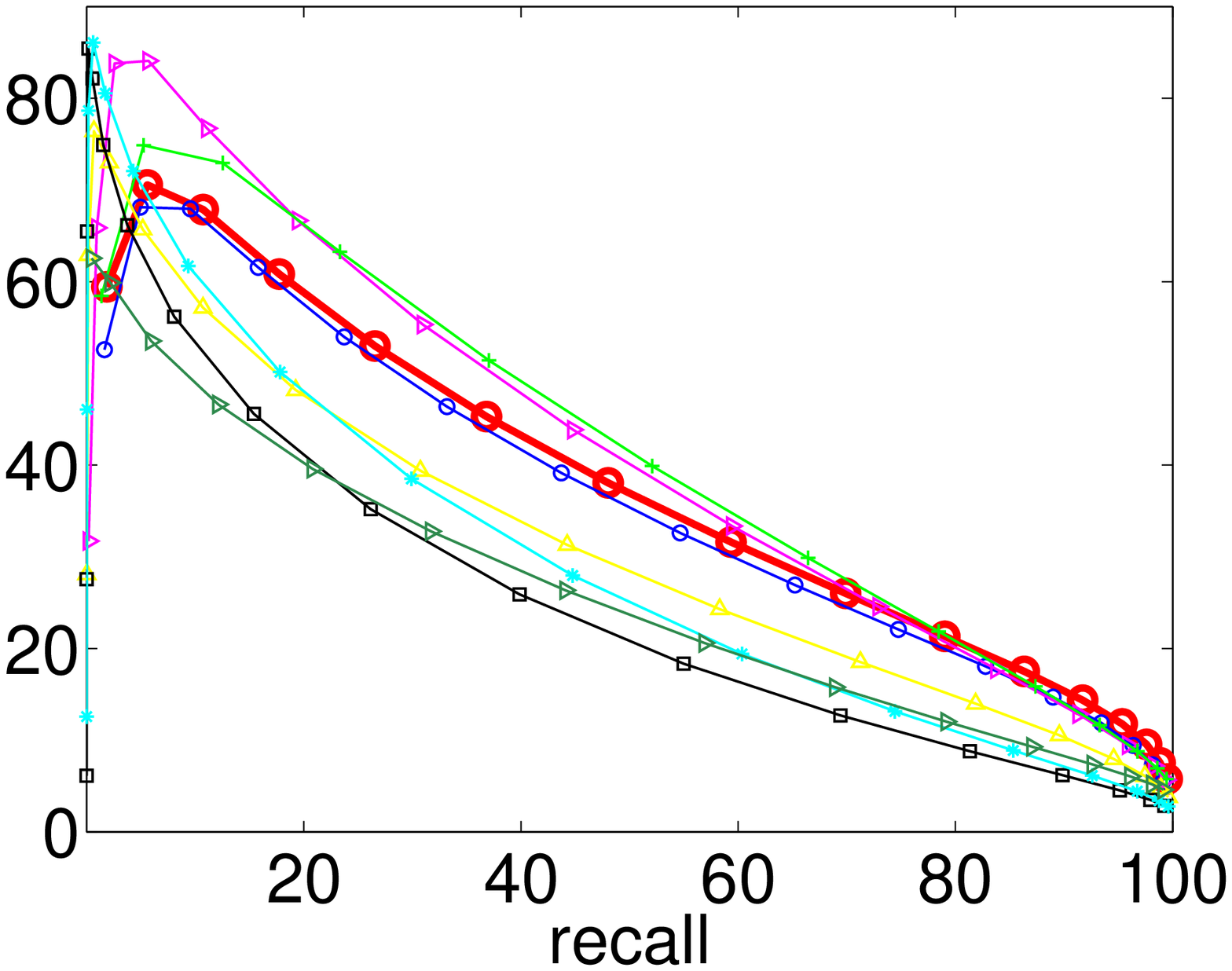}
  \end{tabular}
  \caption{As in fig.~\ref{f:unsup-comparison} but using the cosine similarity instead of the Euclidean distance to find neighbors (i.e., all the points are centered and normalized before training and testing), on SIFT1M.}
  \label{f:unsup-cosine}
\end{figure}

\begin{figure}[b!]
  \centering
  \psfrag{bits}[][b]{$b$}
  \psfrag{entropy}[][t]{$b_{\text{eff}}$}
  \begin{tabular}{@{}c@{\hspace{1em}}c@{}}
    \includegraphics[width=0.32\linewidth]{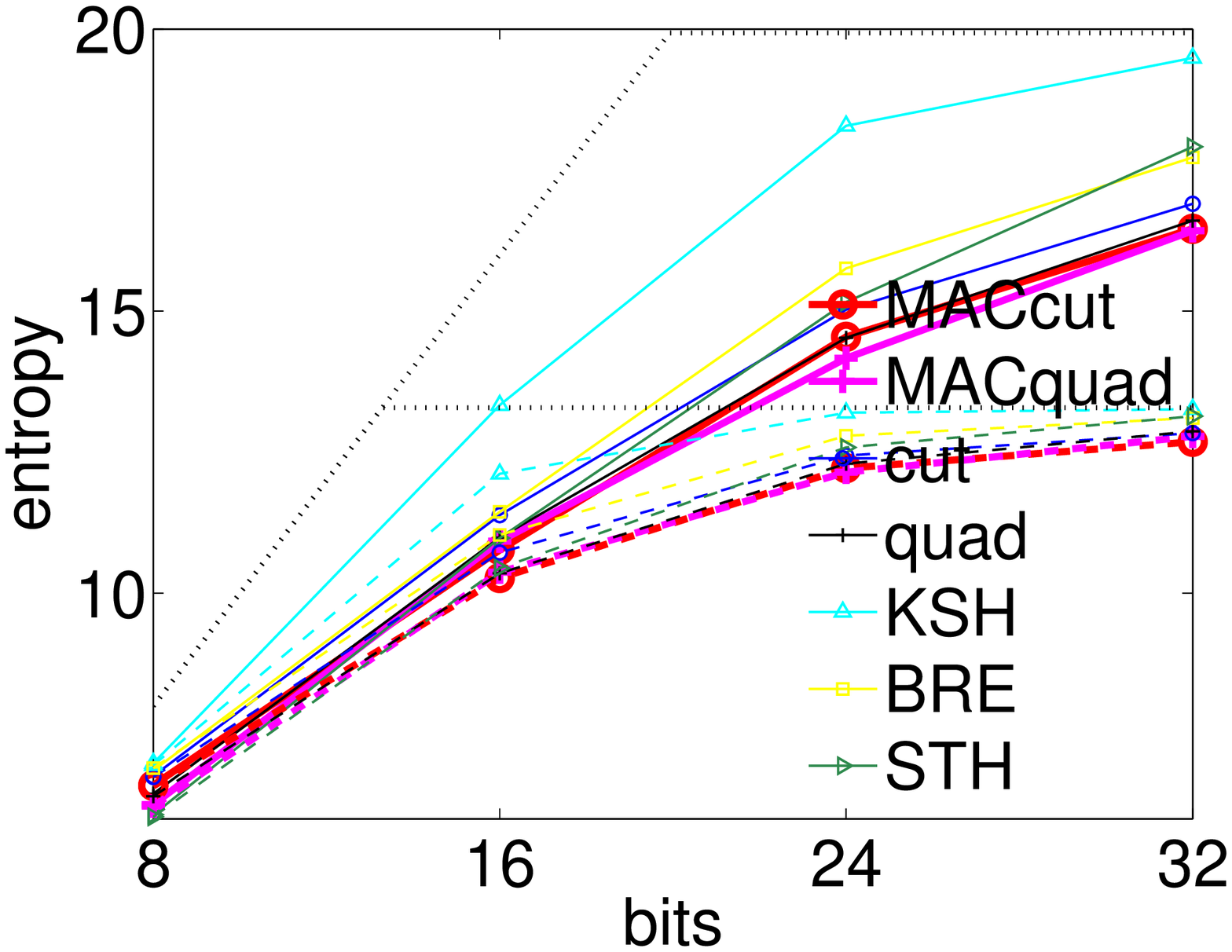} &
    \includegraphics[width=0.296\linewidth]{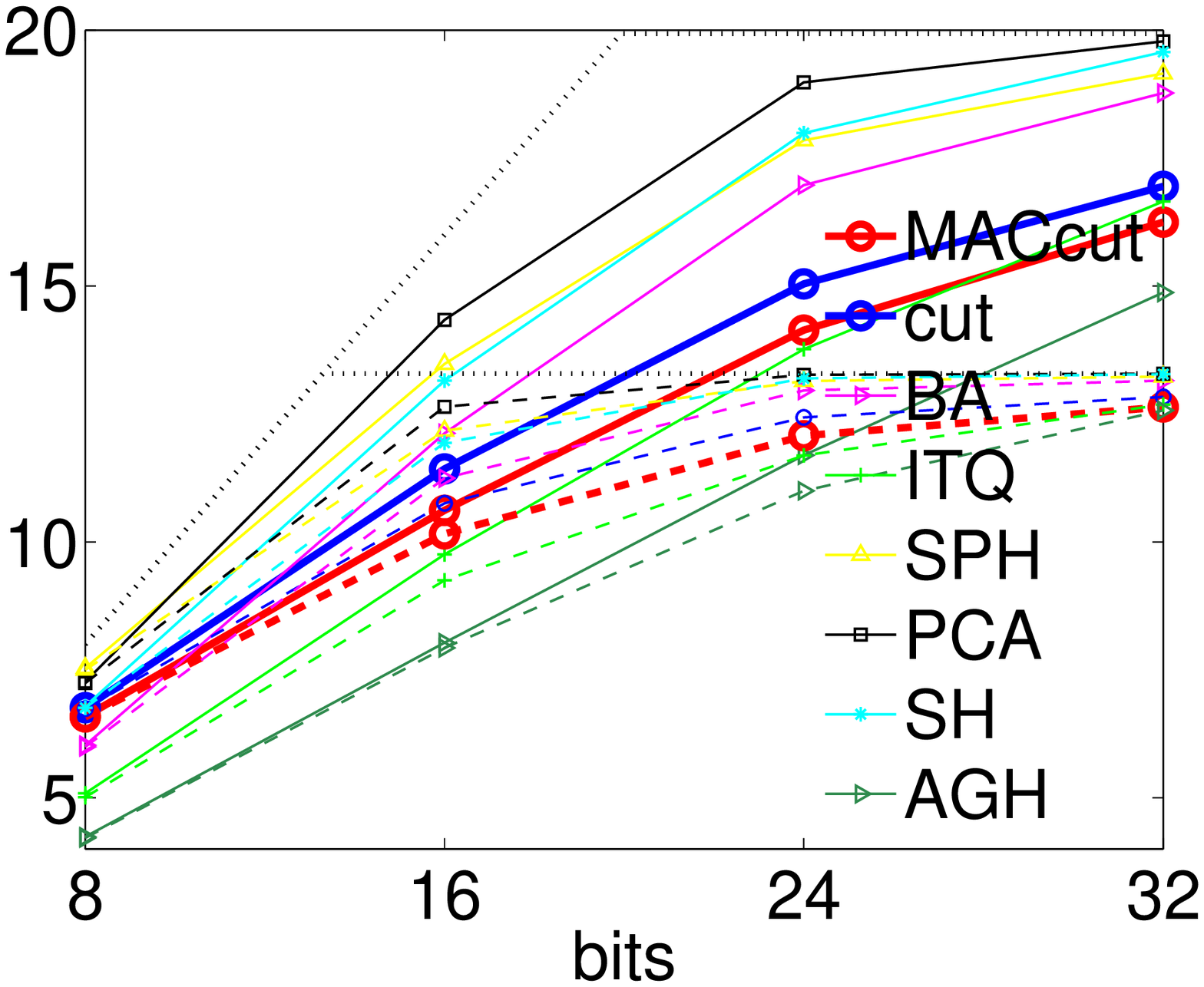}
  \end{tabular}
  \caption{Code utilization in effective number of bits $b_{\text{eff}}$ (entropy of code distribution) of different hashing algorithms, using $b = 8$ to $32$ bits, for the SIFT1M dataset. The plots correspond to the codes obtained by the algorithms in figure~\ref{f:unsup-comparison}, with solid lines for the training set and dashed lines for the test set. The two diagonal-horizontal black dotted lines give the upper bound (maximal code utilization) $\min(b,\log_2{N})$ on $b_{\text{eff}}$ of any algorithm for the training and test sets (where $N$ is the size of the training or test set).}
  \label{f:unsup-entropy}
\end{figure}

\paragraph{Comparison using code utilization}

Fig.~\ref{f:unsup-entropy} shows the results (for all methods on SIFT1M) in effective number of bits $b_{\text{eff}}$. This is a measure of code utilization of a hash function introduced by \citet{CarreirRaziper15a}, defined as the entropy of the code distribution. That is, given the $N$ codes $\z_1,\dots,\z_N \in \{0,1\}^b$ for the training set, we consider them as samples of a distribution over the $2^b$ possible codes. The entropy of this distribution, measured in bits, is between $0$ (when all $N$ codes are equal) and $\min(b,\log_2{N})$ (when all $N$ codes are distributed as uniformly as possible). We do the same for the test set. Although code utilization correlates to some extent with precision/recall when ranking different methods, a large $b_{\text{eff}}$ does not guarantee a good hash function, and indeed, tPCA (which usually achieves a low precision compared to the state-of-the-art) typically achieves the largest $b_{\text{eff}}$; see the discussion in \citet{CarreirRaziper15a}. However, a large $b_{\text{eff}}$ does indicate a better use of the available codes (and fewer collisions if $N < 2^b$), and $b_{\text{eff}}$ has the advantage over precision/recall that it does not depend on any user parameters (such as ground truth size or retrieved set size), so we can compare all binary hashing methods with a single number $b_{\text{eff}}$ (for a given number of bits $b$). It is particularly useful to compare methods that are optimizing the same objective function. With this in mind, we can compare \emph{MACcut} with \emph{cut} and \emph{MACquad} with \emph{quad} because these pairs of methods optimize the same objective function.

\clearpage

\subsubsection*{Acknowledgments}

Work supported by NSF award IIS--1423515. We thank Ming-Hsuan Yang, Yi-Hsuan Tsai and Mehdi Alizadeh (UC Merced) for useful discussions.


\end{document}